%% file: top.tex
\documentclass[10pt,twocolumn,letterpaper]{article}

\usepackage{iccv}
\usepackage{times}
\usepackage{epsfig}
\usepackage{graphicx}
\usepackage{amsmath}
\usepackage{amssymb}
\usepackage[pagebackref=true,breaklinks=true,letterpaper=true,colorlinks,bookmarks=false]{hyperref}

\iccvfinalcopy
\def\iccvPaperID{****}

\ificcvfinal\fi

\input{commands} 
\begin{document}
\title{Relating Adversarially Robust Generalization to Flat Minima}
\author{David Stutz$^{1}$ \qquad Matthias Hein$^{2}$ \qquad Bernt Schiele$^{1}$\\
$^1$Max Planck Institute for Informatics, Saarland Informatics Campus, Saarbr\"{u}cken\\
$^2$University of T\"{u}bingen, T\"{u}bingen\\
{\tt\small \{david.stutz,schiele\}@mpi-inf.mpg.de, matthias.hein@uni-tuebingen.de}	
}

\maketitle

\ificcvfinal\thispagestyle{empty}\fi

\input{sec_abstract}
\input{sec_introduction}
\input{sec_related_work}
\input{sec_main}
\input{sec_conclusion}

{\small
\bibliographystyle{ieee_fullname}
\bibliography{bibliography}
}

\clearpage
\begin{appendix}
\input{supp_introduction}
\input{supp_related_work}
\input{supp_main}
\end{appendix}
\end{document}

%% file: commands.tex
\usepackage{amsmath}
\usepackage{xspace}
\usepackage{subcaption}
\usepackage{bbm}
\usepackage{nicefrac}
\usepackage{tabularx}
\usepackage{colortbl}
\usepackage{multirow}
\usepackage{amsmath}
\usepackage{amsthm}
\usepackage{amsfonts}
\usepackage{xspace}
\usepackage{mathtools}
\usepackage{algorithm}
\usepackage{algorithmicx}
\usepackage{algpseudocode}
\algrenewcommand\algorithmicindent{0.75em}%
\algtext*{EndWhile}
\algtext*{EndIf}
\algtext*{EndFor}
\algtext*{EndProcedure}
\usepackage{caption}
\usepackage{wrapfig}
\usepackage{array}
\usepackage{placeins}

\newcolumntype{L}[1]{>{\raggedright\let\newline\\\arraybackslash\hspace{0pt}}m{#1}}
\newcolumntype{C}[1]{>{\centering\let\newline\\\arraybackslash\hspace{0pt}}m{#1}}
\newcolumntype{R}[1]{>{\raggedleft\let\newline\\\arraybackslash\hspace{0pt}}m{#1}}

\def\Vhrulefill{\leavevmode\hrule height 0.7ex depth \dimexpr0.4pt-0.7ex with 1cm\kern0pt}

\PassOptionsToPackage{dvipsnames}{xcolor}
\usepackage{tcolorbox}
\tcbuselibrary{skins,breakable}
\definecolor{colorbrewer0}{RGB}{45,45,45}
\definecolor{colorbrewer1}{RGB}{228,26,28}
\definecolor{colorbrewer2}{RGB}{55,126,184}
\definecolor{colorbrewer3}{RGB}{77,175,74}
\definecolor{colorbrewer4}{RGB}{152,78,163}
\definecolor{colorbrewer5}{RGB}{255,127,0}
\definecolor{colorbrewer6}{RGB}{255,255,51}
\definecolor{colorbrewer7}{RGB}{166,86,40}
\definecolor{colorbrewer8}{RGB}{247,129,191}
\definecolor{colorbrewer9}{RGB}{153,153,153}
\definecolor{colorbrewer10}{RGB}{24,167,181}

\definecolor{plot0}{RGB}{166, 206, 227}
\definecolor{plot1}{RGB}{31, 120, 180}
\definecolor{plot2}{RGB}{251, 154, 153}
\definecolor{plot3}{RGB}{178, 223, 138}
\definecolor{plot4}{RGB}{51, 160, 44}
\definecolor{plot5}{RGB}{227, 26, 28}
\definecolor{plot6}{RGB}{253, 191, 111}
\definecolor{plot7}{RGB}{255, 127, 0}
\definecolor{plot8}{RGB}{202, 178, 214}
\definecolor{plot9}{RGB}{106, 61, 154}
\definecolor{plot10}{RGB}{245, 245, 145}
\definecolor{plot11}{RGB}{177, 89, 40}

\definecolor{gray}{RGB}{65, 65, 65}
\definecolor{darkred}{RGB}{255, 0, 0}
\definecolor{darkgreen}{RGB}{25, 100, 25}
\definecolor{darkmagenta}{RGB}{255, 0, 255}
\definecolor{darkblue}{RGB}{0, 0, 255}

\makeatletter
\DeclareRobustCommand\onedot{\futurelet\@let@token\@onedot}
\def\@onedot{\ifx\@let@token.\else.\null\fi\xspace}

\def\eg{\emph{e.g}\onedot} 
\def\ie{\emph{i.e}\onedot} 
\def\cf{\emph{c.f}\onedot} 
\def\etc{\emph{etc}\onedot} 
\def\wrt{w.r.t\onedot} 

\makeatother

\makeatletter
\newcommand\bigDiamond{\mathop{\mathpalette\bigDi@mond\relax}}
\newcommand\bigDi@mond[2]{%
  \vcenter{\hbox{\m@th
    \scalebox{\ifx#1\displaystyle 2\else1.2\fi}{$#1\Diamond$}%
  }}%
}
\newcommand\bigLozenge{\mathop{\mathpalette\bigL@zenge\relax}}
\newcommand\bigL@zenge[2]{%
  \vcenter{\hbox{\m@th
    \scalebox{\ifx#1\displaystyle 2\else1.2\fi}{$#1\blacklozenge$}%
  }}%
}
\makeatother

\newcommand{\figref}[1]{\Fig~\ref{#1}}
\newcommand{\secref}[1]{\Sec~\ref{#1}}

\newcommand{\eqnref}[1]{\Eq~\eqref{#1}}
\newcommand{\tabref}[1]{\Tab~\ref{#1}}

\makeatletter
\DeclareRobustCommand\onedot{\futurelet\@let@token\@onedot}
\def\@onedot{\ifx\@let@token.\else.\null\fi\xspace}
\def\eg{e.g\onedot} 
\def\ie{i.e\onedot} 
 
\def\cf{cf\onedot} 
\def\etc{etc\onedot}

\def\Fig{Fig\onedot} \def\Eq{Eq\onedot}
\def\Sec{Sec\onedot} 
\def\Tab{Tab\onedot} 
\makeatother

\DeclareRobustCommand{\RTE}{%
    \ifmmode
    \text{RErr}
    \else
    RErr\xspace
    \fi
}
\DeclareRobustCommand{\TE}{%
    \ifmmode
    \text{Err}
    \else
    Err\xspace
    \fi
}
\DeclareRobustCommand{\RCE}{%
    \ifmmode
    \text{RLoss}
    \else
    RLoss\xspace
    \fi
}
\DeclareRobustCommand{\CE}{%
    \ifmmode
    \text{Loss}
    \else
    Loss\xspace
    \fi
}
\DeclareRobustCommand{\wmax}{%
    \ifmmode
    w_{\text{max}}
    \else
    $w_{\text{max}}$\xspace
    \fi
}
\DeclareRobustCommand{\Vmin}{%
    \ifmmode
    V_{\text{min}}
    \else
    $V_{\text{min}}$\xspace
    \fi
}
\DeclareRobustCommand{\Vt}{%
    \ifmmode
    V_{\text{th}}
    \else
    $V_{\text{th}}$\xspace
    \fi
}
\DeclareRobustCommand{\biterror}{%
    \ifmmode
    \text{BErr}
    \else
    BErr\xspace
    \fi
}
\DeclareRobustCommand{\pfault}{%
    \ifmmode
    p_{\text{flt}}
    \else
    $p_{\text{{flt}}}$\xspace
    \fi
}
\DeclareRobustCommand{\perror}{%
    \ifmmode
    p_{\text{err}}
    \else
    $p_{\text{err}}$\xspace
    \fi
}

\def\min{\mathop{\rm min}\nolimits}
\def\max{\mathop{\rm max}\nolimits}

\newcommand{\CifarT}{CIFAR10\xspace}

\newcommand{\red}[1]{{\color{red}#1}}
\newcommand{\david}[1]{\noindent{\color{red}{David: #1}}}
\newcommand{\matthias}[1]{\noindent{\color{blue}{Matthias: #1}}}
\newcommand{\bernt}[1]{\noindent{\textcolor[rgb]{0.08, 0.38, 0.74}{\textbf{Bernt: #1}}}}

\renewcommand{\red}[1]{#1}
\renewcommand{\david}[1]{}
\renewcommand{\matthias}[1]{}
\renewcommand{\bernt}[1]{}

%% file: sec_abstract.tex
\begin{abstract}
	Adversarial training (AT) has become the de-facto standard to obtain models robust against adversarial examples.
	However, AT exhibits severe robust overfitting: cross-entropy loss on adversarial examples, so-called robust loss, decreases continuously on training examples, while eventually increasing on test examples. In practice, this leads to poor robust generalization, \ie, adversarial robustness does not generalize well to new examples.
	In this paper, we study the relationship between robust generalization and flatness of the robust loss landscape in weight space, \ie, whether robust loss changes significantly when perturbing weights.
	To this end, we propose average- and worst-case metrics to measure flatness in the robust loss landscape and show a \textbf{correlation between good robust generalization and flatness}.
	For example, throughout training, flatness reduces significantly during overfitting such that early stopping effectively finds flatter minima in the robust loss landscape.
	Similarly, AT variants achieving higher adversarial robustness also correspond to flatter minima. This holds for many popular choices, \eg, AT-AWP, TRADES, MART, AT with self-supervision or additional unlabeled examples, as well as simple regularization techniques, \eg, AutoAugment, weight decay or label noise.
	For fair comparison across these approaches, our flatness measures are specifically designed to be scale-invariant and we conduct extensive experiments to validate our findings.
\end{abstract}

%% file: sec_introduction.tex
\section{Introduction}
\label{sec:introduction}

\begin{figure}[t]
	\centering
	\vspace*{-0.2cm}
	\hspace*{-0.25cm}
	\includegraphics[width=0.5\textwidth]{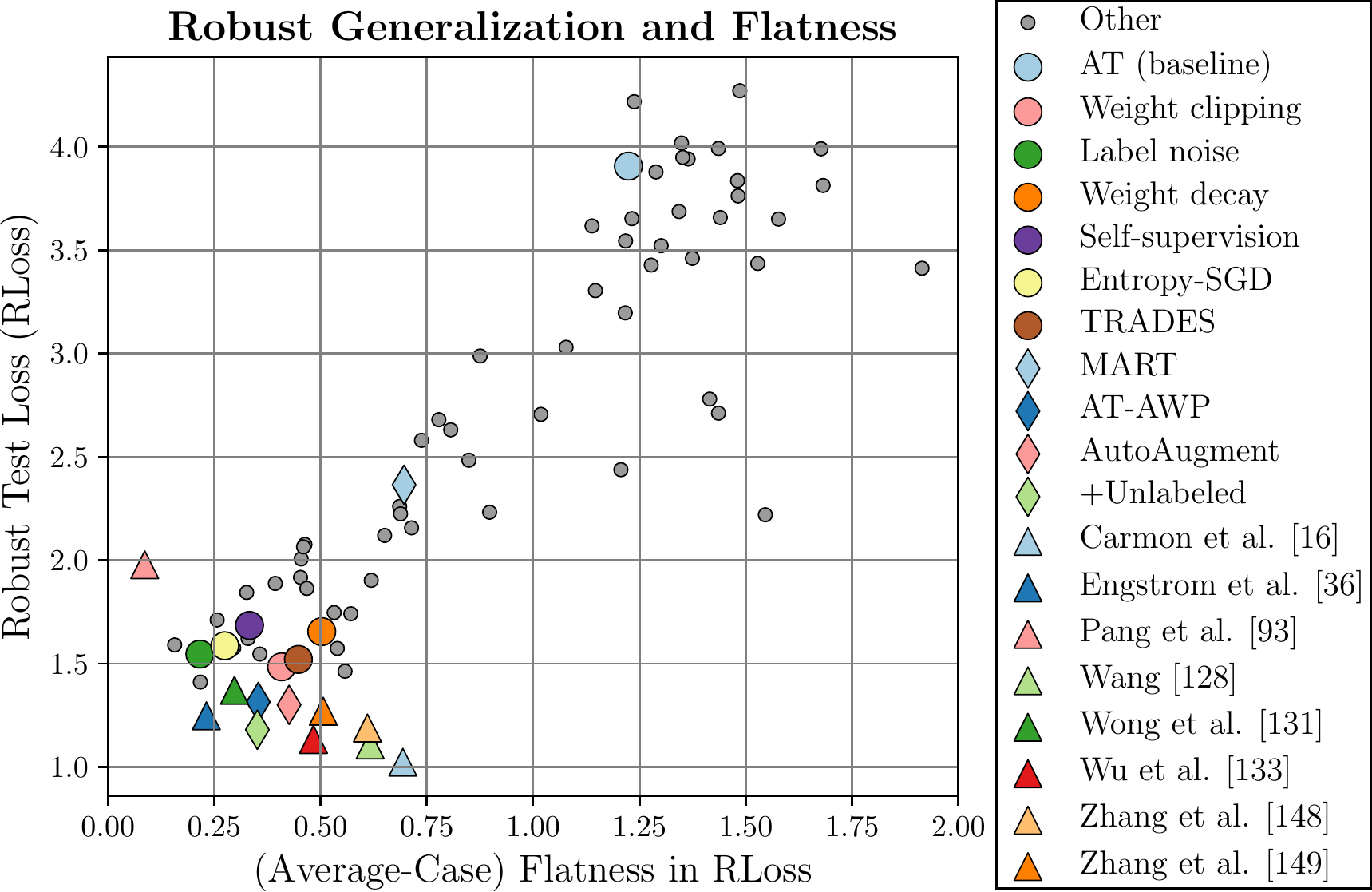}
	\vspace*{-18px}
	\caption{\textbf{Robust Generalization and Flatness:} Robust loss (\RCE, lower is more robust, y-axis), \ie, cross-entropy loss on PGD adversarial examples \cite{MadryICLR2018}, against our \emph{average-case flatness} measure of \RCE in weight space (lower is ``flatter'', x-axis).
	Popular AT variants improving adversarial robustness on \CifarT, \eg, TRADES \cite{ZhangICML2019}, AT-AWP \cite{WuNIPS2020}, MART \cite{WangICLR2020} or AT with self-supervision \cite{HendrycksNIPS2019}/unlabeled examples \cite{CarmonNIPS2019}, also correspond to flatter minima. Vice-versa, regularization explicitly improving flatness, \eg, Entropy-SGD \cite{ChaudhariICLR2017}, weight decay or weight clipping \cite{StutzMLSYS2021}, also improve robustness.
	Across all models, there is a \textbf{clear relationship between good robust generalization and flatness in \RCE.}
	{\LARGE$\bullet$},\raisebox{0.5mm}{$\mathbin{\blacklozenge}$} Our models, \underline{w/o} early stopping.
	{\large$\blacktriangle$} RobustBench \cite{CroceARXIV2020b} models \emph{w/} early stopping.}
	\label{fig:introduction}
	\vspace*{-6px} 
\end{figure}

\begin{figure}[t]
	\centering
	\vspace*{-0.2cm}
	\includegraphics[width=0.225\textwidth]{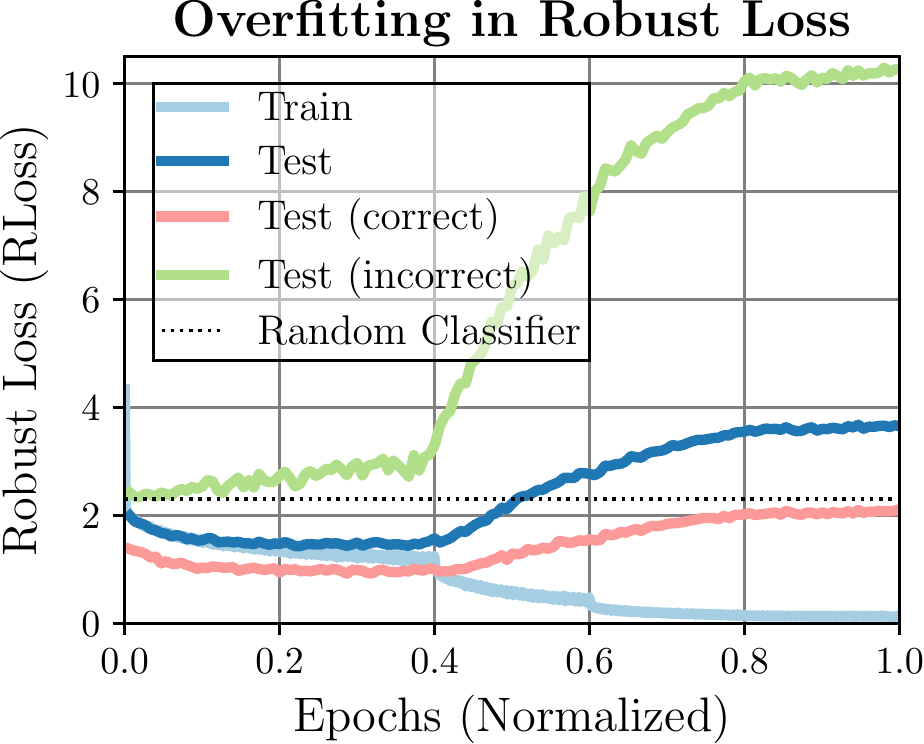}
	\includegraphics[width=0.225\textwidth]{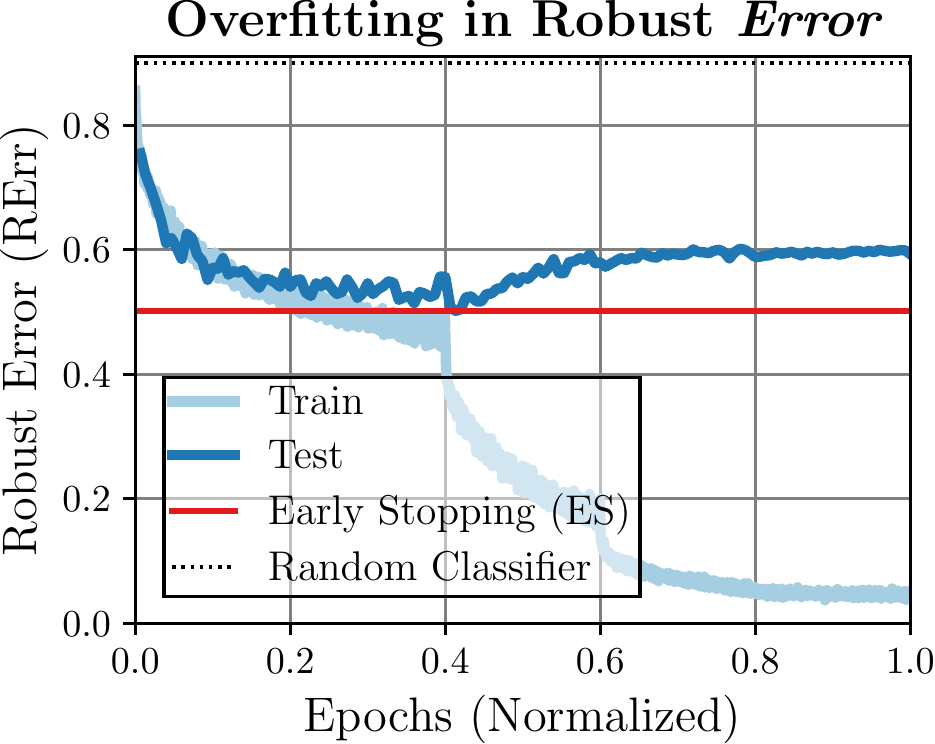}
	\vspace*{-8px}
	\caption{\textbf{Robust Overfitting:} Robust (cross-entropy) loss (\RCE) and robust error (\RTE) over epochs (normalized by $150$ epochs) for AT, \red{using a ResNet-18 on \CifarT (\cf \secref{sec:experiments})}, to illustrate \emph{robust} overfitting. \textbf{Left:} Training \RCE ({\color{plot0}light blue}) reduces continuously throughout training, while test \RCE ({\color{plot1}dark blue}) eventually increases again.
	We also highlight that robust overfitting is \emph{not} limited to incorrectly classified examples ({\color{plot4}green}), but also affects correctly classified ones ({\color{plot2}rose}). \textbf{Right:} Similar behavior, but less pronounced, can be observed considering \RTE. We also show \RTE obtained through early stopping ({\color{plot5}red}).}
	\label{fig:main-overfitting}
	\vspace*{-6px}
\end{figure}

In order to obtain robustness against adversarial examples \cite{SzegedyICLR2014}, \emph{adversarial training (AT)} \cite{MadryICLR2018} augments training with adversarial examples that are generated on-the-fly. While many different variants have been proposed, AT is known to require more training data \cite{KhouryARXIV2018,SchmidtNIPS2018}, generally leading to generalization problems \cite{FarniaICLR2019}. In fact, \emph{robust overfitting} \cite{RiceICML2020} has been identified as the main problem in AT:  adversarial robustness on test examples eventually starts to decrease, while robustness on training examples continues to increase (\cf \figref{fig:main-overfitting}). This is typically observed as increasing \emph{robust loss (\RCE)} or \emph{robust test error (\RTE)}, \ie, (cross-entropy) loss and test error on adversarial examples. As a result, the \emph{robust generalization gap}, \ie, the difference between test and training robustness, tends to be very large. In \cite{RiceICML2020}, early stopping is used as a simple and effective strategy to avoid robust overfitting. However, despite recent work tackling robust overfitting \cite{SinglaARXIV2021,WuNIPS2020,HwangARXIV2020}, it remains an open and poorly understood problem.

In \emph{``clean''} generalization (\ie, on natural examples), overfitting is well-studied and commonly tied to flatness of the loss landscape in weight space, both visually \cite{LiNIPS2018} and empirically \cite{NeyshaburNIPS2017,KeskarICLR2017,JiangICLR2020}.
In general, the optimal weights on test examples do not coincide with the minimum found on training examples. Flatness ensures that the loss does \emph{not} increase significantly in a neighborhood around the found minimum. Therefore, flatness leads to good generalization because the loss on test examples does not increase significantly (\ie, small generalization gap, \cf \figref{fig:main-illustration}, right).
\cite{LiNIPS2018} showed that \emph{visually} flatter minima correspond to better generalization. \cite{NeyshaburNIPS2017} and \cite{KeskarICLR2017} formalize this idea by measuring the change in loss within a local neighborhood around the minimum considering random \cite{NeyshaburNIPS2017} or ``adversarial'' weight perturbations \cite{KeskarICLR2017}.
These measures are shown to be effective in predicting generalization in a recent large-scale empirical study \cite{JiangICLR2020} and explicitly encouraging flatness during training has been shown to be successful in practice \cite{ZhengARXIV2020c,CicekICCVWOR2019,TinICLR2020,ChaudhariICLR2017,IzmailovUAI2018}.

Recently, \cite{WuNIPS2020} applied the idea of flat minima to AT: through \emph{adversarial weight perturbations}, AT is regularized to find flatter minima of the \emph{robust} loss landscape. This reduces the impact of robust overfitting and improves robust generalization, but does not \emph{avoid} robust overfitting. As result, early stopping is still necessary. Furthermore, flatness is only assessed \emph{visually} and it remains unclear whether flatness does actually improve in these adversarial weight directions.
Similarly, \cite{GowalARXIV2020} shows that weight averaging \cite{IzmailovUAI2018} can improve robust generalization, indicating that flatness might be beneficial in general. This raises the question whether other ``tricks'' \cite{PangARXIV2020b,GowalARXIV2020}, \eg, different activation functions \cite{SinglaARXIV2021} or label smoothing \cite{SzegedyCVPR2016}, or approaches such as AT with self-supervision \cite{HendrycksNIPS2019}/unlabeled examples \cite{CarmonNIPS2019} are successful \emph{because of} finding flatter minima.

\textbf{Contributions:} In this paper, we study \textbf{whether flatness of the robust loss (\RCE) in weight space improves robust generalization}. To this end,
we propose both average- and worst-case flatness measures for the \emph{robust} case, \red{thereby addressing challenges such as scale-invariance \cite{DinhICML2017}, estimation of \RCE on top or jointly with weight perturbations, and the discrepancy between \RCE and \RTE}. We show that \textbf{robust generalization generally improves alongside flatness} and vice-versa: \figref{fig:introduction} plots \RCE (lower is more robust, y-axis) against our average-case flatness in \RCE (lower is flatter, x-axis), showing a clear relationship. 
\red{In contrast to \cite{WuNIPS2020}, not providing empirical flatness measures, our results show that this relationship is stronger for average-case flatness.}
This trend covers a wide range of AT variants on \CifarT, \eg,  AT-AWP \cite{WuNIPS2020}, TRADES \cite{ZhangICML2019}, MART \cite{WangICLR2020}, AT with self-supervision \cite{HendrycksNIPS2019} or additional unlabeled examples \cite{CarmonNIPS2019,UesatoNIPS2019}, as well as various regularization schemes, including AutoAugment \cite{CubukARXIV2018}, label smoothing \cite{SzegedyCVPR2016} and noise or weight clipping \cite{StutzMLSYS2021}. Furthermore, we consider hyper-parameters, \eg, learning rate schedule, weight decay, batch size, or different activation functions \cite{ElfwingNN2018,MisraBMVC2020,HendrycksARXIV2016}, and methods explicitly improving flatness, \eg, Entropy-SGD \cite{ChaudhariICLR2017} or weight averaging \cite{IzmailovUAI2018}.

%% file: sec_related_work.tex
\section{Related Work}
\label{sec:related-work}

\begin{figure}[t]
	\centering
	\vspace*{-0.3cm}
	\hspace*{-0.4cm}
	\begin{minipage}[t]{0.3\textwidth}
		\includegraphics[width=0.9\textwidth]{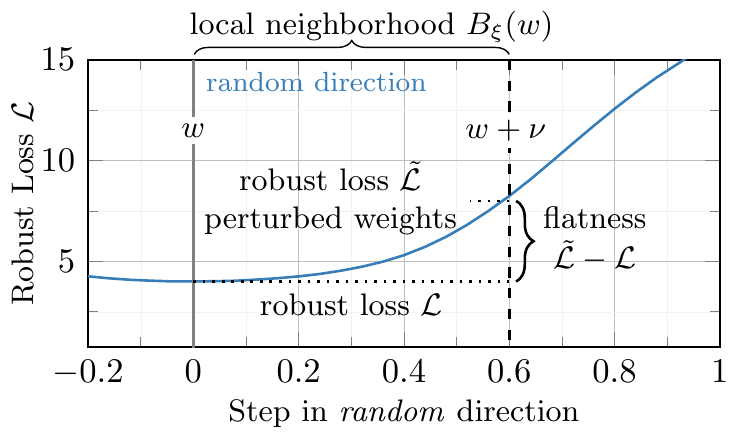}
	\end{minipage}
	\begin{minipage}[t]{0.12\textwidth}
		\includegraphics[width=1.2\textwidth]{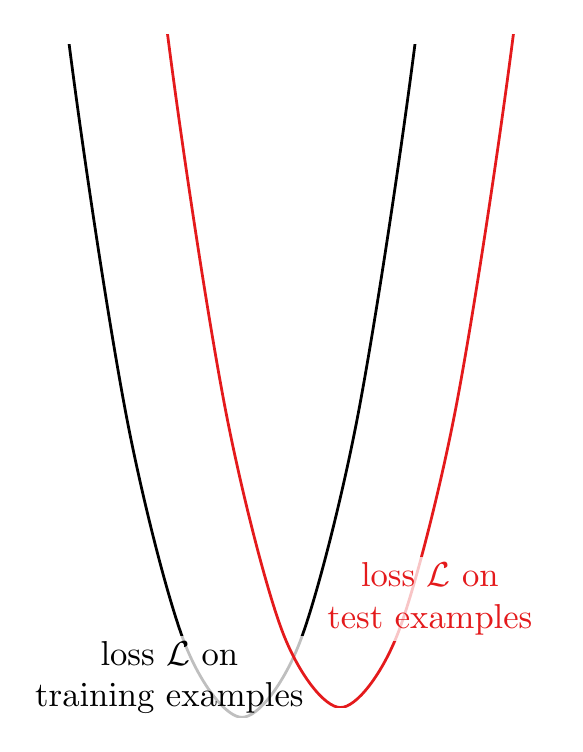} 
	\end{minipage}
	\vspace*{-10px}
	\caption{\textbf{Measuring Flatness.} \textbf{Left:} Illustration of measuring flatness in a random (\ie, average-case, {\color{colorbrewer2}blue}) direction by computing the difference between \RCE $\tilde{\mathcal{L}}$ \emph{after} perturbing weights (\ie, $w + \nu$) and the ``reference'' \RCE $\mathcal{L}$ given a local neighborhood $B_\xi(w)$ around the found weights $w$, see \secref{subsec:main-flatness}. In practice, we average across/take the worst of several random/adversarial directions.
	\textbf{Right:} Large changes in \RCE around the ``sharp'' minimum causes poor generalization from training ({\color{colorbrewer0}black}) to test examples ({\color{colorbrewer1}red}).
	}
	\label{fig:main-illustration}
	\vspace*{-6px}
\end{figure}

\textbf{Adversarial Training (AT):}
Despite a vast amount of work on adversarial robustness, \eg, see \cite{SilvaARXIV2020,YuanARXIV2017,AkhtarACCESS2018,BiggioCCS2018,XuARXIV2019}, adversarial training (AT) has become the de-facto standard for (empirical) robustness. Originally proposed in different variants in \cite{SzegedyICLR2014,MiyatoICLR2016,HuangARXIV2015}, it received considerable attention in \cite{MadryICLR2018,robustness} and has been extended in various ways:
\cite{LambAISEC2019,CarmonNIPS2019,UesatoNIPS2019} utilize interpolated or unlabeled examples, \cite{TramerNIPS2019,MainiICML2020} achieve robustness against multiple threat models, \cite{StutzICML2020,LaidlawARXIV2019,WuICML2018} augment AT with a reject option, \cite{YeNIPS2018,LiuICLR2019b} use Bayesian networks, \cite{TramerICLR2018,GrefenstetteARXIV2018} build ensembles, \cite{BalajiARXIV2019,DingICLR2020} adapt the threat model for each example, \cite{Wong2020ICLR,AndriushchenkoNIPS2020,VivekCVPR2020} perform AT with single-step attacks, \cite{HendrycksNIPS2019} uses self-supervision and \cite{PangNIPS2020} additionally regularizes features  -- to name a few directions. However, AT is slow \cite{ZhangNIPS2020} and suffers from increased sample complexity \cite{SchmidtNIPS2018} as well as reduced (clean) accuracy \cite{TsiprasICLR2019,StutzCVPR2019,ZhangICML2019,RaghunathanARXIV2019}. Furthermore, progress is slowing down. In fact, ``standard'' AT is shown to perform surprisingly well on recent benchmarks \cite{CroceICML2020,CroceARXIV2020b} when tuning hyper-parameters properly \cite{PangARXIV2020b,GowalARXIV2020}. In our experiments, we consider several popular variants \cite{WuNIPS2020,WangICLR2020,ZhangICML2019,CarmonNIPS2019,HendrycksNIPS2019}.

\textbf{Robust Overfitting:} Recently, \cite{RiceICML2020} identified \emph{robust} overfitting as a crucial problem in AT and proposed early stopping as an effective mitigation strategy. This motivated work \cite{SinglaARXIV2021,WuNIPS2020} trying to mitigate robust overfitting. While \cite{SinglaARXIV2021} studies the use of different activation functions, \cite{WuNIPS2020} proposes AT with \emph{adversarial weight perturbations} (AT-AWP) explicitly aimed at finding flatter minima in order to reduce overfitting. While the results are promising, early stopping is still necessary. Furthermore, flatness is merely assessed visually, leaving open whether AT-AWP \emph{actually} improves flatness in adversarial weight directions. We consider both average- and worst-case flatness, \ie, random and adversarial weight perturbations, to answer this question.

\textbf{Flat Minima} in the loss landscape, \wrt changes in the weights, are generally assumed to improve \emph{standard} generalization \cite{HochreiterNC1997}. \cite{LiNIPS2018} shows that residual connections in ResNets \cite{HeCVPR2016} or weight decay lead to \emph{visually} flatter minima. \cite{NeyshaburNIPS2017,KeskarICLR2017} formalize this concept of flatness in terms of \emph{average-case} and \emph{worst-case} flatness. \cite{KeskarICLR2017,JiangICLR2020} show that worst-case flatness correlates well with better generalization, \eg, for small batch sizes, while \cite{NeyshaburNIPS2017} argues that generalization can be explained using both an average-case flatness measure and an appropriate capacity measure. Similarly, batch normalization is argued to improve generalization by allowing to find flatter minima \cite{SanturkarNIPS2018,BjorckNIPS1018}. These insights have been used to explicitly regularize flatness \cite{ZhengARXIV2020c}, improve semi-supervised learning \cite{CicekICCVWOR2019} and develop novel optimization algorithms such as Entropy-SGD \cite{ChaudhariICLR2017}, local SGD \cite{TinICLR2020} or weight averaging \cite{IzmailovUAI2018}.
\cite{DinhICML2017}, in contrast, criticizes some of these flatness measures as not being scale-invariant.
We transfer the intuition of flatness to the \emph{robust} loss landscape, showing that flatness is desirable for adversarial robustness, while using scale-invariant measures.

%% file: sec_main.tex
\section{Robust Generalization and Flat Minima}
\label{sec:main}

\begin{figure}[t]
	\centering
	\vspace*{-0.2cm}
	\hspace*{-0.4cm}
	\begin{minipage}[t]{0.16\textwidth}
		\vspace*{0px}
		\centering
		\scriptsize 
		\begin{tabularx}{1\textwidth}{|@{\hspace*{1px}}X@{\hspace*{1px}}|@{\hspace*{1px}}c@{\hspace*{1px}}|}
			\hline
			\textbf{Model} & \RTE $\downarrow$\\
			\hline
			\hline
			AT (baseline) {\color{plot0}$\bullet$} & 62.8\\
			Scaled $\times0.5$ {\color{plot1}$\bullet$} & 62.8\\
			Scaled $\times2$ {\color{plot2}$\bullet$} & 62.8\\
			MiSH {\color{plot5}$\bullet$} & 59.8\\
			Batch size $8$ {\color{plot3}$\bullet$} & 58.2\\
			Adam {\color{plot4}$\bullet$} & 57.5\\
			\hline
			\hline
			Label smoothing {\color{plot1}$\bullet$} & 61.2\\
			MART {\color{plot5}$\bullet$} & 61\\
			Entropy-SGD {\color{plot3}$\bullet$} & 58.6\\
			Self-supervision {\color{plot2}$\bullet$} & 57.1\\
			TRADES {\color{plot4}$\bullet$} & 56.7\\
			AT-AWP {\color{plot6}$\bullet$} & 54.3\\
			\hline
		\end{tabularx}
	\end{minipage}
	\begin{minipage}[t]{0.29\textwidth}
		\vspace*{7px}
		
		\begin{minipage}[t]{0.625\textwidth}
			\vspace*{0px}
			
			\includegraphics[height=1.85cm]{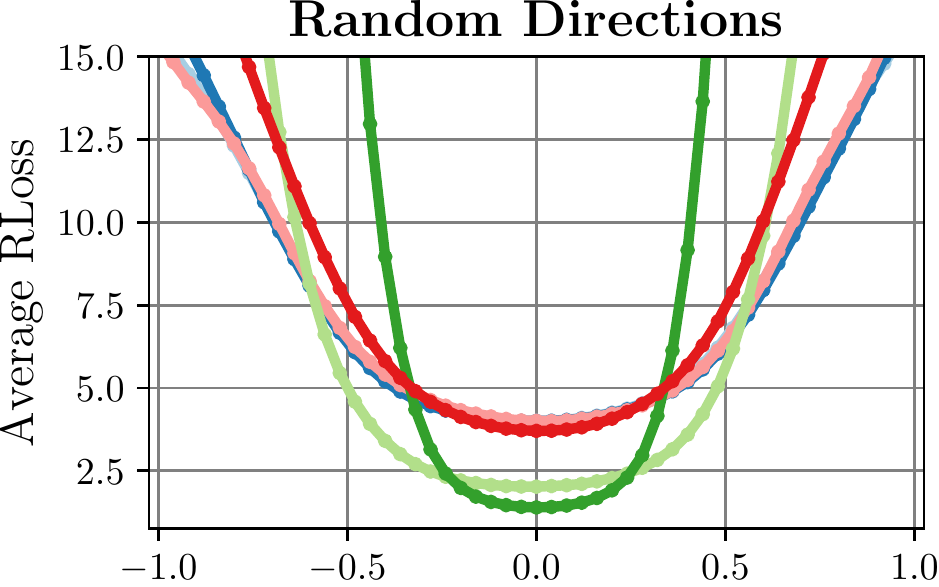}
		\end{minipage}
		\begin{minipage}[t]{0.3\textwidth}
			\vspace*{0px}
			
			\includegraphics[height=1.85cm]{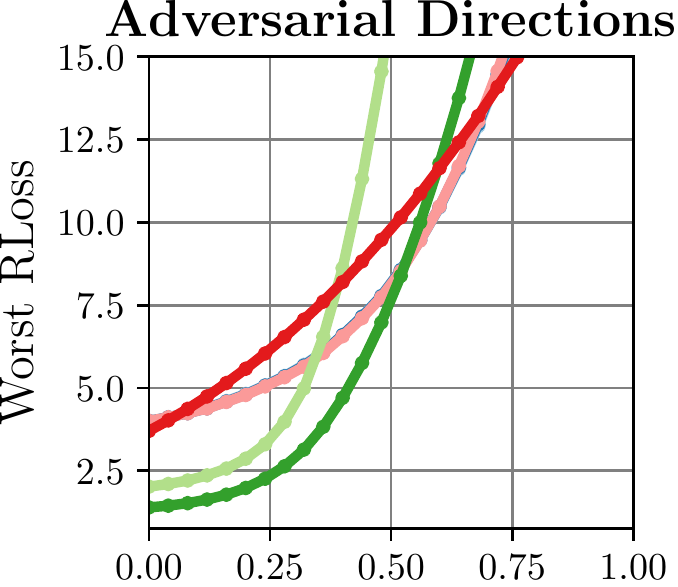}
		\end{minipage}
		
		{\rule{5.5cm}{0.25px}}
		\\[-9px]
		\begin{minipage}[t]{0.625\textwidth}
			\vspace*{0px}
			
			\includegraphics[height=1.9cm]{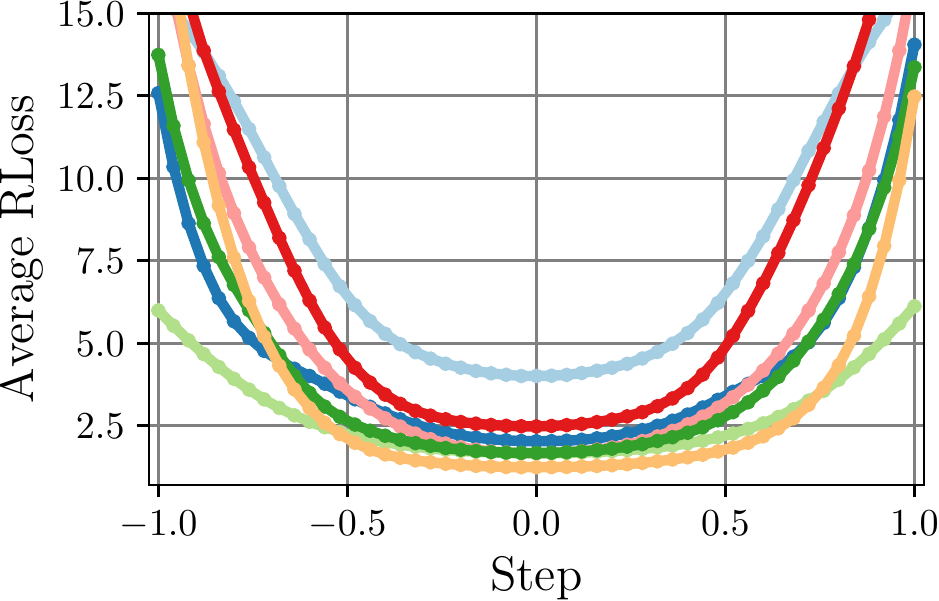}
		\end{minipage}
		\begin{minipage}[t]{0.3\textwidth}
			\vspace*{0px}
			
			\includegraphics[height=1.9cm]{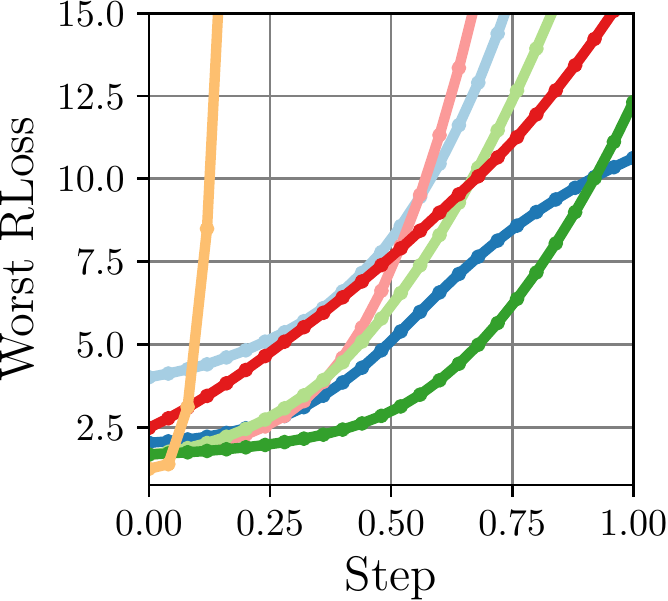}
		\end{minipage}
	\end{minipage}
	
	\vspace*{-10px}
	\caption{\textbf{Visualizing Flatness:} \RCE landscape across $10$ random or adversarial directions. \textbf{Top:} Our AT baseline (ResNet-18) and scaled variants ($\times2$ and $\times0.5$). Training with smaller batch size or Adam \cite{KingmaICLR2015} improves adversarial robustness (lower \RTE vs\onedot AutoAttack \cite{CroceICML2020}) but does \emph{not} result in \emph{visually} flatter minima. \textbf{Bottom:} AT-AWP \cite{WuNIPS2020} or Entropy-SGD \cite{ChaudhariICLR2017} improve robustness \emph{and} visual flatness in random directions. In adversarial directions, however, AT-AWP looks very sharp. Overall, visual inspection does \emph{not} provide a clear, objective picture of flatness.}
	\label{fig:main}
	\vspace*{-6px}
\end{figure}
\begin{figure*}[t]
	\centering
	\vspace*{-0.2cm}
	\begin{minipage}[t]{0.2\textwidth}
		\vspace*{0px}
		
		\includegraphics[width=\textwidth]{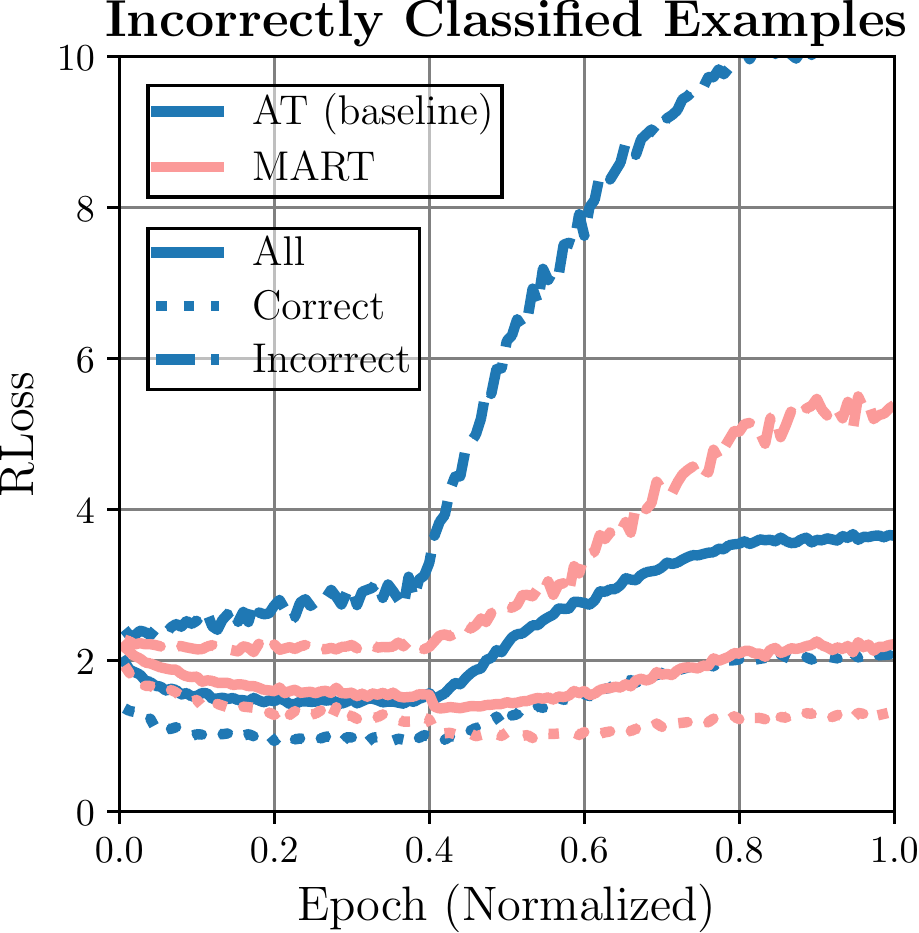}
	\end{minipage}
	\hspace*{1px}
	\begin{minipage}[t]{0.175\textwidth}
		\vspace*{0px}
		
		\includegraphics[width=\textwidth]{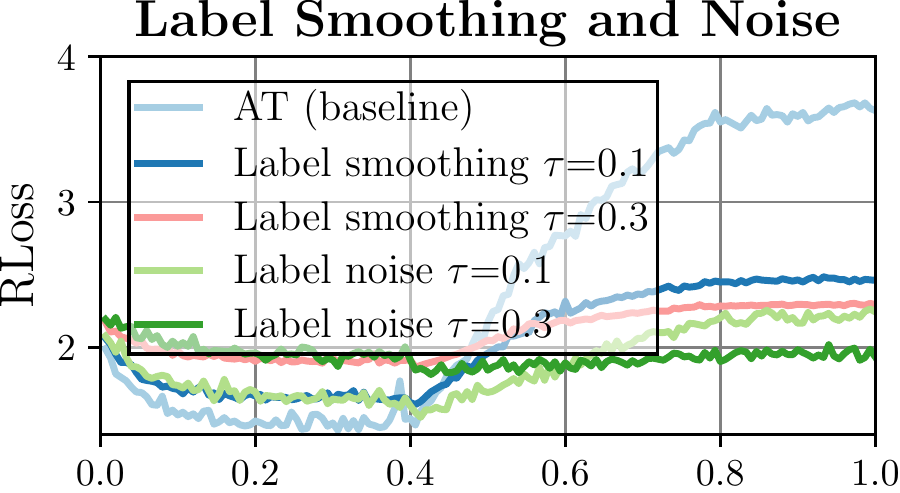}
		\vspace*{-9px}
		
		\includegraphics[width=\textwidth]{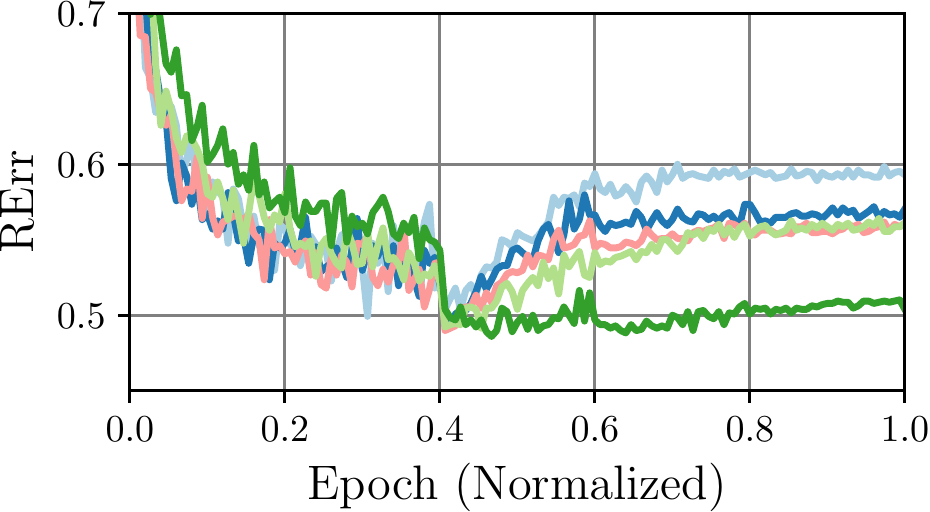}
	\end{minipage}
	\hspace*{1px} 
	\begin{minipage}[t]{0.2\textwidth}
		\vspace*{0px}
		 
		\includegraphics[width=\textwidth]{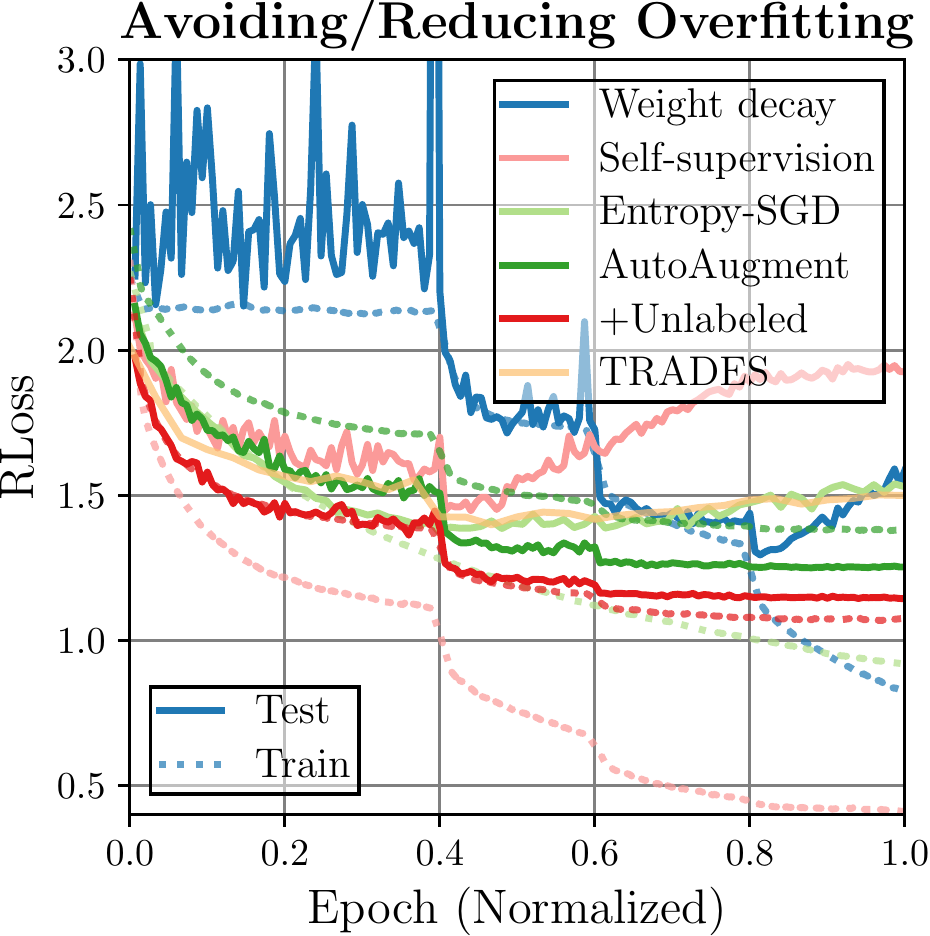}
	\end{minipage}
	\hspace*{1px}
	\begin{minipage}[t]{0.175\textwidth}
		\vspace*{0px}
		
		\includegraphics[width=1.025\textwidth]{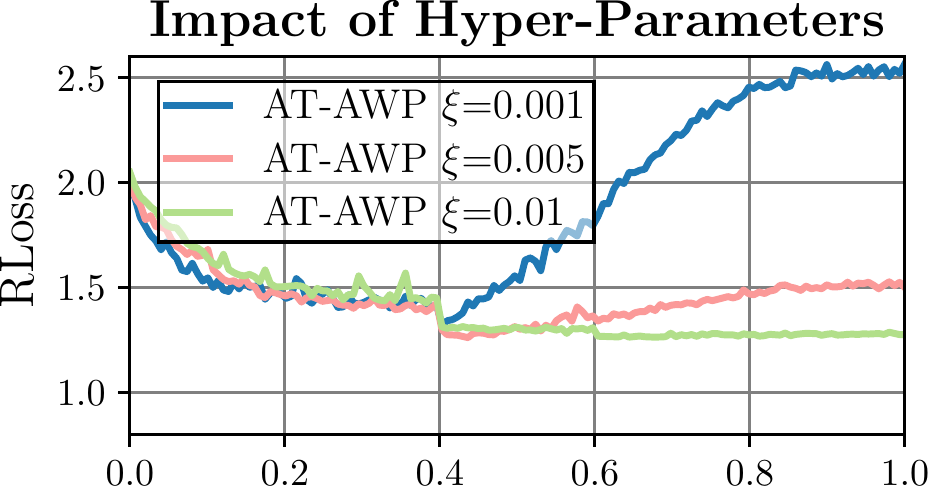}
		\vspace*{-9px}
				
		\includegraphics[width=\textwidth]{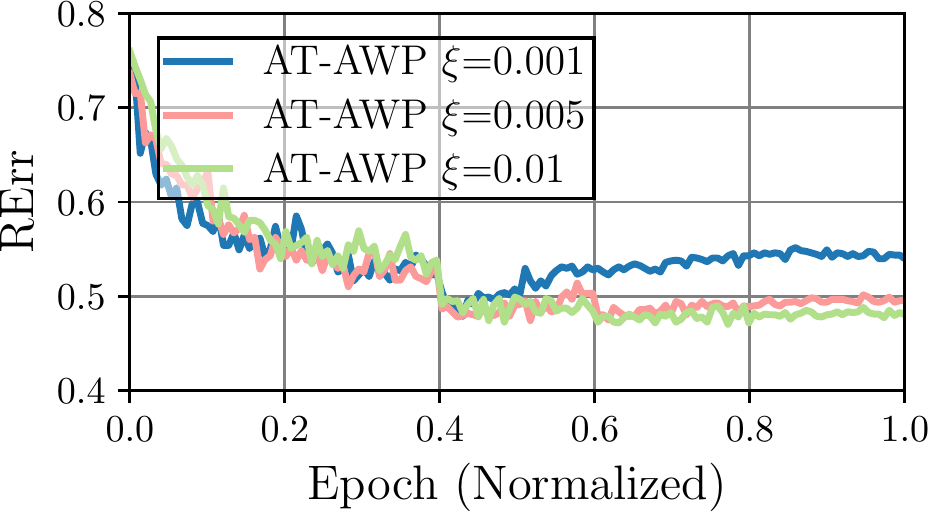}
	\end{minipage}
	\hspace*{1px}
	\begin{minipage}[t]{0.2\textwidth}
		\vspace*{0px}
	
		\includegraphics[width=\textwidth]{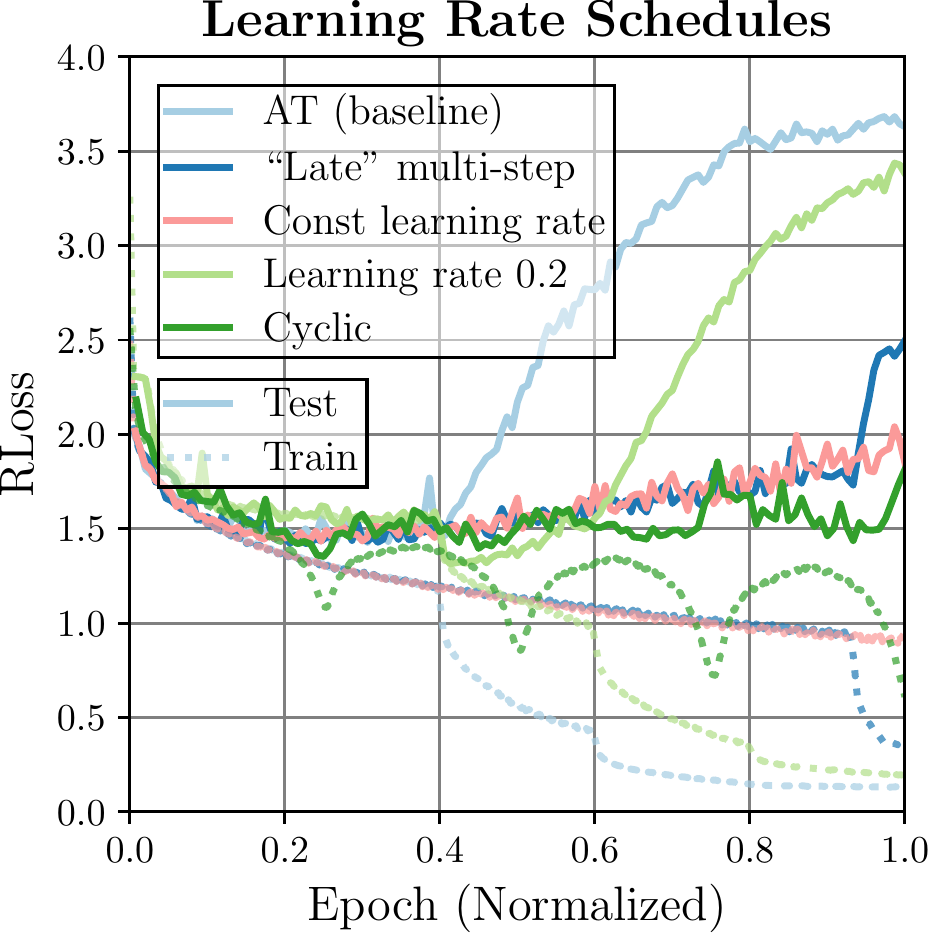}
	\end{minipage}
	\vspace*{-6px}
	\caption{\textbf{Understanding Robust Overfitting:} Training curves plotted over (normalized) epochs, \red{see \secref{subsec:main-discussion} for detailed discussion}. \textbf{First column:} \RCE, split for correct/incorrect test examples, for AT and MART, which successfully dampens the effect of overfitting using a weighted loss on incorrectly classified examples. \textbf{Second column:} Both label smoothing and label noise reduce robust overfitting \wrt \RCE. However, the reduction in \RCE does not translate to a similar reduction of \RTE. \textbf{Third to fifth column:} \RCE (test solid and train dotted) for various approaches improving adversarial robustness and different learning rate schedules. While some approaches avoid robust overfitting altogether (\eg, AT-AWP), others (\eg, weight decay) merely reduce its impact (third column). But the success depends strongly on hyper-parameters (fourth column). Robust overfitting occurs using all tested learning rate schedules (fifth column), confirming \cite{RiceICML2020}.}
	\label{fig:experiments-understanding}
	\vspace*{-6px}
\end{figure*}

We study robust generalization and overfitting in the context of flatness of the \emph{robust} loss landscape in weight space, \ie, \wrt changes in the weights. While flat minima have consistently been linked to standard generalization \cite{HochreiterNC1997,LiNIPS2018,NeyshaburNIPS2017,KeskarICLR2017}, this relationship remains unclear for adversarial robustness.
We start by briefly introducing the robust overfitting phenomenon (\secref{subsec:main-overfitting}). Then, we discuss problems in judging flatness visually \cite{LiNIPS2018} (\secref{subsec:main-visualization}). Thus, we are inspired by \cite{KeskarICLR2017,NeyshaburNIPS2017} and introduce average- and worst-case flatness measures based on the change in robust loss along random or adversarial weight directions in a local neighborhood (\secref{subsec:main-flatness}), \cf \figref{fig:main-illustration}.
We also discuss the connection of flatness to the Hessian eigenspectrum \cite{YaoNIPS2018} and the importance of scale-invariance as in \cite{DinhICML2017}.

\subsection{Background}
\label{subsec:main-overfitting}

\textbf{Adversarial Training (AT):}
Let $f$ be a (deep) neural network taking input $x \in [0,1]^D$ and weights $w \in \mathbb{R}^W$ and predicting a label $f(x;w)$. Given a true label $y$, an adversarial example is a perturbation $\tilde{x} = x + \delta$ such that $f(\tilde{x};w) \neq y$. The perturbation $\delta$ is intended to be nearly invisible which is, in practice, enforced using a $L_p$ constraint: $\|\delta\|_p \leq \epsilon$. To obtain robustness against these perturbations, AT injects adversarial examples during training:
\begin{align}
	\min_w \mathbb{E}_{x,y}\left[\max_{\|\delta\|_p \leq \epsilon} \mathcal{L}(f(x + \delta;w), y)\right]
\end{align}
where $\mathcal{L}$ denotes the cross-entropy loss. The outer minimization problem can be solved using regular stochastic gradient descent (SGD) on mini-batches. To compute adversarial examples, the inner maximization problem is tackled using projected gradient descent (PGD) \cite{MadryICLR2018}. Here, we focus on $p = \infty$ as this constrains the maximum change per feature/pixel, \eg, $\epsilon = \nicefrac{8}{255}$ on \CifarT. For evaluation (at test time), we consider both robust loss (\RCE) $\max_{\|\delta\|_\infty \leq \epsilon} \mathcal{L}(f(x + \delta;w), y)$, approximated using PGD, and robust test error (\RTE), which we approximate using AutoAttack \cite{CroceICML2020}. Note that AutoAttack \red{stops when adversarial examples are found} and does \emph{not} maximize cross-entropy loss, rendering it unfit to estimate \RCE.

\textbf{Robust Overfitting:}
Following \cite{RiceICML2020}, \figref{fig:main-overfitting} illustrates the problem of \emph{robust} overfitting, plotting \RCE (left) and \RTE (right) over epochs, which we normalize by the total number of epochs for clarity. Shortly after the first learning rate drop (at epoch $60$, \ie, $40\%$ of training), test \RCE and \RTE start to increase significantly, while robustness on training examples continues to improve.
Robust overfitting was shown to be independent of the learning rate schedule \cite{RiceICML2020} and, as we show (\secref{subsec:experiments-overfitting}), occurs across various different activation functions as well as many popular AT variants.
In contrast to \cite{RiceICML2020}, mostly focusing on \RTE, \figref{fig:main-overfitting} shows that \RCE overfits more severely, indicating a ``disconnectedness'' between \RCE and \RTE that we consider in detail later. For now, \RCE and \RTE do clearly not move ``in parallel'' and \RCE, reaching values around~$4$, is higher than for a random classifier (which is possible considering \emph{adversarial} examples). This is primarily due to an extremely high \RCE on incorrectly classified test examples (which are ``trivial'' adversarial examples). We emphasize, however, that robust overfitting also occurs on correctly classified test examples.

\subsection{Intuition and Visualizing Flatness}
\label{subsec:main-visualization}

\begin{figure*}[t]
	\centering
	\vspace*{-0.2cm}
	\hspace*{-0.1cm}
	\begin{minipage}[t]{0.29\textwidth}
		\vspace*{0px}
		
		\includegraphics[width=1\textwidth]{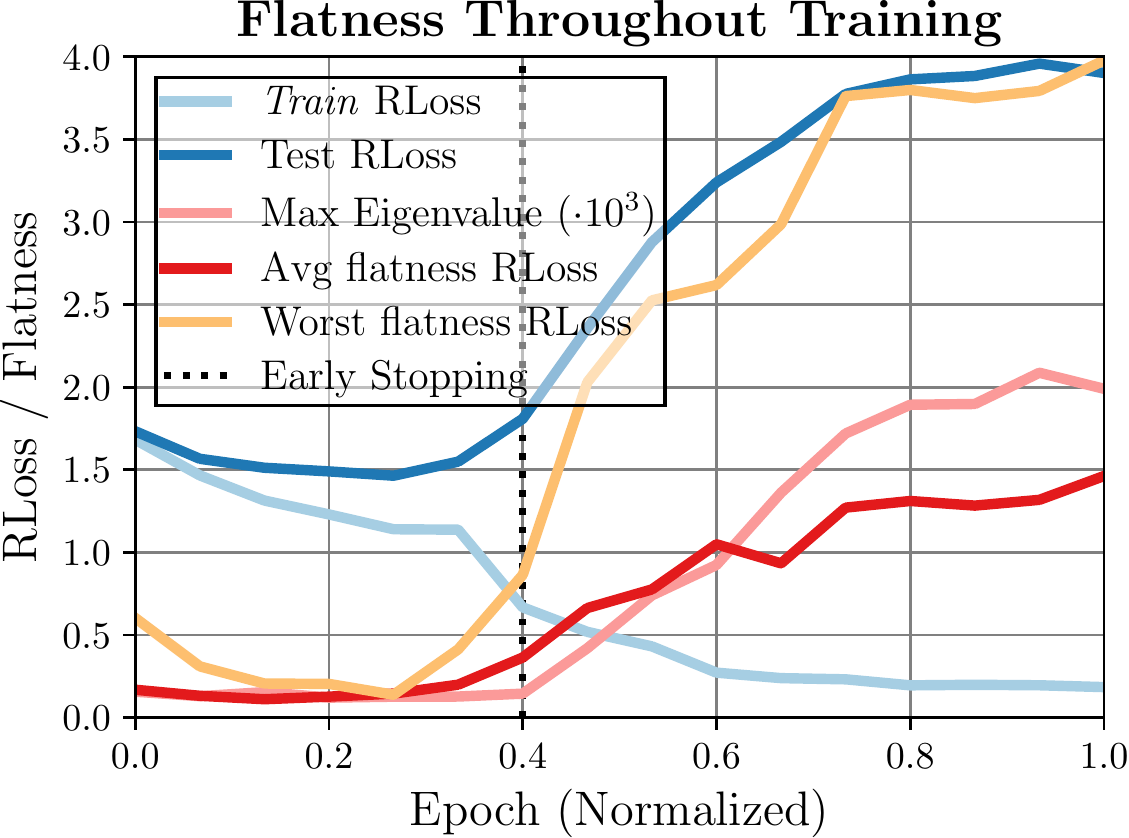}
	\end{minipage}
	\begin{minipage}[t]{0.015\textwidth}
		\vspace*{0px}
		
		\hspace*{4px}{\color{black!75}\rule{0.65px}{3.8cm}}
	\end{minipage}
	\begin{minipage}[t]{0.12\textwidth}
		\vspace*{0px}
		
		\includegraphics[height=3.7cm]{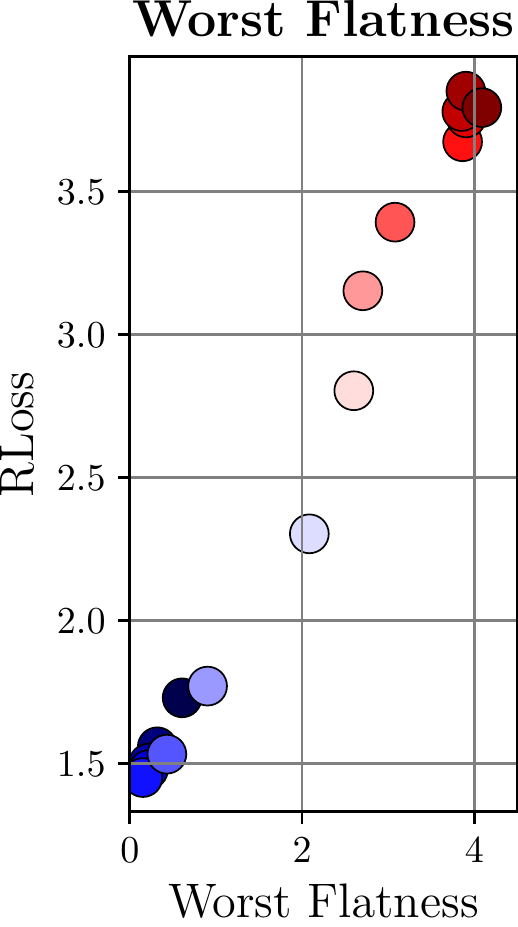}
	\end{minipage}
	\begin{minipage}[t]{0.12\textwidth}
		\vspace*{0px}
		
		\includegraphics[height=3.7cm]{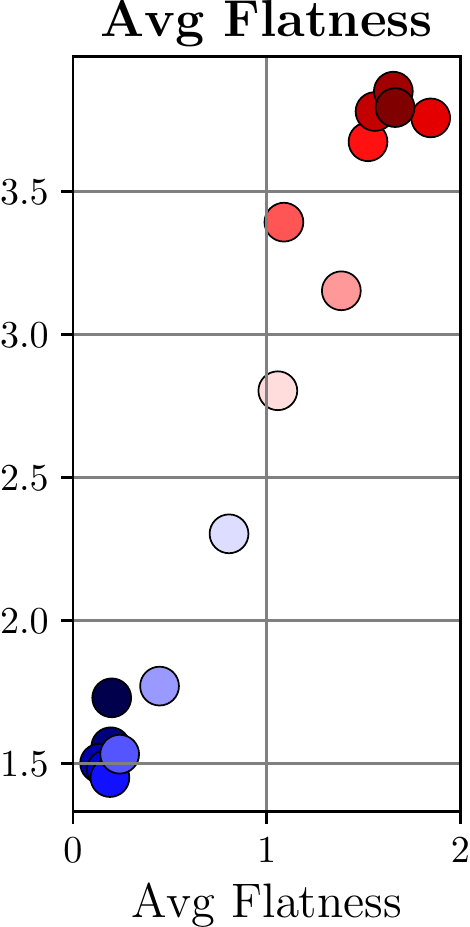}
	\end{minipage}
	\begin{minipage}[t]{0.01\textwidth}
		\vspace*{0px}
		
		\hspace*{4px}{\color{black!75}\rule{0.65px}{3.8cm}}
	\end{minipage}
	\begin{minipage}[t]{0.12\textwidth}
		\vspace*{0px}
		
		\includegraphics[height=3.7cm]{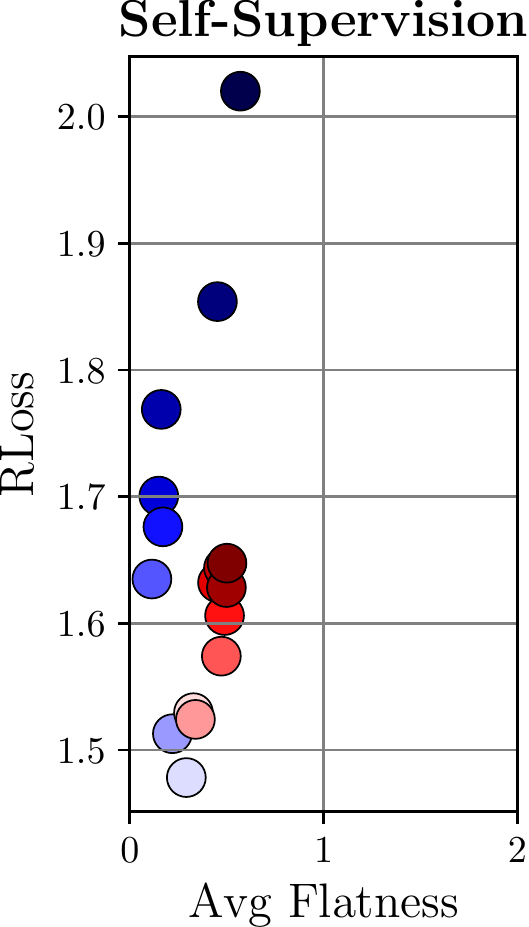}
	\end{minipage}
	\begin{minipage}[t]{0.112\textwidth}
		\vspace*{0px}
		
		\includegraphics[height=3.7cm]{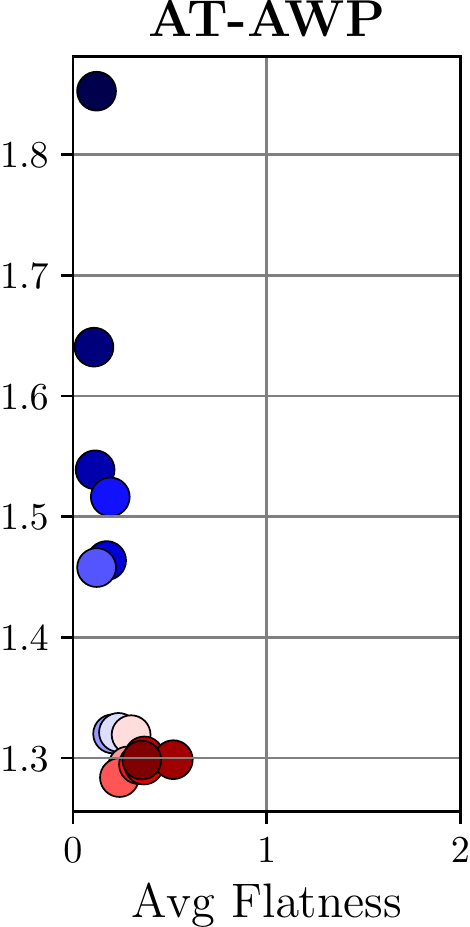}
	\end{minipage}
	\begin{minipage}[t]{0.17\textwidth}
		\vspace*{-2px}
		
		\includegraphics[height=3.7cm]{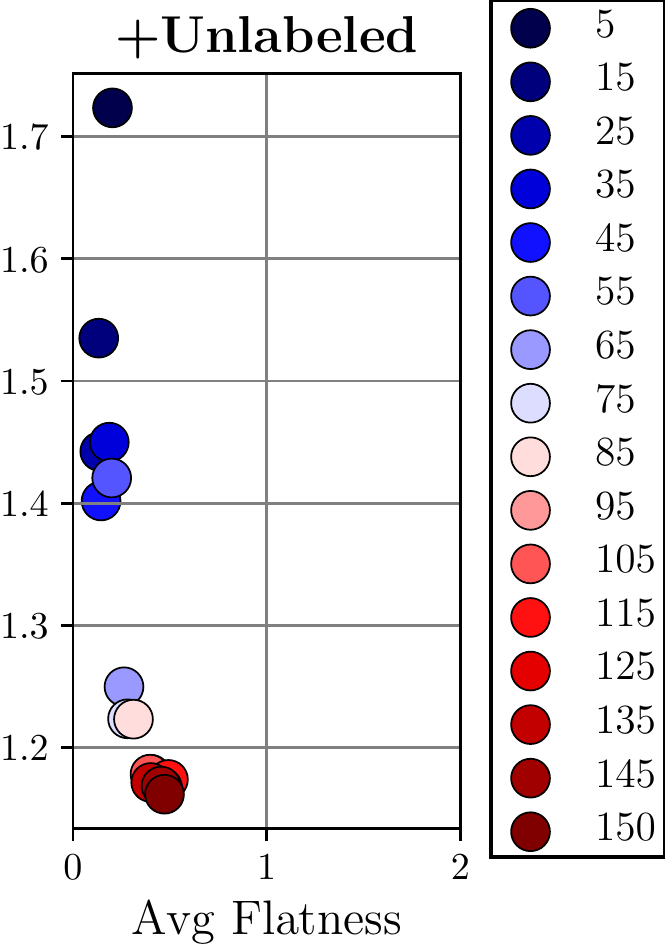}
	\end{minipage}
	\vspace*{-8px}
	\caption{\textbf{Flatness Throughout Training.} \textbf{Left:} Flatness in \RCE throughout training, showing that flatness reduces when the model overfits (\ie, {\color{plot1}test \RCE} increases, while {\color{plot0}train \RCE} decreases). 
	\textbf{Middle:} Test \RCE (y-axis) plotted against flatness in \RCE (x-axis) during training \red{(early epochs in {\color{blue!60!black}dark blue}, late epochs in {\color{red!60!black}dark red})}, showing a clear correlation, for both average- and worst-case flatness.
	\textbf{Right:} AT with self-supervision reduces the impact of robust overfitting (\RCE increases less) and simultaneously favors flatter minima. This behavior is pronounced for AT-AWP, explicitly optimizing flatness, and AT with additional unlabeled examples, generally resulting in the highest adversarial robustness, \cf \tabref{tab:experiments-flatness}.}
	\label{fig:experiments-flatness-epochs}
\end{figure*}

For judging robust flatness, we consider how \RCE changes \wrt random or adversarial perturbations in the weights $w$. Generally, we expect flatter minima to generalize better as the loss does not change significantly within a small neighborhood around the minimum, \ie, the found weights. Then, even if the loss landscape on test examples does not coincide with the loss landscape on training examples, loss remains small, ensuring good generalization. The contrary case, \ie, that sharp minima generalize poorly is illustrated in \figref{fig:main-illustration} (right). Before considering to \emph{measure} flatness, we discuss the easiest way to ``judge'' flatness: visual inspection of the \RCE landscape along random or adversarial directions in weight space.

In \cite{LiNIPS2018}, loss landscape is visualized along \emph{normalized} random directions. Normalization is important to handle different scales, \ie, weight distributions, and allow comparison across models.
We follow \cite{WuNIPS2020} and perform \emph{per-layer} normalization: Letting $\nu \in \mathbb{R}^W$ be a direction in weight space, it is normalized as
\begin{align}
	\hat{\nu}^{(l)} = \frac{\nu^{(l)}}{\|\nu^{(l)}\|_2} \|w^{(l)}\|_2 \quad\text{ for layer }l.\label{eq:main-normalization}
\end{align}
\vskip -2px
In contrast to \cite{LiNIPS2018}, we also consider biases and treat them as individual layer, but we exclude batch normalization parameters.
Then, the loss landscape is visualized in discrete steps along this direction, \ie, $w + s\hat{\nu}$ for $s \in [-1, 1]$.
Adversarial examples are computed ``on-the-fly'', \ie, for each $w + s\hat{\nu}$ individually, to avoid underestimating \RCE as in \cite{YuARXIV2018,PrabhuARXIV2019}.
The result is indeed scale-invariant: \figref{fig:main} (top) shows that the loss landscapes for scaled versions (factors $0.5$ or $2$, see supplementary material) of our AT baseline coincide with the original landscape.
However, \figref{fig:main} also illustrates that judging flatness visually is difficult: Considering random weight directions, AT with Adam \cite{KingmaICLR2015} or small batch size improves adversarial robustness, but the found minima look less flat (top). For other approaches, \eg, TRADES \cite{ZhangICML2019} or AT-AWP \cite{WuNIPS2020}, results look indeed flatter while also improving robustness (bottom). In adversarial directions, in contrast, AT-AWP looks particularly sharp. Furthermore, not only flatness but also the vertical ``height'' of the loss landscape matters and it is impossible to tell ``how much'' flatness is necessary.

\subsection{Average- and Worst-Case Flatness Measures}
\label{subsec:main-flatness}

In order to objectively measure and compare flatness, we draw inspiration from \cite{NeyshaburNIPS2017,KeskarICLR2017} and propose average- and worst-case flatness measures adapted to the robust loss. \red{We emphasize that measuring flatness in \RCE is non-trivial and flatness in (clean) \CE \emph{cannot} be expected to correlate with robustness (see supplementary material). For example, we need to ensure scale-invariance \cite{DinhICML2017} and estimate \RCE \emph{on top} of random or adversarial weight perturbations:}

\textbf{Average-Case / Random Flatness:}
Considering random weight perturbations $\nu \in B_\xi(w)$ within the $\xi$-neighborhood of $w$, average-case flatness is computed as
\vspace*{-12px}
\begin{align}
	\hspace*{-8px}
	\begin{split}
		\mathbb{E}_{\nu}[\max\limits_{\|\delta\|_\infty \leq \epsilon} \mathcal{L}(f(x{+}\delta; w{+}\nu), y)]\text{\hphantom{aaaaaaaaaaa}}\\[-2px]
		\text{\hphantom{aaaaaaaaaaa}}- \max\limits_{\|\delta\|_\infty \leq \epsilon} \mathcal{L}(f(x{+}\delta;w), y)
	\end{split}\label{eq:main-average}
\end{align}
\vskip -4px
\noindent averaged over test examples $x,y$, as illustrated in \figref{fig:main-illustration}. We define $B_\xi(w)$ using \emph{relative} $L_2$-balls per layer (\cf \eqnref{eq:main-normalization}): 
\begin{align}
	B_\xi(w) = \{w + \nu : \|\nu^{(l)}\|_2 \leq \xi \|w^{(l)}\|_2 \forall\text{ layers }l\}.\label{eq:main-ball}
\end{align}
\vskip -2px
\red{This ensures scale-invariance \wrt the weights as $B_\xi(w)$ scales with the weights on a \emph{per-layer} basis.}
Note that the second term in \eqnref{eq:main-average}, \ie, the ``reference'' robust loss, is important to make the measure independent of the absolute loss (\ie, corresponding to the vertical shift in \figref{fig:main}, left).
In practice, $\xi$ can be as large as $0.5$. We refer to \eqnref{eq:main-average} as \textbf{average-case flatness in \RCE}.

\textbf{Worst-Case / Adversarial Flatness:} \cite{WuNIPS2020} explicitly optimizes flatness in \emph{adversarial weight} directions and shows that average-case flatness is not sufficient to improve adversarial robustness. As it is unclear whether \cite{WuNIPS2020} actually improves worst-case flatness, we define
\begin{align}
	\begin{split}
		\max\limits_{\nu \in B_\xi(w)}&\left[\max\limits_{\|\delta\|_\infty \leq \epsilon} \mathcal{L}(f(x{+}\delta; w{+}\nu), y)\right]\hphantom{aaaaaa}\\[1px]
		&\hphantom{aaaaaaaa}- \max\limits_{\|\delta\|_\infty \leq \epsilon} \mathcal{L}(f(x{+}\delta;w), y)
	\end{split}\label{eq:main-worst}
\end{align}
\vskip -2px
as \textbf{worst-case flatness in \RCE}. Here, we use the same definition of $B_\xi(w)$ as above (aligned with \cite{WuNIPS2020}), but for smaller values of $\xi$.
Regarding \emph{standard} performance, this worst-case notion of flatness has been shown to be a reliable predictor of generalization \cite{JiangICLR2020,KeskarICLR2017}. For computing \eqnref{eq:main-worst} in practice, we jointly optimize over $\nu$ and $\delta$ (for each batch individually) using PGD.
As illustrated in \figref{fig:main}, \RCE increases quickly along adversarial directions, even for very small values of $\xi$, \eg, $\xi = 0.005$.

\subsection{Discussion}
\label{subsec:main-discussion}

In the context of flatness, there has also been some discussion concerning the meaning of Hessian eigenvalues \cite{LiNIPS2018,YaoNIPS2018} as well as concerns regarding the scale-invariance of flatness measures \cite{DinhICML2017}. First, regarding the Hessian eigenspectrum,
\cite{YaoNIPS2018} shows that large Hessian eigenvalues indicate poor adversarial robustness. However, Hessian eigenvalues are generally \emph{not} scale-invariant (which is acknowledged in \cite{YaoNIPS2018}): Our AT baseline has a maximum eigenvalue of $1990$ which reduces to $505$ when \emph{up-scaling} the model and increases to $7936$ when \emph{down-scaling}, without affecting robustness (\cf $\times0.5$ and $\times2$ in \figref{fig:main}). We also found that the largest eigenvalue is \emph{not} correlated with adversarial robustness. Second, following a similar train of thought, \cite{DinhICML2017} criticizes the flatness measures of \cite{NeyshaburNIPS2017,KeskarICLR2017} as not being scale-invariant. That is, through clever scaling of weights, without changing predictions, arbitrary flatness values can be ``produced''. However, the analysis in \cite{DinhICML2017} does not take into account the relative neighborhood as defined in \cite{KeskarICLR2017}, which renders the measure explicitly scale-invariant. This also applies to our definition of $B_\xi(w)$ in \eqnref{eq:main-ball} and is shown in \figref{fig:main} where normalization is performed relative \red{(per-layer)} to the weights; \red{empirical validation can be found in the supplementary material.}

\section{Experiments}
\label{sec:experiments}

\begin{figure}[t]
	\centering
	\vspace*{-0.2cm}
	\includegraphics[width=0.4\textwidth]{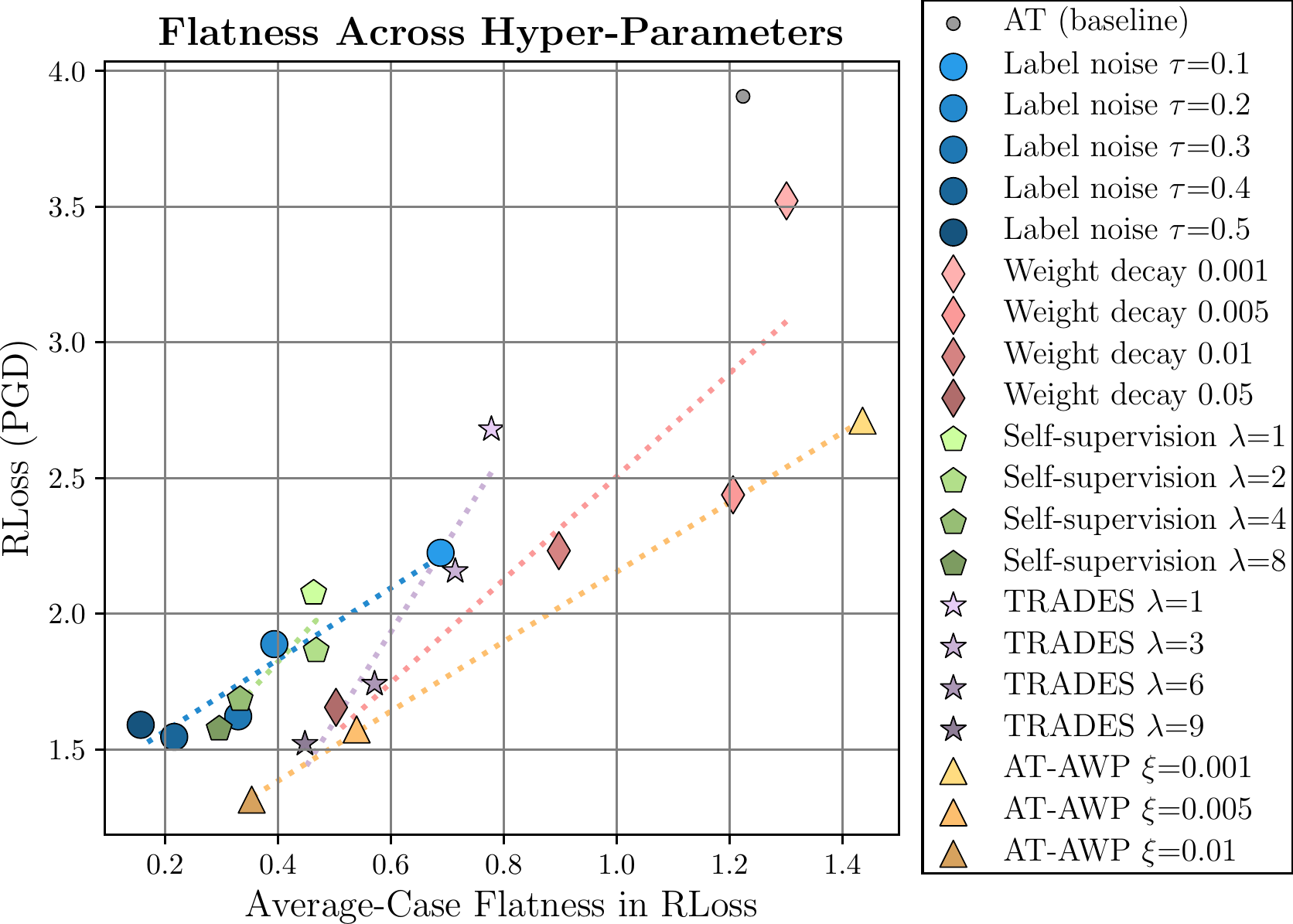}
	\vspace*{-6px}
	\caption{\textbf{Flatness Across Hyper-Parameters:} \RCE (y-axis) vs\onedot average-case flatness (x-axis) for selected methods and hyper-parameters (\cf supplementary material). For example, we consider different strengths of weight decay ({\color{plot2}rose}) or sizes $\xi$ of adversarial weight perturbations for AT-AWP ({\color{plot10!80!black}orange}). For clarity, we plot (dotted) lines representing the trend per method. Clearly, improved adversarial robustness, \ie, low \RCE, is related to improved flatness.}
	\label{fig:experiments-flatness-methods}
	\vspace*{-6px}
\end{figure}

\begin{figure*}[t]
	\centering
	\vspace*{-0.2cm}
	\hspace*{-0.4cm}
	\begin{minipage}[t]{0.29\textwidth}
		\vspace*{0px}
		\includegraphics[height=3.8cm]{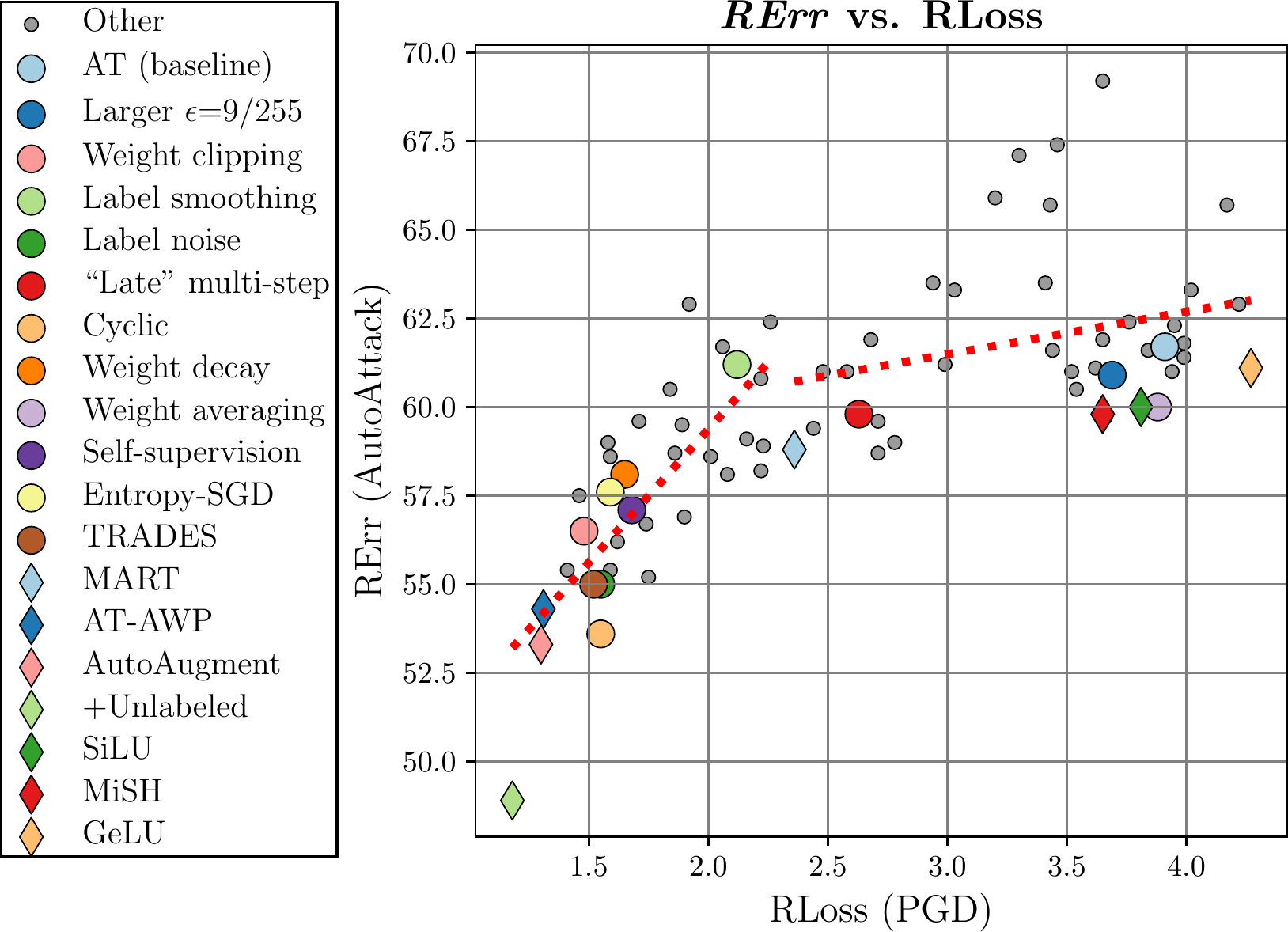}
	\end{minipage}
	\hspace*{2px}
	\begin{minipage}[t]{0.01\textwidth}
		\vspace*{0px}
		
		\hspace*{4px}{\color{black!75}\rule{0.65px}{3.8cm}}
	\end{minipage}
	\hspace*{2px}
	\begin{minipage}[t]{0.21\textwidth}
		\vspace*{0px}
		
		\includegraphics[height=3.8cm]{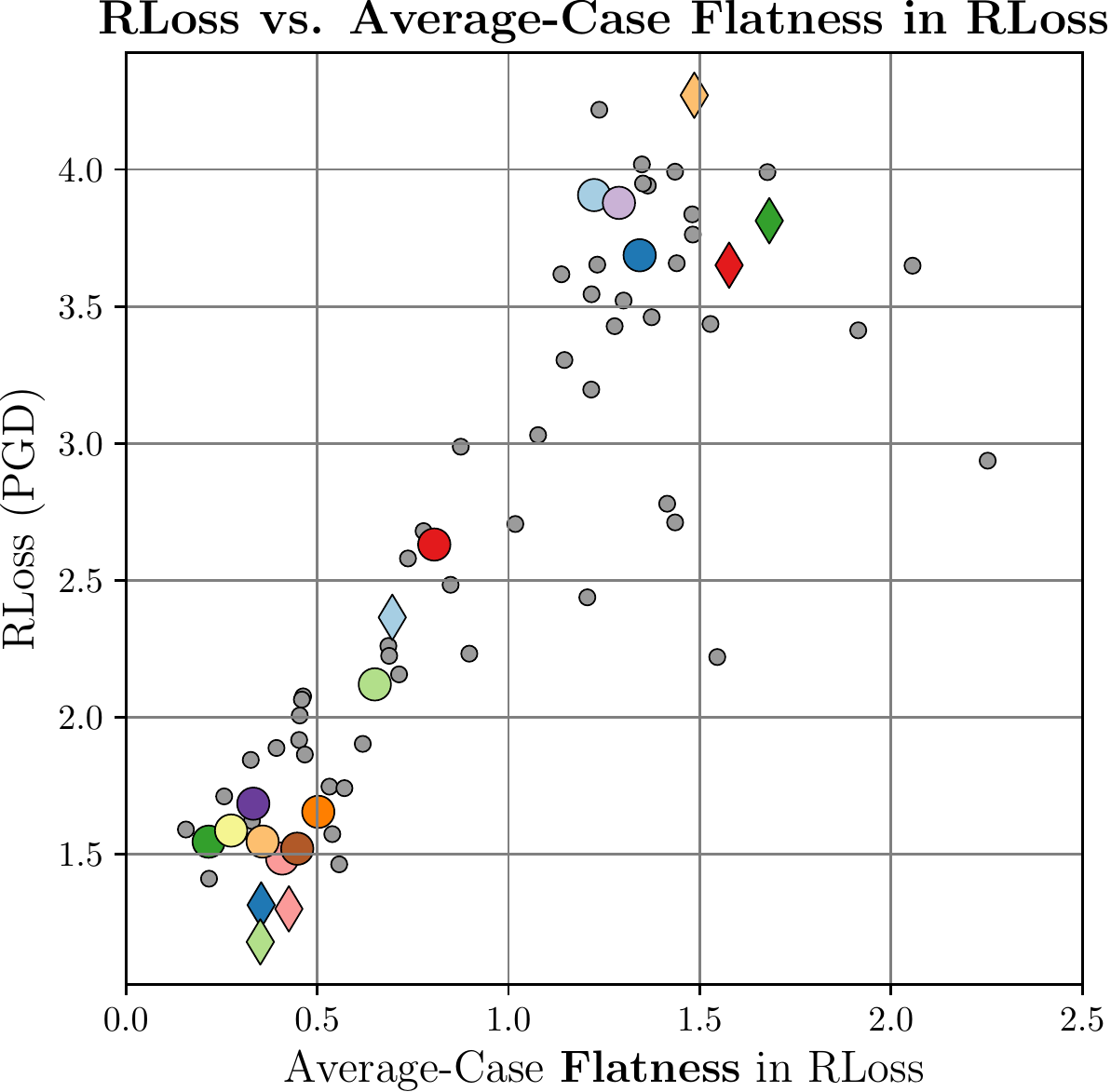}
	\end{minipage}
	\hspace*{2px}
	\begin{minipage}[t]{0.21\textwidth}
		\vspace*{0px}
		
		\includegraphics[height=3.8cm]{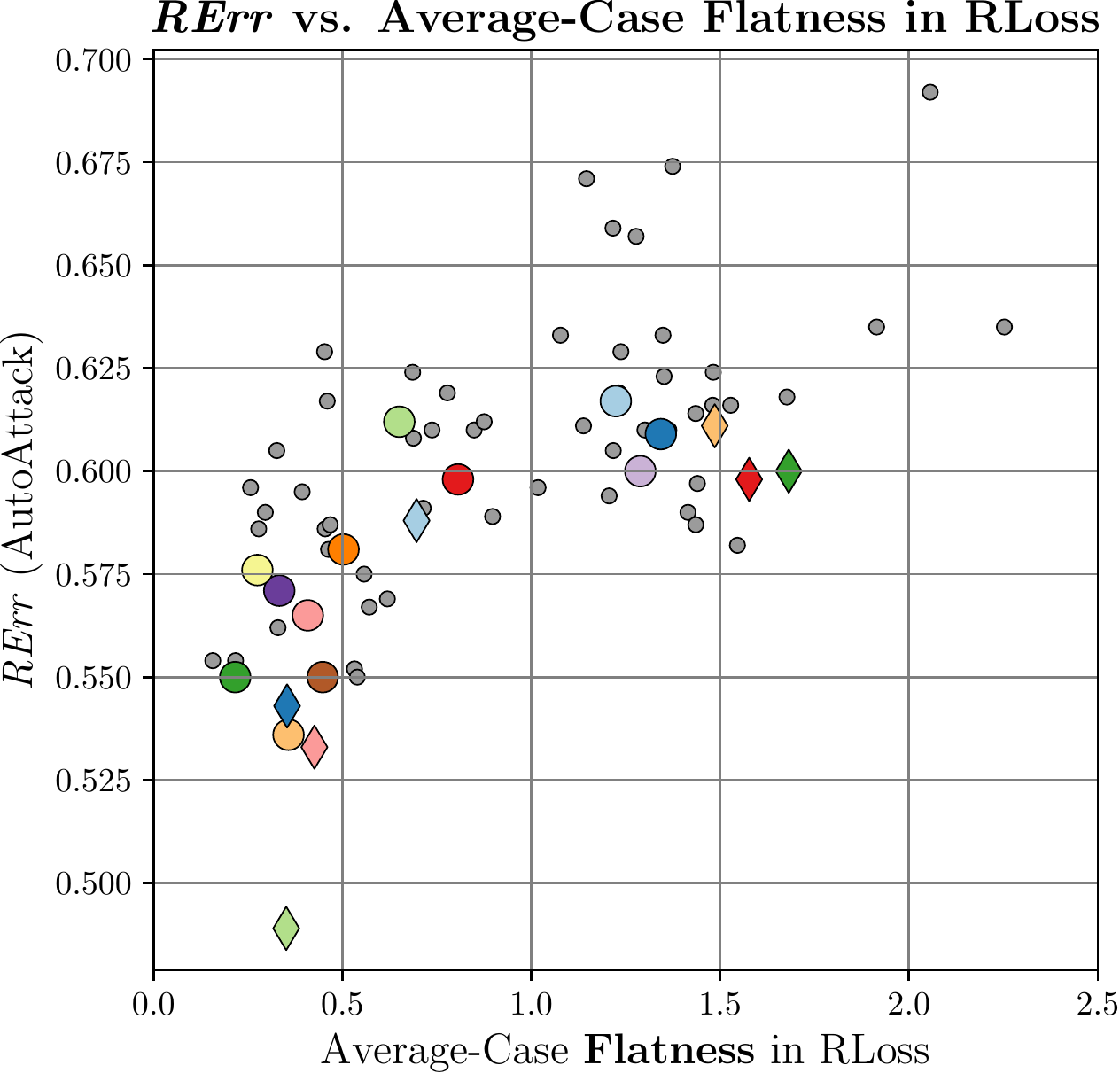}
	\end{minipage}
	\hspace*{2px}
	\begin{minipage}[t]{0.01\textwidth}
		\vspace*{0px}
		
		\hspace*{4px}{\color{black!75}\rule{0.65px}{3.8cm}}
	\end{minipage}
	\hspace*{2px}
	\begin{minipage}[t]{0.2\textwidth}
		\vspace*{0px}
		
		\includegraphics[height=3.8cm]{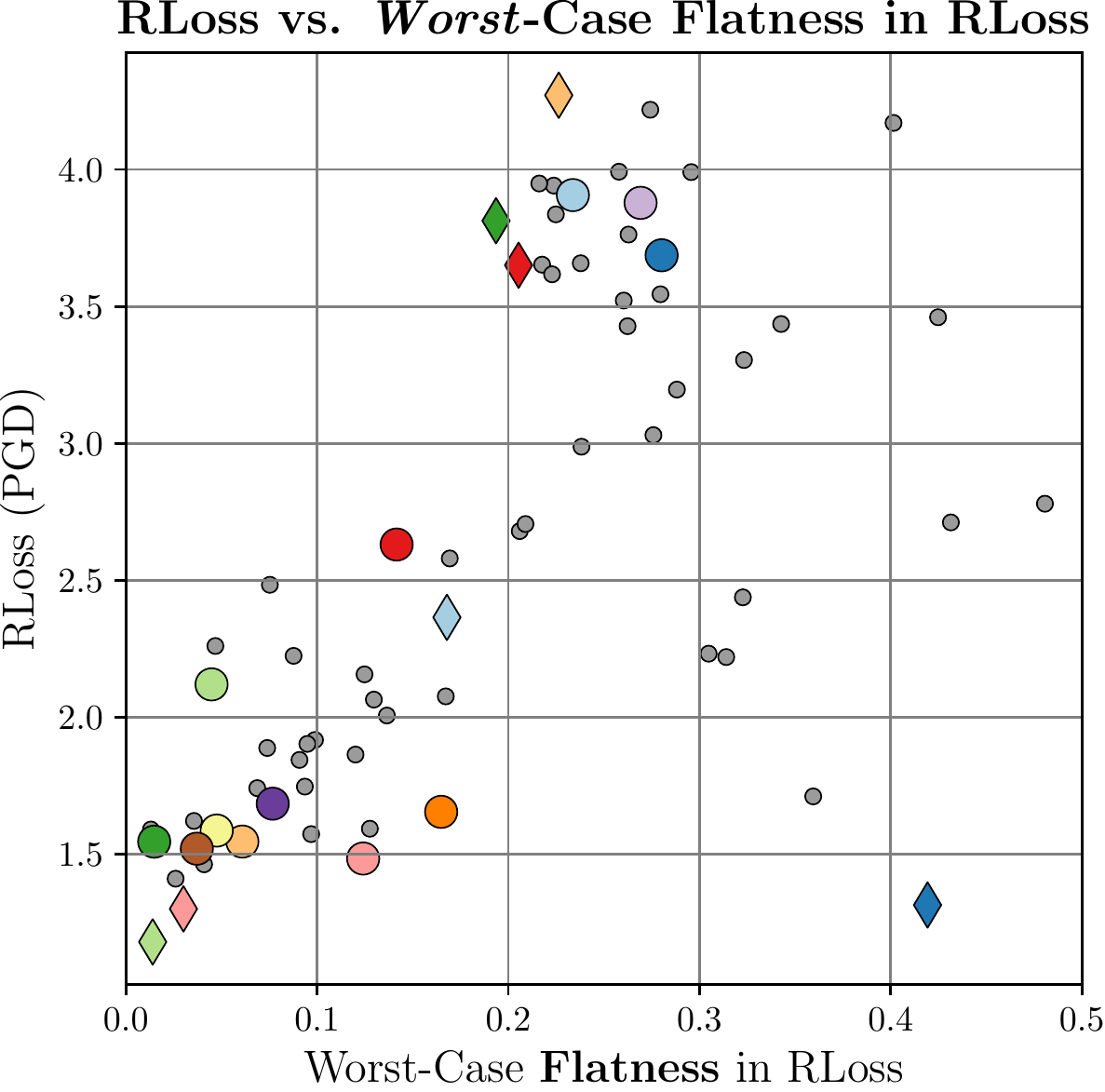}
	\end{minipage}
	
	\vspace*{-6px}
	\caption{\textbf{Robustness and Flatness:} \textbf{Left:} \RTE plotted against \RCE, showing that improved \RCE does not directly translate to reduced \RTE for large \RCE. In these cases, reducing \RCE mainly means reducing the confidence of adversarial examples, which is necessary to improve adversarial robustness. \textbf{Middle:} \RCE or \RTE (y-axis) plotted against our \emph{average-case} flatness in \RCE. We highlight selected models, as in \tabref{tab:experiments-flatness}. Considering \RCE, we reveal a striking correlation between adversarial robustness and flatness. Popular AT variants improving robustness (\eg, {\color{plot11}TRADES}, {\color{plot0}MART}, \etc) also correspond to flatter minima. Vice versa, methods improving flatness (\eg, {\color{plot10}Entropy-SGD}, {\color{plot7}weight decay}, \etc) improve robustness obtained through AT. Subject to the non-trivial interplay between \RTE and \RCE (\cf left), this relationship is also visible using \RTE to quantify robustness. \textbf{Right:} \RCE (y-axis) plotted against \emph{worst-case} flatness (x-axis) shows a less clear relationship. Still, improved flatness remains a necessity for better robust generalization, \red{see \secref{subsec:experiments-flatness} for discussion.}}
	\label{fig:experiments-flatness-correlation}
	\vspace*{-6px}
\end{figure*}

\begin{figure*}[t]
	\centering 
	\vspace*{-0.2cm}
	\begin{minipage}[t]{0.2\textwidth}
		\vspace*{0px}
				
		\includegraphics[height=3.6cm,clip,trim={0cm 0cm 3.75cm 0cm}]{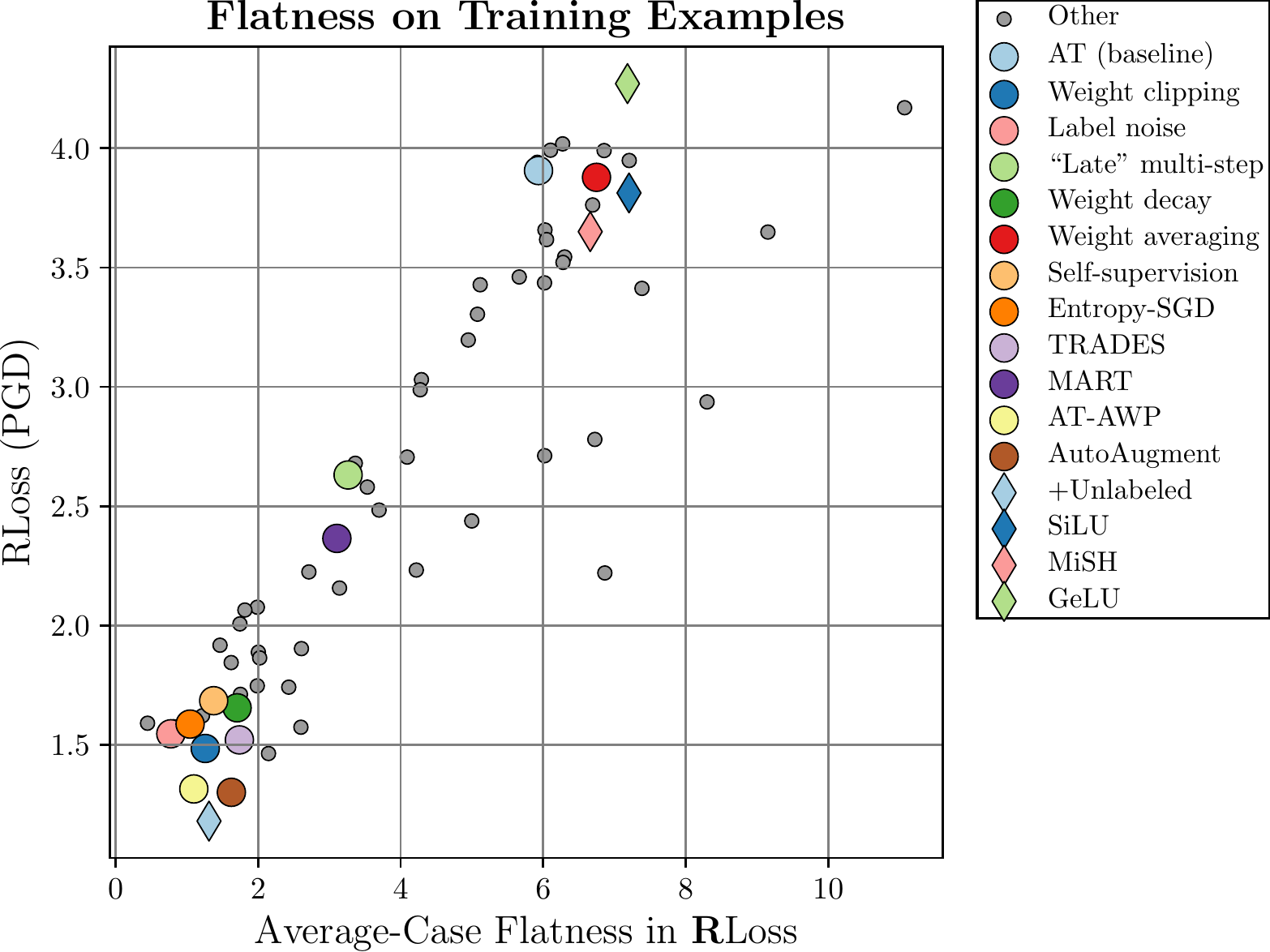}
	\end{minipage}
	\hspace*{2px}
	\begin{minipage}[t]{0.2\textwidth}
		\vspace*{0px}
		
		\includegraphics[height=3.6cm]{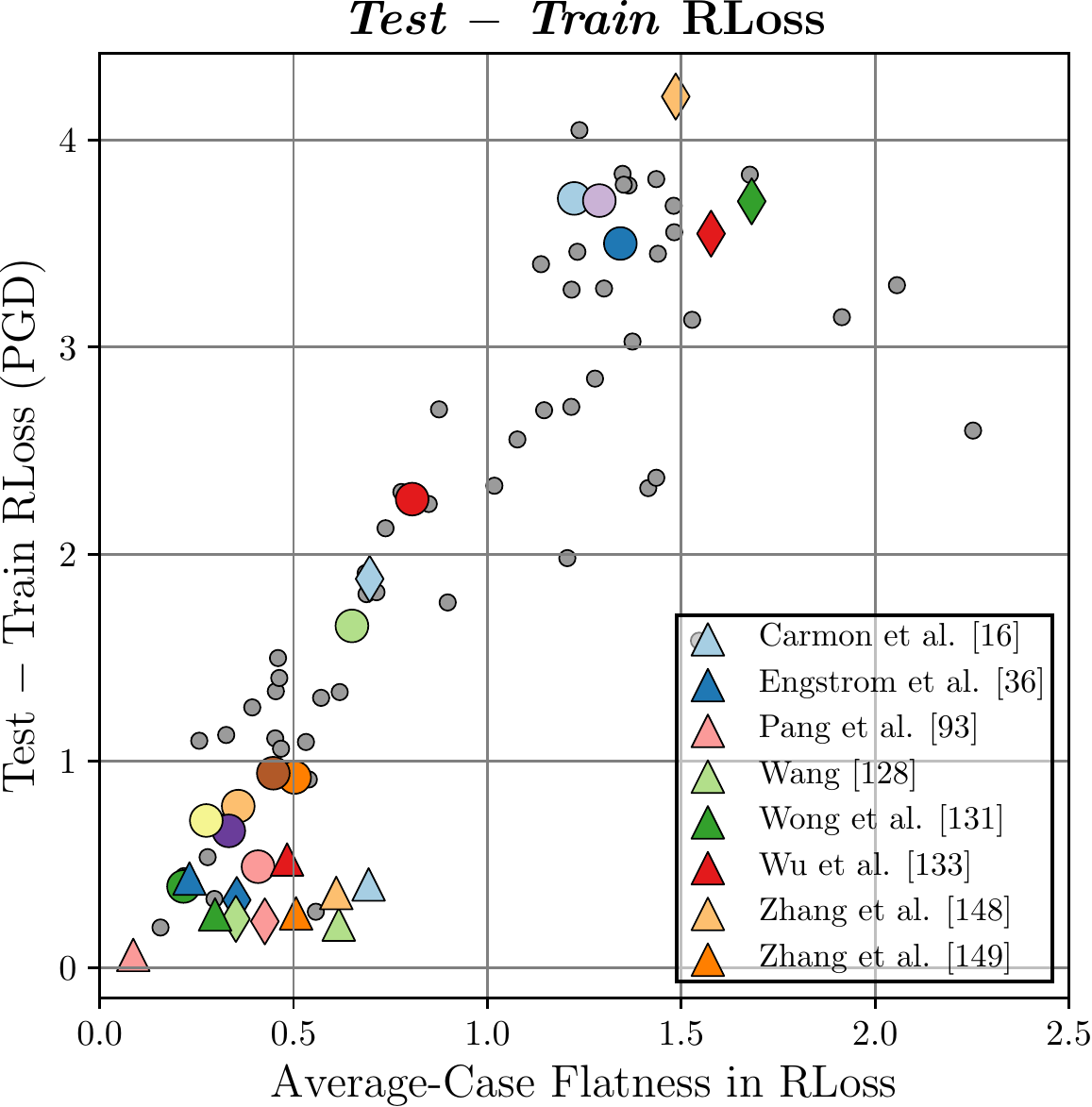}
	\end{minipage}
	\hspace*{2px}
	\begin{minipage}[t]{0.27\textwidth}
		\vspace*{-2.5px}
		
		\includegraphics[height=3.7cm]{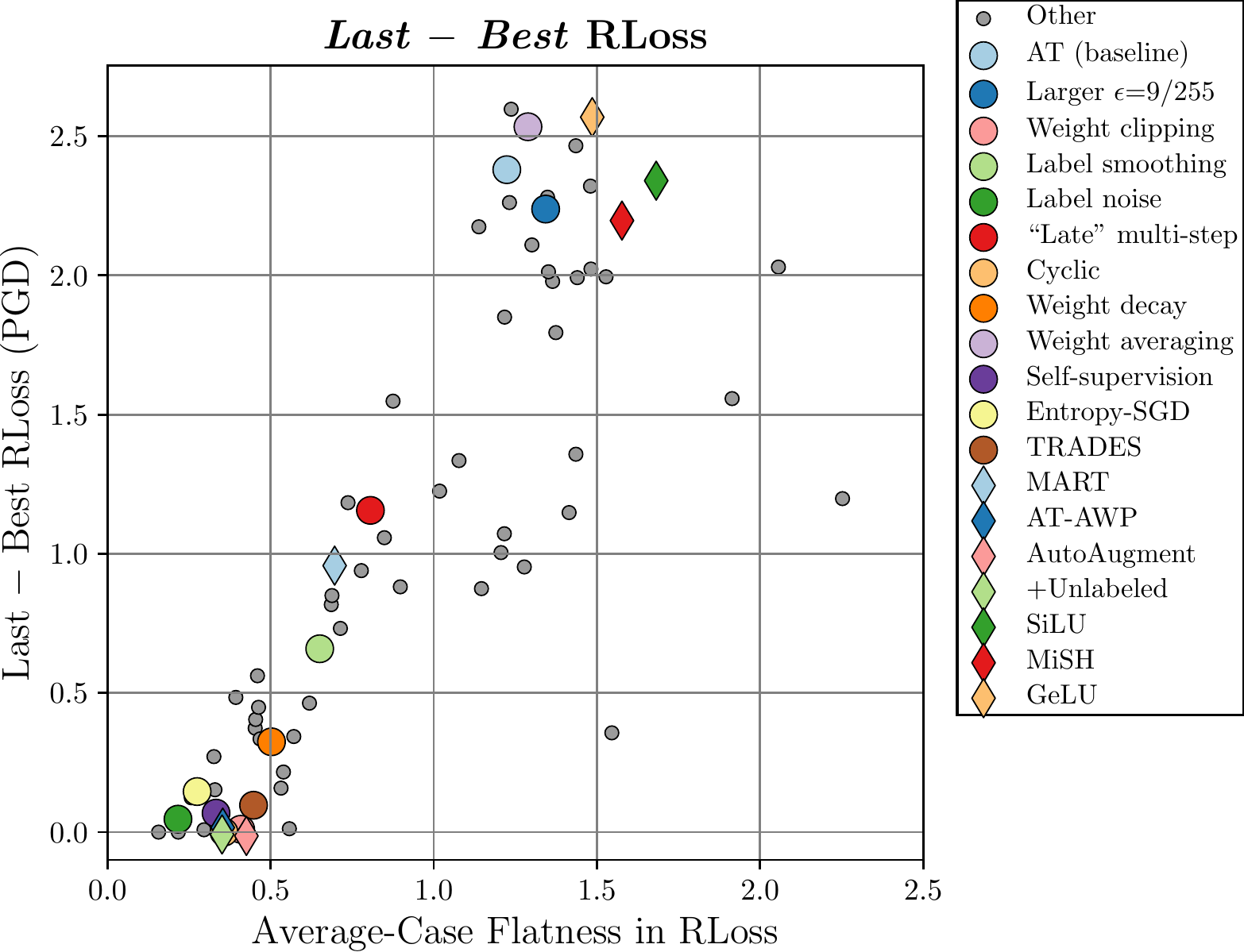}
	\end{minipage}
	\hspace*{2px}
	\begin{minipage}[t]{0.01\textwidth}
		\vspace*{0px}
		
		\hspace*{4px}{\color{black!75}\rule{0.65px}{3.65cm}}
	\end{minipage}
	\hspace*{2px}
	\begin{minipage}[t]{0.265\textwidth}
		\vspace*{0px}
		
		\includegraphics[height=3.6cm]{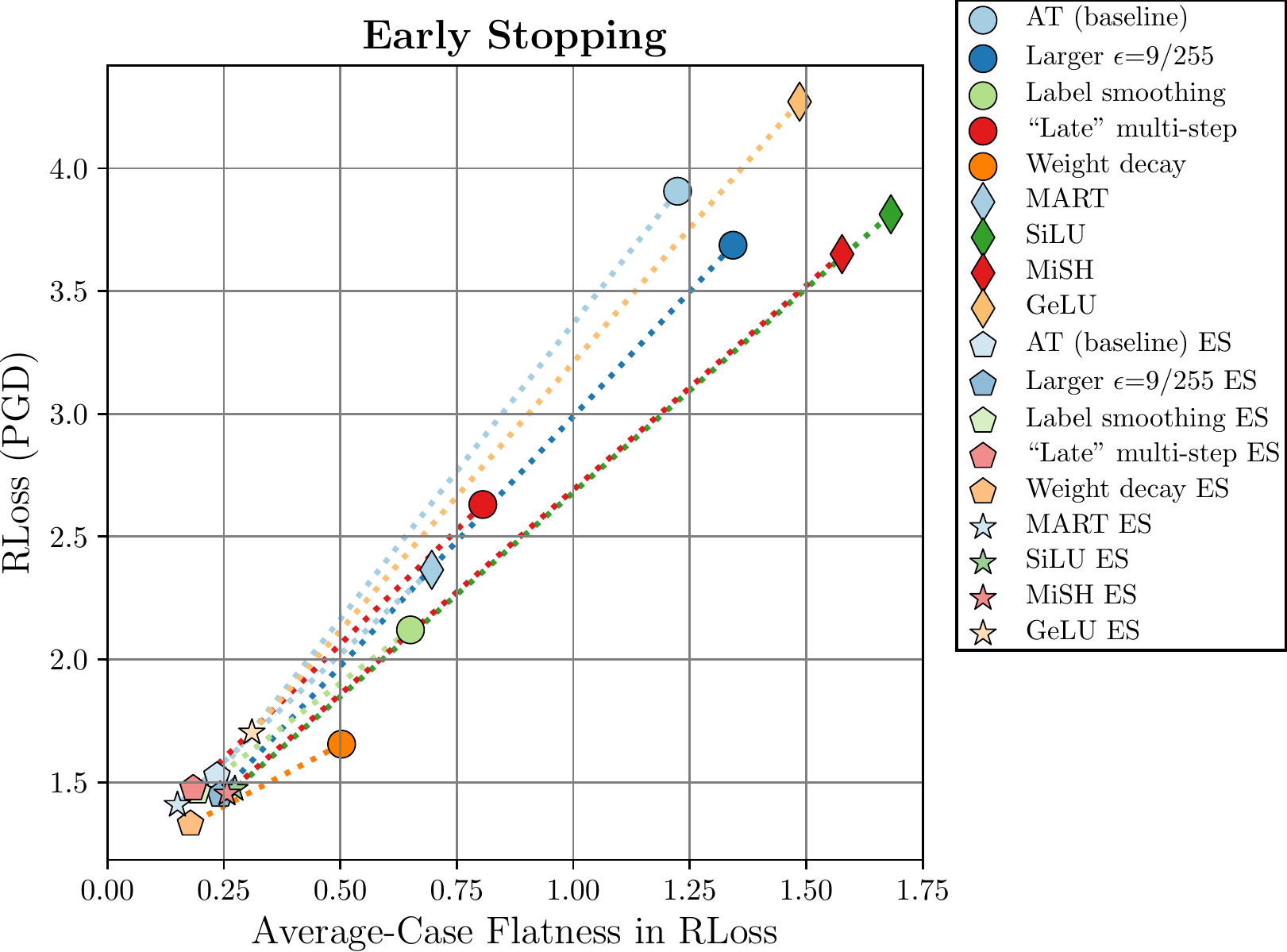}
	\end{minipage}	
	\vspace*{-8px}
	\caption{\textbf{Robust Generalization and Early Stopping.} \textbf{Left:} \red{\RCE plotted vs. average-case flatness measured \emph{on training examples}. Even on training examples, flatness is a good indicator for robust generalization.} \textbf{Middle:} Robust generalization (\RCE) decomposed into the test-train difference and the last-best (epoch) improvement (y-axis), both plotted against average-case flatness in \RCE(x-axis). In both cases, flatness seems to play an important role, \ie, flatness clearly reduces both the robust generalization gap \emph{and} robust overfitting. \textbf{Right:} \RCE vs\onedot average-case flatness in \RCE for selected models (all in supplementary material) with and without early stopping (``ES''). Early stopping consistently leads to improved adversarial robustness \emph{and} better flatness.}
	\label{fig:experiments-flatness-gap}
	\vspace*{-6px}
\end{figure*}

We start with a closer look at \RCE in robust overfitting (\secref{subsec:experiments-overfitting}, \figref{fig:experiments-understanding}). Then, we show a strong correlation between good robust generalization and flatness (\secref{subsec:experiments-flatness}). \red{For example, robust overfitting causes sharper minima (\figref{fig:experiments-flatness-epochs}). More importantly, more robust models generally find flatter minima and, vice-versa, methods encouraging flatness improve adversarial robustness (\figref{fig:experiments-flatness-methods}, \ref{fig:experiments-flatness-correlation}).} In fact, flatness improves robust generalization by \emph{both} lowering the robust generalization gap \red{(incl. a reduction in robust overfitting, \cf \figref{fig:experiments-flatness-gap})}.

\textbf{Setup:} 
On \CifarT \cite{Krizhevsky2009}, our \emph{AT baseline} uses ResNet-18 \cite{HeCVPR2016} and is trained for $150$ epochs, batch size $128$, learning rate $0.05$, reduced by factor $0.1$ at $60$, $90$ and $120$ epochs, using weight decay $0.005$ and momentum $0.9$ with standard SGD. We use random flips and cropping as data augmentation. During training, we use $7$ iterations PGD, with learning rate $0.007$, signed gradient and $\epsilon = \nicefrac{8}{255}$ for $L_\infty$ adversarial examples. PGD-$7$ is also used for early stopping (every $5$th epoch) on the last $500$ test examples. We do \emph{not} use early stopping by default. For evaluation on the first $1000$ test examples, we run PGD with $20$ iterations, $10$ random restarts to estimate \RCE and AutoAttack \cite{CroceICML2020} to estimate \RTE\xspace\red{(\cf \secref{subsec:main-overfitting})}. For \emph{average-case flatness of \RCE}, we \red{take the average of} $10$ random weight perturbations with $\xi{=}0.5$. For \emph{worst-case flatness}, we maximize \RCE jointly over adversarial examples and adversarial weights with $\xi{=}0.00075$, \red{taking the worst of $10$ restarts.}
 
\textbf{Methods:}
Besides our AT baseline, we consider AT-AWP \cite{WuNIPS2016}, TRADES \cite{ZhangICML2019}, MART \cite{WangICLR2020}, AT with self-supervision \cite{HendrycksNIPS2019} or additional unlabeled examples \cite{CarmonNIPS2019,UesatoNIPS2019}, weight averaging \cite{IzmailovUAI2018} and AT with ``early-stopped'' PGD \cite{ZhangICML2020}. We investigate different hyper-parameters and ``tricks'' recently studied in \cite{PangARXIV2020b,GowalARXIV2020}: learning rate schedules, batch size, weight decay, label smoothing \cite{SzegedyCVPR2016} as well as SiLU/Mish/GeLU \cite{ElfwingNN2018,MisraBMVC2020,HendrycksARXIV2016} activation functions. Furthermore, we consider Entropy-SGD \cite{ChaudhariICLR2017}, label noise, weight clipping \cite{StutzMLSYS2021} and AutoAugment \cite{CubukARXIV2018}. We emphasize that weight averaging, Entropy-SGD and weight clipping are known to improve flatness of the (clean) loss.
\red{\emph{If not stated otherwise, these methods are applied on top or as replacement of our AT baseline.}}
We report results using the best hyper-parameters per method. Finally, we also use pre-trained models from RobustBench \cite{CroceARXIV2020b}, which were obtained using early stopping. 

\begin{table}[t]
	\centering
	\vspace*{-0.25cm}
	\hspace*{-0.4cm}
	{
	\scriptsize
	\setlength{\tabcolsep}{0pt}
    \newcolumntype{C}[1]{@{}>{\centering\arraybackslash}p{#1}@{}}
	\begin{tabularx}{0.5\textwidth}{|X|C{0.85cm}|C{1.4cm}|C{0.85cm}|C{0.85cm}||C{1.4cm}|}
		\hline
		\hspace*{2px} \bfseries Model & \multicolumn{2}{c|}{\bfseries Robustness $\downarrow$} & \multicolumn{2}{c||}{\bfseries Flatness $\downarrow$} & \bfseries Early Stop.\\
		\hline
		\hspace*{2px}\textcolor{colorbrewer1}{(sorted asc. by test \RTE)} & \RTE & \RTE & Avg & Worst & \RTE $\downarrow$\\
		\hspace*{2px}\textcolor{gray}{(split at $70\%$/$30\%$ percentiles)} & (test) & (train) & (\RCE) & (\RCE) & (early stop)\\
		\hline
		\hline		
		\rowcolor{colorbrewer3!15}\hspace*{2px} +Unlabeled & 48.9 & 43.2 (-5.7) & 0.32 & 1.20 & 48.9 (-0.0)\\
		\rowcolor{colorbrewer3!15}\hspace*{2px} Cyclic & 53.6 & 35.4 (-18.2) & 0.35 & 1.50) & 53.6 (-0.0)\\
		\rowcolor{colorbrewer3!15}\hspace*{2px} AutoAugment & 54.0 & 47.9 (-6.1) & 0.49 & 0.69 & 53.5 (-0.5)\\
		\rowcolor{colorbrewer3!15}\hspace*{2px} AT-AWP & 54.3 & 43.1 (-11.2) & 0.35 & 2.68 & 53.6 (-0.7)\\
		\rowcolor{colorbrewer3!15}\hspace*{2px} Label noise & 56.2 & 30.0 (-26.2) & 0.33 & 0.93) & 55.5 (-0.7)\\
		\rowcolor{colorbrewer3!15}\hspace*{2px} Weight clipping & 56.5 & 39.0 (-17.5) & 0.41 & 4.57 & 56.5 (-0.0)\\
		\rowcolor{colorbrewer3!15}\hspace*{2px} TRADES & 56.7 & 15.8 (-40.9) & 0.57 & 2.25 & 53.4 (-3.3)\\
		\hline
		\rowcolor{colorbrewer5!15}\hspace*{2px} Self-supervision & 57.1 & 45.0 (-12.1) & 0.33 & 2.63 & 56.8 (-0.3)\\
		\rowcolor{colorbrewer5!15}\hspace*{2px} Weight decay & 58.1 & 32.8 (-25.3) & 0.50 & 3.93 & 54.8 (-3.3)\\
		\rowcolor{colorbrewer5!15}\hspace*{2px} Entropy-SGD & 58.6 & 46.1 (-12.5) & 0.28 & 1.80 & 56.9 (-1.7)\\
		\rowcolor{colorbrewer5!15}\hspace*{2px} MiSH & 59.8 & 5.3 (-54.5) & 1.56 & 3.54 & 53.7 (-6.1)\\
		\rowcolor{colorbrewer5!15}\hspace*{2px} ``Late'' multi-step & 59.8 & 18.4 (-41.4) & 0.80 & 2.96 & 57.8 (-2.0)\\
		\hline
		\rowcolor{colorbrewer1!15}\hspace*{2px} SiLU & 60.0 & 5.6 (-54.4) & 1.71 & 4.20 & 53.7 (-6.3)\\
		\rowcolor{colorbrewer1!15}\hspace*{2px} Weight averaging & 60.0 & 10.0 (-50.0) & 1.28 & 5.98 & 53.0 (-7.0)\\
		\rowcolor{colorbrewer1!15}\hspace*{2px} Larger $\epsilon{=}\nicefrac{9}{255}$ & 60.9 & 11.1 (-49.8) & 1.33 & 5.84 & 53.8 (-7.1)\\
		\rowcolor{colorbrewer1!15}\hspace*{2px} MART & 61.0 & 20.8 (-40.2) & 0.73 & 3.17 & 54.7 (-6.3)\\
		\rowcolor{colorbrewer1!15}\hspace*{2px} GeLU & 61.1 & 3.2 (-57.9) & 1.55 & 4.12 & 56.7 (-4.4)\\
		\rowcolor{colorbrewer1!15}\hspace*{2px} Label smoothing & 61.2 & 8.0 (-53.2) & 0.65 & 2.72 & 54.0 (-7.2)\\
		\rowcolor{colorbrewer1!15}\hspace*{2px} AT (baseline) & 62.8 & 10.7 (-52.1) & 1.21 & 6.48 & 54.6 (-8.2)\\
		\hline
	\end{tabularx}
	}
	
	\hspace*{-0.38cm}
	{
	\tiny
	\setlength{\tabcolsep}{0pt}
 	\newcolumntype{C}[1]{@{}>{\centering\arraybackslash}p{#1}@{}}
	\begin{tabularx}{0.5\textwidth}{|X|C{0.85cm}|C{1.4cm}|C{0.85cm}|C{0.85cm}||C{1.4cm}|}
		\hline
		\hspace*{2px} \bfseries Robustness & \multicolumn{5}{c|}{\bfseries Averages (across models)}\\
		\hline
		\rowcolor{colorbrewer3!15}\hspace*{2px} Good ($\RTE{<}57\%{\approx}30\%$ percentile) & 54.3 & 36.3 (-18.0) & 0.40 & 2.00 & 53.6 (-0.7)\\
		\rowcolor{colorbrewer5!15}\hspace*{2px} Average ($57\%{\geq}\RTE < 60\%$) & 58.7 & 29.5 (-29.2) & 0.69 & 2.9 & 56.0 (-2.7)\\
		\rowcolor{colorbrewer1!15}\hspace*{2px} Poor ($\RTE{\geq}60\%{\approx}70\%$ percentile) & 61.0 & 9.9 (-51.1) & 1.21 & 4.67 & 54.4 (-6.6)\\
		\hline
	\end{tabularx}
	}
	\vspace*{-6px}
	\caption{\textbf{Robustness and Flatness, Quantitative Results:} Test and train \RTE (first, second column, early stopping in fifth column) as well as average-/worst-case flatness in \RCE (third, fourth column) for selected methods, \cf \figref{fig:experiments-flatness-correlation}.
	We split methods into \colorbox{colorbrewer3!15}{good}, \colorbox{colorbrewer5!15}{average}, and \colorbox{colorbrewer1!15}{poor} robustness using the $30\%$ and $70\%$ percentiles.
	Most methods improve adversarial robustness alongside both average- and worst-case flatness.
	}
	\label{tab:experiments-flatness}
	\vspace*{-6px}
\end{table}

Our \textbf{supplementary material} includes additional details on the experimental setup and the evaluated methods. Furthermore, it contains an ablation regarding our average- and worst-case flatness measure and hyper-parameter ablation for individual methods, including training curves.
 
\subsection{Understanding Robust Overfitting}
\label{subsec:experiments-overfitting}

In contrast to related work \cite{RiceICML2020}, we take a closer look at \RCE during robust overfitting because \RTE is ``blind'' to many improvements in \RCE, especially on incorrectly classified examples. \figref{fig:experiments-understanding} shows training curves for various methods, \ie, \RCE/\RTE over (normalized) epochs. For example, explicitly handling incorrectly classified examples during training, using MART, helps but does not \emph{prevent} overfitting: \RCE for {\color{plot2}MART} reduces compared to {\color{plot1}AT} (first column). Unfortunately, this improvement does \emph{not} translate to significantly better \RTE, \cf \tabref{tab:experiments-flatness}. This discrepancy between \RCE and \RTE can be reproduced for other methods, as well: label smoothing and label noise enforce, in expectation, the same target distribution. Thus, both reduce \RCE during overfitting (second column, top, {\color{plot2}rose} and {\color{plot4}dark green}). Label smoothing, however, does not improve \RTE as significantly as label noise, \ie, does not \emph{prevent} mis-classification. This illustrates an important aspect: against adversarial examples, ``merely'' improving \RCE does not translate to improved \RTE if \RCE is high to begin with, \ie, ``above'' $-\ln(\nicefrac{1}{K}){\approx}2.3$ for $K{=}10$ classes. However, this is usually the case during robust overfitting. \RTE, on the other hand, does not take into account the confidence of wrong predictions, \ie, it is ``blind'' for these improvements in \RCE. Label noise, in contrast, also improves \RTE, which might be due to the additional randomness.

Similar to established methods, many ``simple'' regularization schemes prove surprisingly effective in tackling robust overfitting. For example, strong {\color{plot1}weight decay} delays robust overfitting and {\color{plot4}AutoAugment} prevents overfitting entirely, \cf \figref{fig:experiments-understanding} (third column). This indicates that popular AT variants, \eg, {\color{plot6}TRADES}, AT with {\color{plot2}self-supervision} or {\color{plot5}unlabeled} examples, improve adversarial robustness by avoiding robust overfitting through regularization. This is achieved by preventing convergence on training examples (dotted).
In regularization, however, hyper-parameters play a key role: even AT-AWP does not prevent robust overfitting if regularization is ``too weak'' ({\color{plot1}blue}, fourth column). This is particularly  prominent in terms of \RCE (top). Finally, learning rate schedules play an important role in how and \emph{when} robust overfitting occurs (fifth column). However, as in \cite{RiceICML2020}, all schedules are subject to robust overfitting.

\subsection{Robust Generalization and Flatness in \RCE}
\label{subsec:experiments-flatness}

As robust overfitting is primarily avoided through strong regularization, we hypothesize that this is because strong regularization finds flatter minima in the \RCE landscape. These flat minima help to improve robust generalization.

\textbf{Flatness in \RCE ``Explains'' Overfitting:}
Using our average- and worst-case flatness measures in \RCE, we find that flatness reduces significantly during robust overfitting. Namely, flatness ``explains'' the increased \RCE caused by overfitting very well. \figref{fig:experiments-flatness-epochs} (left) plots \RCE, alongside average- and worst-case flatness and the maximum Hessian eigenvalue throughout training of our AT baseline. Clearly, flatness increases alongside (test) \RCE as soon as robust overfitting occurs. Note that the best epoch is $60$, meaning $0.4$ (black dotted). For further illustration, \figref{fig:experiments-flatness-epochs} (middle) explicitly plots \RCE (y-axis) against flatness in \RCE (x-axis) across epochs ({\color{blue!50!black}dark blue} to {\color{red!50!black}dark red}): \RCE and flatness clearly worsen ``alongside'' each other during overfitting, for both average- and worst-case flatness. Methods such as AT with self-supervision, AT-AWP or AT with unlabeled examples avoid both robust overfitting \emph{and} sharp minima (right).
This relationship generalizes to different hyper-parameter choices of these methods: \figref{fig:experiments-flatness-methods} plots \RCE (y-axis) vs\onedot average-case flatness (x-axis) across different hyper-parameters. Again, \eg, for {\color{plot9}TRADES} or {\color{plot10!80!black}AT-AWP}, hyper-parameters with lower \RCE also correspond to flatter minima.
In fact, \figref{fig:experiments-flatness-methods} indicates that the connection between robustness and flatness also generalizes \emph{across} different methods (and individual models).

\textbf{Improved Robustness Through Flatness:}
Indeed, across all trained models, we found a \textbf{strong correlation between robust generalization and flatness}, using \RCE as measure for robust generalization.
As discussed in \secref{subsec:experiments-overfitting}, we mainly consider \RCE to assess robust generalization as improvements in \RCE above ${\sim}2.3$ have, on average, only small impact on \RTE. Pushing \RCE below $2.3$, in contrast, directly translates to better \RTE. This is illustrated in \figref{fig:experiments-flatness-correlation} (left) which plots \RTE vs\onedot \RCE for all evaluated models. To avoid this ``kink'' in the {\color{red}dotted red} lines around $\RCE{\approx}2.3$, \figref{fig:experiments-flatness-correlation} (middle left) plots \emph{\RCE} (y-axis) against \emph{average-case} flatness in \RCE (x-axis), highlighting selected models. This reveals a \emph{clear correlation between robustness and flatness}:
More robust methods, \eg, AT with unlabeled examples or AT-AWP, correspond to flatter minima. Methods improving flatness, \eg, Entropy-SGD, weight decay or weight clipping, improve adversarial robustness. This also translates to \RTE (middle right), subject to the described bend at $\RCE{\approx}2.3$. While many robust methods still obtain better flatness, activation functions such as SiLU, MiSH or GeLU also seem to improve flatness, without clear advantage in terms of robustness. Similarly, weight decay or clipping improve robustness considerably. \red{Overall, with Pearson/Spearman correlation coefficients of $0.85$/$0.87$ ($p$-values ${<}10^{-21}$), we revealed a strong relationship between robustness and flatness.}

\red{\figref{fig:experiments-flatness-correlation} (right) shows that this relationship is less clear when considering \emph{worst-case} flatness in \RCE (Pearson coefficient $0.54$). This is in contrast to \cite{WuNIPS2020} suggesting that worst-case flatness, in particular, is important to improve robustness of AT. However, worst-case flatness is more sensitive to $\xi$ and, thus, less comparable across methods. Note that worst-case robustness is still a good indicator for overfitting, \cf \figref{fig:experiments-flatness-epochs}.} All results are summarized in tabular form in \tabref{tab:experiments-flatness}: Grouping methods by \colorbox{colorbrewer3!15}{good}, \colorbox{colorbrewer5!15}{average} or \colorbox{colorbrewer1!15}{poor} robustness, we find that methods need at least ``some'' flatness, average- \emph{or} worst-case, to be successful.

\textbf{Decomposing Robust Generalization:}
So far, we used (absolute) \RCE on test examples as proxy of robust generalization. This is based on the assumption that deep models are generally able to obtain nearly zero \emph{train} \RCE. However, this is not the case for many methods in \tabref{tab:experiments-flatness} (second column). Thus, we also consider the robust generalization \emph{gap} and the \RCE difference between last and best (early stopped) epoch. \red{First, however, \figref{fig:experiments-flatness-gap} (left) shows that flatness, when measured on \emph{training examples}, is also a good predictor of (test) robustness.} Then, \figref{fig:experiments-flatness-gap} (middle left) explicitly plots the \RCE generalization gap (test$-$train \RCE, y-axis) against average-case flatness in \RCE (x-axis). Robust methods generally reduce this gap by \emph{both} reducing test \RCE \emph{and} avoiding convergence in train \RCE. Furthermore, \figref{fig:experiments-flatness-gap} (middle right) considers the difference between last and best epoch, essentially quantifying the extent of robust overfitting. Again, methods with small difference, \ie, little robust overfitting, generally correspond to flatter minima. This is also confirmed in \figref{fig:experiments-flatness-gap} (right) showing that early stopping essentially finds flatter minima along the training trajectory, thereby improving adversarial robustness.
Altogether, flatness improves robust generalization by reducing both the robust generalization gap \emph{and} the impact of robust overfitting.

\textbf{More Results:} \figref{fig:introduction} shows that the pre-trained models from RobustBench \cite{CroceARXIV2020b} confirm our observations so far (also see \figref{fig:experiments-flatness-gap}, middle left). While detailed analysis is not possible as only early stopped models are provided, they are consistently more robust \emph{and} correspond to flatter minima compared to our models. This is despite using different architectures (commonly Wide ResNets \cite{ZagoruykoBMVC2016}).

%% file: sec_conclusion.tex
\section{Conclusion}
\label{sec:conclusion}

In this paper, we studied the relationship between adversarial robustness, specifically considering robust overfitting \cite{RiceICML2020}, and flatness of the robust loss (\RCE) landscape \wrt perturbations in the weight space. We introduced both average- and worst-case measures for flatness in \RCE that are scale-invariant and allow comparison across models. Considering adversarial training (AT) and several popular variants, including TRADES \cite{ZhangICML2019}, AT-AWP \cite{WuNIPS2020} or AT with additional unlabeled examples \cite{CarmonNIPS2019}, we show a \textbf{clear relationship between adversarial robustness and flatness} in \RCE. More robust methods predominantly find flatter minima. Vice versa, approaches known to improve flatness, \eg, Entropy-SGD \cite{ChaudhariICLR2017} or weight clipping \cite{StutzMLSYS2021} can help AT become more robust, as well. Moreover, even simple regularization methods such as AutoAugment \cite{CubukARXIV2018}, weight decay or label noise, are effective in increasing robustness by improving flatness. These observations also generalize to pre-trained models from RobustBench \cite{CroceARXIV2020b}. 

%% file: supp_introduction.tex
\section{Overview}

In the main paper, we empirically studied the connection between adversarial robustness (in terms of the \emph{robust} loss \RCE, \ie, the cross-entropy loss on adversarial examples) and flatness of the \RCE landscape \wrt changes in the weight space. In this context, we also consider the phenomenon of robust overfitting \cite{RiceICML2020}, \ie, that robustness on training examples increases consistently throughout training while robustness on test examples eventually \emph{decreases}. Based on average- and worst-case metrics of flatness in \RCE, which we ensure to be scale-invariant,  we show a \textbf{clear relationship between adversarial robustness and flatness}. This takes into account many popular variants of adversarial training (AT), \ie, training on adversarial examples: TRADES \cite{ZhangICML2019}, MART \cite{WangICLR2020}, AT-AWP \cite{WuNIPS2020} AT with self-supervision \cite{HendrycksNIPS2019} or additional unlabeled examples \cite{CarmonNIPS2019}. All of them improve adversarial robustness \emph{and} flatness. Vice versa, approaches known to improve flatness, \eg, Entropy-SGD \cite{ChaudhariICLR2017}, weight clipping \cite{StutzMLSYS2021} or weight averaging \cite{GowalARXIV2020} also improve adversarial robustness. Finally, we found that even simple regularization schemes, \eg, AutoAugment \cite{CubukARXIV2018}, weight decay or label noise, also improve robustness by finding flatter minima.

\subsection{Contents}

This supplementary material is organized as follows:
\begin{itemize}
	\item \secref{sec:supp-related-work}: additional discussion of \textbf{related work}.
	\item \secref{sec:supp-visualization}: details on \textbf{\RCE landscape visualization} and comparison to \cite{LiNIPS2018} (\cf \figref{fig:supp-visualization} and \ref{fig:supp-visualization-li}).
	\item \secref{sec:supp-flatness-computation}: details on how to \textbf{compute our average- and worst-case flatness measures}, including ablation studies in \secref{sec:supp-flatness-ablation} (\cf \figref{fig:supp-flatness-epochs}, \ref{fig:supp-flatness-misc} and \ref{fig:supp-flatness-ablation}).
	\item \secref{sec:supp-scaling}: discussion of \textbf{scale-invariance} of our visualization and flatness measures (\cf \figref{fig:supp-scaling} and \tabref{tab:supp-convexity}).
	\item \secref{sec:supp-setup}: specifics on our \textbf{experimental setup} (training and evaluation details).
	\item \secref{sec:supp-methods}: discussion of all individual \textbf{methods}, including ablation regarding hyper-parameters in \secref{sec:supp-methods-ablation} (\cf \figref{fig:supp-ii-pll} and \ref{fig:supp-training-ablation}) and flatness in \secref{sec:supp-methods-flatness} (\cf \figref{fig:supp-methods-flatness-epochs} and \ref{fig:supp-flatness-methods}). Training curves for all methods in \figref{fig:supp-training-curves}.
	\item \secref{sec:supp-results}: all \textbf{results in tabular form} (\tabref{tab:supp-table-robustbench-error}, \ref{tab:supp-table-robustbench-loss}, \ref{tab:supp-table-loss}, \ref{tab:supp-table-error})
\end{itemize}

%% file: supp_related_work.tex
\section{Related Work}
\label{sec:supp-related-work}

\textbf{Adversarial Examples and Defenses:} Adversarial examples, first reported in \cite{SzegedyARXIV2013}, can be generated using a wide-range of white-box attacks \cite{SzegedyARXIV2013,GoodfellowARXIV2014,KurakinARXIV2016b,PapernotSP2016b,MoosaviCVPR2016,MadryARXIV2017,CarliniSP2017,DongCVPR2018,LuoAAAI2018}, with full access to the network, or black-box attacks \cite{ChenAISEC2017,BrendelARXIV2017a,SuARXV2017,IlyasARXIV2018,SarkarARXIV2017,NarodytskaCVPRWORK2017}, with limited access to model queries.
Besides certified and provable defenses \cite{CohenARXIV2019,YangARXIV2020,KumarARXIV2020b,ZhangNIPS2018,ZhangARXIV2019,WongICML2018,GowalARXIV2019,GehrSP2018,MirmanICML2018,SinghNIPS2018,LeeARXIV2019,CroceARXIV2018}, adversarial training (AT) has become the de-facto standard, as discussed in the main paper. However, there are also many detection/rejection approaches \cite{GrosseARXIV2017,FeinmanARXIV2017,LiaoCVPR2018,MaARXIV2018,AmsalegWIFS2017,MetzenARXIV2017}, so-called manifold-projection methods \cite{IlyasARXIV2017,SamangoueiICLR2018,SchottARXIV2018,ShenARXIV2017}, several methods based on pre-processing, quantization and/or dimensionality reduction \cite{BuckmanICLR2018,PrakashDCC2018,BhagojiARXIV2017}, methods based on randomness, regularization or adapted architectures \cite{ZantedschiAISEC2017,BhagojiARXIV2017,NayebiARXIV2017,SimonGabrielARXIV2018,HeinNIPS2017,JakubovitzARXIV2018,RossAAAI2018,KannanARXIV2018,LambARXIV2018,XieICLR2018} or ensemble methods \cite{LiuARXIV2017,StraussARXIV2017,HeUSENIXWORK2017,TramerICLR2018}, to name a few directions. However, often these defenses can be broken by considering adaptive attacks \cite{CarliniAISec2017,CarliniARXIV2016,AthalyeARXIV2018b,AthalyeARXIV2018}.

\textbf{Weight Robustness:} Flatness, \wrt the clean or robust loss surface, is also related to robustness in the weights. However, only few works explicitly study this ``weight robustness'': \cite{WengAAAI2020} considers robustness \wrt $L_\infty$ weight perturbations, while \cite{CheneyARXIV2017} studies Gaussian noise on weights. \cite{RakinICCV2019,HeCVPR2020}, in contrast, adversarially flip bits in (quantized) weights to reduce performance. Recently, \cite{StutzMLSYS2021} shows that robustness in weights can improve energy efficiency of neural network accelerators (\ie, specialized hardware for inference). This type of weight robustness is also relevant for some backdooring attacks that explicitly manipulate weights \cite{JiCCS2018,DumfordARXIV2018}. Fault tolerance is also a related concept, as it often involved changes in units or weights. It has been studied in early works \cite{NetiTNN1992,Chiu1994,DeodhareTNN1998}, obtaining fault tolerance using approaches similar to adversarial training NNs using approaches similar to adversarial training. However, there are also more recent works, \eg, weight dropping regularization \cite{RahmanICIP2018} or GAN-based training \cite{DudduARXIV2019}. We refer to \cite{TorreshuitzilIEEEACCESS2017} for a comprehensive survey.

%% file: supp_main.tex
\section{Visualization Details and Discussion}
\label{sec:supp-visualization}

\begin{figure}[t]
	\centering
	\vspace*{-0.2cm}
	\includegraphics[width=0.425\textwidth]{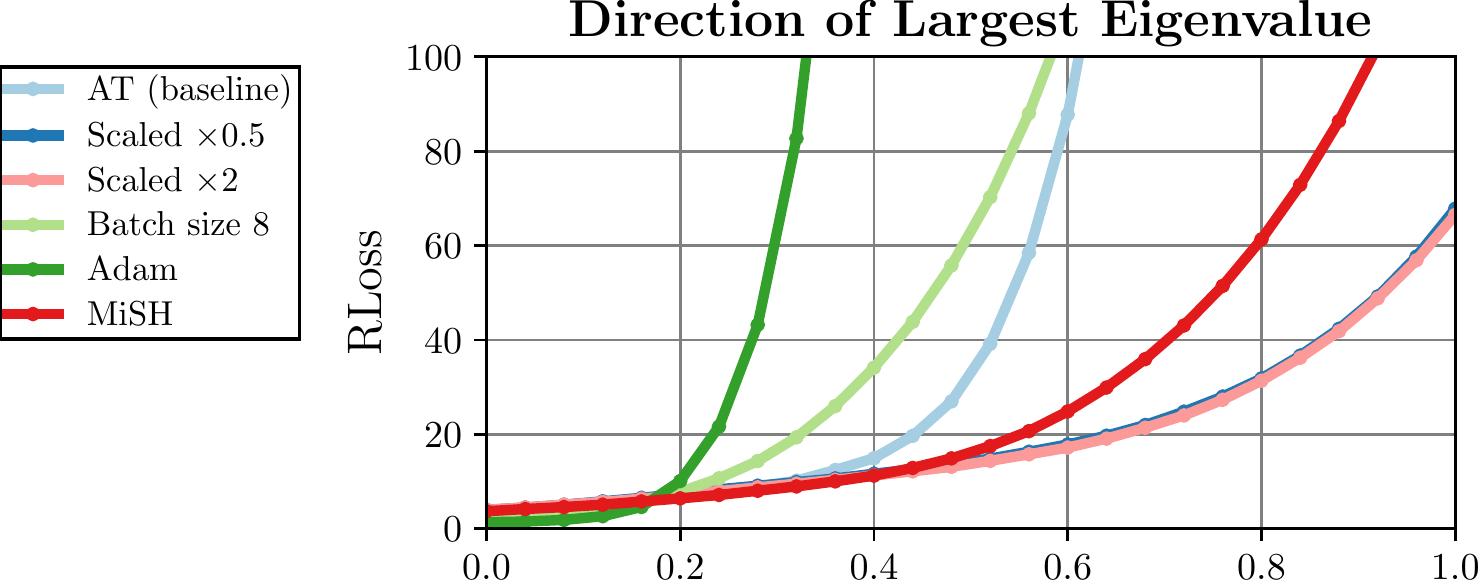}
	\vspace*{2px}
	
	\hspace*{-0.35cm}
	\includegraphics[width=0.44\textwidth]{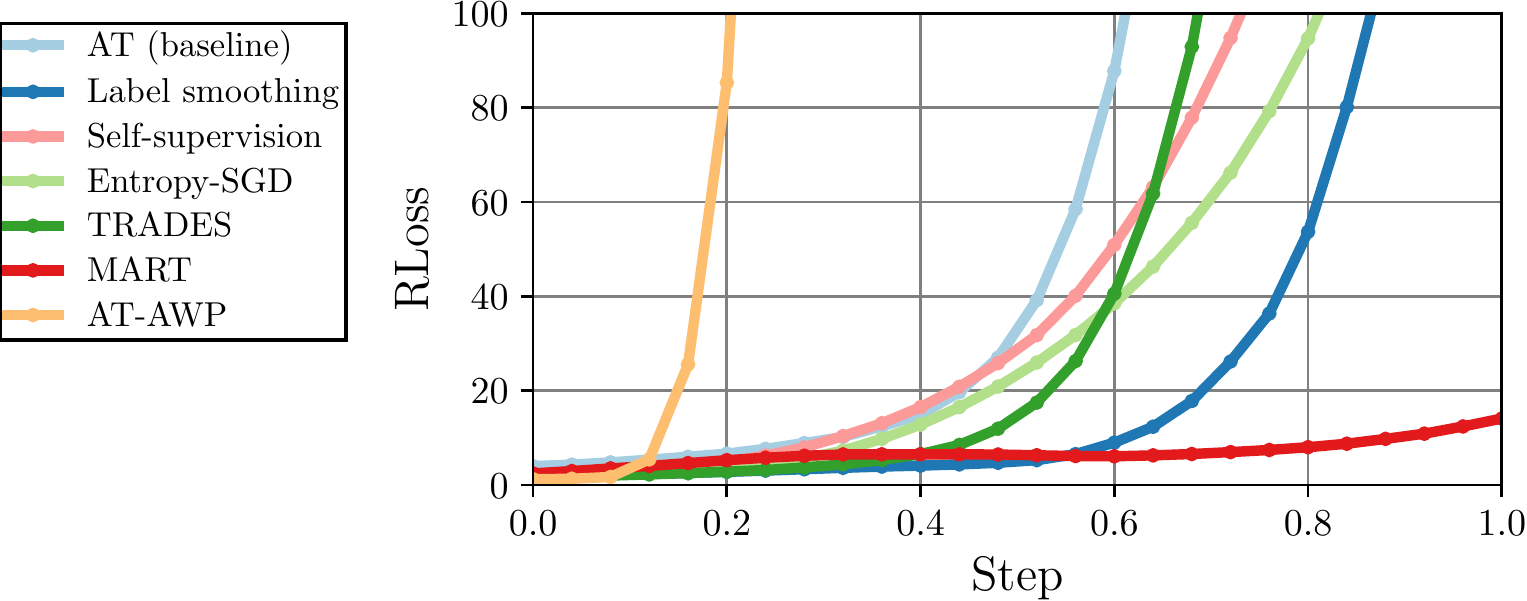}
	\vspace*{-4px}
	\caption{\textbf{Visualization in ``Hessian'' Direction:} \RCE visualized in the direction of the largest Hessian eigenvalue (\ie, the corresponding eigenvector). The eigenvalues quantify the ``rate of change'' along the corresponding eigenvector. Thus, the largest eigenvalue represents a worst-case direction in weight space. Clearly, \RCE increases significantly in these directions.}
	\label{fig:supp-visualization}
	\vspace*{-6px}
\end{figure}

\textbf{Visualization Details:}
For the plots in the main paper, we compute the mean \RCE across $10$ random, normalized directions; for adversarial directions, we plot max \RCE over $10$ adversarial directions. After normalization, we re-scale the weight directions to have length $0.5$ for random directions and $0.025$ for adversarial directions. This essentially ``zooms in'' and is particularly important when visualizing along adversarial weight directions. In all cases, we estimate \RCE on one batch of $128$ test examples for $51$ evenly spaced  step sizes in $[-1, 1]$. We found that using more test examples does not change the \RCE landscape significantly. \figref{fig:supp-visualization} shows additional visualizations along the direction of the largest Hessian eigenvalue (also using per-layer normalization, multiplied by $0.5$).

\begin{figure}[t]
	\centering
	\vspace*{-0.2cm}
	
	\includegraphics[width=0.425\textwidth]{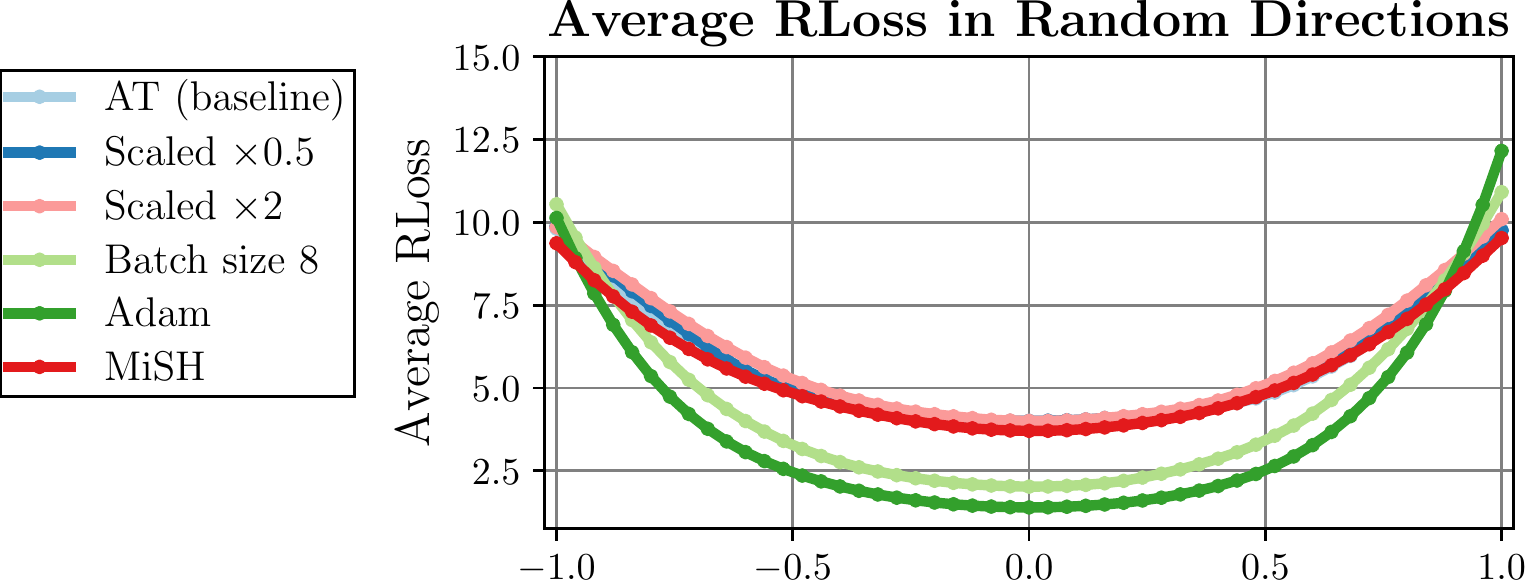}
	\vspace*{2px}
	
	\hspace*{-0.35cm}
	\includegraphics[width=0.44\textwidth]{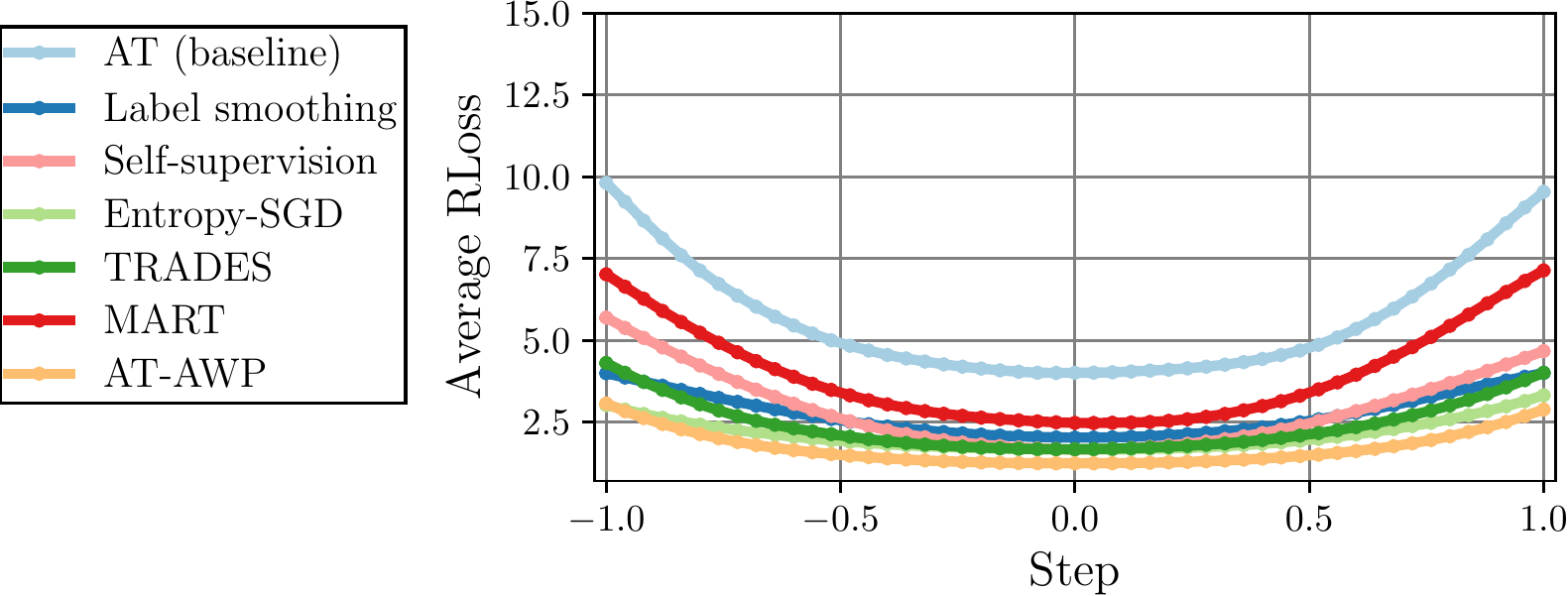}
	\vspace*{-4px}
	\caption{\textbf{Filter-Wise Normalization:} Compared to the \RCE landscape visualizations in the main paper, using per-layer normalization in \eqnref{eq:supp-normalization}, we follow \cite{LiNIPS2018} and use filter-wise normalization in \eqnref{eq:supp-normalization-filter}. Again, we plot mean \RCE across $10$ random directions. However, this does not change results significantly, flatness remains difficult to judge and compare in an objective way. Filter-wise normalization, however, ``looks'' generally flatter.}
	\label{fig:supp-visualization-li}
	\vspace*{-6px}
\end{figure}

\textbf{Discussion of \cite{LiNIPS2018}:}
Originally, \cite{LiNIPS2018} uses a per-filter normalization instead of our per-layer normalization. Specifically, this means
\begin{align}
	\hat{\nu}^{(l,i)} &= \frac{\nu^{(l)}}{\|\nu^{(l,i)}\|_2} \|w^{(l,i)}\|_2\quad\text{ for layer }l\text{, filter }i,\label{eq:supp-normalization-filter}
\end{align}
instead of our normalization outlined in the main paper:
\begin{align}
	\hat{\nu}^{(i)} &= \frac{\nu^{(l)}}{\|\nu^{(l)}\|_2} \|w^{(l)}\|_2\quad\text{ for layer }l.\label{eq:supp-normalization}
\end{align}
Furthermore, \cite{LiNIPS2018} does not consider changes in the biases or batch normalization parameters. Instead, we also normalize the biases as above and take them into account for visualization (but not the batch normalization parameters). More importantly, \cite{LiNIPS2018} considers only (clean) \CE, while we focus on \RCE. Compared to the plots from the main paper, \figref{fig:supp-visualization-li} shows that the difference between filter-wise and layer-wise normalization has little impact in visually judging flatness. Generally, filter-wise normalization makes the \RCE landscape ``look'' flatter. However, this is mainly because the absolute step size, \ie, $\|\hat{\nu}\|_2$, is smaller compared to layer-wise normalization: for our AT baseline, this is (on average) $\|\hat{\nu}\|_2 \approx 33.13$ for layer-wise and $\|\hat{\nu}\|_2 \approx 21.49$ for filter-wise normalization.

\section{Computing Flatness in \RCE}
\label{sec:supp-flatness-computation}

\begin{figure*}[t]
	\centering
	\vspace*{-0.2cm}
	\begin{minipage}[t]{0.31\textwidth}
		\includegraphics[width=\textwidth]{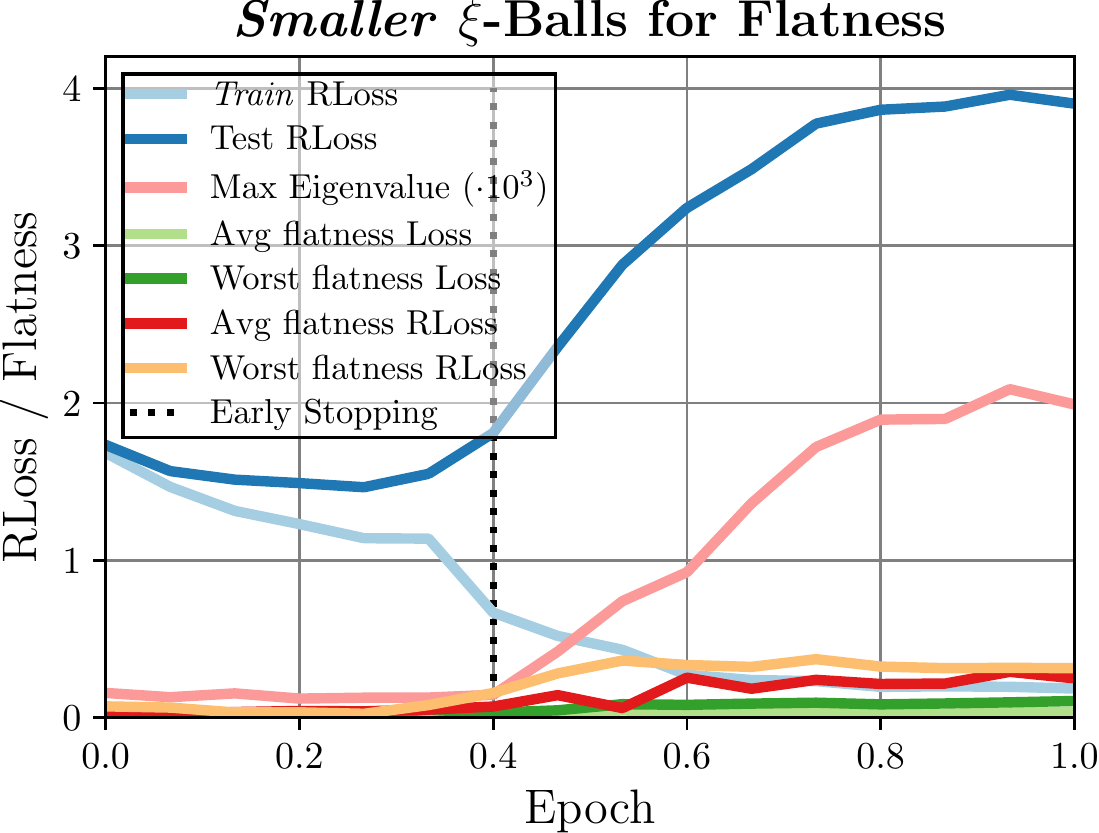}
	\end{minipage}
	\hspace*{2px}
	\begin{minipage}[t]{0.31\textwidth}
		\includegraphics[width=\textwidth]{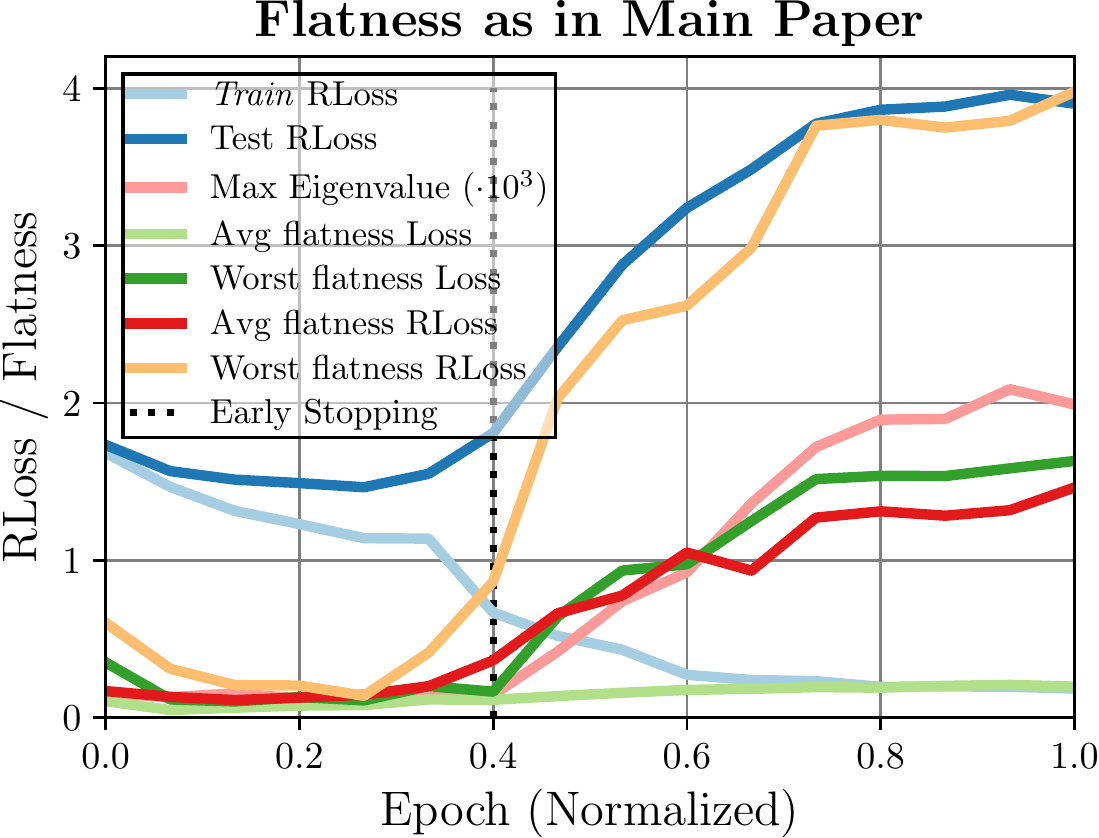}
	\end{minipage}
	\hspace*{2px}
	\begin{minipage}[t]{0.31\textwidth}
		\includegraphics[width=\textwidth]{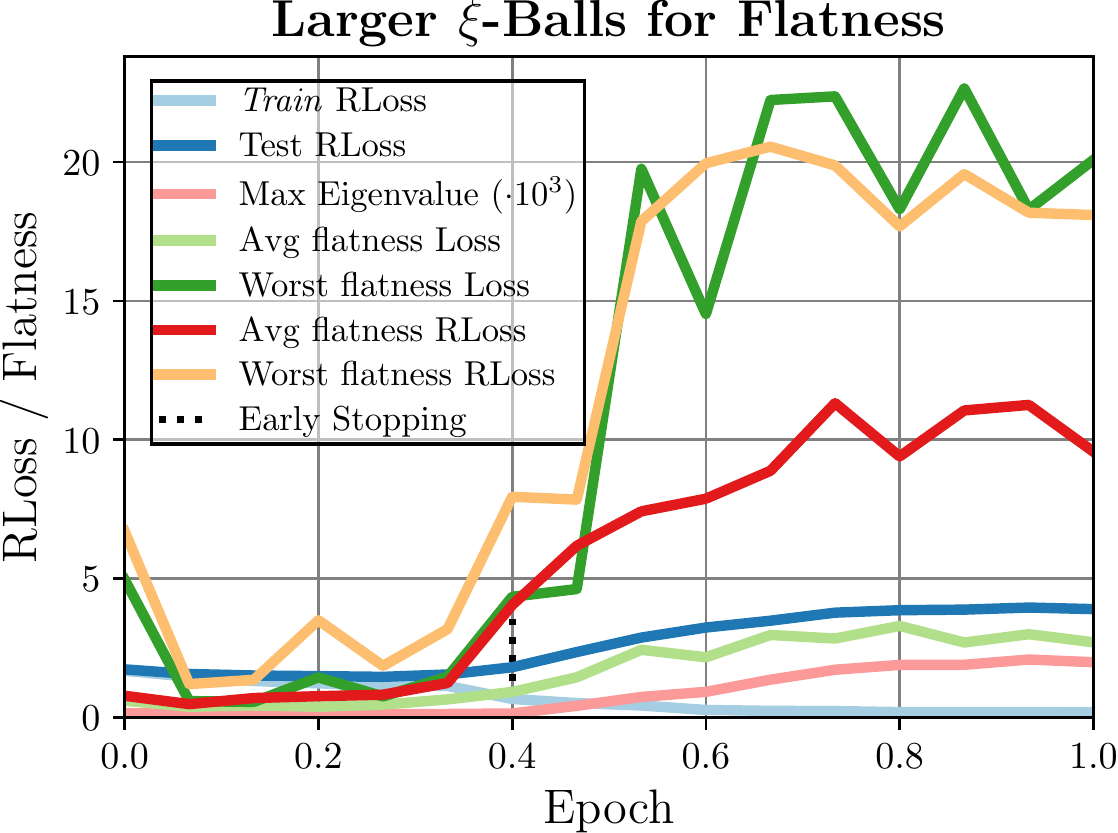}
	\end{minipage}
	\vspace*{-6px}
	\caption{\textbf{Flatness Throughout Training, Ablation:} We plot train and test \RCE, maximum Hessian eigenvalue $\lambda_{\text{max}}$, average-/worst-case flatness of (clean) \CE as well as average-/worst-case flatness on \RCE. We consider $\xi{=}0.25$/$\xi{=}0.001$ (left), $\xi{=}0.5$/$\xi{=}0.003$ (middle and main paper), and $\xi{=}0.75$/$\xi{=}0.005$ for average-/worst-case flatness, respectively. If the neighborhood $b_\xi(w)$ is chosen too small (left), increases/changes in flatness during robust overfitting are difficult to measure due to fluctuations throughout training. Chosen too large (right), in contrast, worst-case flatness (both in \CE and \RCE) quickly reaches unreasonably high loss values. This becomes problematic when comparing across models.}
	\label{fig:supp-flatness-epochs}
	\vspace*{-6px}
\end{figure*}

\textbf{Average-Case Flatness:}
The average-case flatness measure in \RCE is defined as:
\begin{align}
	\hspace*{-0.2cm}\begin{split}
		\mathbb{E}_{\nu}[\max\limits_{\|\delta\|_\infty \leq \epsilon} \mathcal{L}(f(x{+}\delta, w{+}\nu), y)]\text{\hphantom{aaaaaaa}}\\
		\text{\hphantom{aaaaaaa}}- \max\limits_{\|\delta\|_\infty \leq \epsilon} \mathcal{L}(f(x{+}\delta;w), y)
	\end{split}\label{eq:supp-average}
\end{align}
where $\mathbb{E}_{\nu}$ denotes the expectation over random weight perturbations $\nu \in B_\xi(w)$, $\mathcal{L}$ is the cross-entropy loss and $\max_{\|\delta\|_\infty \leq \epsilon}\mathcal{L}(f(x{+}\delta;w), y)$ represents the robust loss (\RCE). The first term is computed by randomly sampling $10$ weight perturbations from
\begin{align}
	B_\xi(w) = \{w + \nu : \|\nu^{(l)}\|_2 \leq \xi \|w^{(l)}\|_2 \forall\text{ layers }l\}.\label{eq:supp-ball}
\end{align}
For each weight perturbation $\nu$, the robust loss, defined as $\max_{\|\delta\|_\infty \leq \epsilon} \mathcal{L}(f(x{+}\delta, w{+}\nu), y)$, is estimated using PGD with $20$ iterations ($\epsilon = \nicefrac{8}{255}$, learning rate $0.007$ and signed gradient). This is done \emph{per-batch} (of size $128$) for the \emph{first} $1000$ test examples. Alternatively, the weights perturbations $\nu$ could also be fixed across batches (\ie, $10$ samples in total for $\lceil\nicefrac{1000}{128}\rceil$ batches). However, this is not possible for our worst-case flatness measure, as discussed next. Thus, for comparability, we sample random weight perturbations \emph{for each} batch individually. The second term is computed using PGD-$20$ with $10$ restarts, choosing the worst-case adversarial examples per test example (\ie, maximizing \RCE).

Sampling in $B_\xi(w)$ is accomplished by sampling individually per layer. That is, for each layer $l$, we compute $\xi' := \xi \cdot \|w^{(l)}\|_2$ given the original weights $w$. Then, a random vector $\nu^{(l)}$ with $\|\nu^{(l)}\|_2 \leq \xi'$ is sampled. This is done for each layer, handling weights and biases as separate layers, but ignoring batch normalization \cite{IoffeICML2015} parameters.

\textbf{Worst-Case Flatness:}
Worst-case flatness is defined as:
\begin{align}
	\begin{split}
		\max_{\nu \in B_\xi(w)} \max\limits_{\|\delta\|_\infty \leq \epsilon} &\mathcal{L}(f(x{+}\delta, w{+}\nu), y)\\
		&- \max\limits_{\|\delta\|_\infty \leq \epsilon} \mathcal{L}(f(x{+}\delta;w), y).
	\end{split}\label{eq:supp-worst}
\end{align}
Here, the expectation over $\nu$ in \eqnref{eq:supp-average} is replaced by a maximum over $\nu \in B_\xi(w)$, considering smaller $\xi$. In practice, the first term in \eqnref{eq:supp-worst} is computed by \emph{jointly} optimizing over weight perturbation $\nu$ and input perturbation(s) $\delta$ on a per-batch basis (of size $B = 128$). This means, after random initialization of $\delta_b$, $\forall b = 1,\ldots,B$, and $\nu \in B_\xi(w)$, each iteration computes and applies updates
\begin{align}
	\Delta_\nu = \nabla_\nu \sum_{b = 1}^B\mathcal{L}(f(x_b + \delta_b; w + \nu), y_b)\\
	\Delta_{\delta_b} = \nabla_{\delta_b} \sum_{b = 1}^B\mathcal{L}(f(x_b + \delta_b; w + \nu), y_b)
\end{align}
before projecting $\delta_b$ and $\nu$ onto the constraints $\|\delta_b\|_\infty \leq \epsilon$ and $\|\nu^{(l)}\|_2 \leq \xi \|w^{(l)}|_2$. The latter projection is applied in a per-layer basis, similar to sampling as described above. For the adversarial weight perturbation $\nu$, we use learning rate $0.001$, after normalizing the update $\Delta_\nu$ per-layer as in \eqnref{eq:supp-normalization}. We run $20$ iterations with $10$ restarts for each batch.

\textbf{Flatness of Clean Loss Landscape:}
We can also consider both \eqnref{eq:supp-average} and \eqnref{eq:supp-worst} on the \emph{clean} (cross-entropy) loss (``\CE''), \ie, $\mathcal{L}(f(x, w{+}\nu), y)$ instead of $\max_{\|\delta\|_\infty \leq \epsilon} \mathcal{L}(f(x{+}\delta, w{+}\nu), y)$. \red{We note that \RCE is an upper bound of (clean) \CE. Thus, flatness in \RCE and \CE are connected. However, Pearson correlation between \RCE and average-case flatness in (clean) \CE is only $0.27$, compared to $0.85$ for average-case flatness in \emph{\RCE}. This indicates that correctly measuring flatness in \RCE is crucial to empirically establish a relationship between robustness and flatness.}

\begin{figure*}[t]
	\centering
	\vspace*{-0.2cm}
	\hspace*{-1.75cm}
	\begin{minipage}[t]{0.28\textwidth}
		\vspace*{0px}
		
		\includegraphics[height=3.5cm]{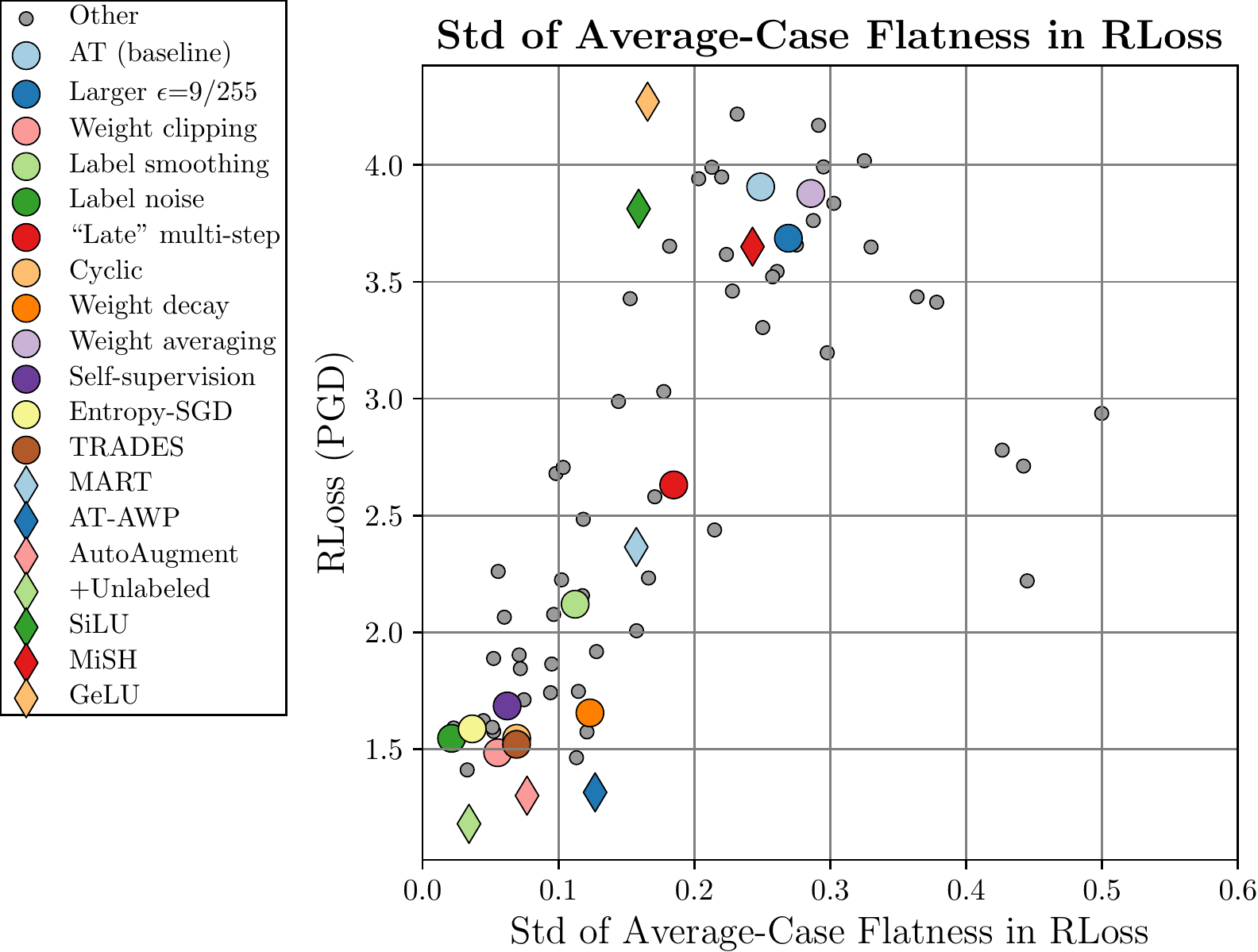}
	\end{minipage}
	\begin{minipage}[t]{0.2\textwidth}
		\vspace*{0px}
		
		\includegraphics[height=3.5cm]{plots_supp_flatness_correlation_seq_train_loss}
	\end{minipage}
	\begin{minipage}[t]{0.2\textwidth}
		\vspace*{0px}
		
		\includegraphics[height=3.5cm]{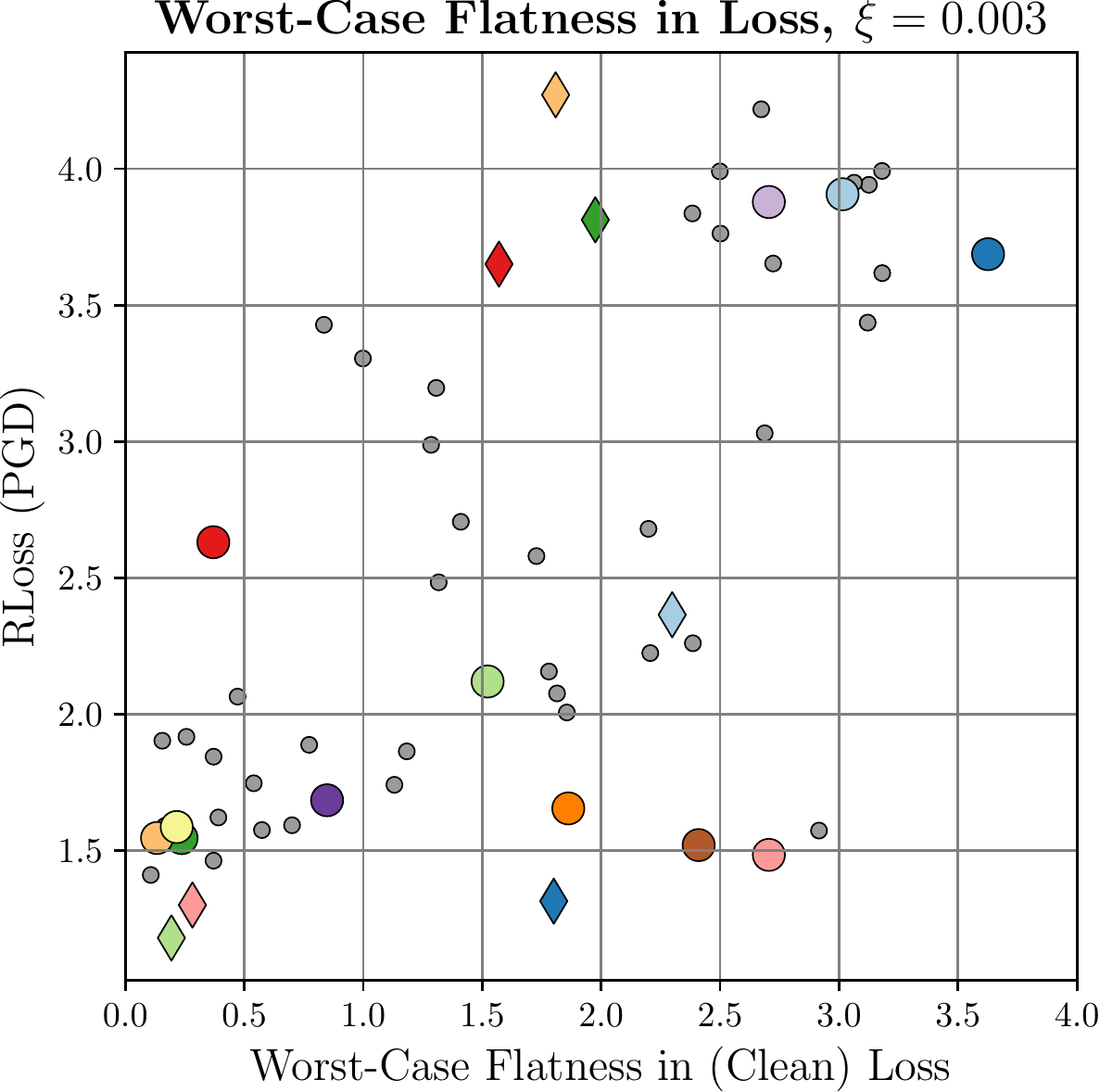}
	\end{minipage}
	\begin{minipage}[t]{0.25\textwidth}
			\vspace*{0px}
			
			\includegraphics[height=3.5cm]{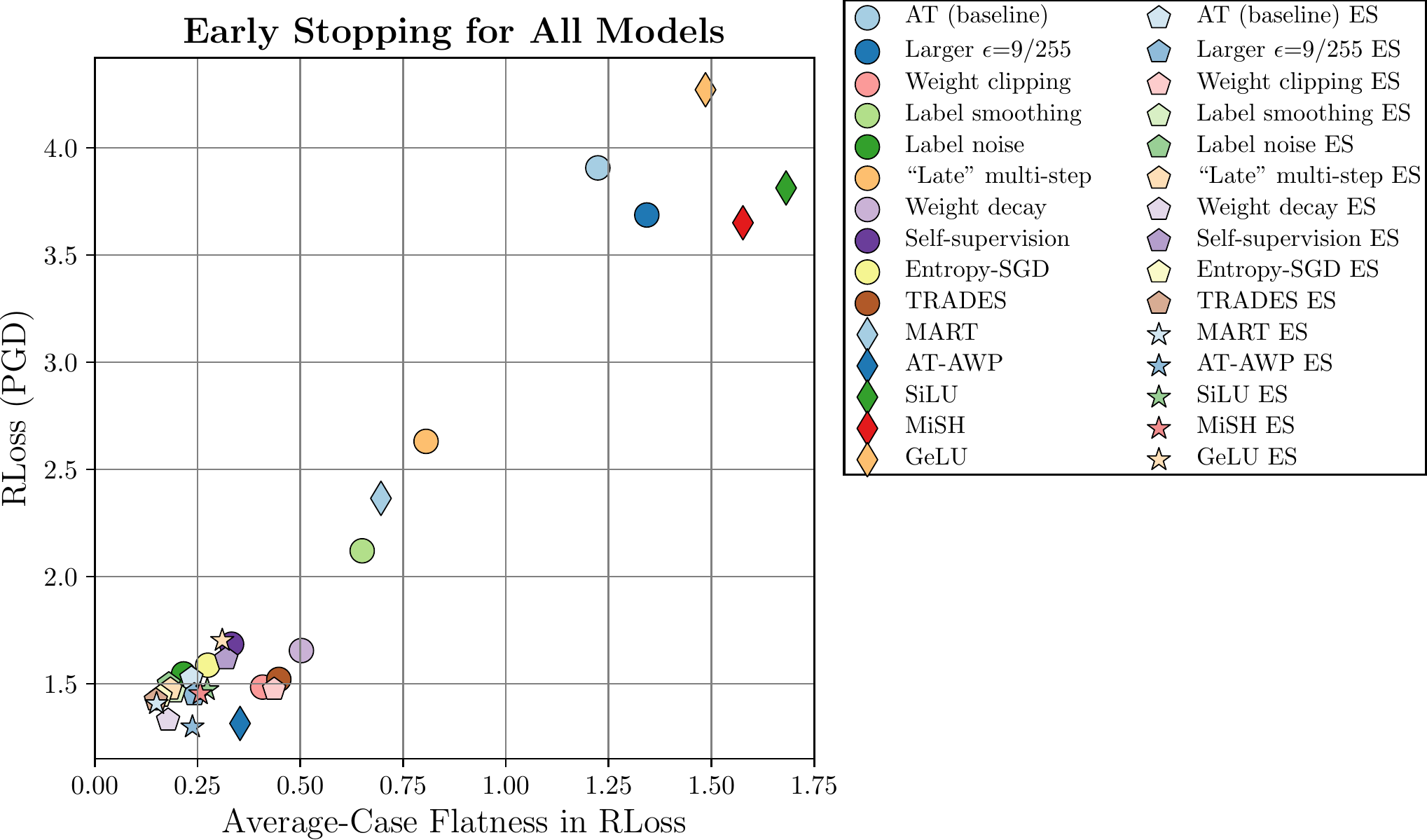}
			
			\begin{tikzpicture}[overlay,remember picture]
				\node[anchor=south west] at (1.8,0.75){\includegraphics[height=1.4cm]{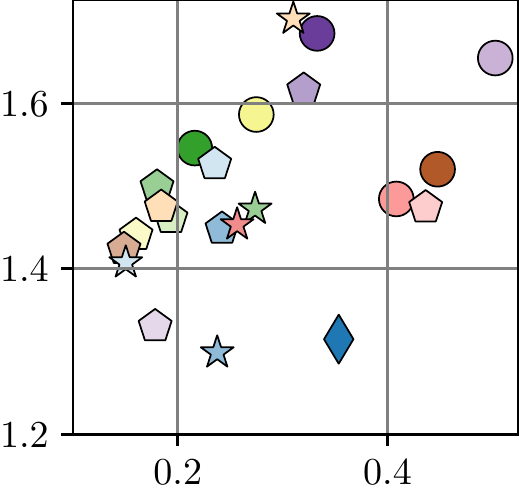}};
			\end{tikzpicture}
		\end{minipage}
	\vspace*{-18px}
	\caption{\textbf{Left: Standard Deviation of Average-Case Flatness:} We plot \RCE (y-axis) against the standard deviation (std) in our average-case flatness measure (x-axis). Note that the standard-deviation is due to the random weight perturbations $\nu$ in \eqnref{eq:supp-average}. Interestingly, more robust methods are not only flatter, but our average-case flatness measure also has lower standard deviation. \textbf{Middle Left: Average-Case Flatness of Train \RCE:} \emph{Test} \RCE plotted against our average-case flatness measure as computed on \emph{training} examples. Even on the training set, flatness is predictive of robust generalization, \ie, adversarial robustness on the test set. The relationship, however, is weaker compared to average-case flatness on test examples. \textbf{Middle Right: Worst-Case Flatness in (Clean) \CE:} As worst-case flatness in the \emph{clean} \CE landscape also mirrors robust overfitting in \figref{fig:supp-flatness-epochs}, we plot \RCE against worst-case flatness in \CE. Even though flatness is measured considering clean \CE, many methods improving robustness (\ie, lower \RCE) exhibit surprisingly good flatness. \textbf{Right: Early Stopping for all Models:} \RCE vs\onedot average-case flatness for all models where early stopping improves adversarial robustness. For example, this is not the case for AutoAugment or AT with unlabeled examples. Across all models, early stopping improves both robustness and flatness. For clarity we provide a zoomed-in plot for the lower left corner.}
	\label{fig:supp-flatness-misc}
\end{figure*}

\subsection{Ablation for Flatness Measures}
\label{sec:supp-flatness-ablation}
 
\textbf{Flatness Throughout Training:}
\figref{fig:supp-flatness-epochs} shows average- and worst-case flatness on both clean as well as robust loss (\CE and \RCE) throughout training of our AT baseline. We consider different sizes of the neighborhood $B_\xi(w)$ for computing our flatness measures: $\xi{=}0.25$/$\xi{=}0.001$ (left), $\xi{=}0.5$/$\xi{=}0.003$ (middle, as in main paper), and $\xi{=}0.75$/$\xi{=}0.005$ for average-/worst-case flatness, respectively. While average-case flatness of \emph{clean} \CE does \emph{not} mirror robust overfitting very well, its worst-case pendant increases during overfitting, even though \RCE is \emph{not} taken into account. Furthermore, if the neighborhood $B_\xi(w)$ is chosen too small, the flatness measures are not sensitive enough to be discriminative (\cf left). \figref{fig:supp-flatness-epochs} also shows that, throughout training of one model, the largest Hessian eigenvalue mirrors robust overfitting. Overall, this means that early stopping essentially improves adversarial robustness by finding flatter minima. This is confirmed in \figref{fig:supp-flatness-misc} (right), showing that early stopping consistently improves robustness and flatness.

\textbf{Standard Deviation in Average-Case Flatness:} 
In \figref{fig:supp-flatness-misc} (left), the x-axis plots the standard deviation in our average-case flatness measure (in \RCE). Note that the standard deviation originates in the random samples $\nu$ used to calculate \eqnref{eq:supp-average}. First of all, standard deviation tends to be small (\ie, $\leq 0.3$) across almost all models. This means that our findings in the main paper, \ie, the strong correlation between flatness and \RCE, is supported by low standard deviation. More importantly, the standard deviation \emph{reduces} for particularly robust methods.

\textbf{Average-Case Flatness on Training Examples:}
\figref{fig:supp-flatness-misc} (middle left) shows that average-case flatness in \RCE is also predictive for robust generalization when computed on \emph{training} examples. However, the correlation between (test) \RCE and (train) flatness is less clear, \ie, there is a larger ``spread'' across methods. Here, we use the first $1000$ training examples to compute average-case flatness.

\textbf{Worst-Case Flatness on \emph{Clean} \CE:}
In \figref{fig:supp-flatness-epochs}, worst-case flatness on clean \CE also correlates with robust overfitting. Thus, in \figref{fig:supp-flatness-misc} (middle right), we plot \RCE against worst-case flatness of \CE, showing that there is no clear relationship across models. Nevertheless, many methods improving adversarial robustness also result in flatter minima in the clean loss landscape. This is sensible as \RCE is generally an upper bound for (clean) \CE. On the other hand, flatness in \CE is \emph{not} discriminative enough to clearly distinguish between robust and less robust models.

\begin{figure*}[t]
	\centering
	\vspace*{-0.2cm}
	\begin{minipage}[t]{0.3\textwidth}
		\vspace*{0px}
		
		\includegraphics[height=3.6cm]{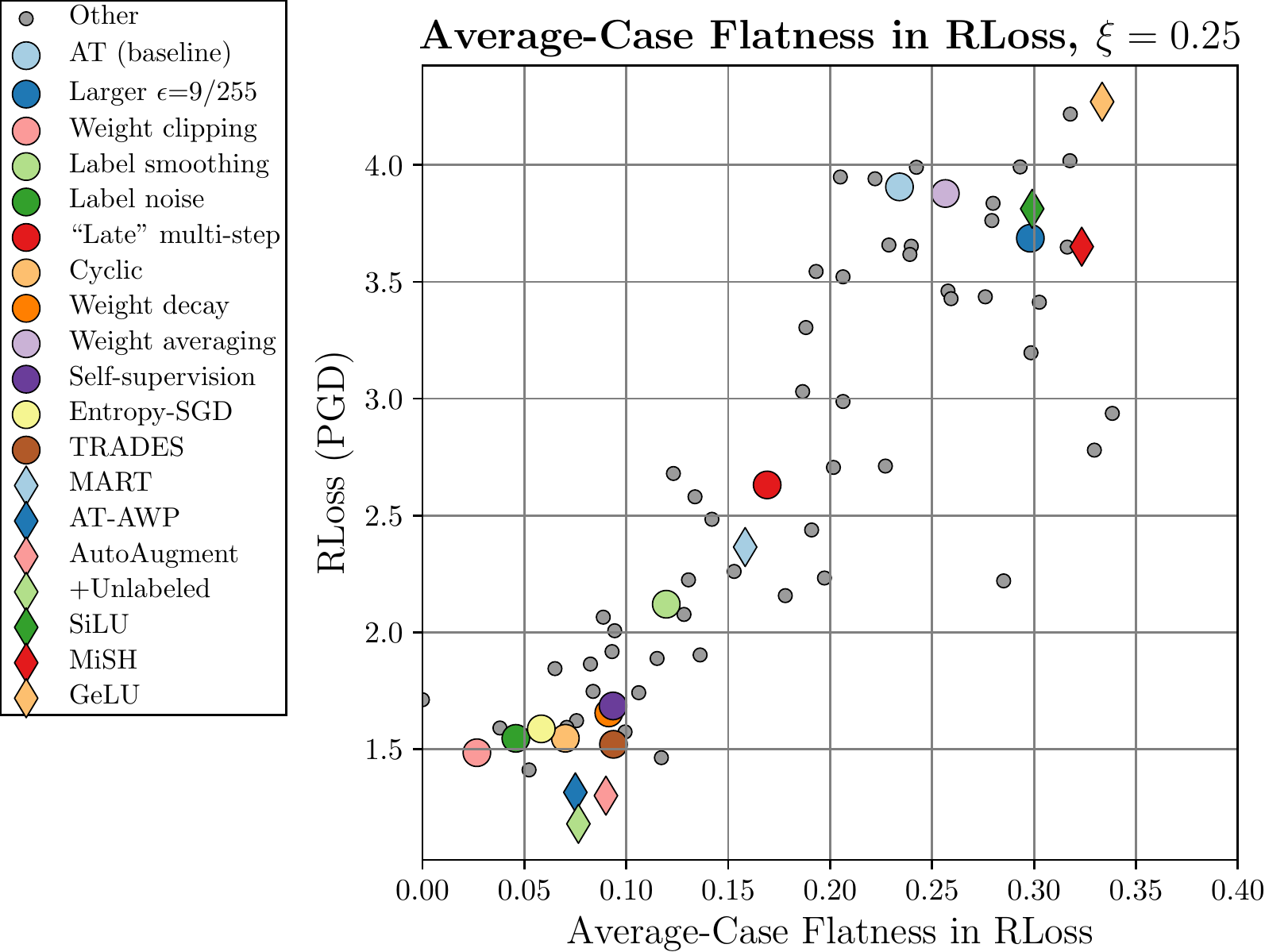}
	\end{minipage}
	\begin{minipage}[t]{0.22\textwidth}
		\vspace*{0px}
		
		\includegraphics[height=3.6cm]{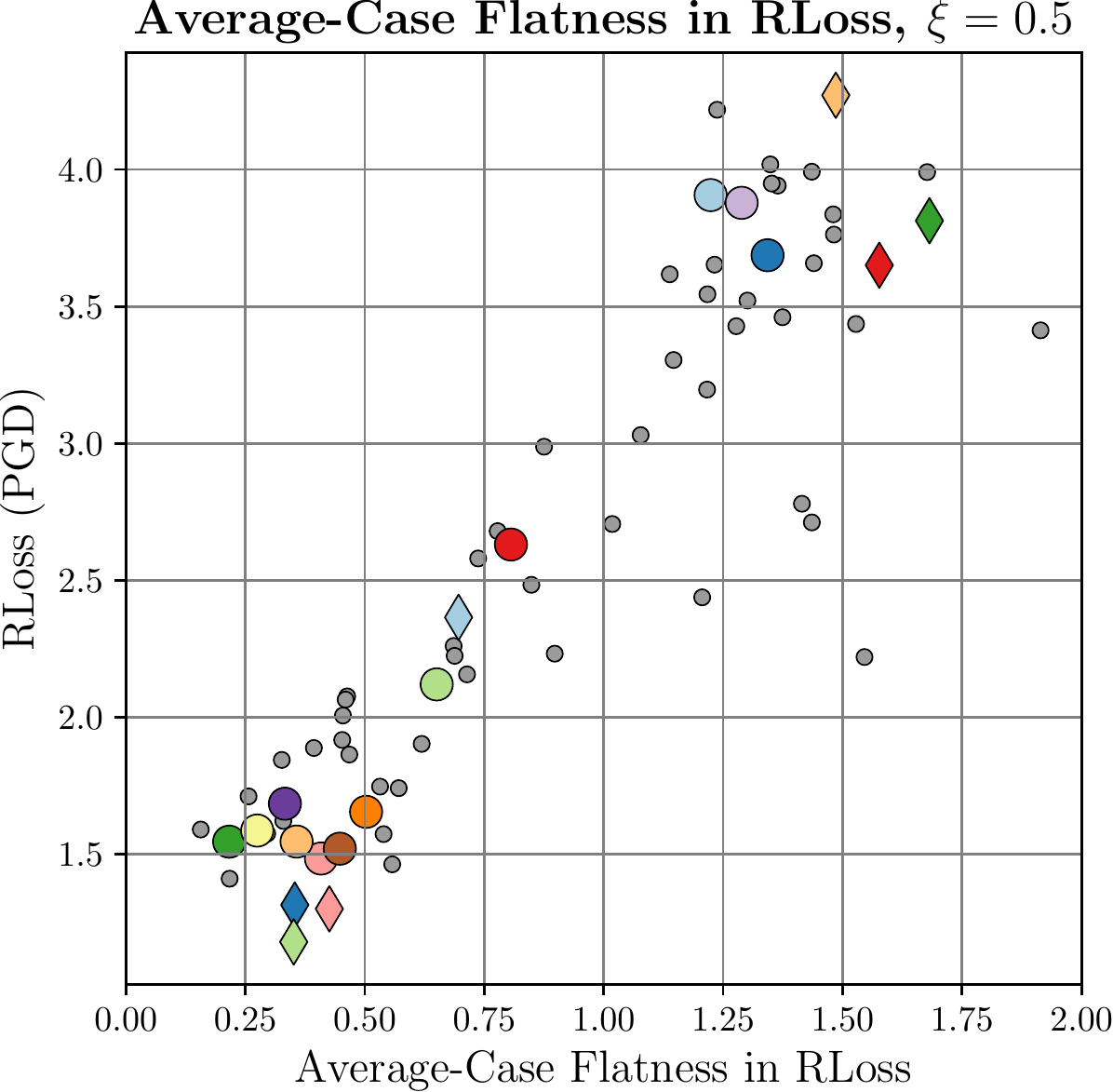}
	\end{minipage}
	\begin{minipage}[t]{0.22\textwidth}
		\vspace*{0px}
		
		\includegraphics[height=3.6cm]{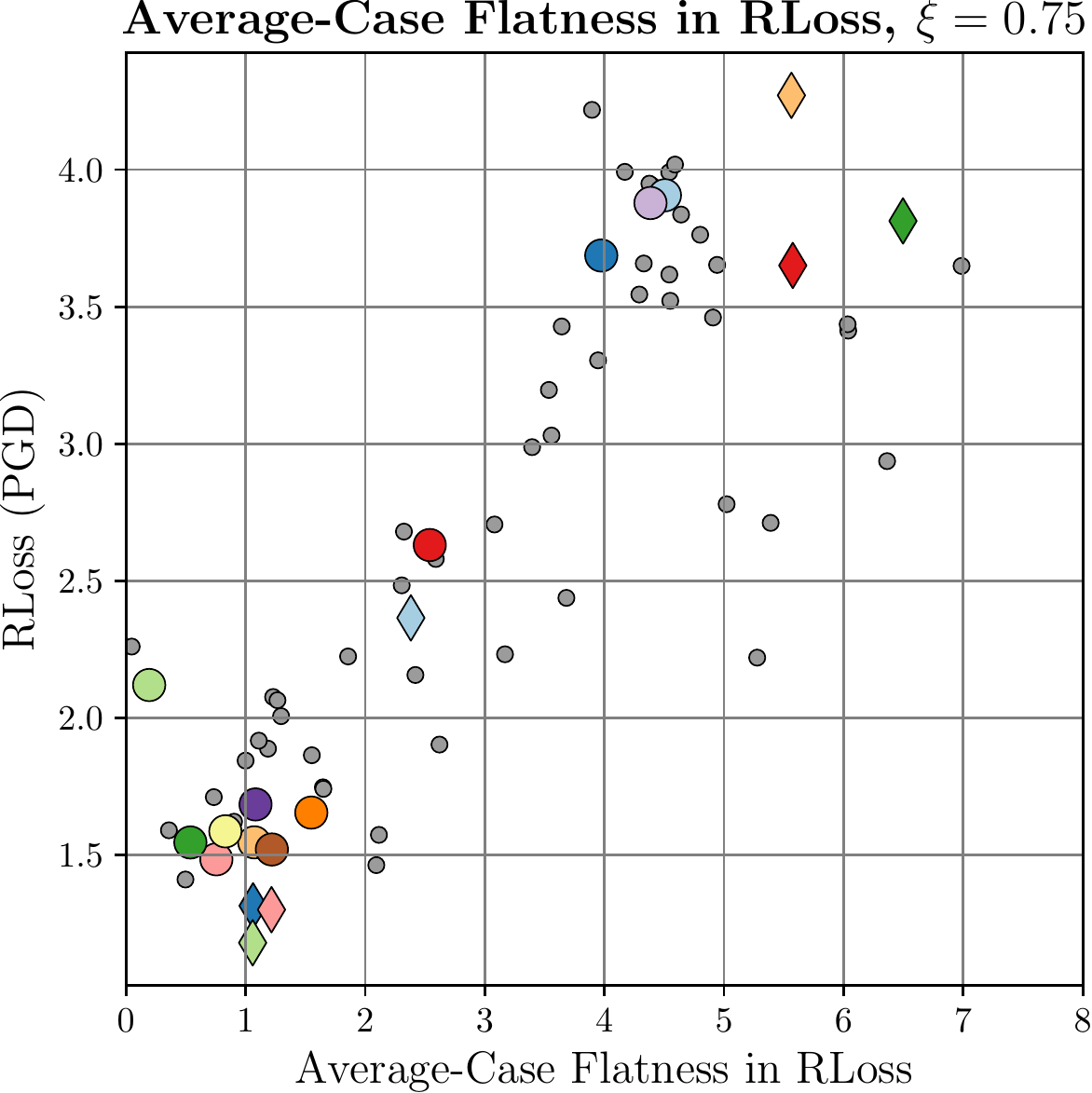}
	\end{minipage}
	\begin{minipage}[t]{0.22\textwidth}
		\vspace*{0px}
		
		\includegraphics[height=3.6cm]{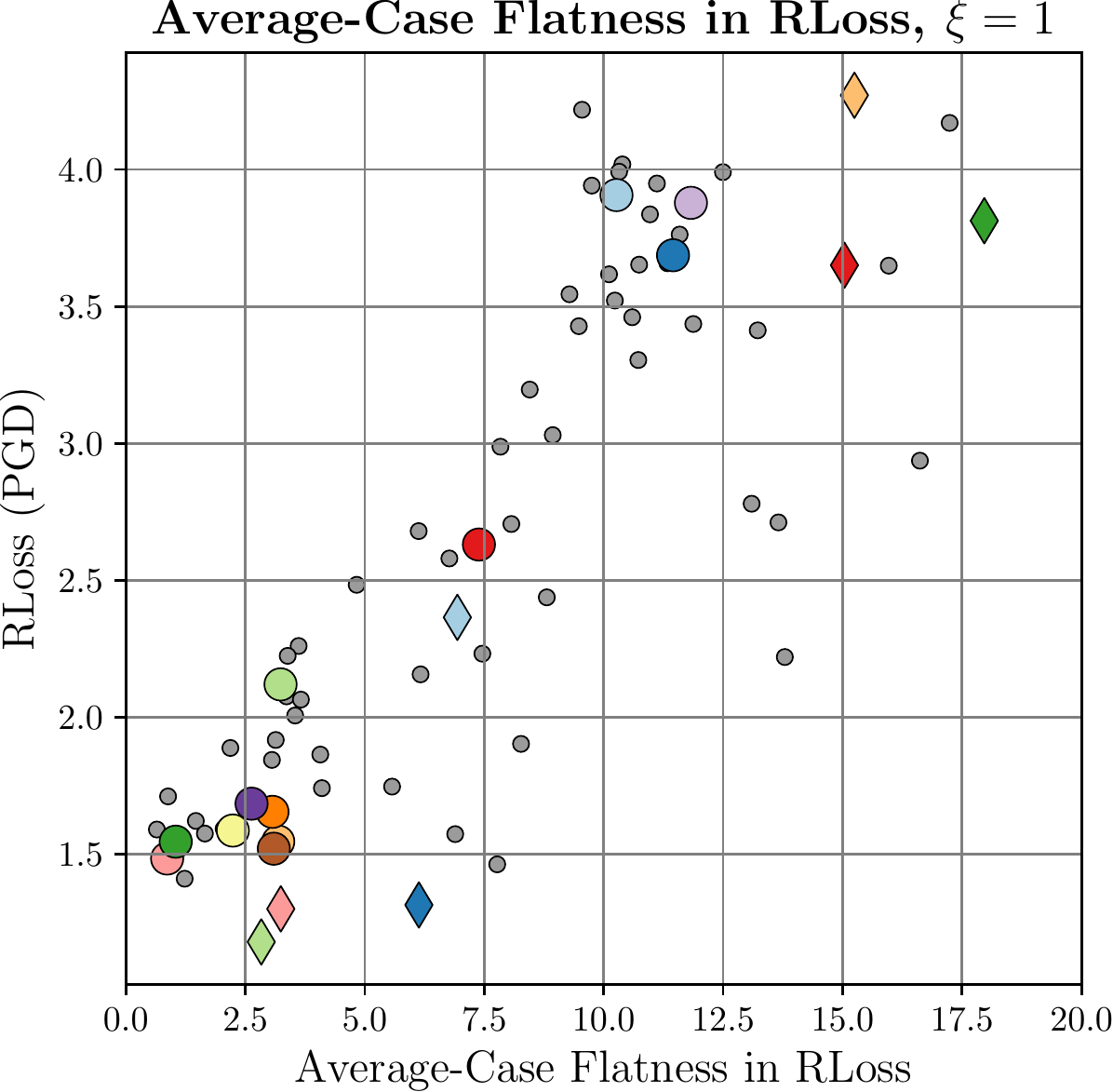}
	\end{minipage}
	\\[6px]
	
	{\color{black!75}\rule{\textwidth}{0.65px}}
	\\[-6px]
	
	\begin{minipage}[t]{0.3\textwidth}
		\vspace*{0px}
		
		\includegraphics[height=3.6cm]{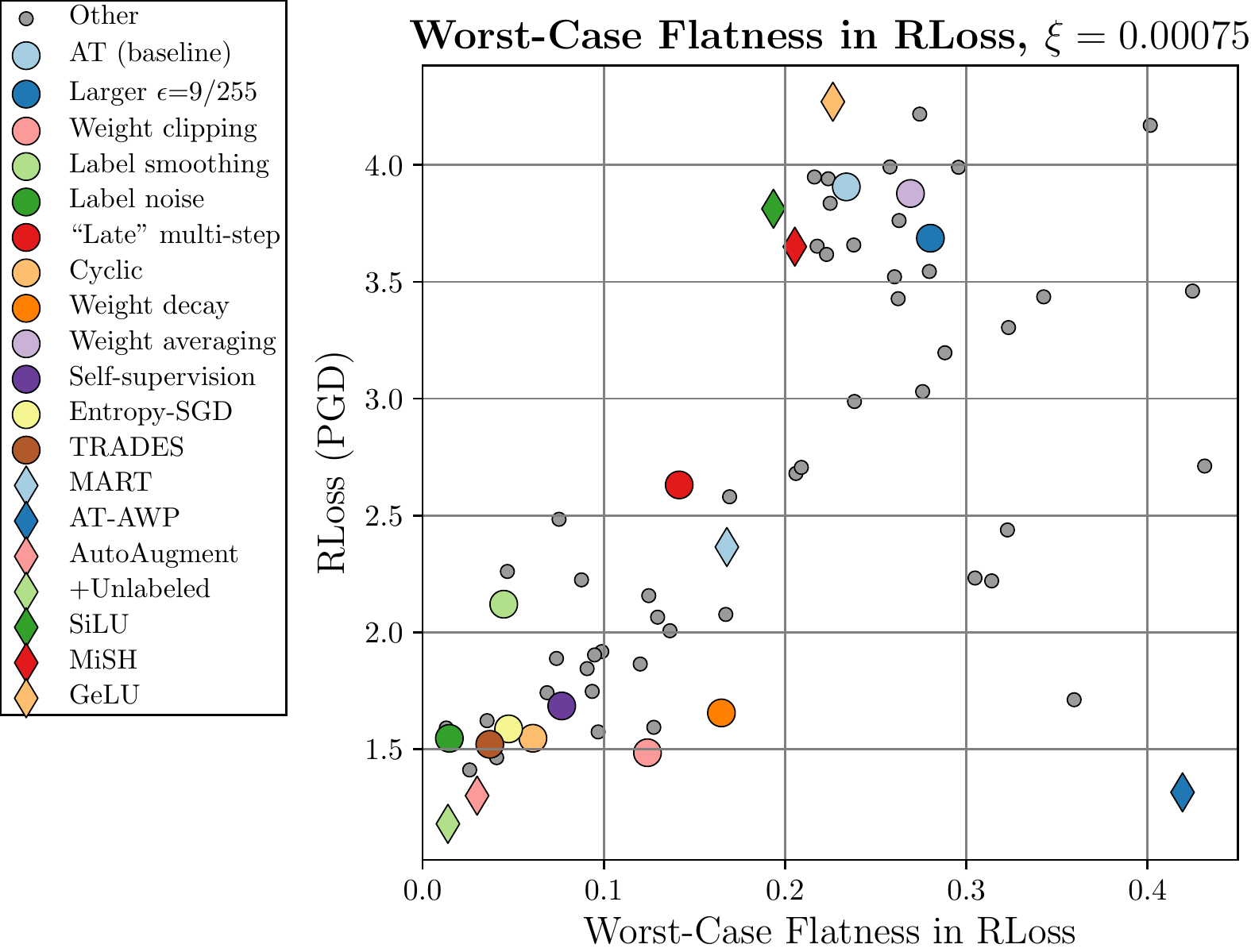}
	\end{minipage}
	\begin{minipage}[t]{0.22\textwidth}
		\vspace*{0px}
		
		\includegraphics[height=3.6cm]{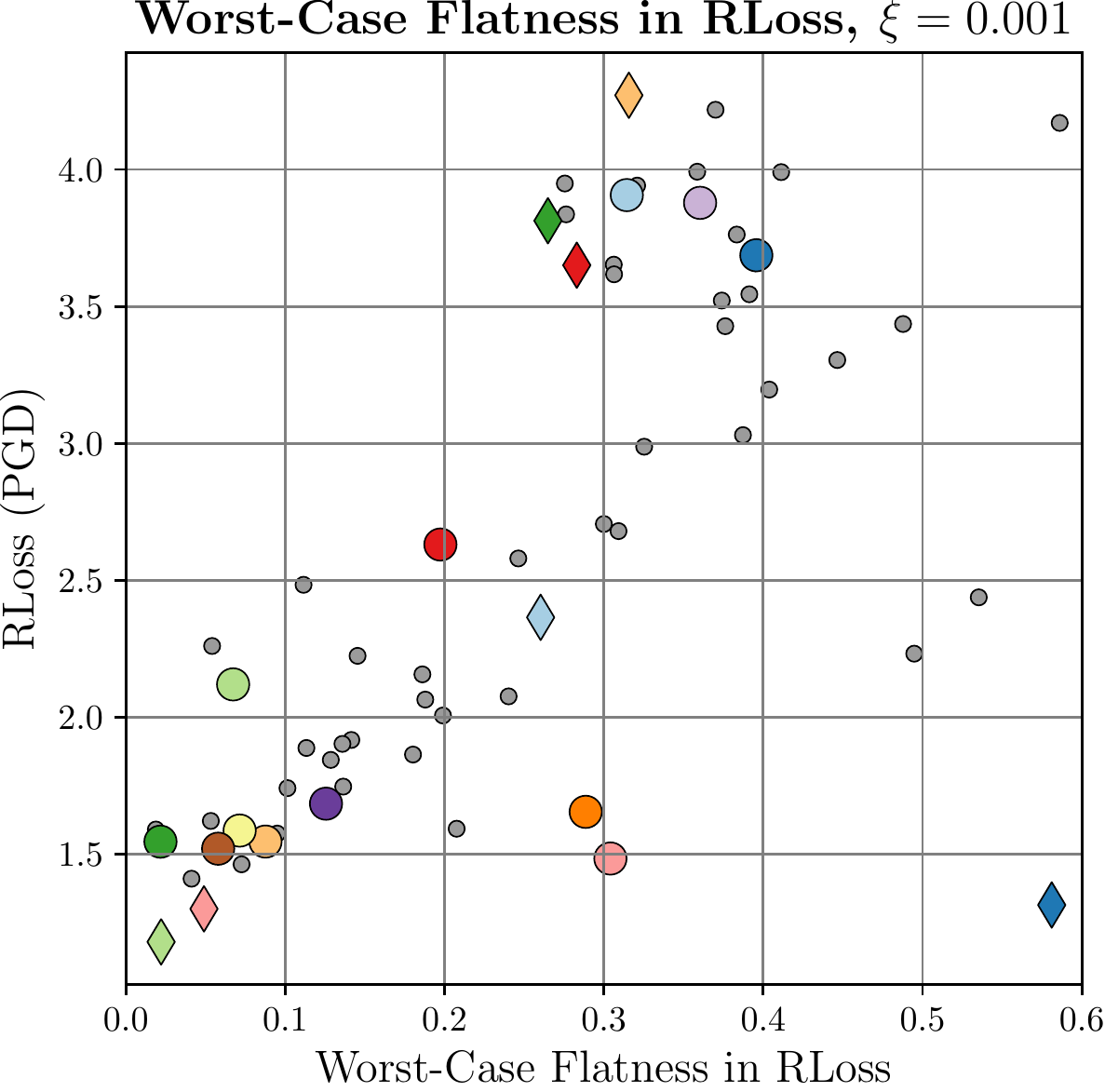}
	\end{minipage}
	\begin{minipage}[t]{0.22\textwidth}
		\vspace*{0px}
		
		\includegraphics[height=3.6cm]{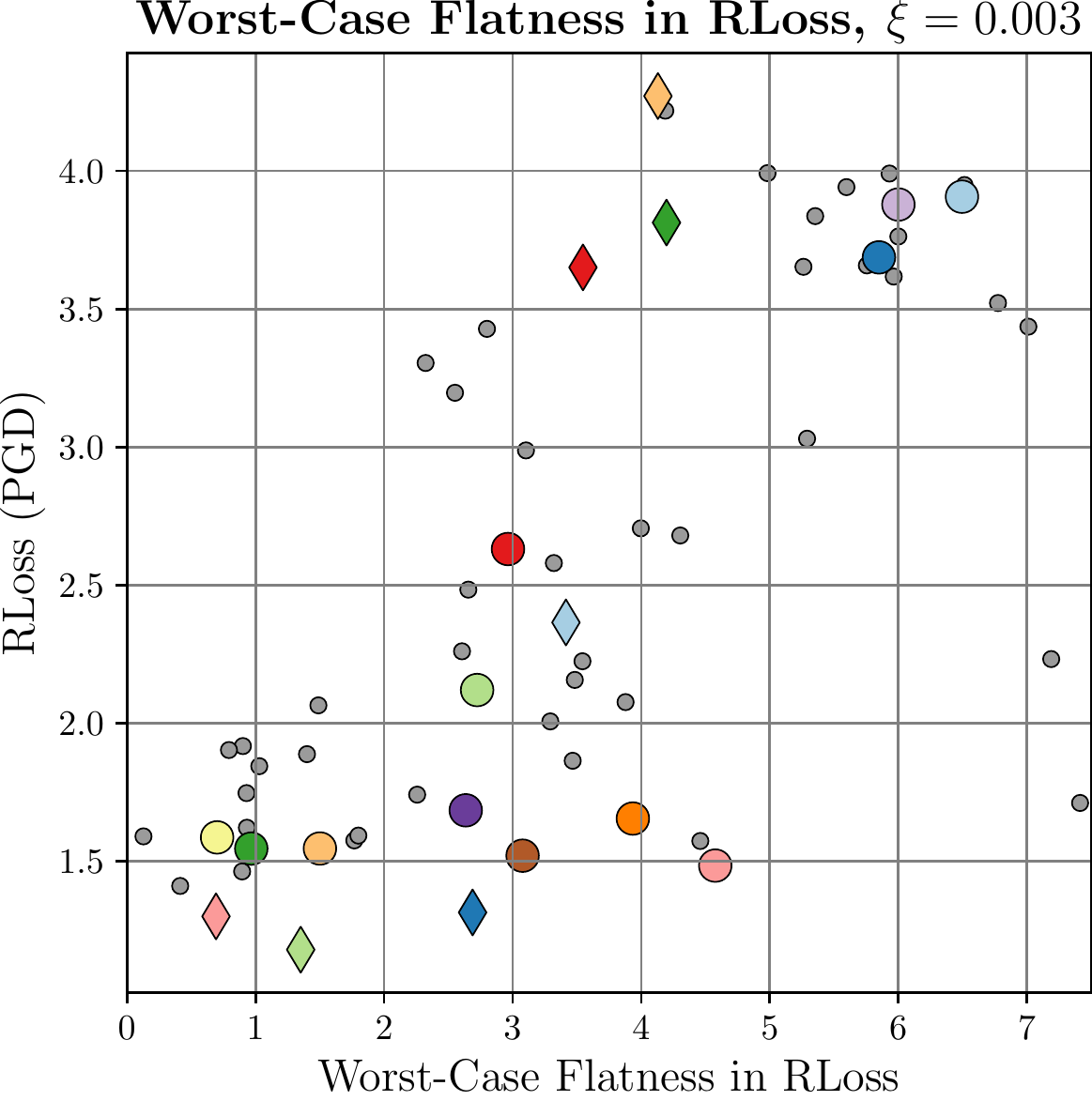}
	\end{minipage}
	\begin{minipage}[t]{0.22\textwidth}
		\vspace*{0px}
		
		\includegraphics[height=3.6cm]{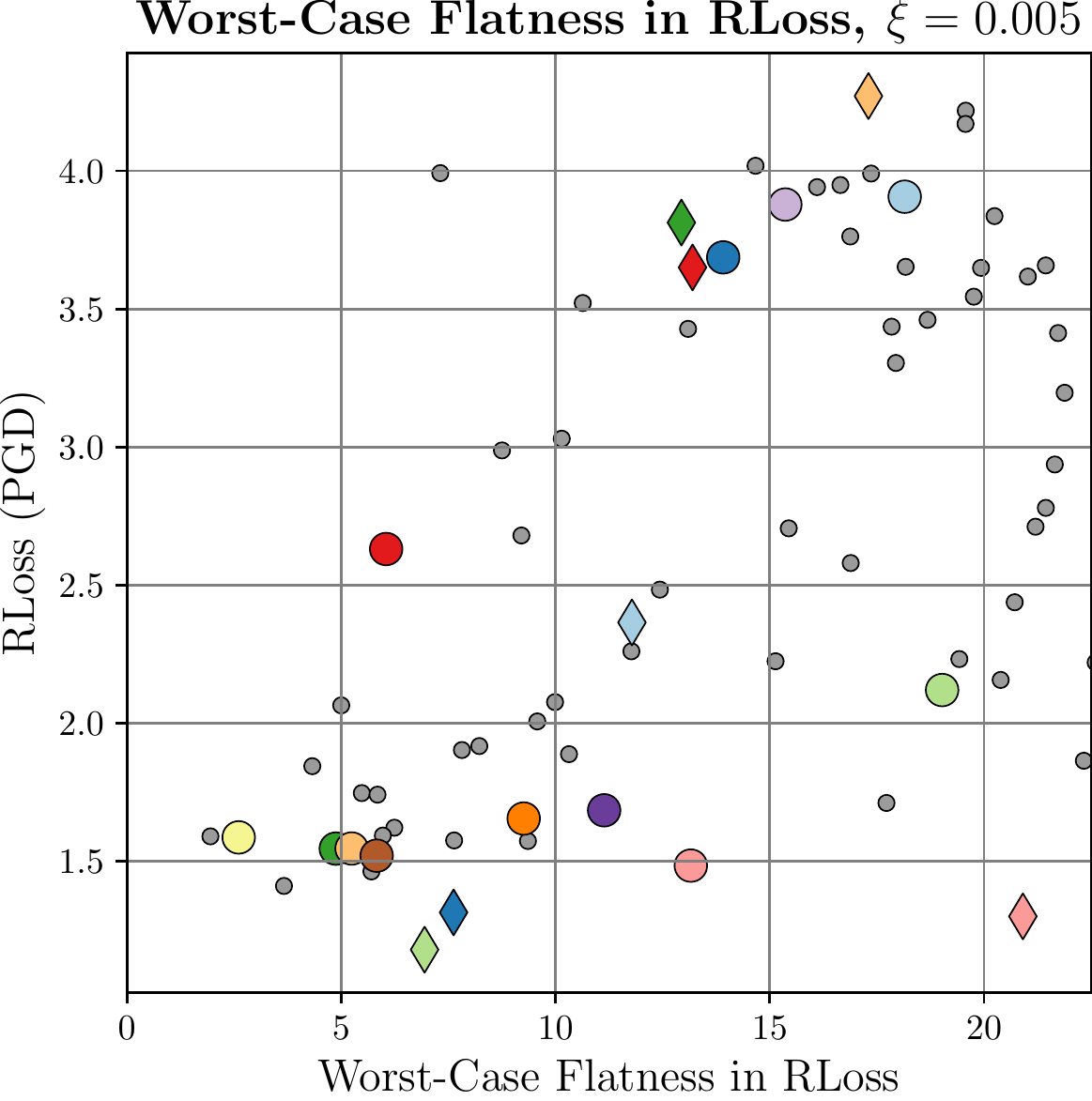}
	\end{minipage}
	\caption{\textbf{Flatness in \RCE, Ablation for $B_\xi(w)$:} \RCE(y-axis) plotted against average-case (top) and worst-case (bottom) flatness in \RCE (x-axis).
	\textbf{Top:} We consider $\xi \in \{0.25, 0.5, 0.75, 1\}$ for average-case flatness. The clear relationship between adversarial robustness, \ie, low \RCE, and flatness shown for $\xi = 0.5$ in the main paper can be reproduced for all cases. \textbf{Bottom:} For worst-case flatness, we consider $\xi \in \{0.00075, 0.001, 0.003, 0.005\}$. When chosen too large, \eg, $\xi = 0.005$, however, variance seems to increase, making the relationship less clear. For small $\xi$, \eg, $\xi = 0.00075$, the correlation between robustness and flatness is pronounced, except for a few outliers, including AT-AWP \cite{WuNIPS2020}.}
	\label{fig:supp-flatness-ablation}
\end{figure*}

\textbf{Ablation for $B_\xi(w)$:}
For computing our average- and worst-case flatness measures (in \RCE), we considered various sizes of neighborhoods in weight space, \ie $B_\xi(w)$ from \eqnref{eq:supp-ball} for different $\xi$. \figref{fig:supp-flatness-ablation} considers $\xi \in \{0.25, 0.5, 0.75, 1\}$ for average-case flatness (top) and $\xi \in \{0.00075, 0.001, 0.003, 0.005\}$ for worst-case flatness (bottom). In both cases, we plot \RCE (y-axis) against flatness in \RCE (y-axis), as known from the main paper. Average-case flatness using small $\xi = 0.25$ results in significantly smaller values, between $0$ and $0.4$, \ie, the increase in \RCE in random weight directions is rather small. Still, the relationship between adversarial robustness and flatness is clearly visible. The same holds for larger $\xi \in \{0.75, 1\}$. Worst-case flatness generally gives a less clear picture regarding the relationship between robustness and flatness.
Additionally, for larger $\xi \in \{0.003, 0.005\}$, variance seems to increase such that this relationship becomes less pronounced. In contrast to average-case flatness, the variance is not induced by the $10$ restarts used for \eqnref{eq:supp-worst}, but caused by training itself. Indeed, re-training our AT baseline leads to a worst-case flatness in \RCE of $5.1$, a significant reduction from $6.49$ as obtained for our original baseline. Overall, however, the observations from the main paper can be confirmed using different sizes of the neighborhood $B_\xi(w)$.

\section{Scaling Networks and Scale-Invariance}
\label{sec:supp-scaling}

\begin{figure*}[t]
	\centering
	\vspace*{-0.2cm}
	\begin{minipage}[t]{0.26\textwidth}
		\vspace*{0px}
		\includegraphics[width=\textwidth]{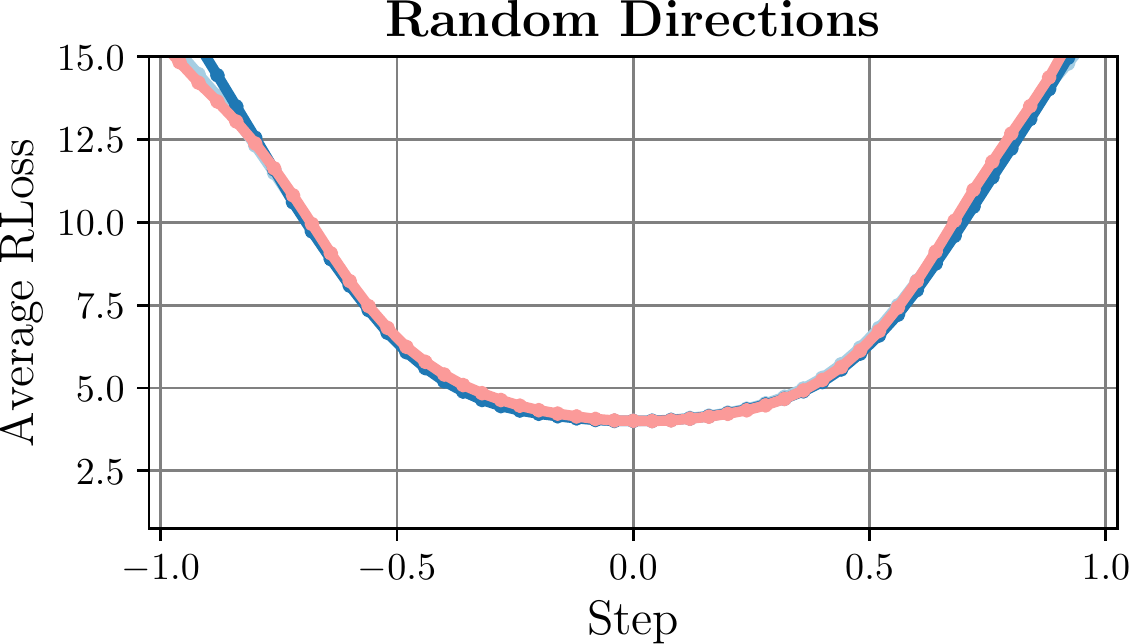}
	\end{minipage}
	\begin{minipage}[t]{0.155\textwidth}
		\vspace*{0px}
		\includegraphics[width=\textwidth]{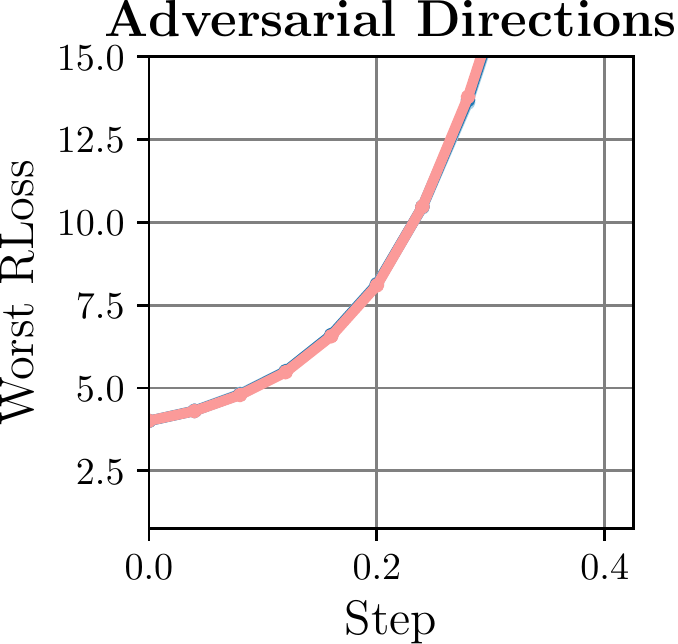}
	\end{minipage}
	\hspace*{0.25cm}
	\begin{minipage}[t]{0.38\textwidth}
		\vspace*{0px}
		\scriptsize
		{
			\setlength{\tabcolsep}{0pt}
		    \newcolumntype{C}[1]{@{}>{\centering\arraybackslash}p{#1}@{}}
			\begin{tabularx}{\textwidth}{|X|C{0.85cm}|C{1.5cm}|C{0.85cm}|C{0.85cm}|C{0.85cm}|}
				\hline
				\hspace*{2px} Model & \multicolumn{2}{c|}{\bfseries Robustness} & \multicolumn{3}{c|}{\bfseries Flatness}\\
				\hline
				& \RTE $\downarrow$ & \RTE $\downarrow$ & Worst & Avg & Worst\\
				& (test) & (train) & \emph{Loss} & \textbf{RLoss} & \textbf{RLoss}\\
				\hline
				\hline		
				\hspace*{2px} Scaled $\times0.5$ & 60.9 & 8.4 (-52.5) &  0.86 & 1.36 & 6.50\\
				\hspace*{2px} AT (baseline) & 61.0 & 8.4 (-52.6) & 0.86 & 1.21 & 6.48\\
				\hspace*{2px} Scaled $\times2$ & 61.0 & 8.3 (-52.7) & 0.86 & 1.27 & 6.49\\
				\hline
			\end{tabularx}
		}
	\end{minipage}
	\vspace*{-6px}
	\caption{\textbf{Flatness and Scale-Invariance.} \textbf{Left:} We plot average \RCE and worst \RCE along random and adversarial directions, as discussed in \secref{sec:supp-visualization}, for AT and its scaled variants, $\times0.5$ and $\times2$. Clearly, \RCE landscape looks nearly identical. \textbf{Right:} Robustness against PGD-$20$ on train and test examples, as well as average- and worst-case flatness measures on \RCE. For completeness, we also include worst-case flatness on clean \CE. All of these measures are nearly invariant to scaling. The shown differences can be attributed to randomness in computing these measures.}
	\label{fig:supp-scaling}

\end{figure*}

\textbf{Scale-Invariance:}
In the main paper, we presented a simple experiment to show that our measures of flatness in \RCE are scale-invariant: we scaled weights \emph{and} biases of \emph{all} convolutional layers in our adversarially trained ResNet-18 \cite{HeCVPR2016} by factor $s \in \{0.5, 2\}$. Note that all convolutional layers in the ResNet are followed by batch normalization layers \cite{IoffeICML2015}. Thus, the effect of scaling is essentially ``canceled out'', \ie, these convolutional layers are scale-invariant. Thus, the prediction stays roughly constant. \figref{fig:supp-scaling} (left) shows \RCE landscape visualizations for AT and its scaled variants in random and adversarial weight directions. Clearly, scaling AT has negligible impact on the \RCE landscape in both cases. \figref{fig:supp-scaling} (right) shows that our flatness measures remain invariant, as well. As $B_\xi(w)$ in \eqnref{eq:supp-ball} is defined \emph{per-layer} (weights and biases separately) and \emph{relative} to $w$, the neighborhood increases alongside the weights, rendering visualization and flatness measures invariant.
\red{When, for example, scaling up specific layers and scaling down others, as discussed in \cite{DinhICML2017}, causes the neighborhood $B_\xi(w)$ to increase or decrease in size for these particular layers. Thus, following \cite{DinhICML2017}, scaling up the first layer of a two-layer ReLU network by $\alpha$ and scaling down the second layer by $\nicefrac{1}{\alpha}$ (keeping the output constant), has no effect in terms of measuring flatness as the per-layer neighborhood $B_\xi(w)$ is scaled accordingly, as well.}
The Hessian eigenspectrum, in contrast, scales with the models, \cf \tabref{tab:supp-convexity}, and is not suited to quantify flatness.

\textbf{Convexity and Flatness:}
\tabref{tab:supp-convexity} also presents the convexity metric introduced in \cite{LiNIPS2018}: $\nicefrac{|\lambda_{\text{min}}|}{|\lambda_{\text{max}}|}$ with $\lambda_{\text{min/max}}$ being largest/smallest Hessian eigenvalue. Note that the Hessian is computed following \cite{LiNIPS2018} \wrt to the \emph{clean} (cross-entropy) loss, not taking into account adversarial examples. The intuition is that negative eigenvalues with large absolute value correspond to non-convex directions in weight space. If these eigenvalues are large in relation to the positive eigenvalues, there is assumed to be significant non-convexity ``around'' the found minimum. \tabref{tab:supp-convexity} shows that this fraction is usually very small, as also found in \cite{LiNIPS2018}. However, \tabref{tab:supp-convexity} also shows that this convexity measure is not clearly correlated with adversarial robustness.

\section{Detailed Experimental Setup}
\label{sec:supp-setup}

We focus our experiments on \CifarT \cite{Krizhevsky2009}, consisting of $50\text{k}$ training examples and $10\text{k}$ test examples of size $32\times32$ (in color) and $K = 10$ class labels. We use all training examples during training, but withhold the \emph{last} $500$ test examples for early stopping. Evaluation is performed on the \emph{first} $1000$ test examples, due to long runtimes of AutoAttack \cite{CroceARXIV2020} and our flatness measures (on \RCE). Any evaluation on the training set is performed on the first $1000$ training examples (\eg, in \figref{fig:supp-flatness-misc}, middle).

As network architecture, we use ResNet-18 \cite{HeCVPR2016} with batch normalization \cite{IoffeICML2015} and ReLU activations. Our AT baseline (\ie, default model) is trained using SGD for $150$ epochs, batch size $128$, learning rate $0.05$, reduced by factor $0.1$ at $60$, $90$ and $120$ epochs, weight decay $0.005$ and momentum $0.9$. We save snapshots every $5$ epochs to perform early stopping, but do \emph{not} use early stopping by default. We whiten input examples by subtracting the (per-channel) mean and dividing by standard deviation. We use standard data augmentation, considering random flips and cropping (by up to $4$ pixels per side). By default, we use $7$ iterations PGD, with learning rate $0.007$, signed gradient and $\epsilon = \nicefrac{8}{255}$ to compute $L_\infty$ adversarial examples. Note that no momentum \cite{DongCVPR2018} or backtracking \cite{StutzICML2020} is used for PGD. The training curves in \figref{fig:supp-training-curves} correspond to robustness measured using the $7$-iterations PGD attack used for training, which we also use for early stopping (with $5$ random restarts).

For evaluation, we run PGD for $20$ iterations and $10$ random restarts, taking the worst-case adversarial example per test example \cite{StutzICML2020}. Our results considering robust loss (\RCE) are based on PGD, while we report robust test error (\RTE) using AutoAttack \cite{CroceARXIV2020}. \red{Note that AutoAttack does \emph{not} maximize cross-entropy loss as it stops when adversarial examples are found. Thus, it is not suitable to estimate \RCE.} Robust test error is calculated as the fraction of test examples that are either mis-classified or successfully attacked. The distinction between PGD-$20$ and AutoAttack is important as AutoAttack does \emph{not} maximize cross-entropy loss, resulting in an under-estimation of \RCE, while PGD-$20$ generally underestimates \RTE. Computation of our average- and worst-case flatness measure is detailed in \secref{sec:supp-flatness-computation}.

Everything is implemented in PyTorch~\cite{PaszkeNIPSWORK2017}.

\section{Methods}
\label{sec:supp-methods}

In the following, we briefly elaborate on the individual methods considered in our experiments.

\textbf{Learning Rate Schedules:}
Besides our default, multi-step learning rate schedule (learning rate $0.05$, reduced by factor $0.1$ after epochs $60$, $90$, and $120$), we followed \cite{PangARXIV2020b} and implemented the following learning rate schedules: First, simply using a constant learning rate of $0.05$. Second, only two ``late'' learning rate reductions at epochs $140$ and $145$, as done in \cite{QinNIPS2019}. Third, using a cyclic learning rate, interpolating between a learning rate of $0.2$ and $0$ for $30$ epochs per cycle, as, \eg, done in \cite{WongARXIV2020}. We consider training for up to $4$ cycles ($=120$ epochs). These learning rate schedules are available as part of PyTorch \cite{PaszkeNIPSWORK2017}.

\textbf{Label Smoothing:}
In \cite{SzegedyCVPR2016}, label smoothing is introduced as regularization to improve (clean) generalization by \emph{not} enforcing one-hot labels in the cross-entropy loss. Instead, for label $y$ and $K = 10$ classes, a target distribution $p \in [0,1]^K$ (subject to $\sum_i p_i = 1$) with $p_y = 1 - \tau$ (correct label) and $p_i = \nicefrac{\tau}{K - 1}$ for $i \neq y$ (all other labels) is enforced. During AT, we only apply label smoothing for the weight update, not for PGD. We consider $\tau \in \{0.1, 0.2, 0.3\}$.

\textbf{Label Noise:}
Instead of explicitly enforcing a ``smoothed'' target distribution, we also consider injecting label noise during training. In each batch, we sample random labels for a fraction of $\tau$ of the examples. Note that the labels are sampled uniformly across all $K = 10$ classes. Thus, in expectation, the enforced target distribution is $p_y = 1 - \tau + \nicefrac{\tau}{K}$ and $p_i = \nicefrac{\tau - \nicefrac{\tau}{K}}{K - 1}$. As result, this is equivalent to label smoothing with $\tau = \tau - \nicefrac{\tau}{K}$. In contrast to label smoothing, this distribution is not enforced explicitly in the cross-entropy loss. As above, adversarial examples are computed against the true labels (without label noise) and label noise is injected for the weight update. We consider $\tau \in \{0.1, 0.2, 0.3, 0.4, 0.5\}$. While label smoothing does not further improve adversarial robustness for $\tau > 0.3$, label noise proved very effective in avoiding robust overfitting, which is why we also consider $\tau = 0.4$ or $0.5$.

\textbf{Weight Averaging:}
To implement weight averaging \cite{IzmailovUAI2018}, we follow \cite{GowalARXIV2020} and keep a ``running'' average $\bar{w}$ of the model's weights throughout training, updated in each iteration $t$ as follows: 
\begin{align}
	\bar{w}^{(t)} = \tau \bar{w}^{(t - 1)} + (1 - \tau) w^{(t)}
\end{align}
where $w^{(t)}$ are the weights in iteration $t$ \emph{after} the gradient update. Weight averaging is motivated by finding the weights $\bar{w}$ in the center of the found local minimum. As, depending on the learning rate, training tends to oscillate, the average of the iterates is assumed to be close to the actual center of the minimum. In our experiments, we consider $\tau \in \{0.98, 0.985, 0.99, 0.9975\}$.

\textbf{Weight Clipping:}
Following \cite{StutzMLSYS2021}, we implement weight clipping by clipping the weights to $[-w_{\text{max}}, w_{\text{max}}]$ after each training iteration. We found that $w_{\text{max}}$ can be chosen as small as $0.005$, which we found to work particularly well. Larger $w_{\text{max}}$ does \emph{not} have significant impact on adversarial robustness for AT. \cite{StutzMLSYS2021} argues that weight clipping together with minimizing cross-entropy loss leads to more redundant weights, improving robustness to random weight perturbations. As result, we also expect weight clipping to improve flatness. We consider $w_{\text{max}} \in \{0.005, 0.01, 0.025\}$.

\begin{figure}[t]
	\centering
	\vspace*{-0.2cm}
	\includegraphics[width=0.425\textwidth]{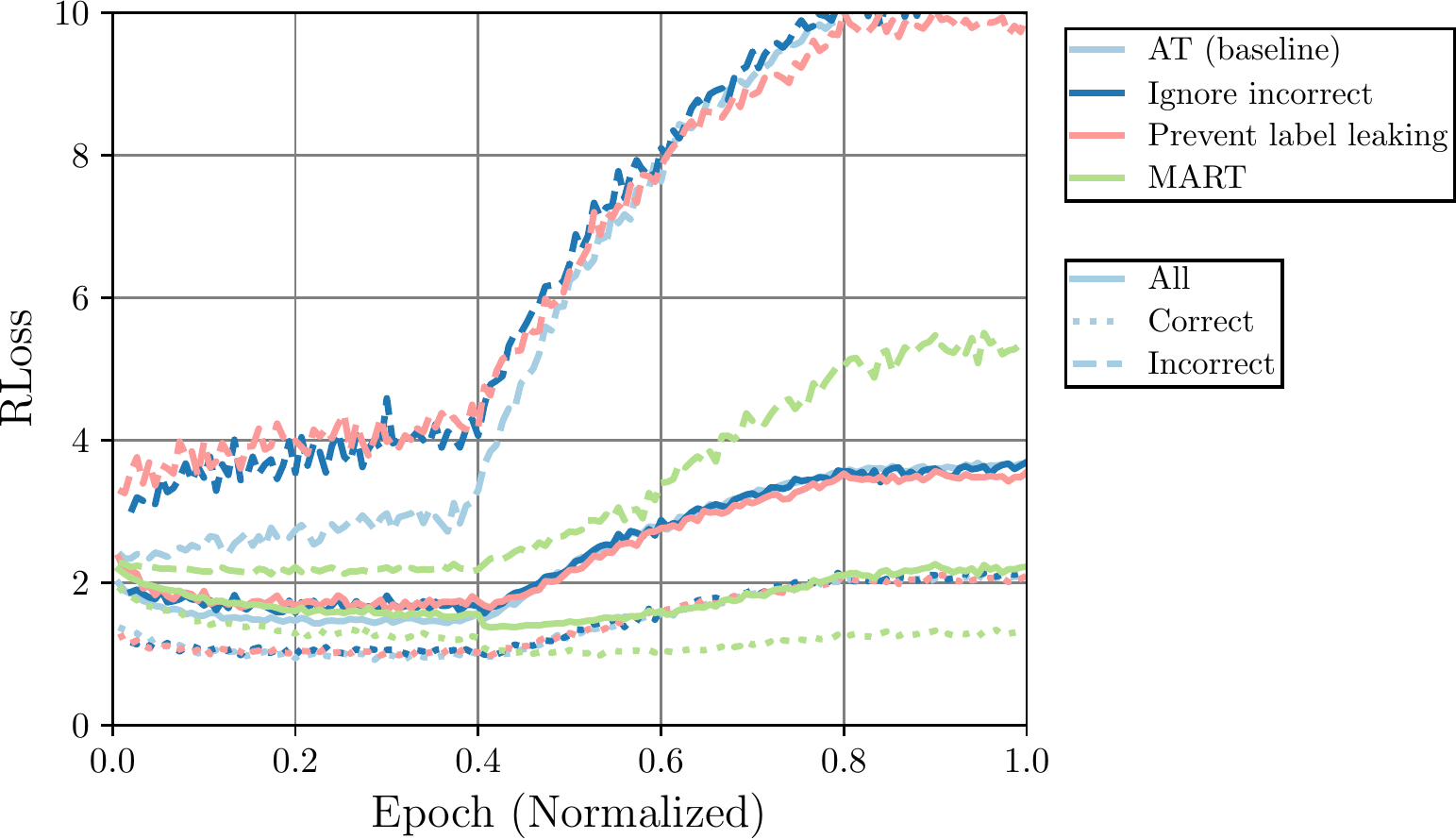}
	\vspace*{-6px}
	\caption{\textbf{Approaches of Handling Incorrect Examples:} We plot test \RCE on all (solid), correctly classified (dotted) and incorrectly classified (dashed) examples throughout training. We consider our AT baseline ({\color{plot0}light blue}), ignoring incorrectly classified training examples in the \RCE computation during training ({\color{plot1}dark blue}) and preventing label leaking by computing adversarial examples against the \emph{predicted} labels during training ({\color{plot2}rose}). However, these ``simple'' approaches of tackling the high \RCE on incorrectly classified test examples are not successful in reducing robust overfitting. As outlined in the main paper, MART \cite{WangICLR2020} ({\color{plot3}green}) is able to dampen overfitting through an additional robust KL-loss weighted by confidence, see text.}
	\label{fig:supp-ii-pll}
	\vspace*{-6px}
\end{figure}

\textbf{Ignoring Incorrect Examples \& Preventing Label Leaking:}
As robust overfitting in AT leads to large \RCE on incorrectly classified test examples, we investigate whether (a) \emph{not} computing adversarial examples on incorrectly classified examples (during training) or (b) computing adversarial examples against the predicted (not true) label (during training) helps to mitigate robust overfitting. These changes can be interpreted as ablations of MART \cite{WangICLR2020} and are easily implemented. Note that option (b) is essentially computing adversarial examples without label leaking \cite{KurakinARXIV2016}. However, as shown in \figref{fig:supp-ii-pll}, these two variants of AT have little to no impact on robust overfitting.

\begin{table}[t]
	\centering
	\vspace*{-0.2cm}
	\scriptsize
	\begin{tabularx}{0.475\textwidth}{|X|@{\hspace*{3px}}c@{\hspace*{3px}}|@{\hspace*{3px}}c@{\hspace*{3px}}|@{\hspace*{3px}}c@{\hspace*{3px}}|}
		\hline
		Model (\RTE against AutoAttack \cite{CroceICML2020}) & \RTE $\downarrow$ & $\lambda_{\text{max}}$ & $\frac{|\lambda_{\text{min}}|}{|\lambda_{\text{max}}|}$\\
		\hline
		\hline
		AT (baseline) & 62.8 & 1990 & 0.088\\
		Scaled $\times0.5$ & 62.8 & 7936 & 0.088\\
		Scaled $\times2$ & 62.8 & 505 & 0.088\\
		\hline
		Batch size $8$ & 58.2 & 3132 & 0.027\\
		Adam & 57.5 & 540 & 0.047\\
		\hline
		Label smoothing & 61.2 & 2484 & 0.085\\
		Self-supervision & 57.1 & 389 & 0.041\\
		Entropy-SGD & 58.6 & 5773 & 0.054\\
		TRADES & 56.7 & 947 & 0.089\\
		MART & 61 & 1285 & 0.087\\
		AT-AWP & 54.3 & 1200 & 0.241\\
		\hline
	\end{tabularx}
	\vspace*{-6px}
	\caption{\textbf{Hessian Eigenvalue $\lambda_{\text{max}}$ and Convexity:} For the models from \figref{fig:supp-visualization} and \ref{fig:supp-visualization-li}, we report \RTE against AutoAttack \cite{CroceARXIV2020}, the maximum Hessian eigenvalue $\lambda_{\text{max}}$ and the convexity measure of \cite{LiNIPS2018} computed as $\nicefrac{|\lambda_{\text{min}}|}{|\lambda_{\text{max}}|}$. This fraction is supposed to quantify the degree of non-convexity around the found minimum. As can be seen, neither $\lambda_{\text{max}}$ nor convexity correlate well with adversarial robustness. Regarding $\lambda_{\text{max}}$ this is due to the Hessian eigenspectrum not being scale-invariant, as shown for scaled versions ($\times0.5$ and $\times2$) of our AT baseline.}
	\label{tab:supp-convexity}
	\vspace*{-6px}
\end{table}

\textbf{AutoAugment:}
In \cite{CubukARXIV2018}, an automatic procedure for finding data augmentation policies is proposed, so-called AutoAugment. We use the found \CifarT policy (\cf \cite{CubukARXIV2018}, appendix), which includes quite extreme augmentations. For example, large translations are possible, rendering the image nearly completely uniform, only leaving few pixels at the border. In practice, AutoAugment usually prevents convergence and, thus, avoids overfitting. We further combine AutoAugment with CutOut \cite{DevriesARXIV2017} (using random $16\times 16$ ``cutouts''). We apply both AutoAugment and CutOut on top of our standard data augmentation, \ie, random flipping and cropping. We use publicly available PyTorch implementations\footnote{\url{https://github.com/DeepVoltaire/AutoAugment}, \url{https://github.com/uoguelph-mlrg/Cutout}}.

\textbf{Entropy-SGD}
\cite{ChaudhariICLR2017} explicitly encourages flatter minima by taking the so-called ``local'' entropy into account. As a result, Entropy-SGD not only finds ``deep'' minima (\ie, low loss values) but also flat ones. In practice, this is done using nested SGD: the inner loop approximates the local entropy using stochastic gradient Langevin dynamics (SGLD), the outer loop updates the weights. The number of inner iterations is denoted by $L$. While the original work \cite{ChaudhariICLR2017} uses $L$ in $[5, 20]$ on \CifarT, we experiment with $L \in \{1, 2, 3, 5\}$. Note that, for fair comparison, we train for $\nicefrac{150}{L}$ epochs.
For details on the Entropy-SGD algorithm, we refer to \cite{ChaudhariICLR2017}. Our implementation follows the official PyTorch implementation\footnote{\url{https://github.com/ucla-vision/entropy-sgd}}.

\begin{figure*}[t]
	\centering
	\begin{minipage}[t]{0.23\textwidth}
		\includegraphics[width=\textwidth]{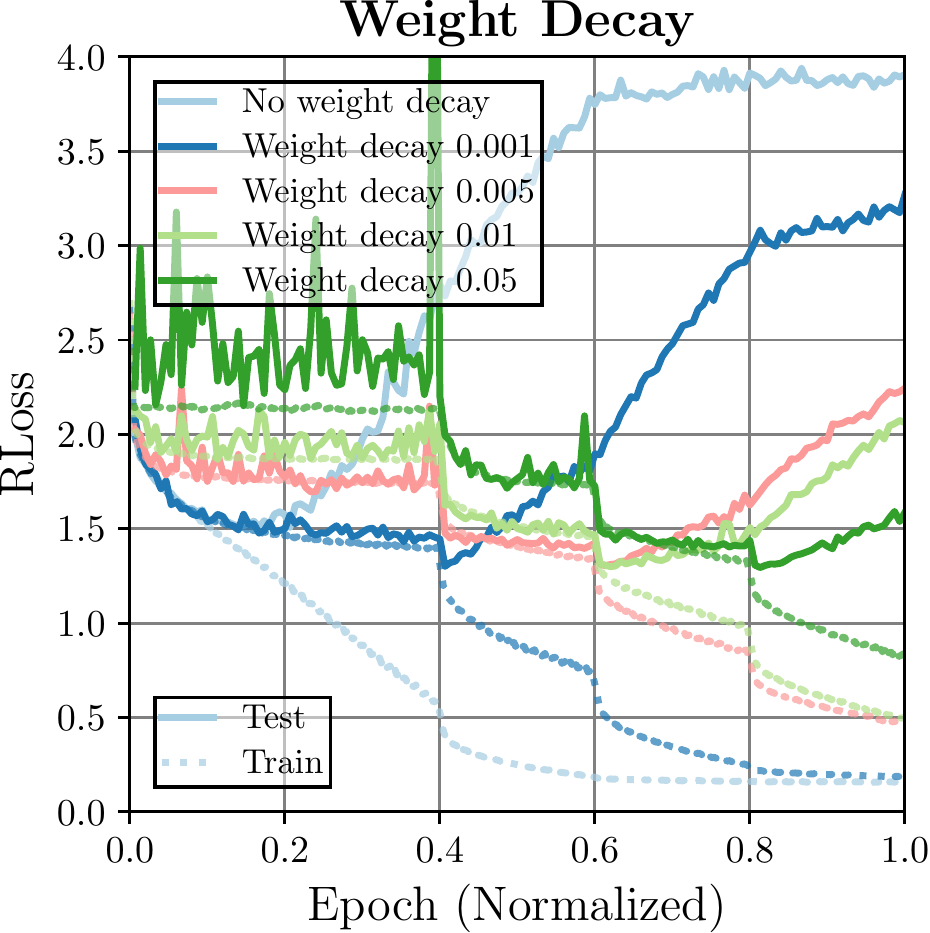}
	\end{minipage}
	\begin{minipage}[t]{0.23\textwidth}
		\includegraphics[width=\textwidth]{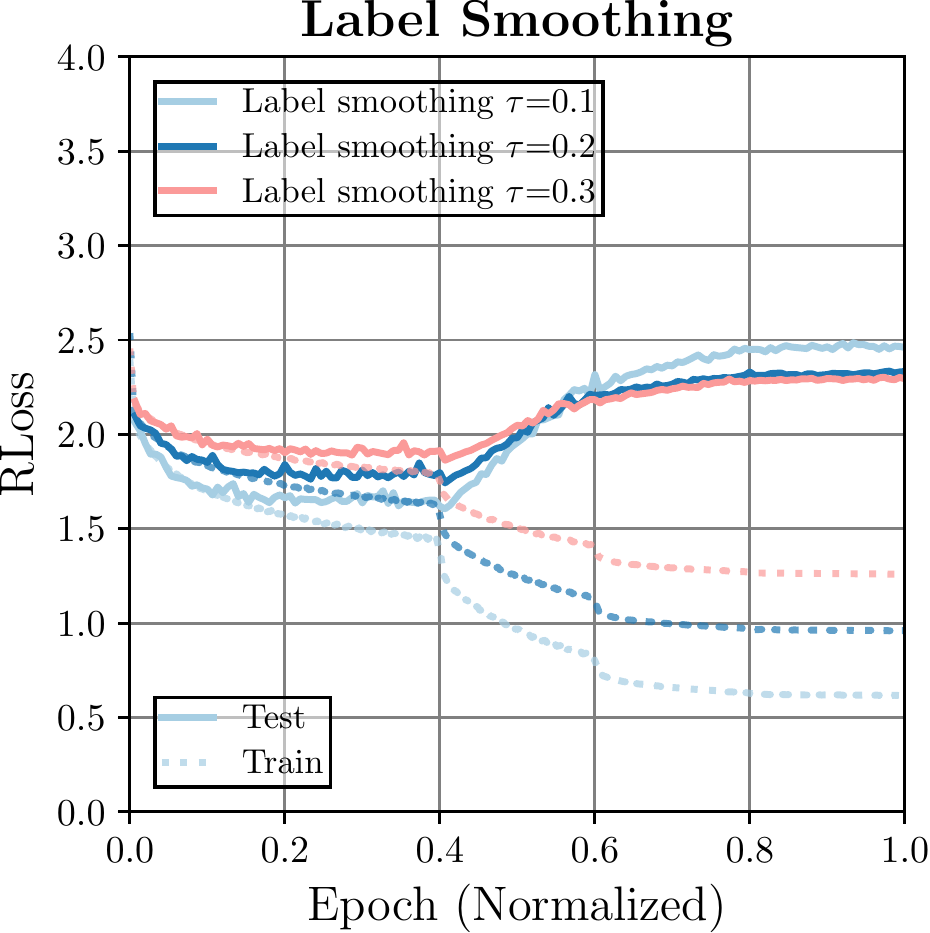}
	\end{minipage}
	\begin{minipage}[t]{0.23\textwidth}
		\includegraphics[width=\textwidth]{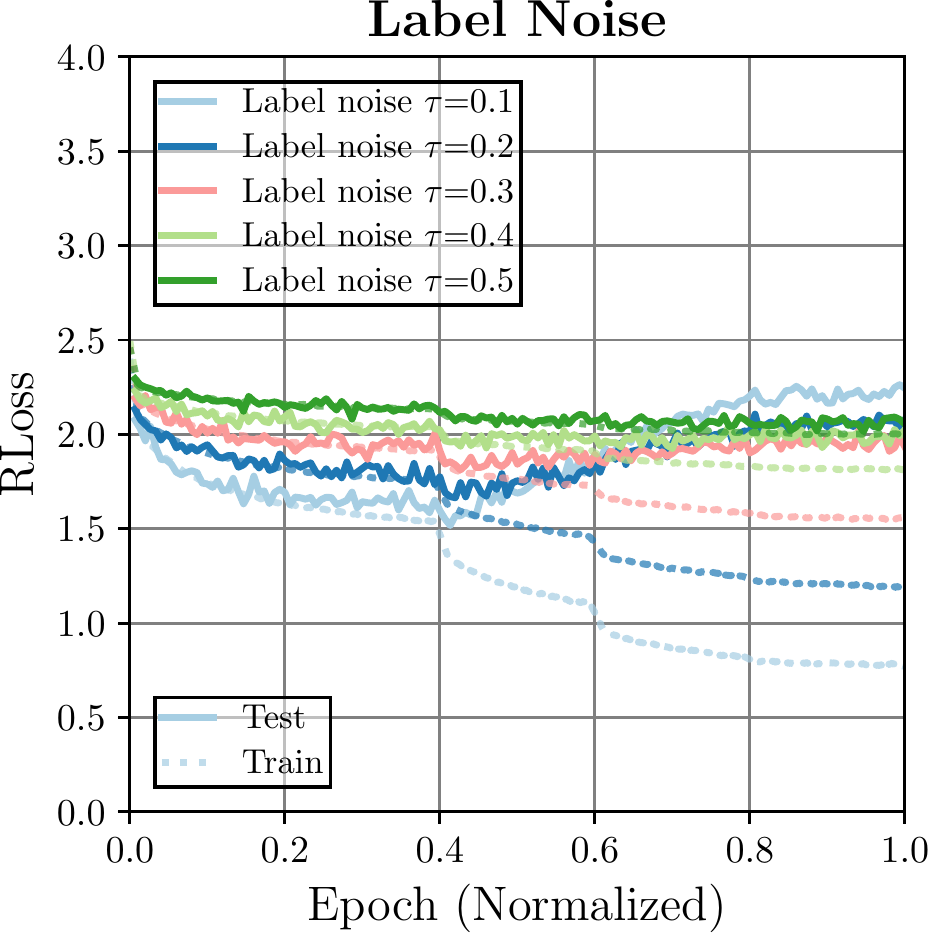}
	\end{minipage}
	\begin{minipage}[t]{0.23\textwidth}
		\includegraphics[width=\textwidth]{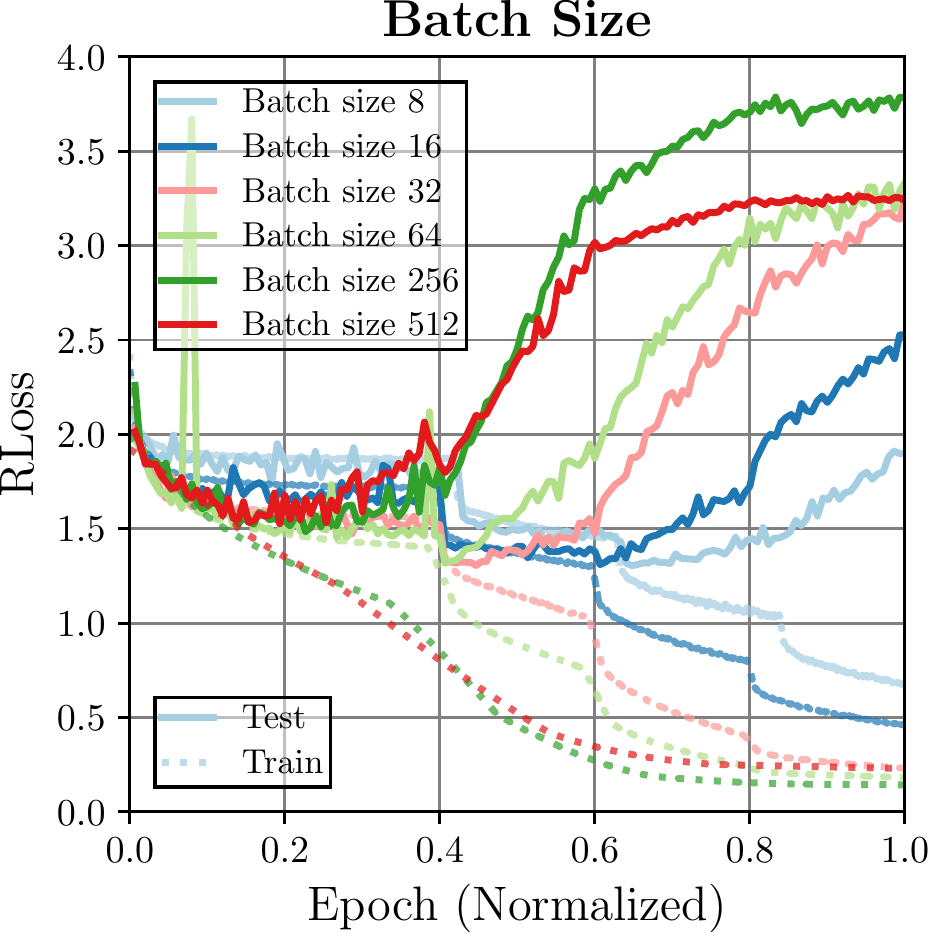}
	\end{minipage}
	\\[2.5px]
	\begin{minipage}[t]{0.23\textwidth}
		\includegraphics[width=\textwidth]{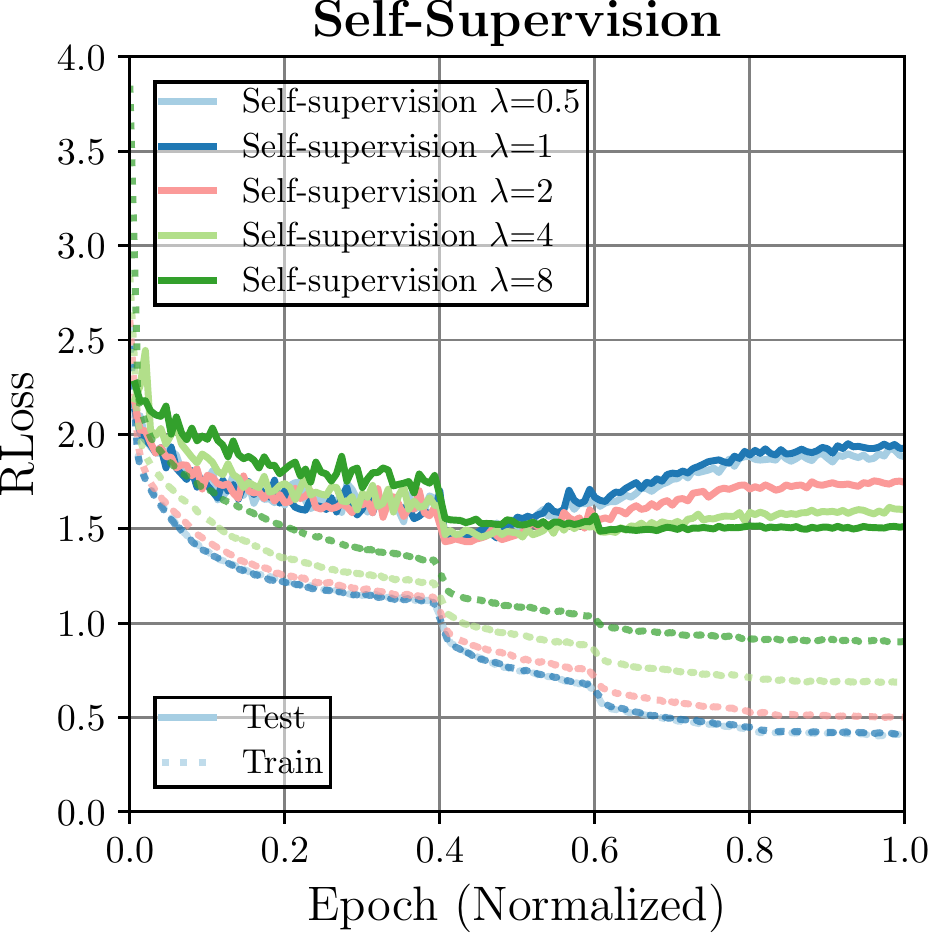}
	\end{minipage}
	\begin{minipage}[t]{0.23\textwidth}
		\includegraphics[width=\textwidth]{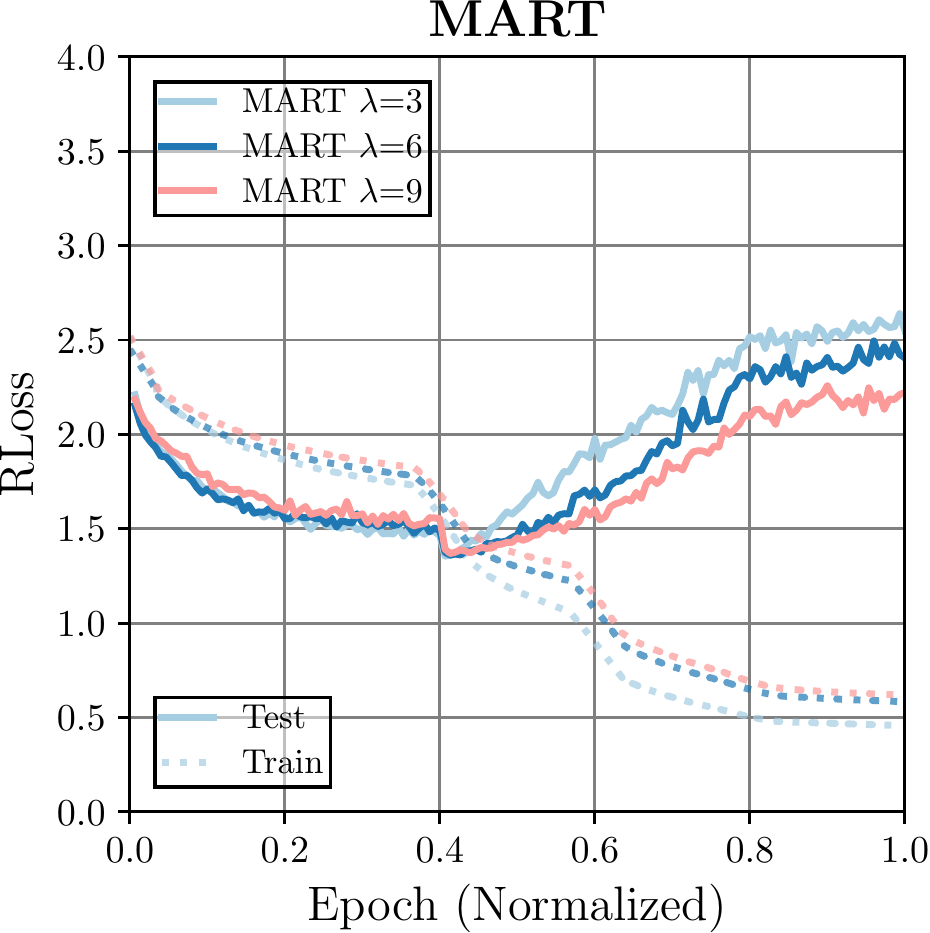}
	\end{minipage}
	\begin{minipage}[t]{0.23\textwidth}
		\includegraphics[width=\textwidth]{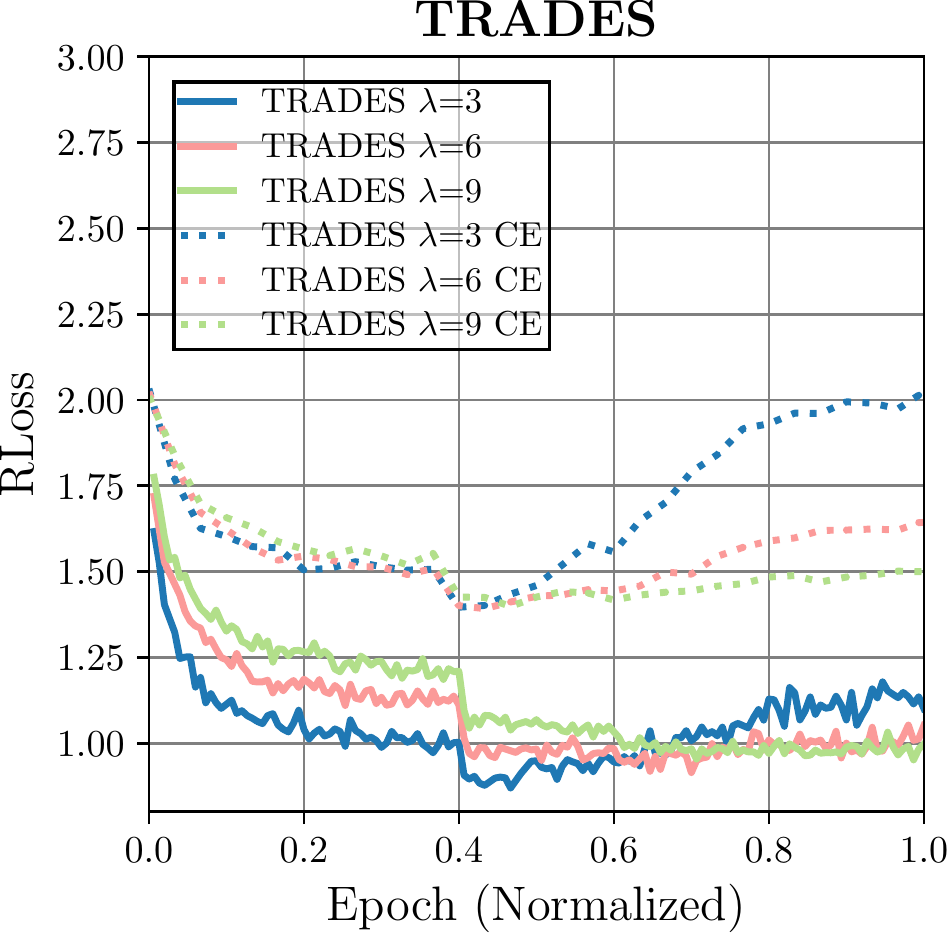}
	\end{minipage}
	\begin{minipage}[t]{0.23\textwidth}
		\includegraphics[width=\textwidth]{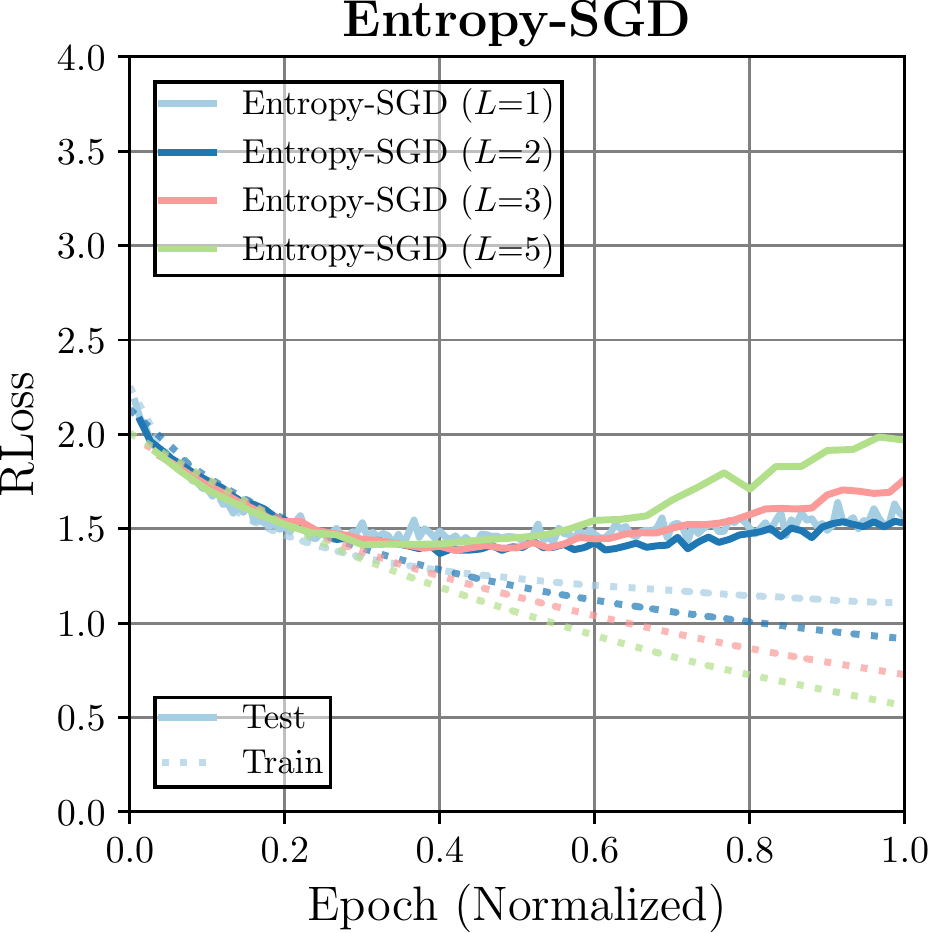}
	\end{minipage}
	\caption{\textbf{Training Curves for Varying Hyper-Parameters:} We plot \RCE for selected methods and hyper-parameters to demonstrate the impact of hyper-parameters on avoiding or reducing robust overfitting. Note that, for TRADES, we show both \RCE on adversarial examples computed by maximizing the KL-divergence in \eqnref{eq:supp-methods-trades} (solid) and on adversarial examples obtained by maximizing cross-entropy loss (``CE'', dotted).}
	\label{fig:supp-training-ablation}
\end{figure*}

\textbf{Activation functions:}
We consider three recently proposed activation functions: SiLU \cite{ElfwingNN2018}, MiSH \cite{MisraBMVC2020} and GeLU \cite{HendrycksARXIV2016}. These are defined as:
\begin{align}
	(\text{SiLU})\quad &x \sigma(x)\text{ with } \sigma(x) = \nicefrac{1}{(1 + \exp(-x))},\\
	(\text{MiSH})\quad &x \tanh(\log(1 + \exp(x))),\\
	(\text{GeLU})\quad &x \sigma(1.702 x).
\end{align}
All of these activation functions can be seen as smooth versions of the ReLU activation.
In \cite{SinglaARXIV2021}, some of these activation functions are argued to avoid robust overfitting due to lower curvature compared to ReLU.

\textbf{AT-AWP:}
AT with adversarial weight perturbations (AT-AWP) \cite{WuNIPS2020} computes adversarial weight perturbations \emph{on top} of adversarial examples to further regularize training. This is similar to our worst-case flatness measure of \RCE, however, adversarial examples and adversarial weights are computed sequentially, not jointly, and only one iteration is used to compute adversarial weights. Specifically, after computing adversarial examples $\tilde{x} = x + \delta$,
an adversarial weight perturbation $\nu$ is computed by solving
\begin{align}
	\max_{\nu \in B_\xi(w)} \mathcal{L}(f(\tilde{x}; w + \nu), y)
\end{align}
with $B_\xi(w)$ as in \eqnref{eq:supp-ball} using one iteration of gradient ascent with fixed step size of $\xi$. The gradient is normalized per layer as in \eqnref{eq:supp-normalization}.
We considered $\xi \in \{0.0005, 0.001, 0.005, 0.01, 0.015, 0.02\}$ and between $1$ and $7$ iterations and found that $\xi = 0.01$ and $1$ iteration works best (similar to \cite{WuNIPS2020}).

\textbf{TRADES:} \cite{ZhangICML2019} proposes an alternative formulation of AT that allows a better trade-off between adversarial robustness and (clean) accuracy. The loss to be minimized is
\begin{align}
	\begin{split}
		\mathcal{L}&(f(x;w), y)\\
		&+ \lambda \max_{\|\delta\|_\infty \leq \epsilon} \text{KL}(f(x;w), f(x + \delta;w)).\label{eq:supp-methods-trades}
	\end{split}
\end{align}
During training, adversarial examples are computed by maximizing the KL-divergence (instead of cross-entropy loss), \ie, using the second term in \eqnref{eq:supp-methods-trades}. Commonly $\lambda = 6$ is chosen, however, we additionally tried $\lambda \in \{1, 3, 6, 9\}$. We follow the official implementation\footnote{\url{https://github.com/yaodongyu/TRADES}}.

\begin{figure*}
	\centering
	\vspace*{-0.2cm}
	\begin{minipage}[t]{0.11\textwidth}
		\includegraphics[height=3.25cm]{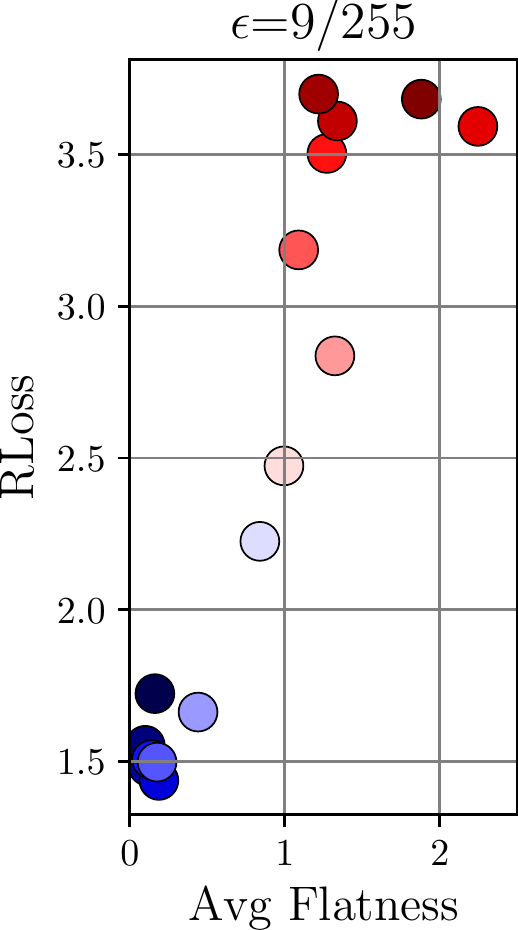}
	\end{minipage}
	\begin{minipage}[t]{0.12\textwidth}
		\includegraphics[height=3.25cm]{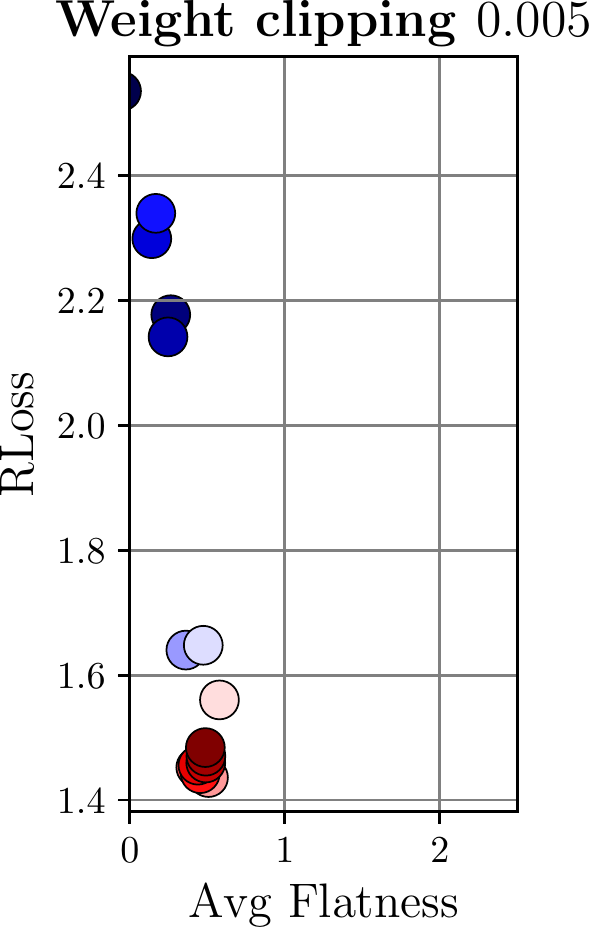}
	\end{minipage}
	\begin{minipage}[t]{0.11\textwidth}
		\includegraphics[height=3.25cm]{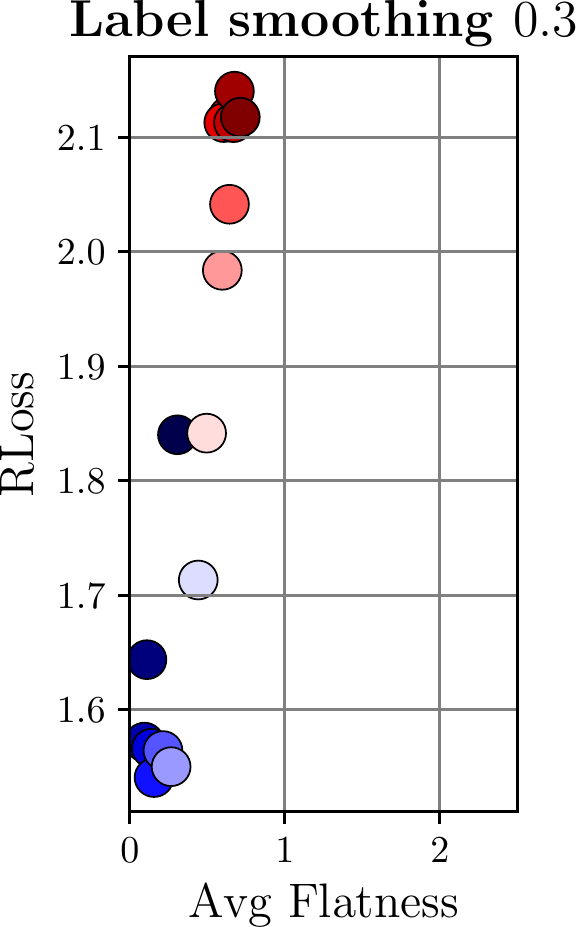}
	\end{minipage}
	\begin{minipage}[t]{0.11\textwidth}
		\includegraphics[height=3.25cm]{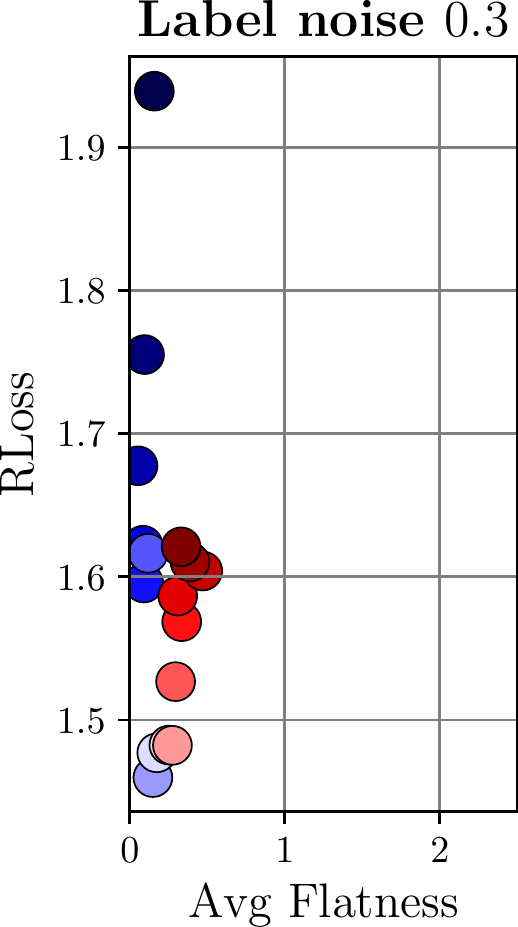}
	\end{minipage}
	\begin{minipage}[t]{0.11\textwidth}
		\includegraphics[height=3.25cm]{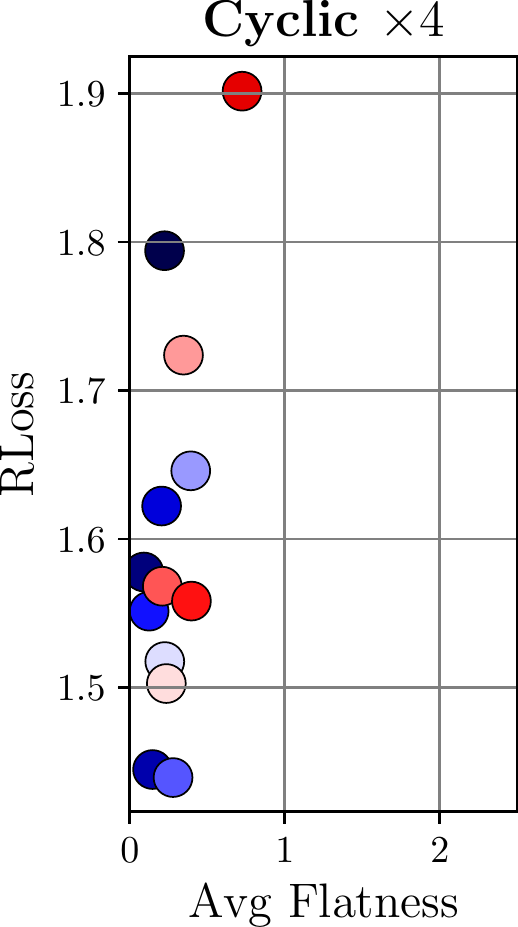}
	\end{minipage}
	\begin{minipage}[t]{0.11\textwidth}
		\includegraphics[height=3.25cm]{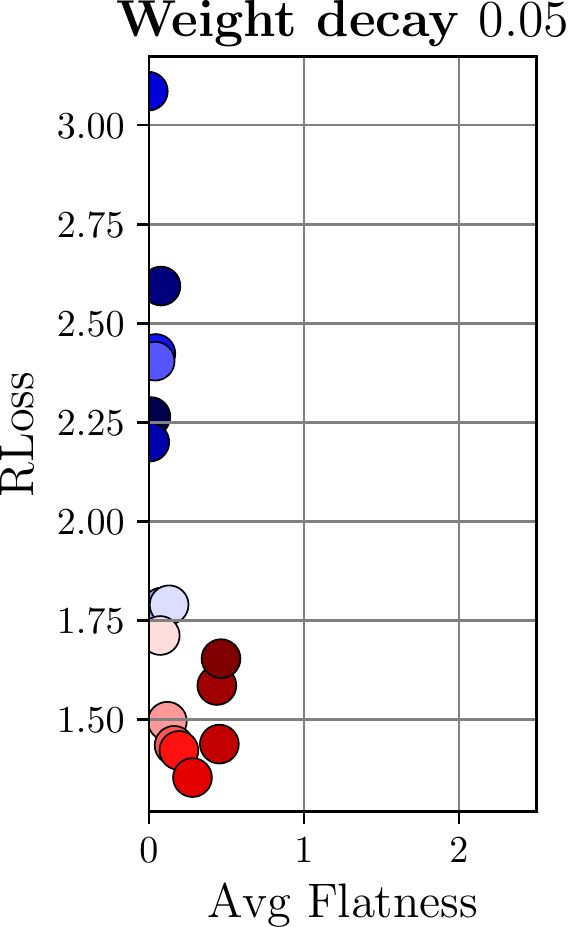}
	\end{minipage}
	\begin{minipage}[t]{0.145\textwidth}
		\includegraphics[height=3.25cm]{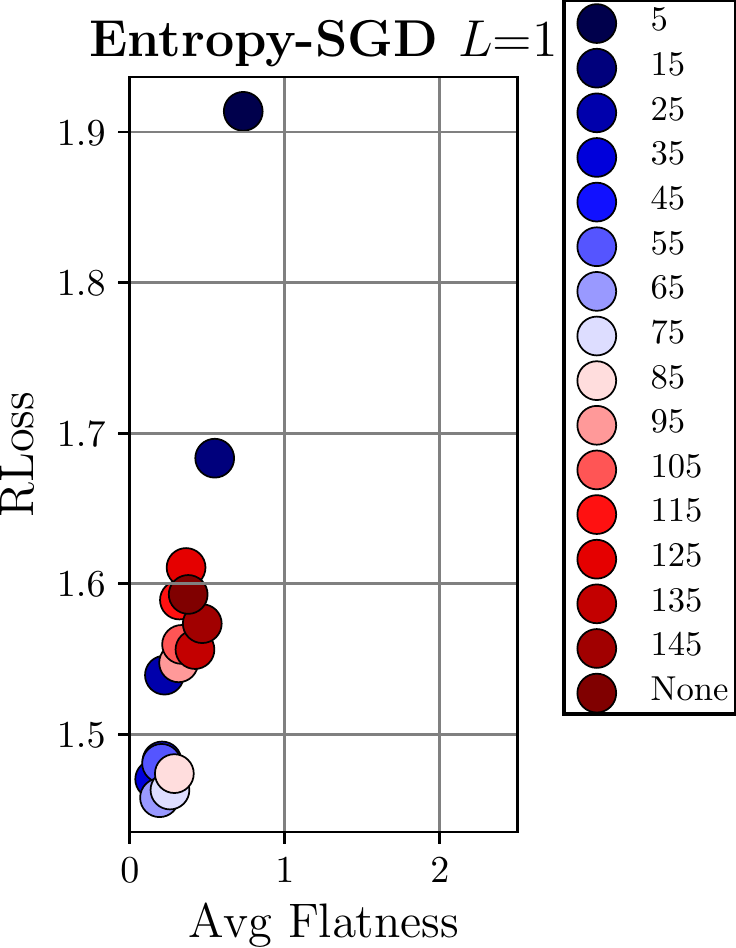}
	\end{minipage}
	\\[2.5px]
	
	\begin{minipage}[t]{0.12\textwidth} 
		\hphantom{\includegraphics[height=3.25cm]{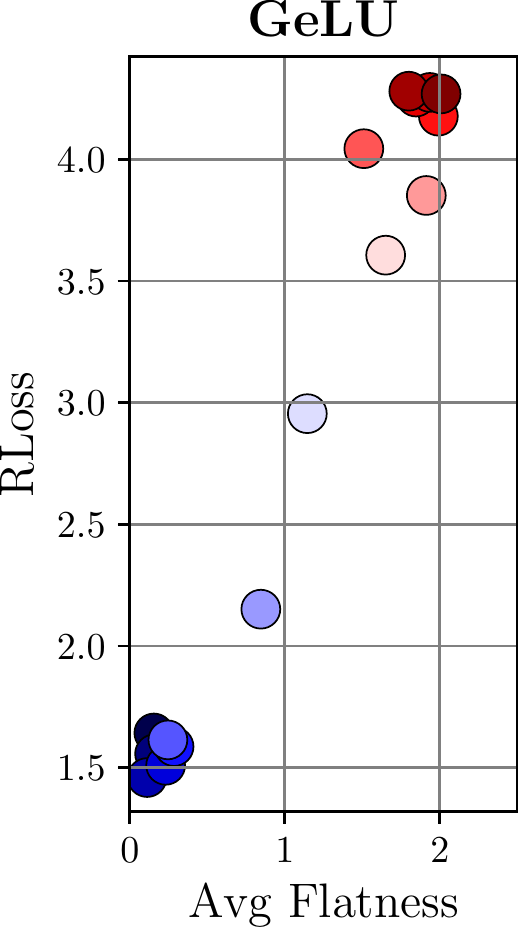}}
	\end{minipage}
	\begin{minipage}[t]{0.11\textwidth}
		\includegraphics[height=3.25cm]{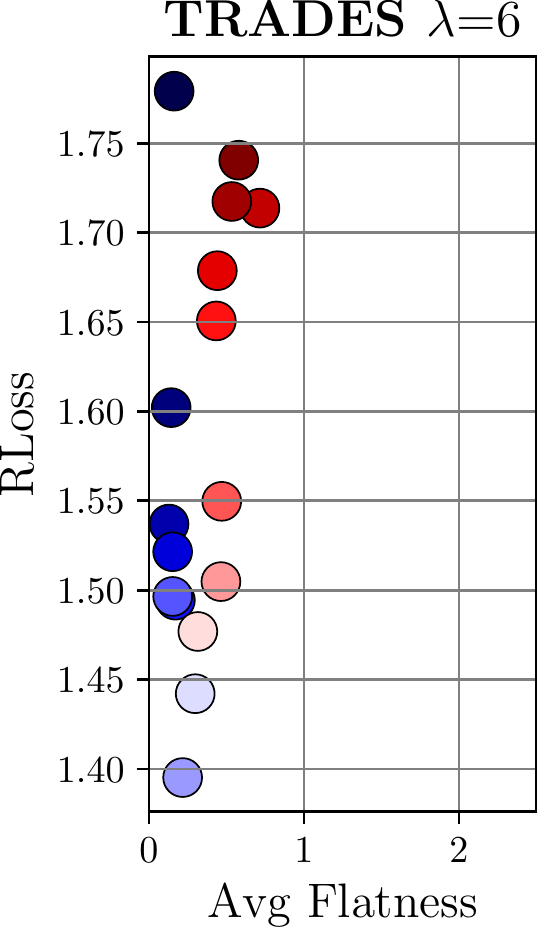}
	\end{minipage}
	\begin{minipage}[t]{0.11\textwidth}
		\includegraphics[height=3.25cm]{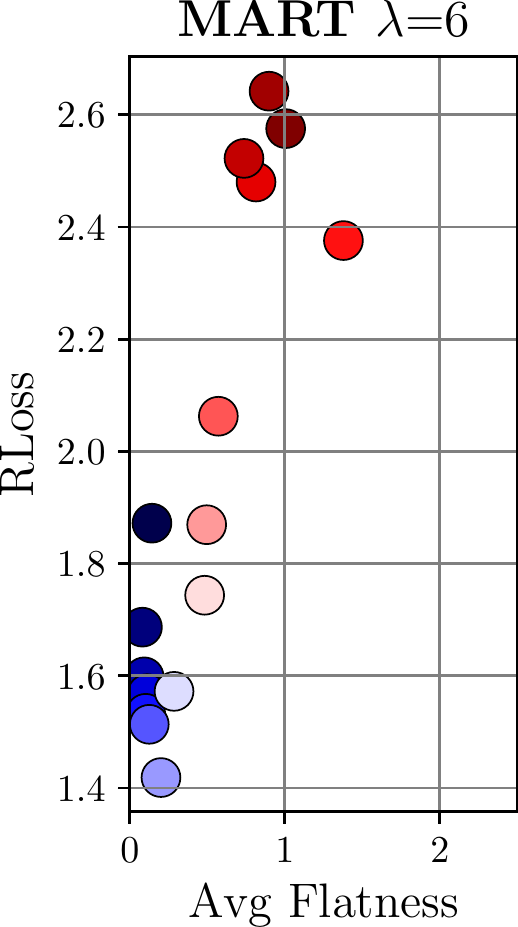}
	\end{minipage}
	\begin{minipage}[t]{0.11\textwidth}
		\includegraphics[height=3.25cm]{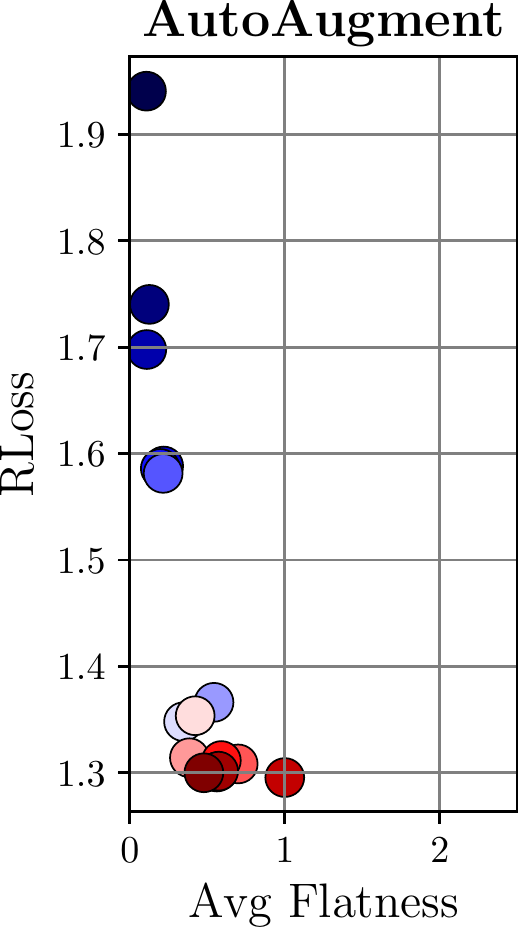}
	\end{minipage}
	\begin{minipage}[t]{0.11\textwidth}
		\includegraphics[height=3.25cm]{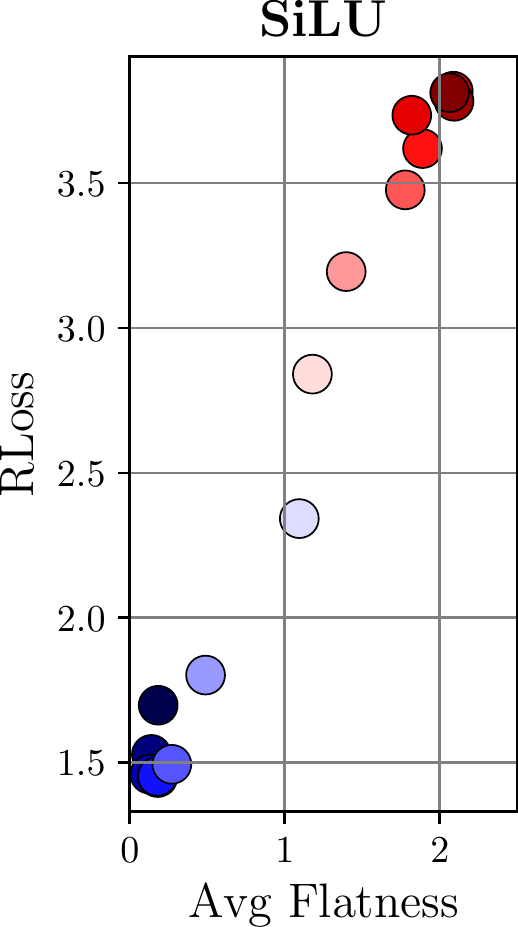}
	\end{minipage}
	\begin{minipage}[t]{0.11\textwidth}
		\includegraphics[height=3.25cm]{plots_supp_flatness_epochs_correlation_joint_gelu}
	\end{minipage}
	\begin{minipage}[t]{0.14\textwidth}
		\includegraphics[height=3.25cm]{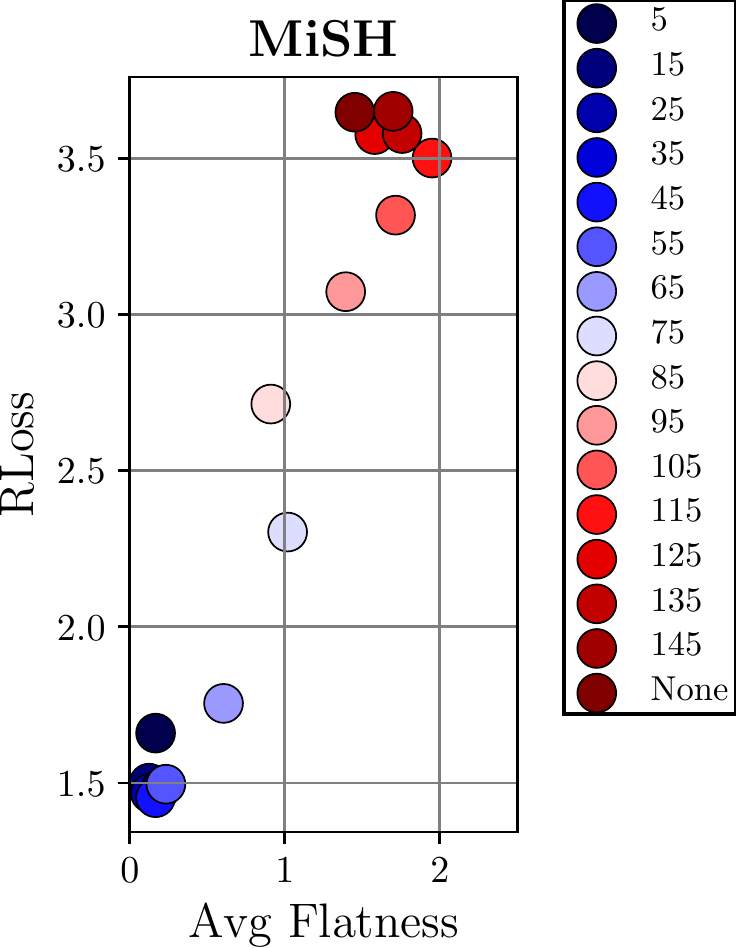}
	\end{minipage}
	\vspace*{-6px}
	\caption{\textbf{Flatness Throughout Training:} Complementary to the main paper, we plot \RCE against average-case flatness in \RCE for selected methods throughout training epochs. \red{Early epochs are shown in {\color{blue!60!black}dark blue}, late epochs are shown in {\color{red!60!black}dark red}.} For cyclic learning rate, we show $4$ cycles with a total of $120$ epochs. For many methods not avoiding robust overfitting, flatness decreases alongside an increase in \RCE during overfitting. Using, \eg, AutoAugment, label noise or Entropy-SGD, in contrast, both effects are reduced.}
	\label{fig:supp-methods-flatness-epochs}
	\vspace*{-6px}
\end{figure*}

\textbf{MART} \cite{WangICLR2020} explicitly addresses the problem of incorrectly classified examples during training. First, the cross-entropy loss $\mathcal{L}$ for training is replaced using a binary cross-entropy loss $\mathcal{L}_{\text{bin}}$, \ie, classifying correct class vs\onedot most-confident ``other'' class:
\begin{align}
	\begin{split}
		\mathcal{L}_{\text{bin}}(f(x; w), y) =& -\log(f_y(x;w))\\
		& - \log(1 - \max_{y' \neq y} f_{y'}(x; w)).
	\end{split}
\end{align}
Second, the KL-divergence used in TRADES in \eqnref{eq:supp-methods-trades} is combined with a confidence-based weight:
\begin{align}
	\begin{split}
		\mathcal{L}_{\text{bin}}&(f(\tilde{x}; w), y)\\ &+ \lambda \text{KL}(f(x;w), f(\tilde{x}; w)() (1 - f_y(x; w))
	\end{split}
\end{align}
Adversarial examples are still computed by maximizing regular cross-entropy loss. We follow the official implementation\footnote{\url{https://github.com/YisenWang/MART}}. MART is successful in reducing robust overfitting on incorrectly classified examples, as shown in \figref{fig:supp-ii-pll}.

\textbf{PGD-$\tau$:}
In \cite{ZhangARXIV2020}, a variant of PGD is proposed for AT: PGD-$\tau$ stops maximization $\tau$ iterations \emph{after} the label flipped. This is supposed to find ``friendlier'' adversarial examples that can be used for AT. Note that $\tau = 0$ also does \emph{not} compute adversarial examples on incorrectly classified training examples. We consider $\tau \in \{0, 1, 2, 3\}$.

\textbf{Self-Supervision:} Following \cite{HendrycksNIPS2019}, we implement AT using rotation-prediction as \emph{additional} self-supervised task. Note, however, that no additional (unlabeled) training examples are used. Specifically, the following learning problem is tackled:
\begin{align}
	\begin{split}
		\max&_{\|\delta\|_\infty \leq \epsilon} \mathcal{L}(f(x + \delta;w), y)\\
		&+ \lambda \max_{\|\delta\|_\infty \leq \epsilon} \mathcal{L}(f(\text{rot}(x + \delta, r); w), y_{r})\\
		&r\in\{0, 90, 180, 270\}, y_r \in \{0, 1, 2, 3\}
	\end{split}
\end{align}
where $\text{rot}(x, r)$ rotates the training example $x$ by $r$ degrees.
In practice, we split every batch in half: The first half uses the original training examples with correct labels. Examples in the second half are rotated randomly by $\{0, 90, 180, 270\}$ degrees, and the labels correspond to the rotation (\ie, $\{0, 1, 2, 3\}$). Adversarial examples are computed against the true or rotation-based labels. Note that, in contrast to common practice \cite{SohnARXIV2020}, we do \emph{not} predict all four possible rotations every batch, but just one randomly drawn per example. We still use $150$ epochs in total. We consider $\lambda \in \{0.5, 1, 2, 4, 8\}$.
 
\textbf{Additional Unlabeled Examples:} As proposed in \cite{CarmonNIPS2019,UesatoNIPS2019}, we also consider additional, pseudo-labeled examples during training. We use the provided pseudo-labeled data from \cite{CarmonNIPS2019} and split each batch in half: using $50\%$ original \CifarT training examples, and $50\%$ pseudo-labeled training examples from \cite{CarmonNIPS2019}. We still use $150$ epochs in total. We follow the official PyTorch implementation\footnote{\url{https://github.com/yaircarmon/semisup-adv}}.

\subsection{Training Curves}
\label{sec:supp-methods-ablation}

\figref{fig:supp-training-ablation} shows (test) \RCE throughout training for selected methods and hyper-parameters. Across all methods, we found that hyper-parameters have a large impact on robust overfitting. For example, weight decay or smaller batch sizes can reduce and delay robust overfitting considerably if regularization is ``strong'' enough, \ie, large weight decay or low batch size (to induce more randomness). For the other methods, difference between hyper-parameters is more subtle. However, across all cases, reduced overfitting generally goes hand in hand with higher \RCE on training examples, \ie, the robust generalization gap is reduced. This indicates that avoiding convergence on training examples plays an important role in avoiding robust overfitting.

\begin{figure*}[t]
	\centering
	\begin{minipage}[t]{0.19\textwidth}
		\vspace*{0px}
		
		\includegraphics[height=1.8cm]{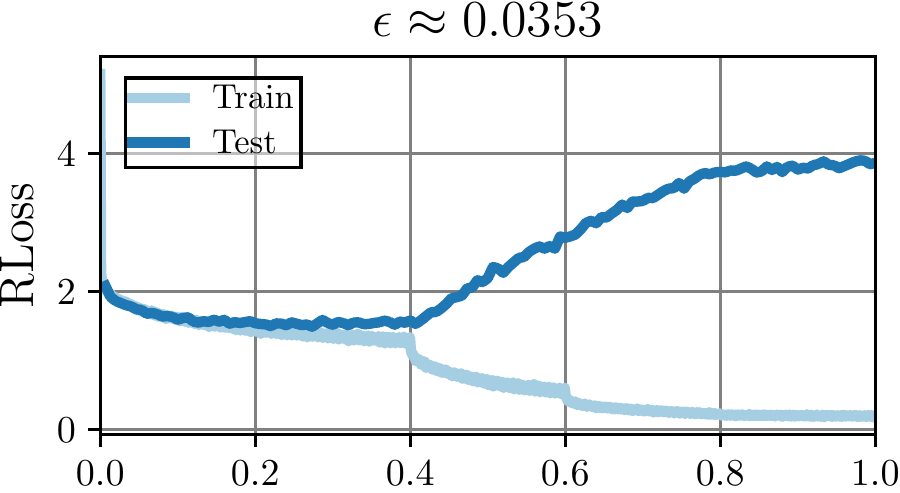}
		
		\hspace*{-0.25cm}
		\includegraphics[height=1.8cm]{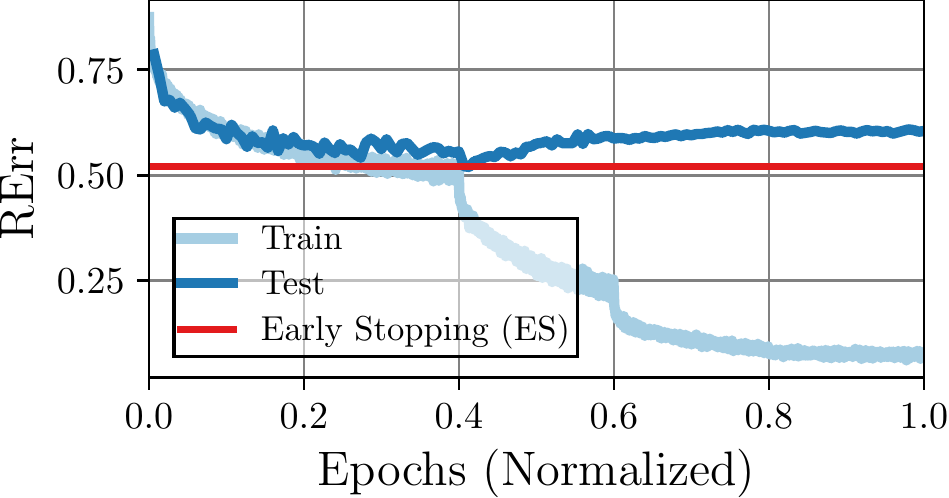}
	\end{minipage}
	\begin{minipage}[t]{0.19\textwidth}
		\vspace*{0px}
		
		\includegraphics[height=1.8cm]{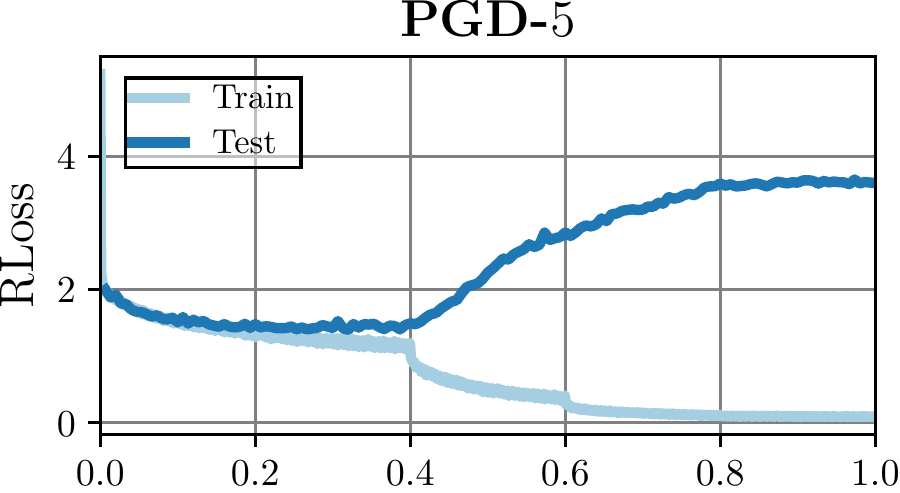}
		
		\hspace*{-0.25cm}
		\includegraphics[height=1.8cm]{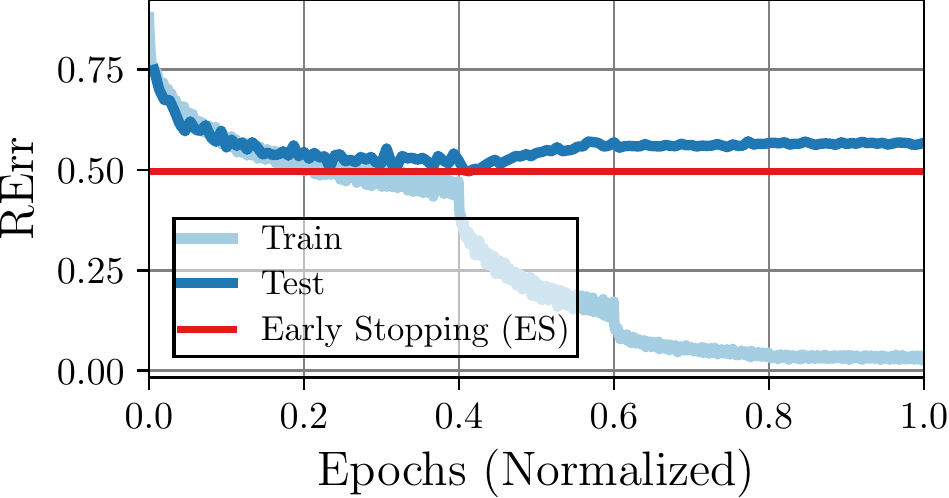}
	\end{minipage}
	\begin{minipage}[t]{0.19\textwidth}
		\vspace*{0px}
		
		\includegraphics[height=1.8cm]{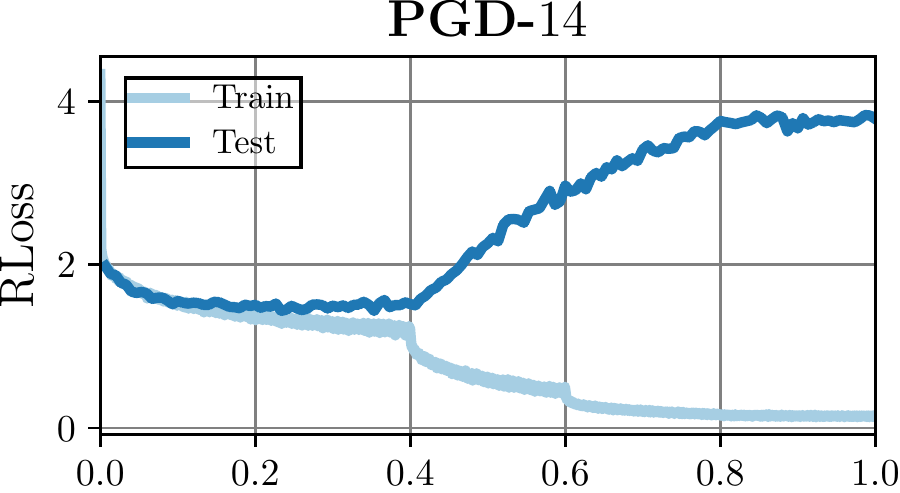}
		
		\hspace*{-0.25cm}
		\includegraphics[height=1.8cm]{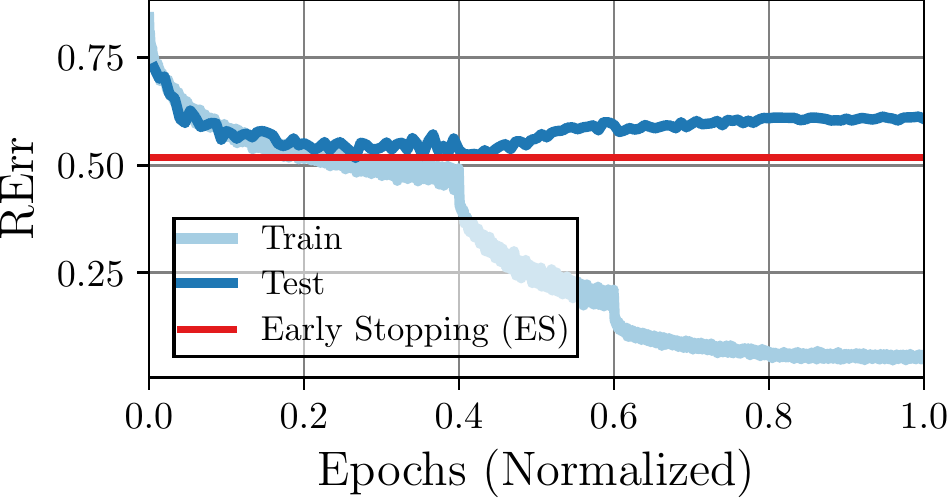}
	\end{minipage}
	\begin{minipage}[t]{0.19\textwidth}
		\vspace*{0px}
		
		\includegraphics[height=1.8cm]{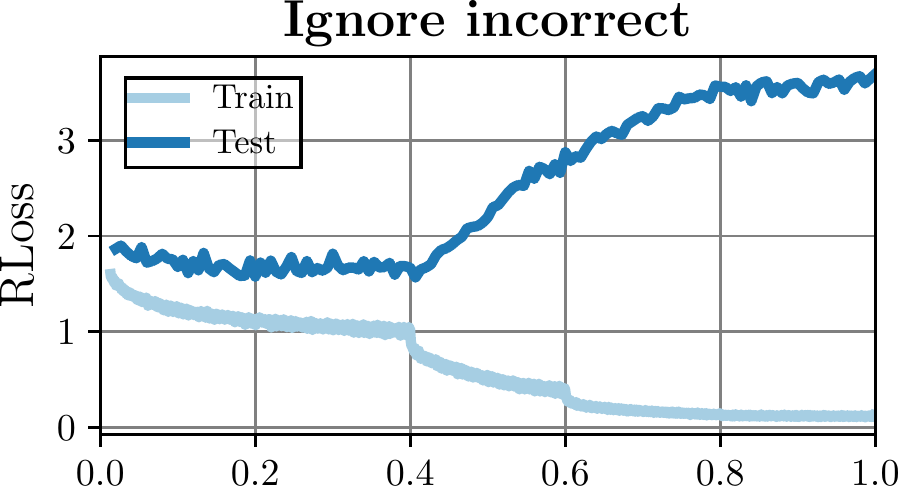}
		
		\hspace*{-0.25cm}
		\includegraphics[height=1.8cm]{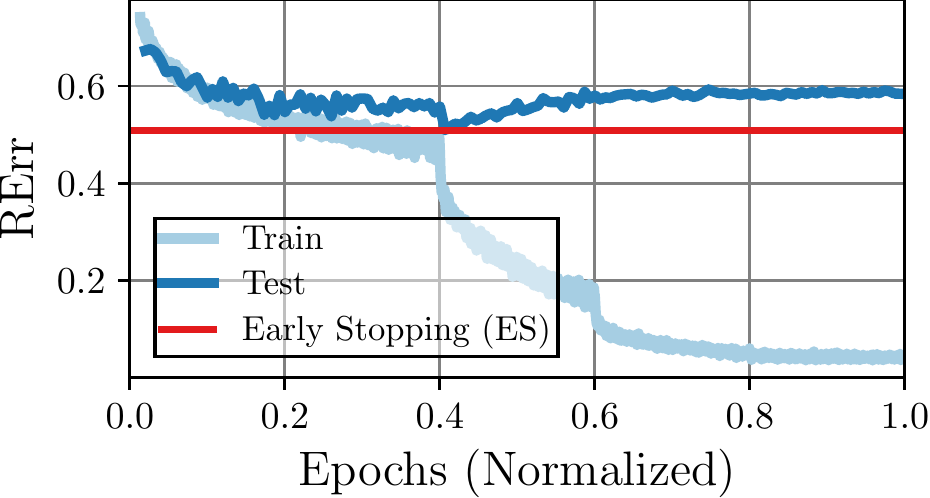}
	\end{minipage}
	\begin{minipage}[t]{0.19\textwidth}
		\vspace*{0px}
		
		\includegraphics[height=1.8cm]{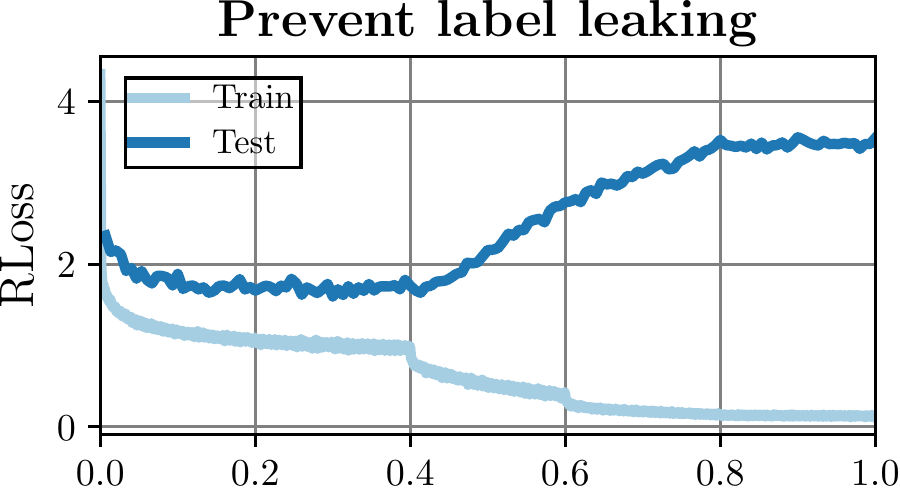}
		
		\hspace*{-0.25cm}
		\includegraphics[height=1.8cm]{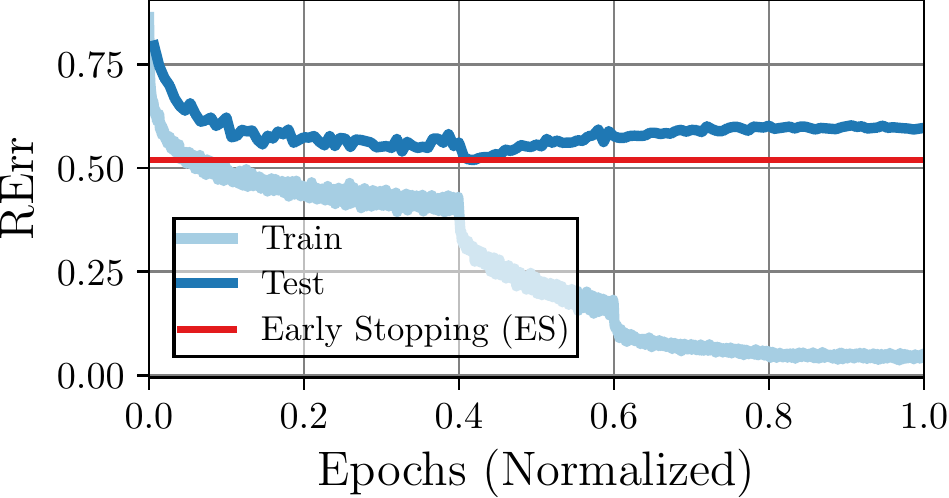}
	\end{minipage}
	\\[2.5px]
	
	\begin{minipage}[t]{0.19\textwidth}
		\vspace*{0px}
		
		\includegraphics[height=1.8cm]{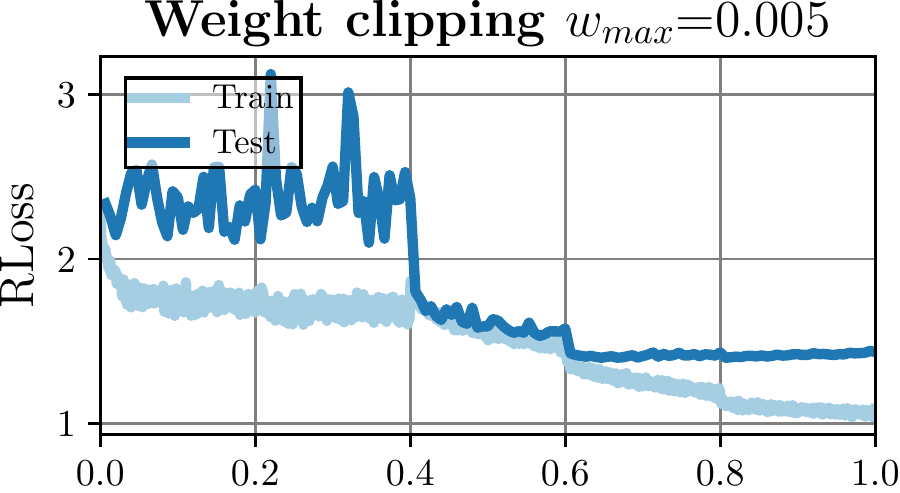}
		
		\hspace*{-0.25cm}
		\includegraphics[height=1.8cm]{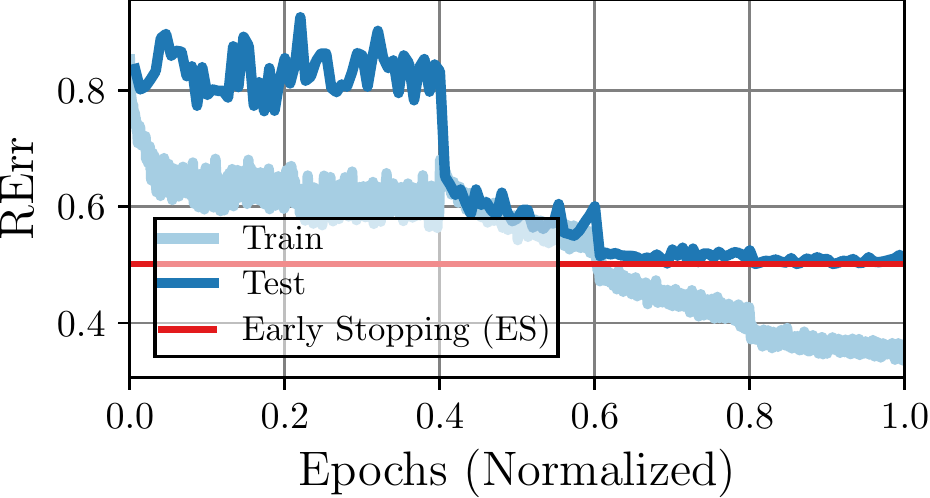}
	\end{minipage}
	\begin{minipage}[t]{0.19\textwidth}
		\vspace*{0px}
		
		\includegraphics[height=1.8cm]{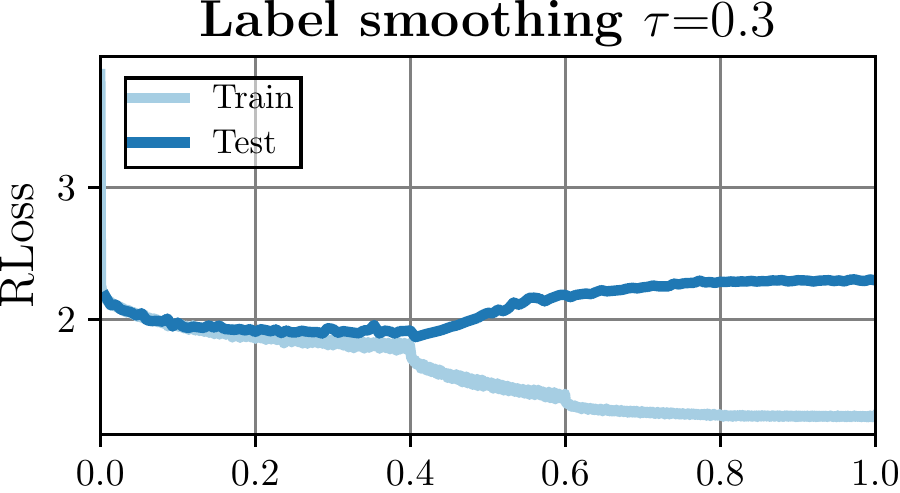}
		
		\hspace*{-0.25cm}
		\includegraphics[height=1.8cm]{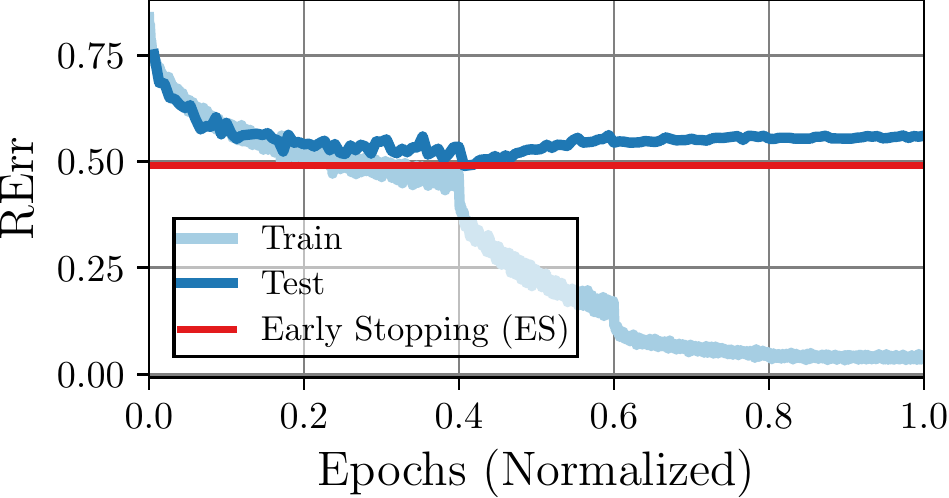}
	\end{minipage}
	\begin{minipage}[t]{0.19\textwidth}
		\vspace*{0px}
		
		\includegraphics[height=1.8cm]{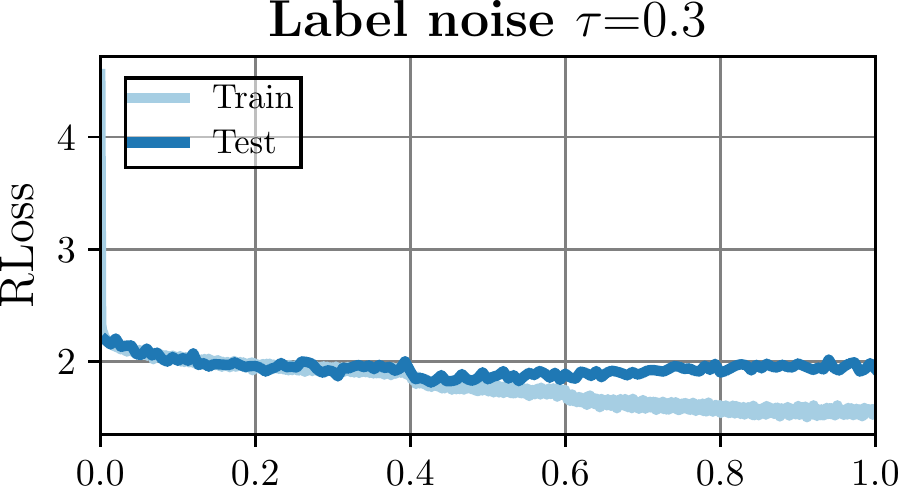}
		
		\hspace*{-0.25cm}
		\includegraphics[height=1.8cm]{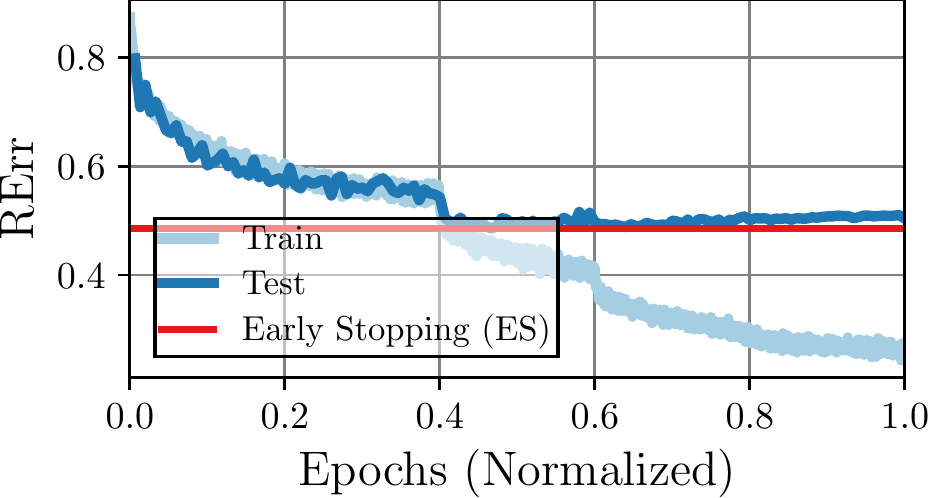}
	\end{minipage}
	\begin{minipage}[t]{0.19\textwidth}
		\vspace*{0px}
		
		\includegraphics[height=1.8cm]{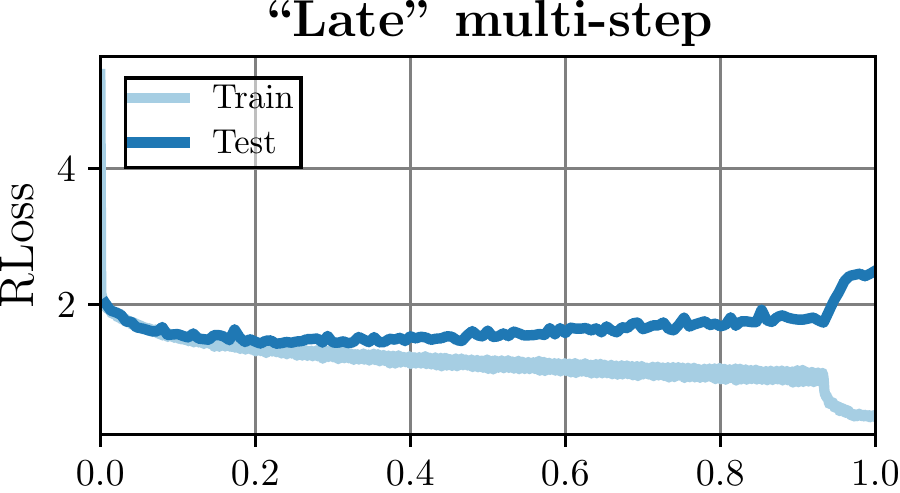}
		
		\hspace*{-0.25cm}
		\includegraphics[height=1.8cm]{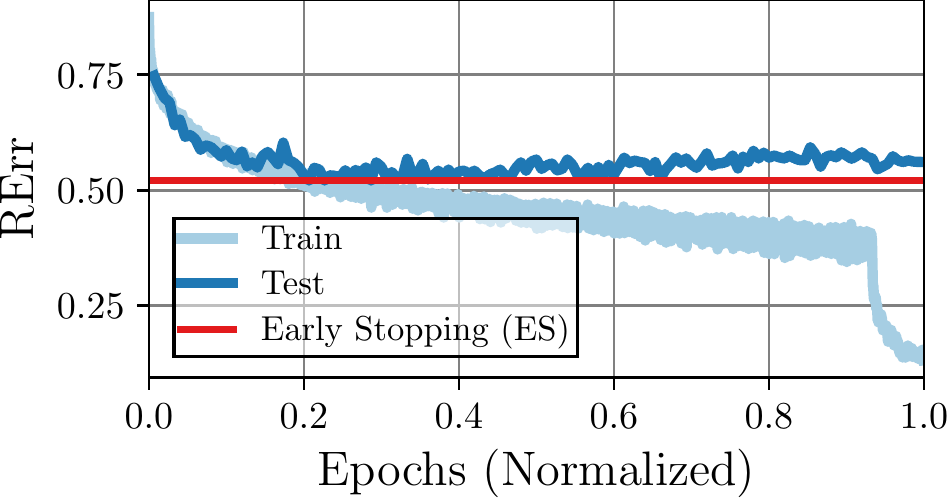}
	\end{minipage}
	\begin{minipage}[t]{0.19\textwidth}
		\vspace*{0px}
		
		\includegraphics[height=1.8cm]{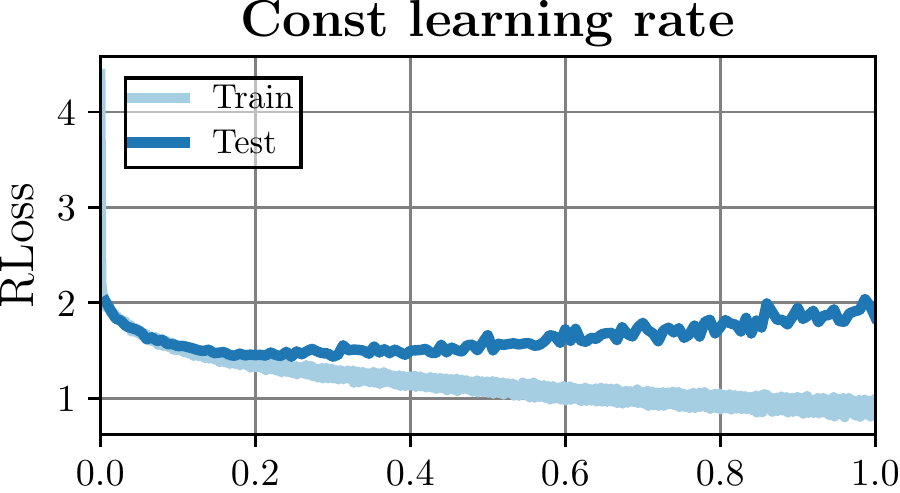}
		
		\hspace*{-0.25cm}
		\includegraphics[height=1.8cm]{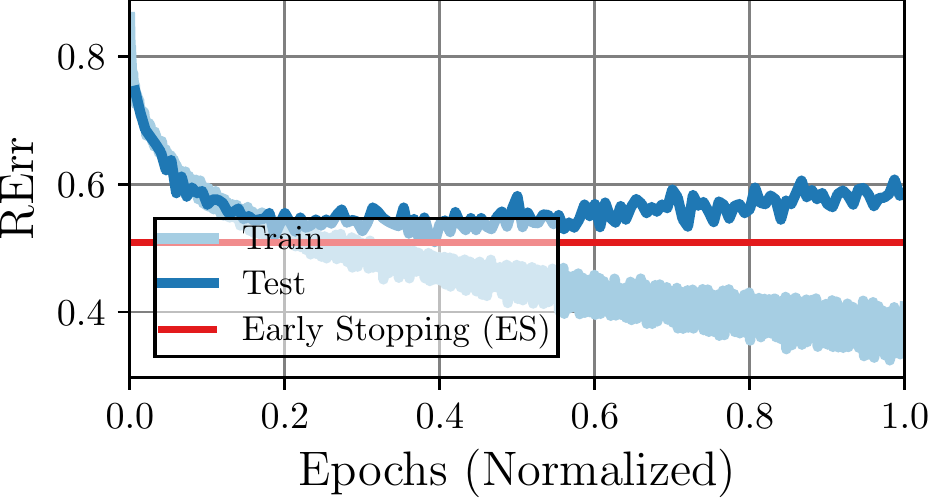}
	\end{minipage}
	\\[2.5px]
		
	\begin{minipage}[t]{0.19\textwidth}
		\vspace*{0px}
		
		\includegraphics[height=1.8cm]{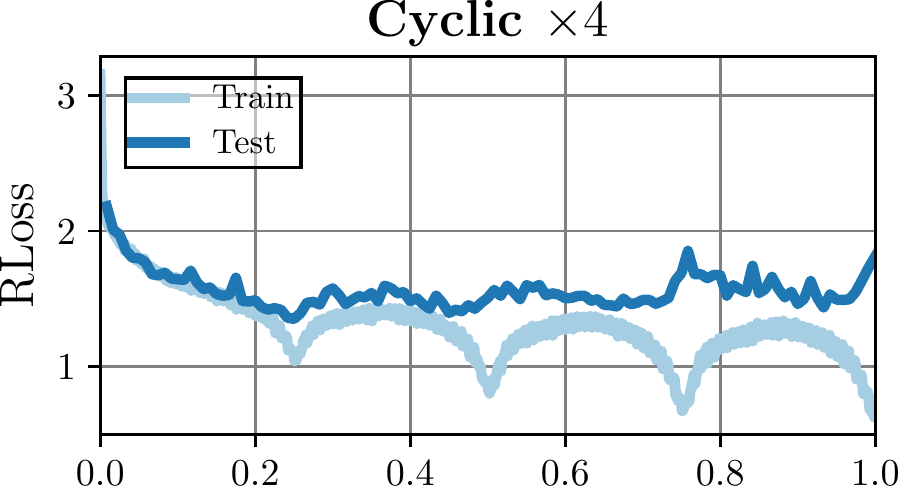}
		
		\hspace*{-0.25cm}
		\includegraphics[height=1.8cm]{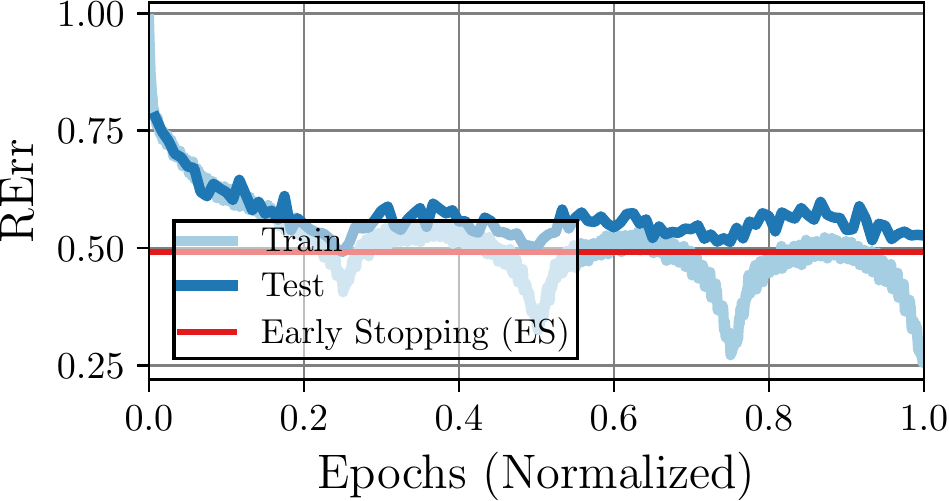}
	\end{minipage}
	\begin{minipage}[t]{0.19\textwidth}
		\vspace*{0px}
		 
		\includegraphics[height=1.8cm]{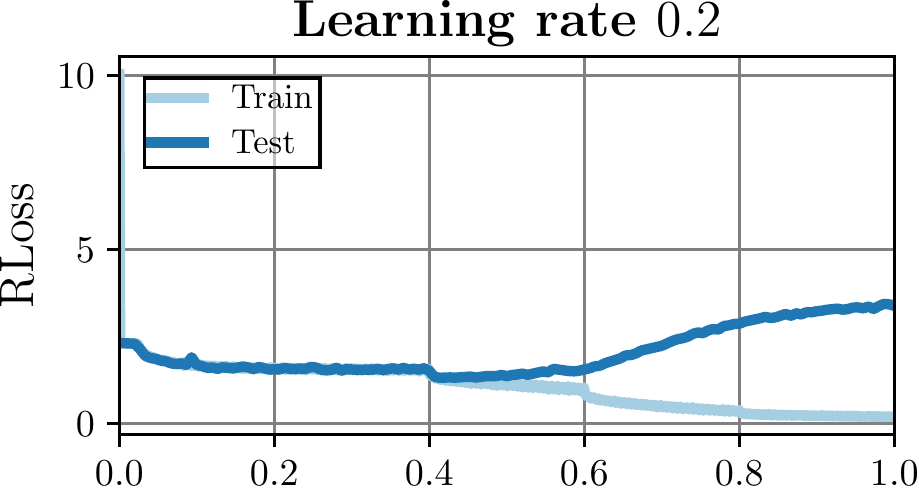}
		
		\hspace*{-0.25cm}
		\includegraphics[height=1.8cm]{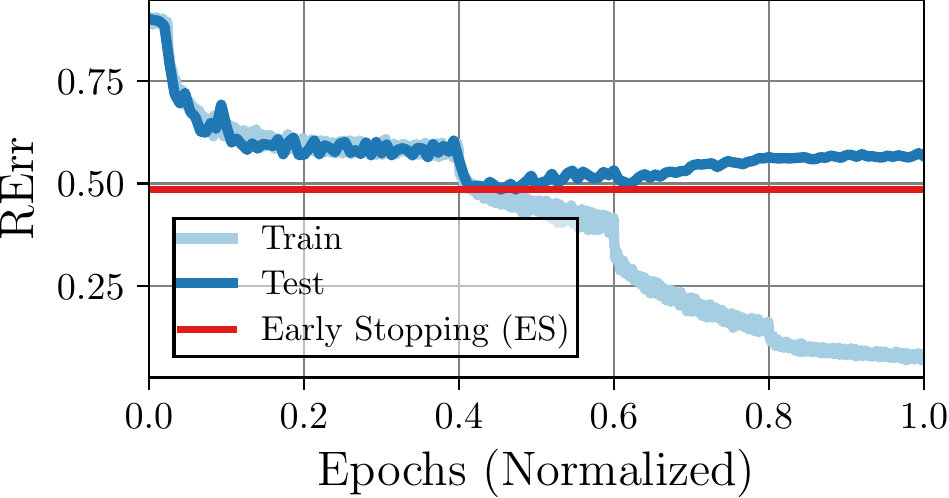}
	\end{minipage}
	\begin{minipage}[t]{0.19\textwidth}
		\vspace*{0px}
		
		\includegraphics[height=1.8cm]{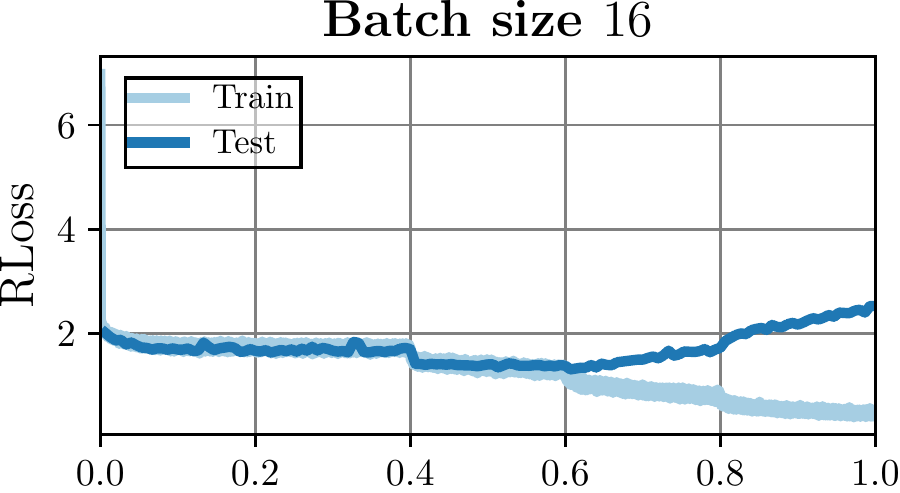}
		
		\hspace*{-0.25cm}
		\includegraphics[height=1.8cm]{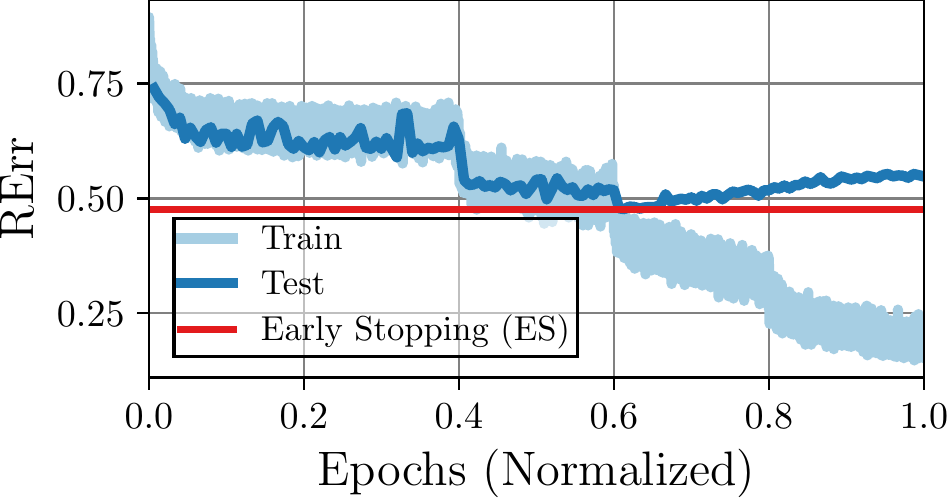}
	\end{minipage}
	\begin{minipage}[t]{0.19\textwidth}
		\vspace*{0px}
		
		\includegraphics[height=1.8cm]{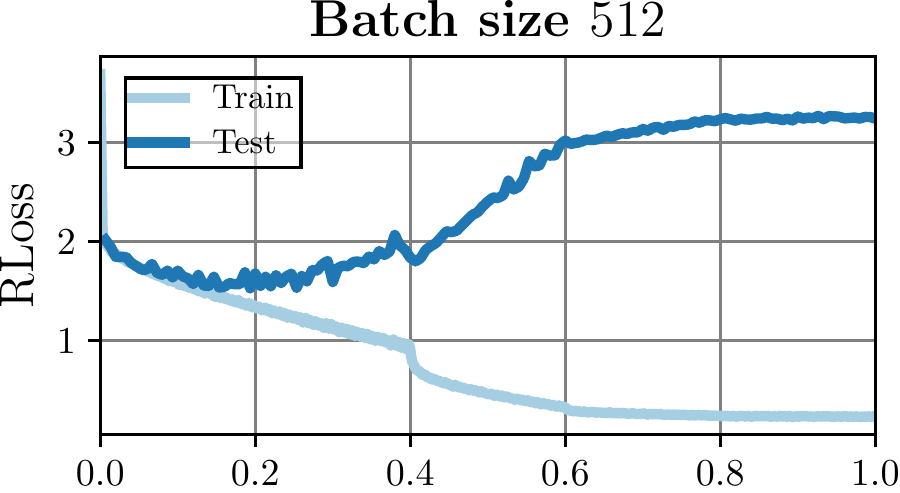}
		
		\hspace*{-0.25cm}
		\includegraphics[height=1.8cm]{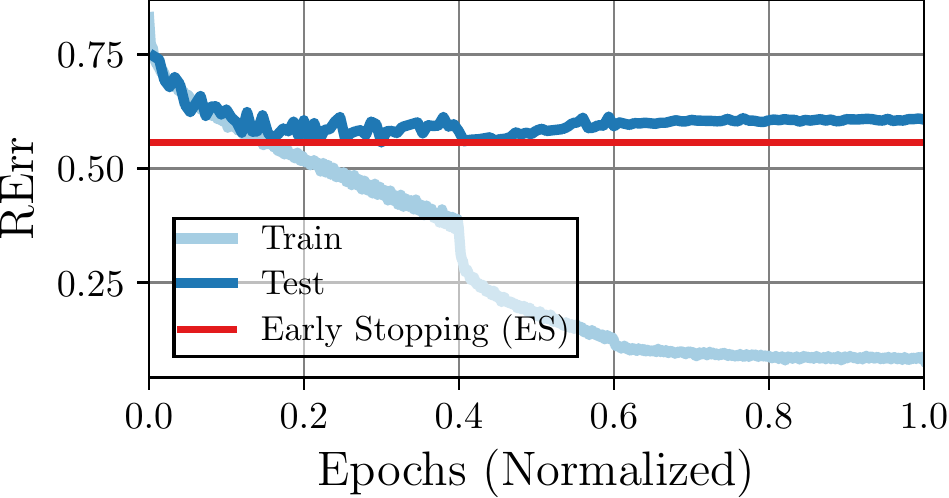}
	\end{minipage}
	\begin{minipage}[t]{0.19\textwidth}
		\vspace*{0px}
		
		\includegraphics[height=1.8cm]{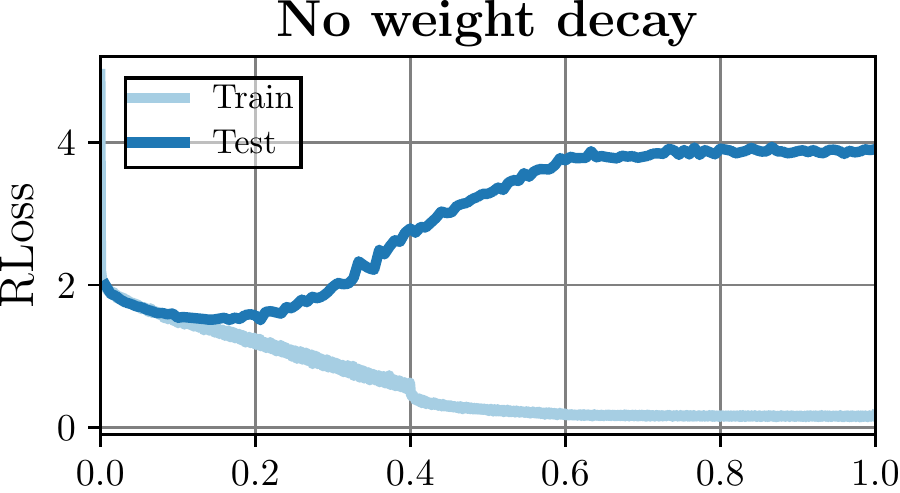}
		
		\hspace*{-0.25cm}
		\includegraphics[height=1.8cm]{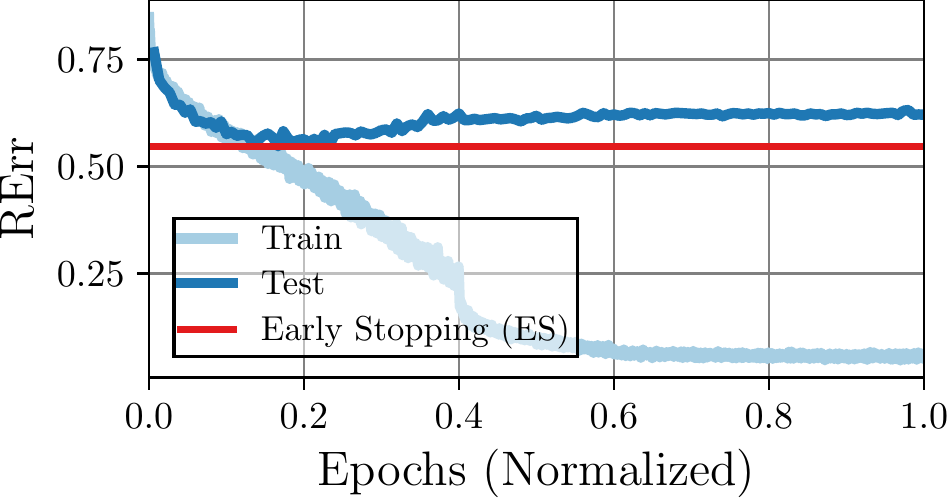}
	\end{minipage}
	\\[2.5px]
	\begin{minipage}[t]{0.19\textwidth}
		\vspace*{0px}
		
		\includegraphics[height=1.8cm]{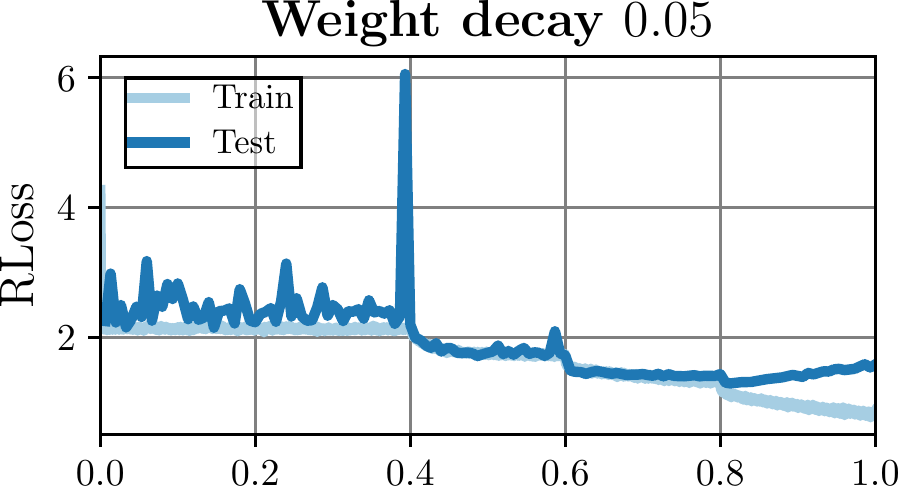}
		
		\hspace*{-0.25cm}
		\includegraphics[height=1.8cm]{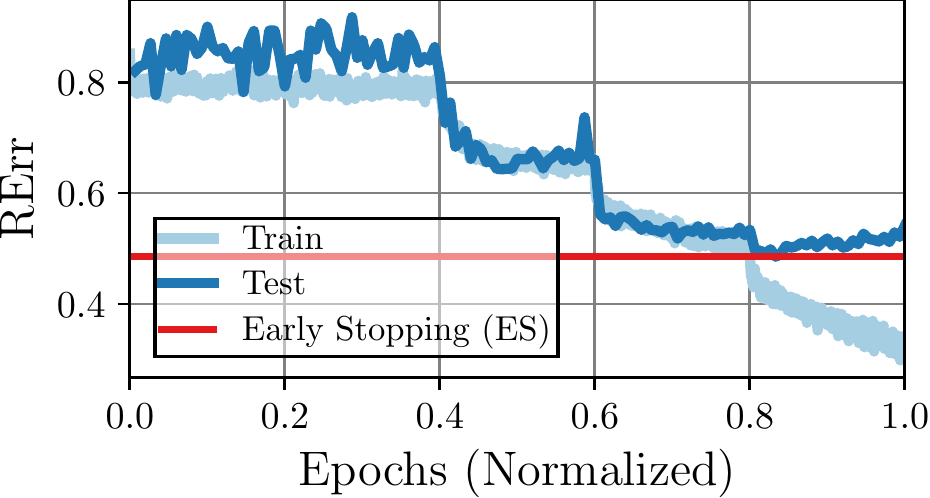}
	\end{minipage}
	\begin{minipage}[t]{0.19\textwidth}
		\vspace*{0px}
		
		\includegraphics[height=1.8cm]{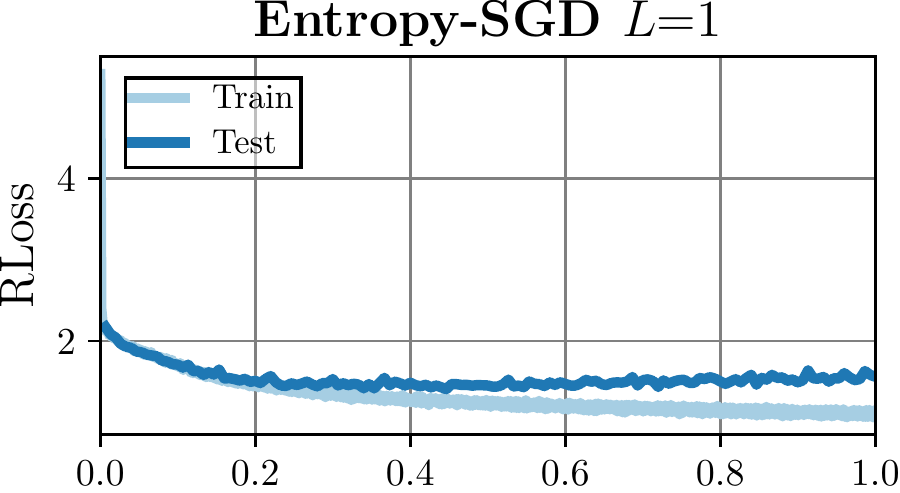}
		
		\hspace*{-0.25cm}
		\includegraphics[height=1.8cm]{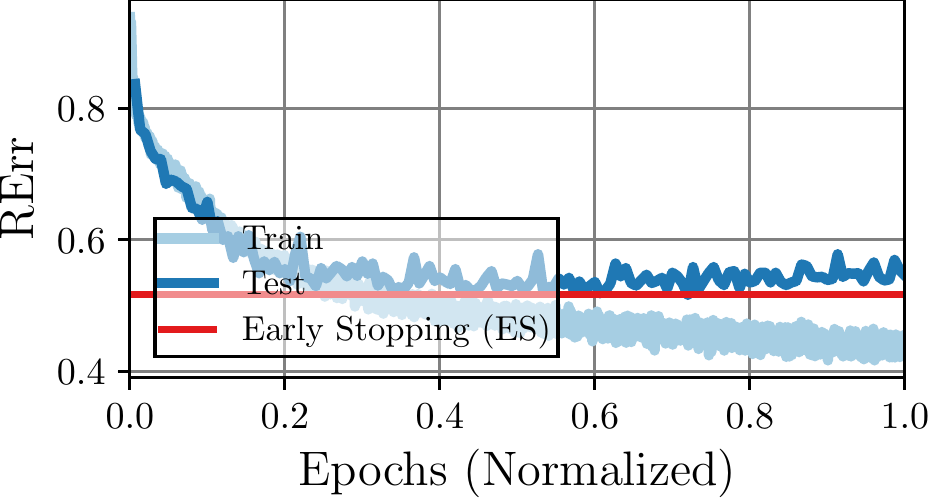}
	\end{minipage}
	\begin{minipage}[t]{0.19\textwidth}
		\vspace*{0px}
		
		\includegraphics[height=1.8cm]{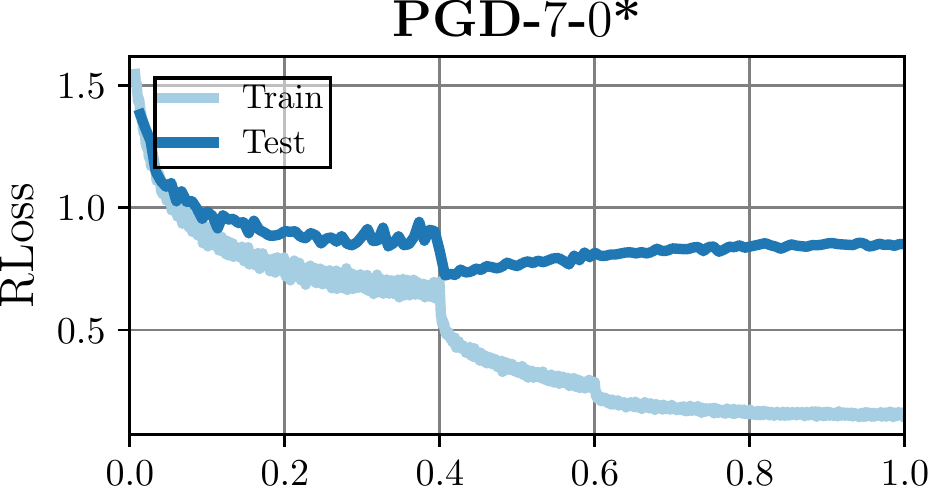}
		
		\hspace*{-0.15cm}
		\includegraphics[height=1.8cm]{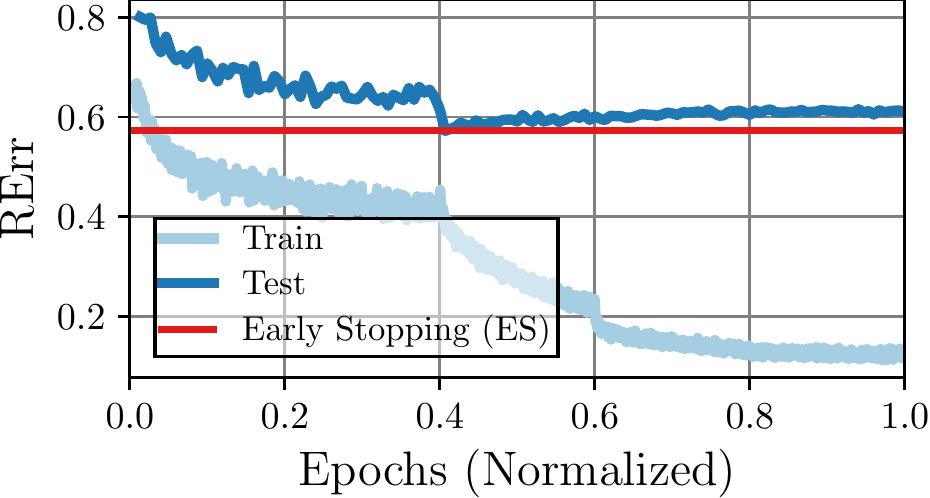}
	\end{minipage}
	\begin{minipage}[t]{0.19\textwidth}
		\vspace*{0px}
		
		\includegraphics[height=1.8cm]{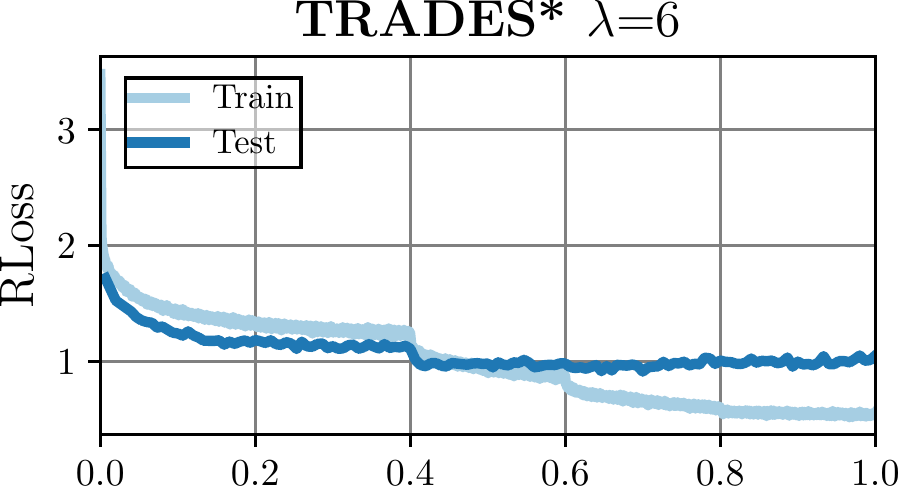}
		
		\hspace*{-0.25cm}
		\includegraphics[height=1.8cm]{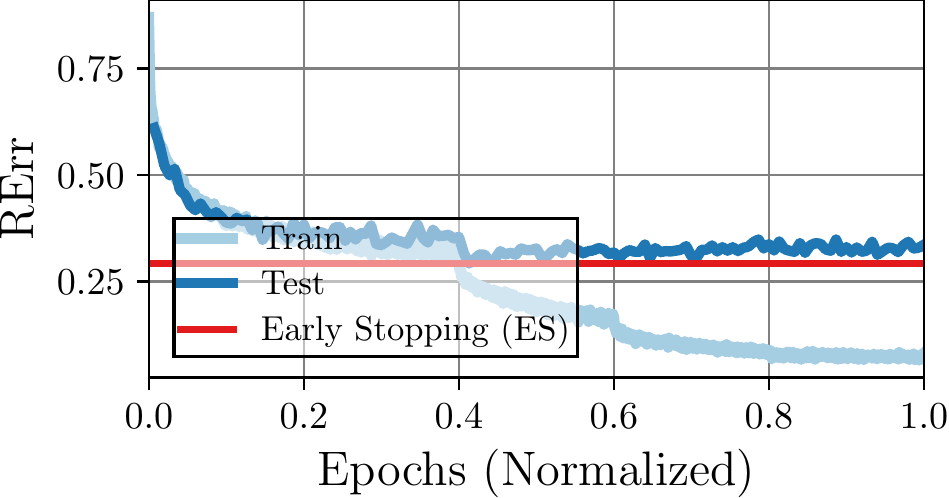}
	\end{minipage}
	\begin{minipage}[t]{0.19\textwidth}
		\vspace*{0px}
		
		\includegraphics[height=1.8cm]{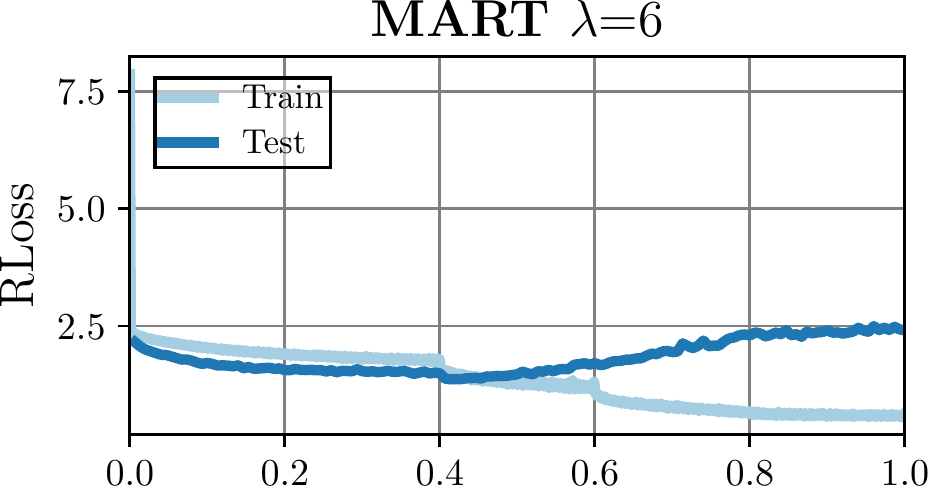}
		
		\hspace*{-0.25cm}
		\includegraphics[height=1.8cm]{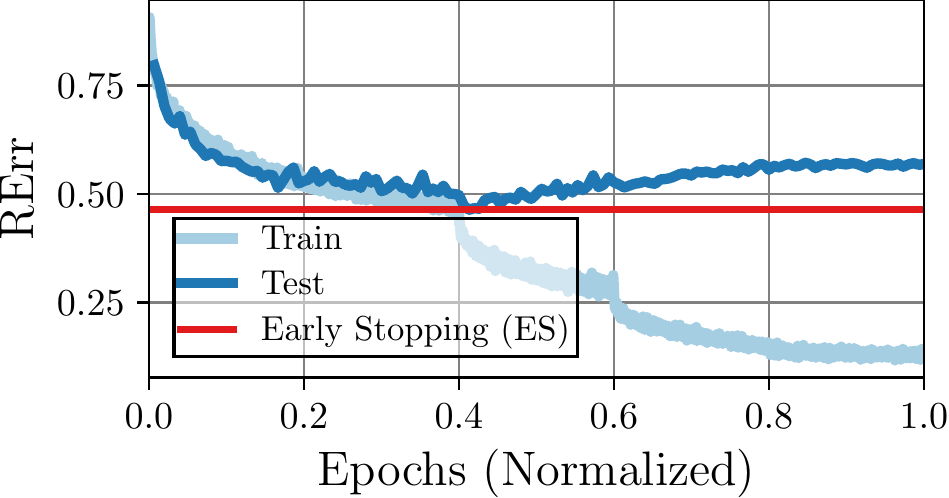}
	\end{minipage}
	\\[2.5px]
	
	\begin{minipage}[t]{0.19\textwidth}
		\vspace*{0px}
		
		\includegraphics[height=1.8cm]{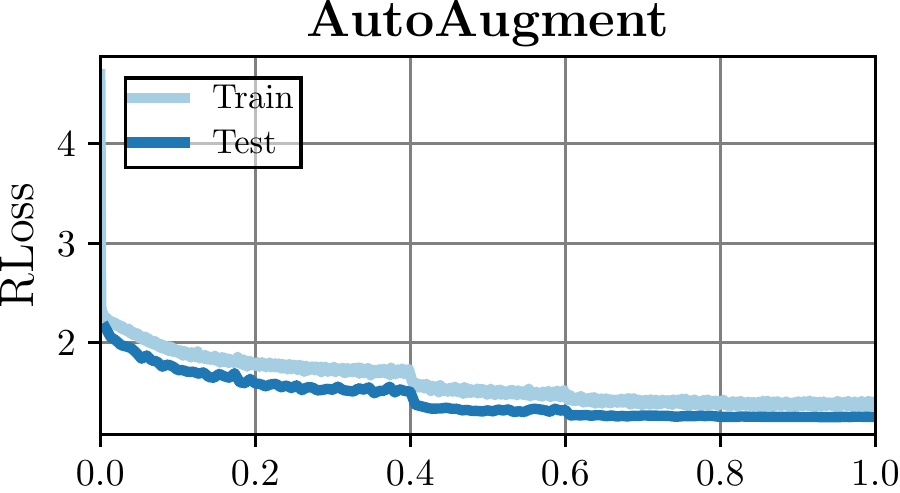}
		
		\hspace*{-0.15cm}
		\includegraphics[height=1.8cm]{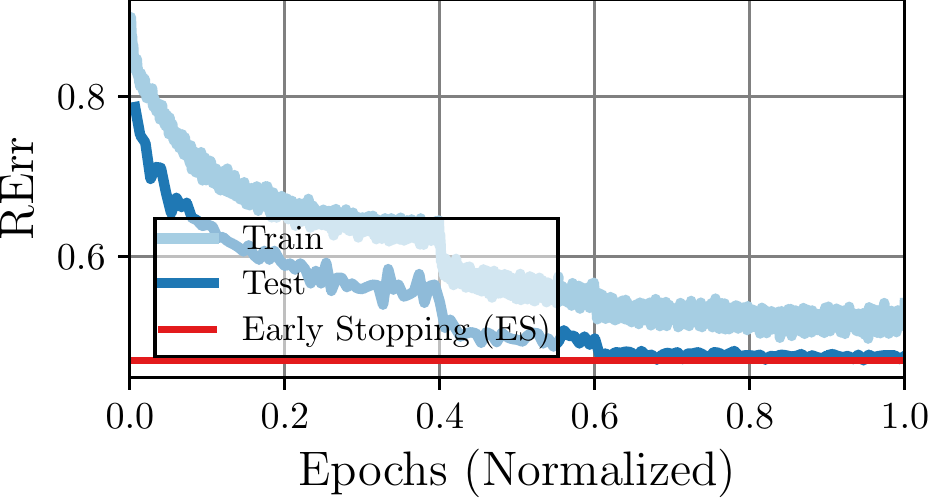}
	\end{minipage}
	\begin{minipage}[t]{0.19\textwidth}
		\vspace*{0px}
		
		\includegraphics[height=1.8cm]{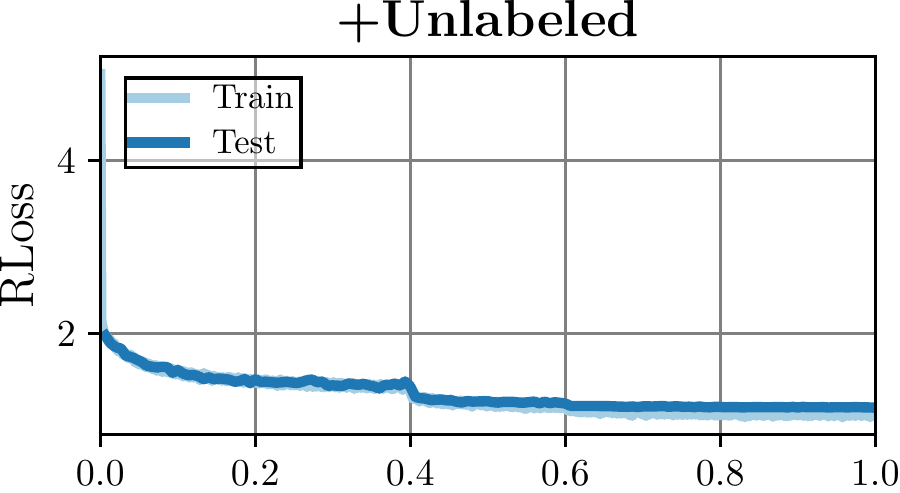}
		
		\hspace*{-0.25cm}
		\includegraphics[height=1.8cm]{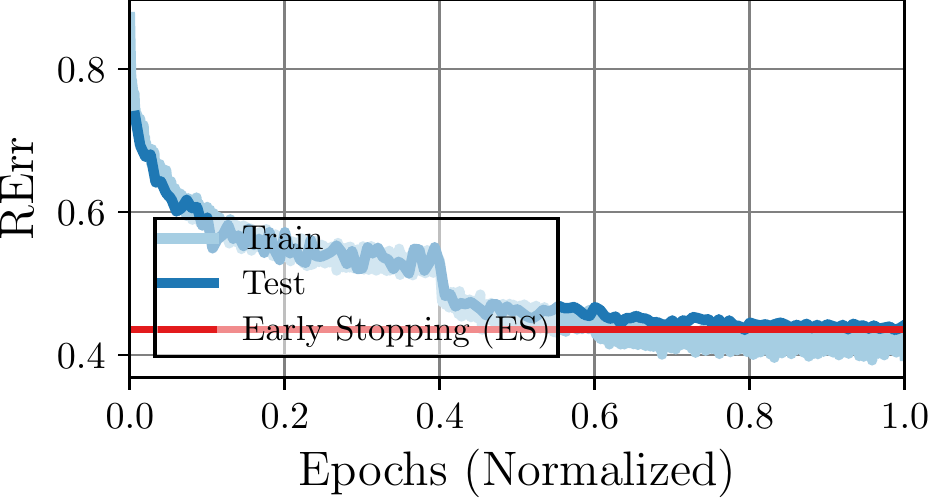}
	\end{minipage}
	\begin{minipage}[t]{0.19\textwidth}
		\vspace*{0px}
		
		\includegraphics[height=1.8cm]{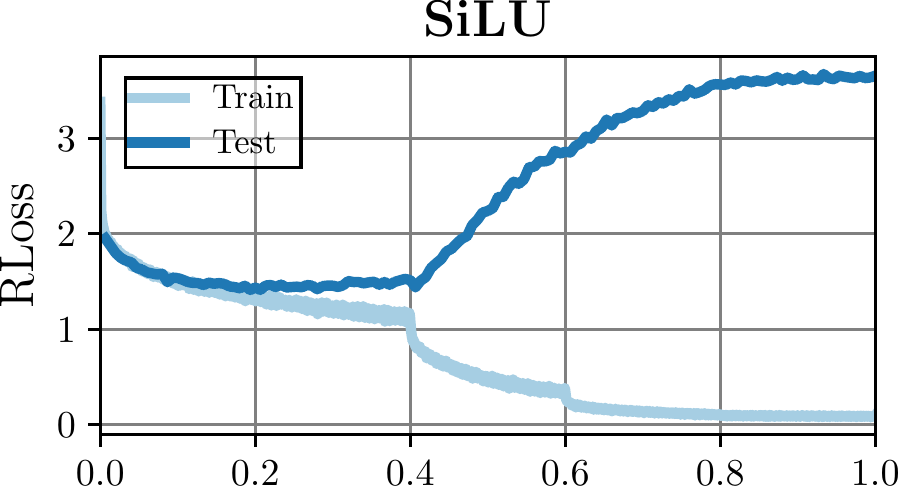}
		
		\hspace*{-0.25cm}
		\includegraphics[height=1.8cm]{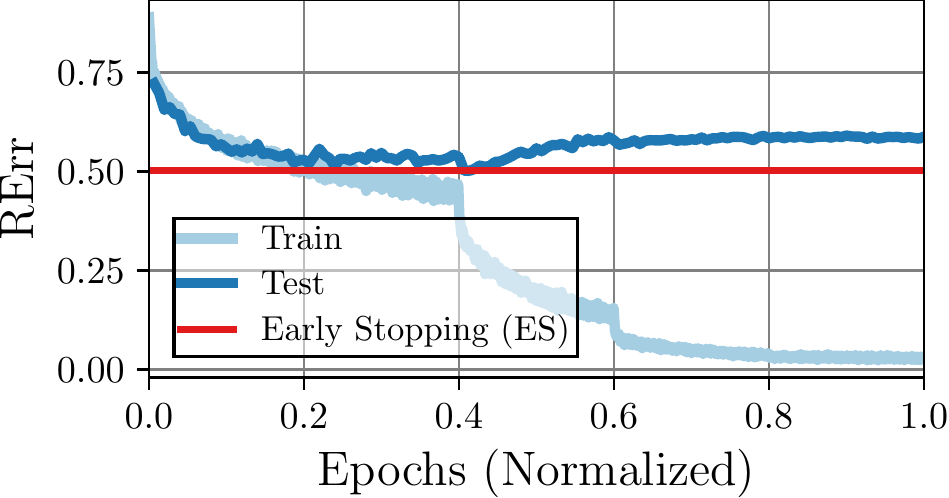}
	\end{minipage}
	\begin{minipage}[t]{0.19\textwidth}
		\vspace*{0px}
		
		\includegraphics[height=1.8cm]{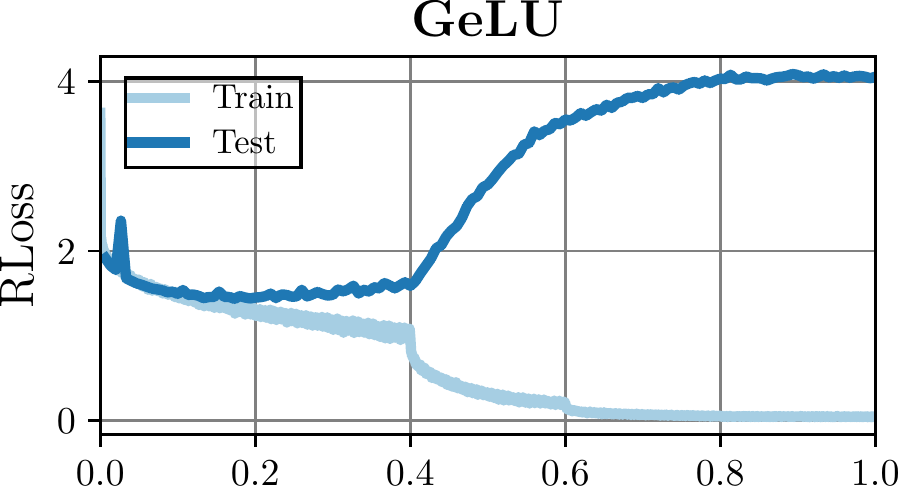}
		
		\hspace*{-0.25cm}
		\includegraphics[height=1.8cm]{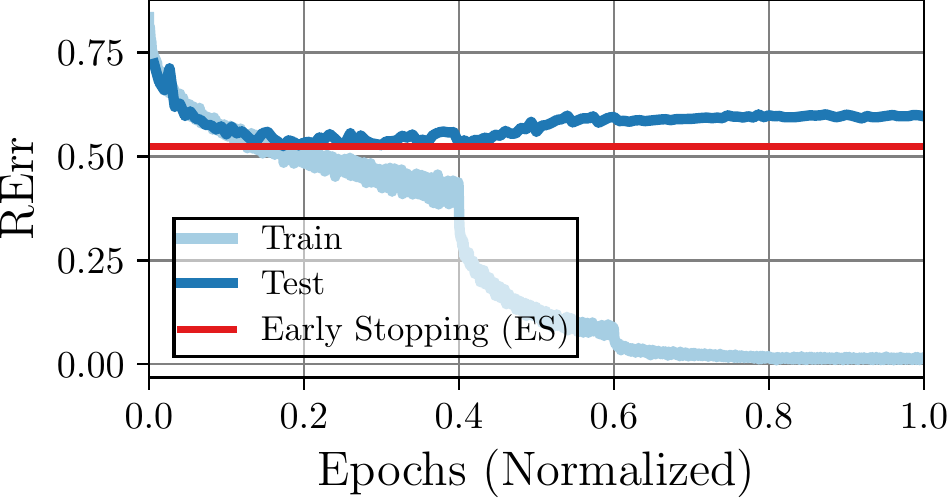}
	\end{minipage}
	\begin{minipage}[t]{0.19\textwidth}
		\vspace*{0px}
		
		\includegraphics[height=1.8cm]{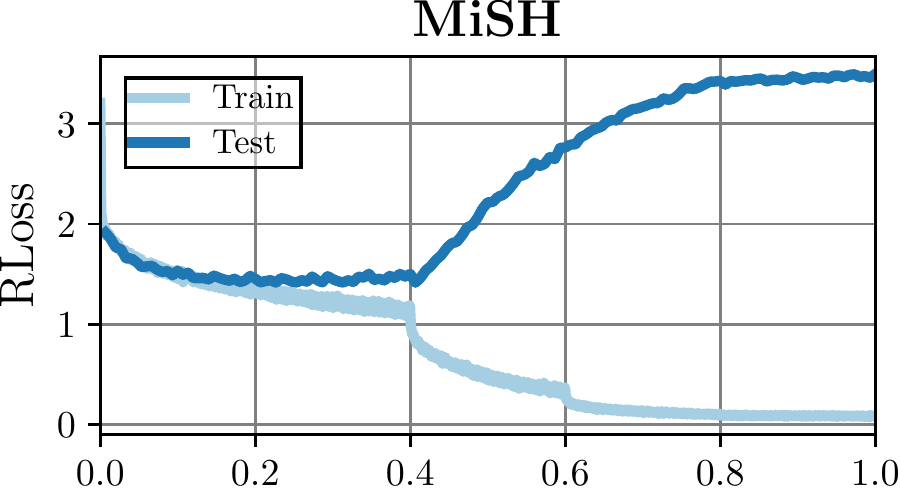}
		
		\hspace*{-0.25cm}
		\includegraphics[height=1.8cm]{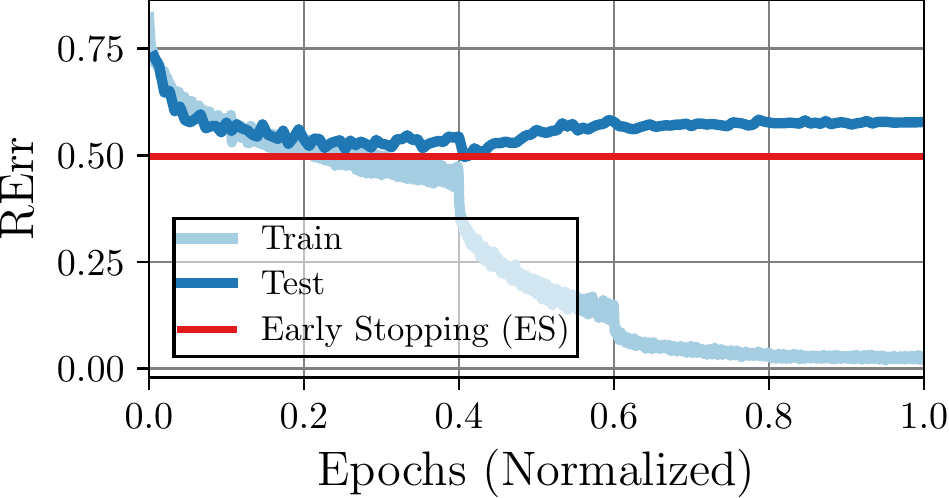}
	\end{minipage}
	\\[2.5px]
	
	\begin{minipage}[t]{0.19\textwidth}
		\vspace*{0px}
		
		\includegraphics[height=1.8cm]{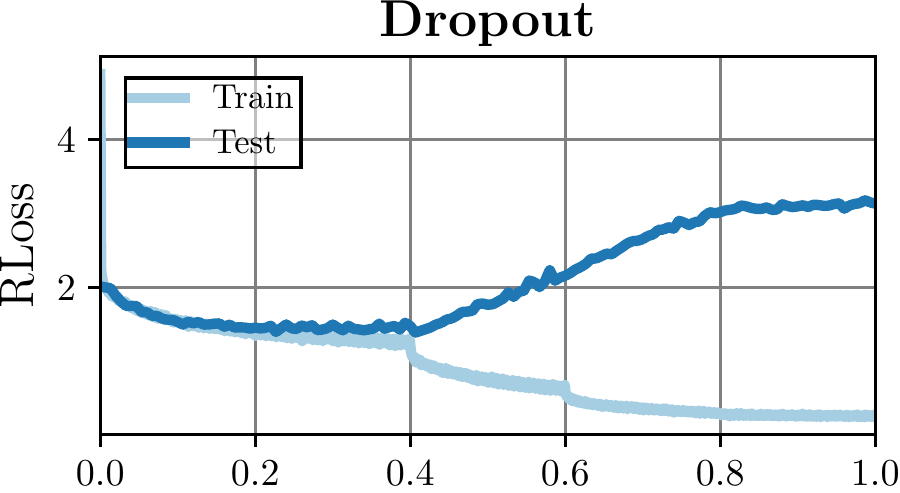}
		
		\hspace*{-0.15cm}
		\includegraphics[height=1.8cm]{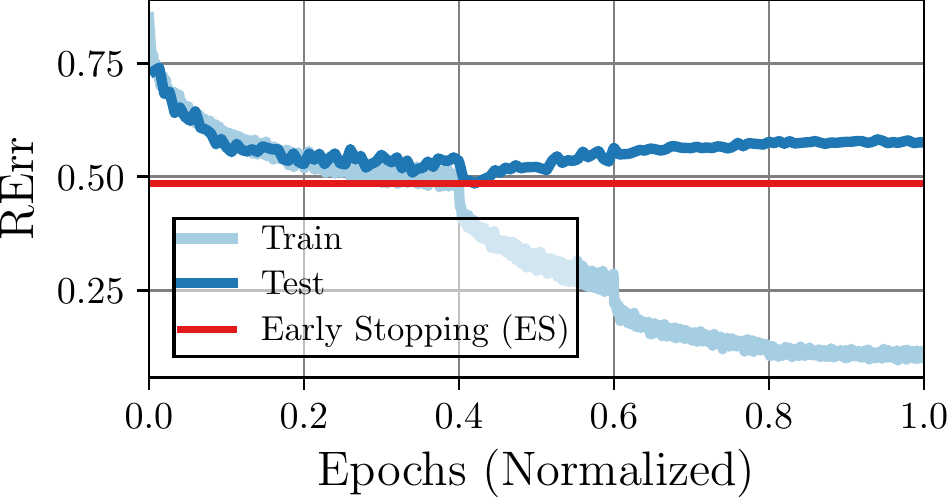}
	\end{minipage}
	\begin{minipage}[t]{0.19\textwidth}
		\vspace*{0px}
		
		\includegraphics[height=1.8cm]{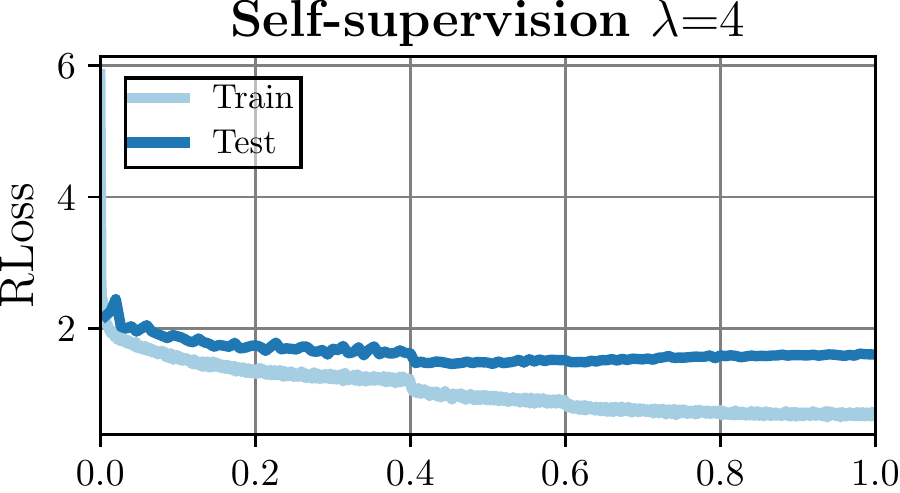}
		
		\hspace*{-0.25cm}
		\includegraphics[height=1.8cm]{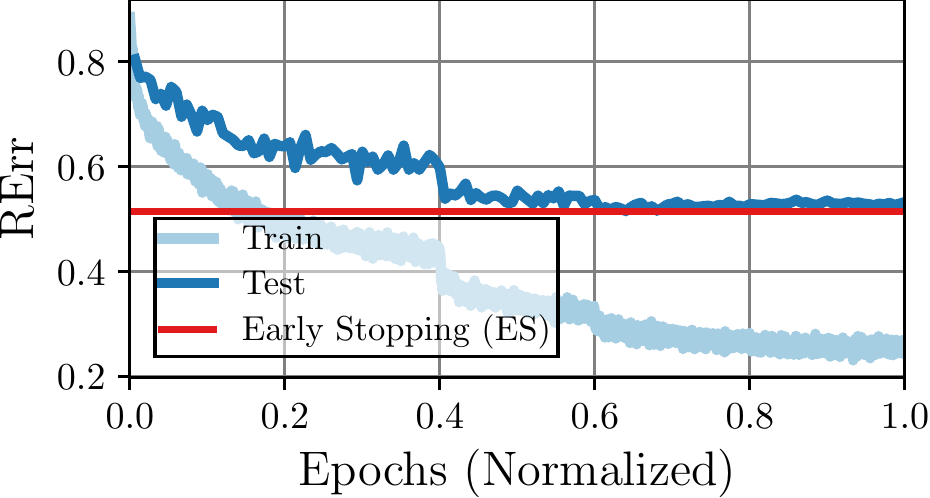}
	\end{minipage}
	\begin{minipage}[t]{0.19\textwidth}
		\vspace*{0px}
		
		\includegraphics[height=1.8cm]{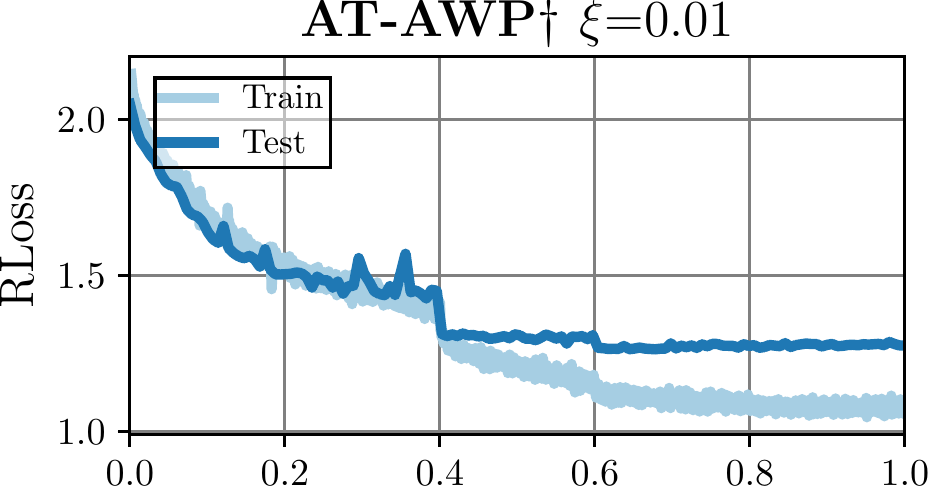}
		
		\hspace*{-0cm}
		\includegraphics[height=1.8cm]{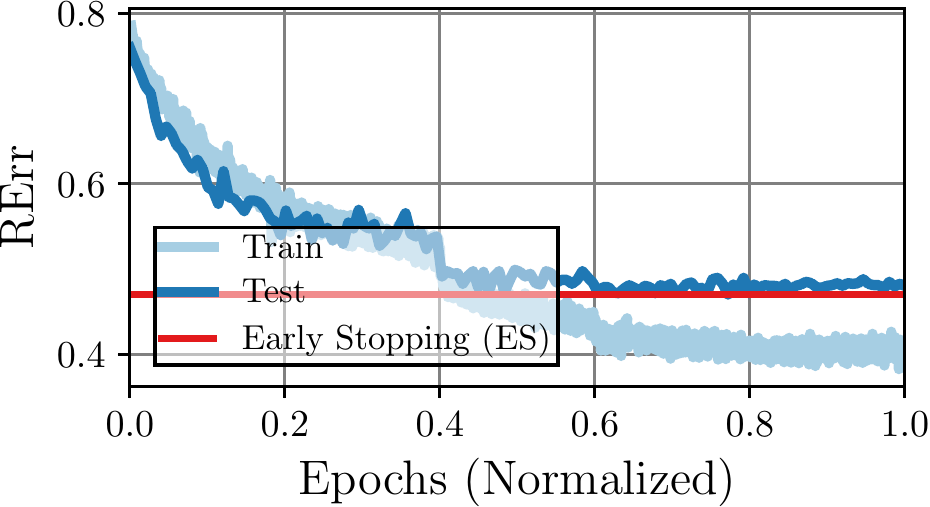}
	\end{minipage}
	\begin{minipage}[t]{0.01\textwidth}
		\vspace*{0px}
		
		\hfill
	\end{minipage}
	\begin{minipage}[t]{0.38\textwidth}
		\vspace*{4px}
		
		\caption{\textbf{Training Curves:} Test and train \RCE (top) and \RTE (bottom), including \RTE for early stopping, for all considered methods with selected hyper-parameters. \textbf{*} Train and test \RCE correspond to the attacks used during training, \eg, PGD-$\tau$ or maximizing KL-divergence for TRADES. \textbf{$\dagger$} Reported \RCE corresponds to \RCE on adversarial examples \emph{without} adversarial weights.}
		\label{fig:supp-training-curves}
	\end{minipage}
\end{figure*}

Training curves for all methods are shown in \figref{fig:supp-training-curves}.

\subsection{Flatness for Methods}
\label{sec:supp-methods-flatness}

\begin{figure*}[t]
\begin{minipage}{\textwidth}
	\centering
	\vspace*{-0.2cm}
	\hspace*{-0.4cm}
	\begin{minipage}[t]{0.18\textwidth}
		\vspace*{0px}
		
		\includegraphics[width=\textwidth]{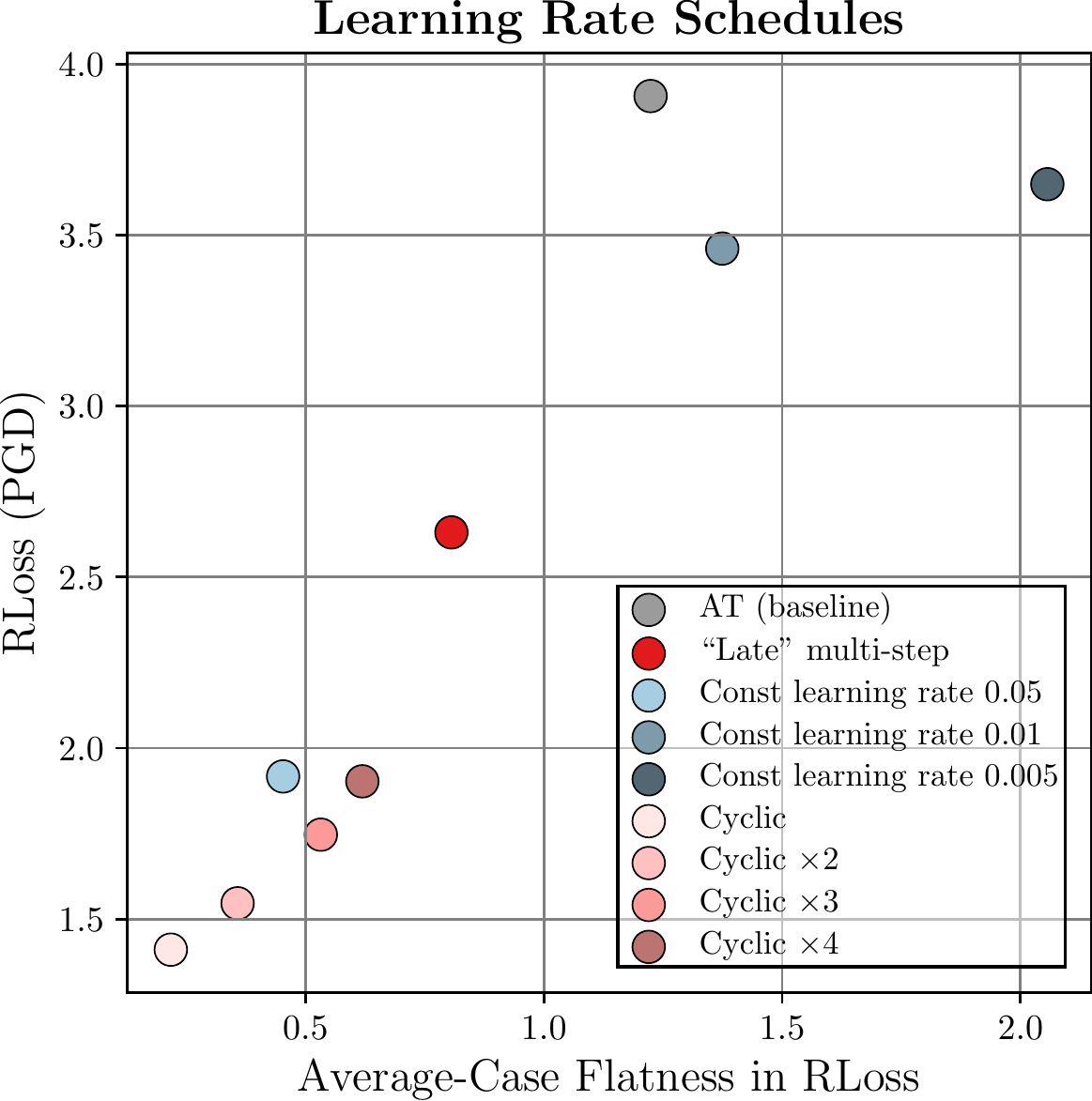}
	\end{minipage}
	\begin{minipage}[t]{0.18\textwidth}
		\vspace*{0px}
				
		\includegraphics[width=\textwidth]{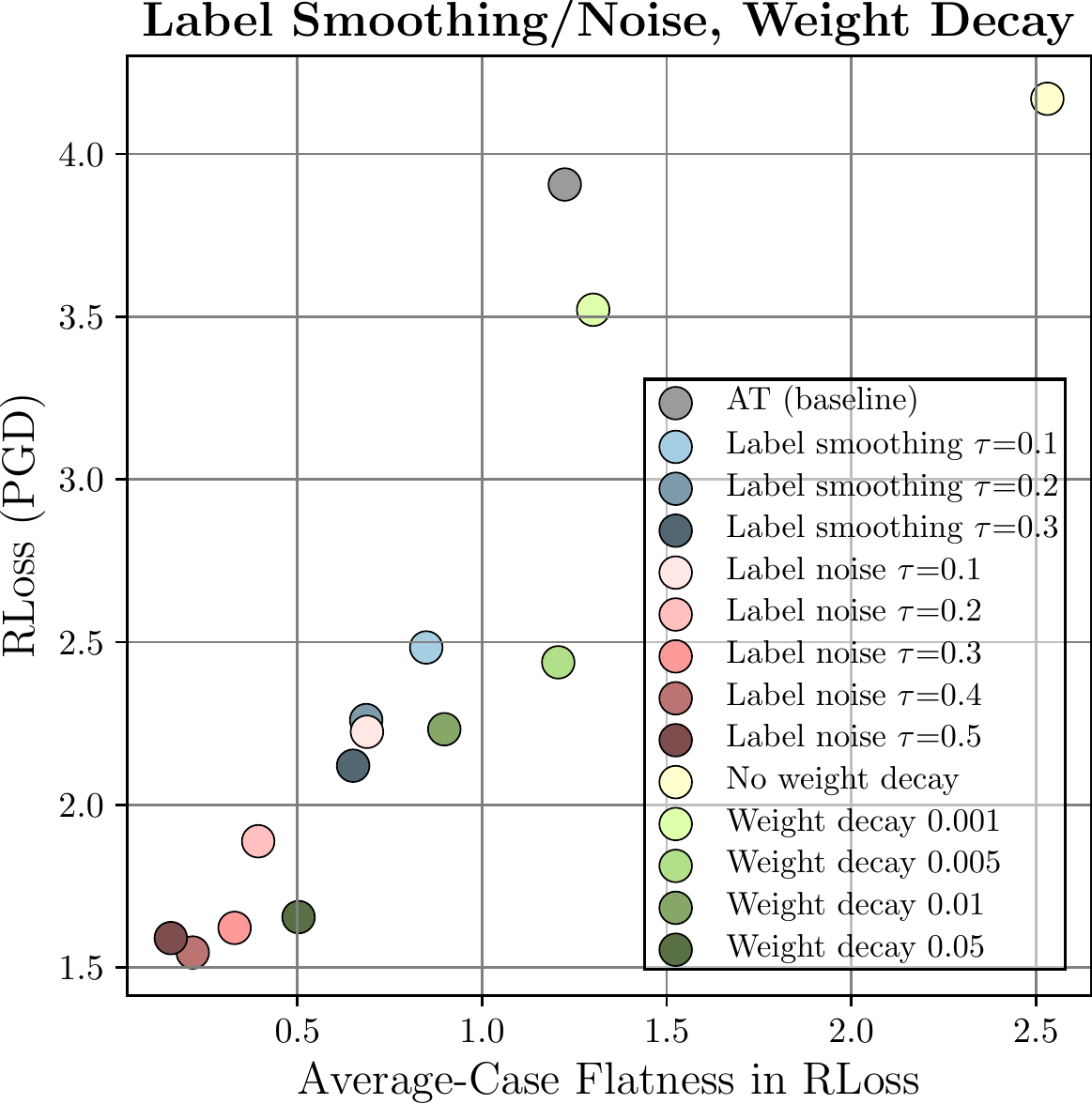}
	\end{minipage}
	\begin{minipage}[t]{0.18\textwidth}
		\vspace*{0px}
				
		\includegraphics[width=\textwidth]{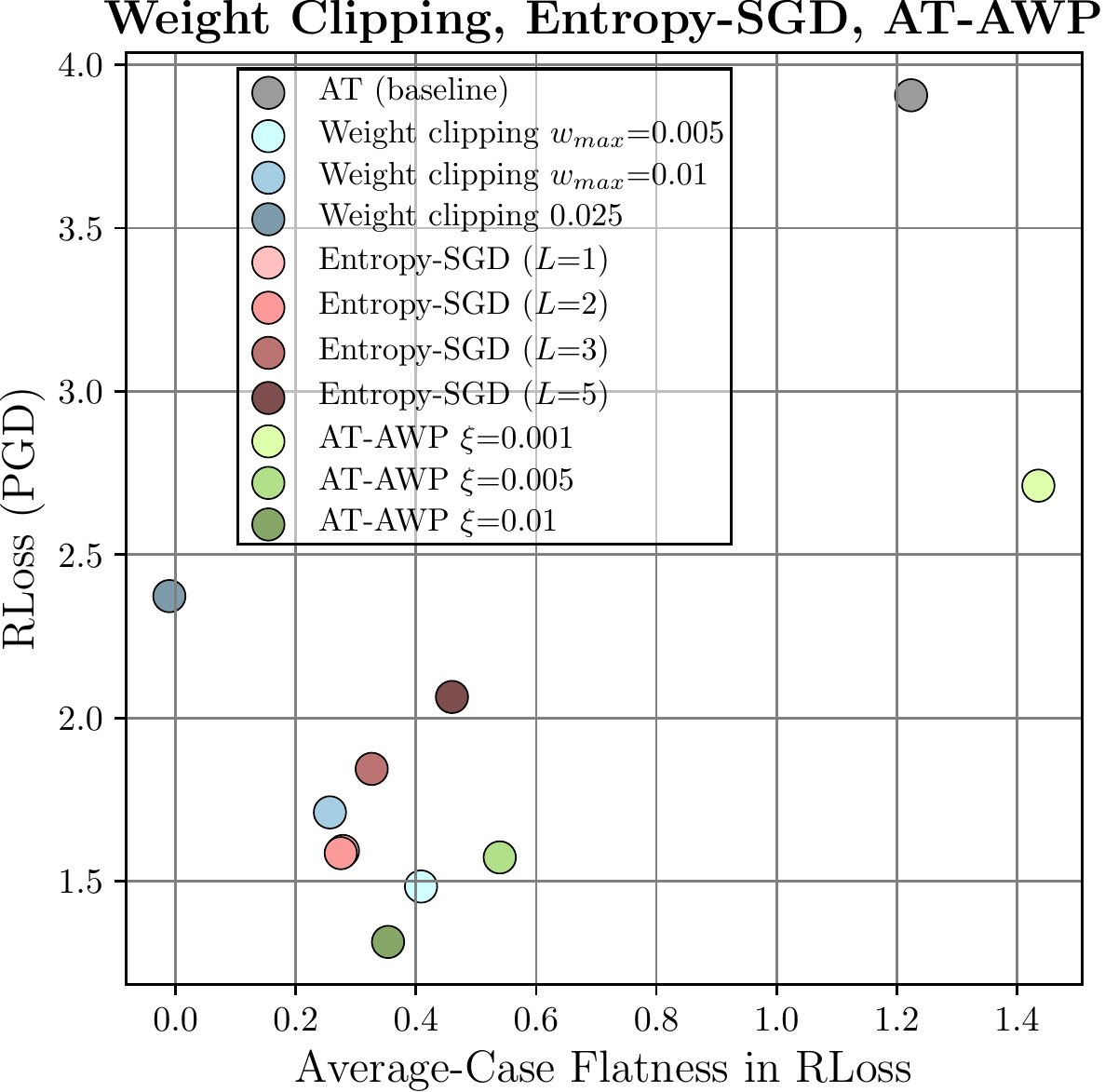}
	\end{minipage}
	\begin{minipage}[t]{0.26\textwidth}
		\vspace*{0px}
				
		\includegraphics[width=1\textwidth]{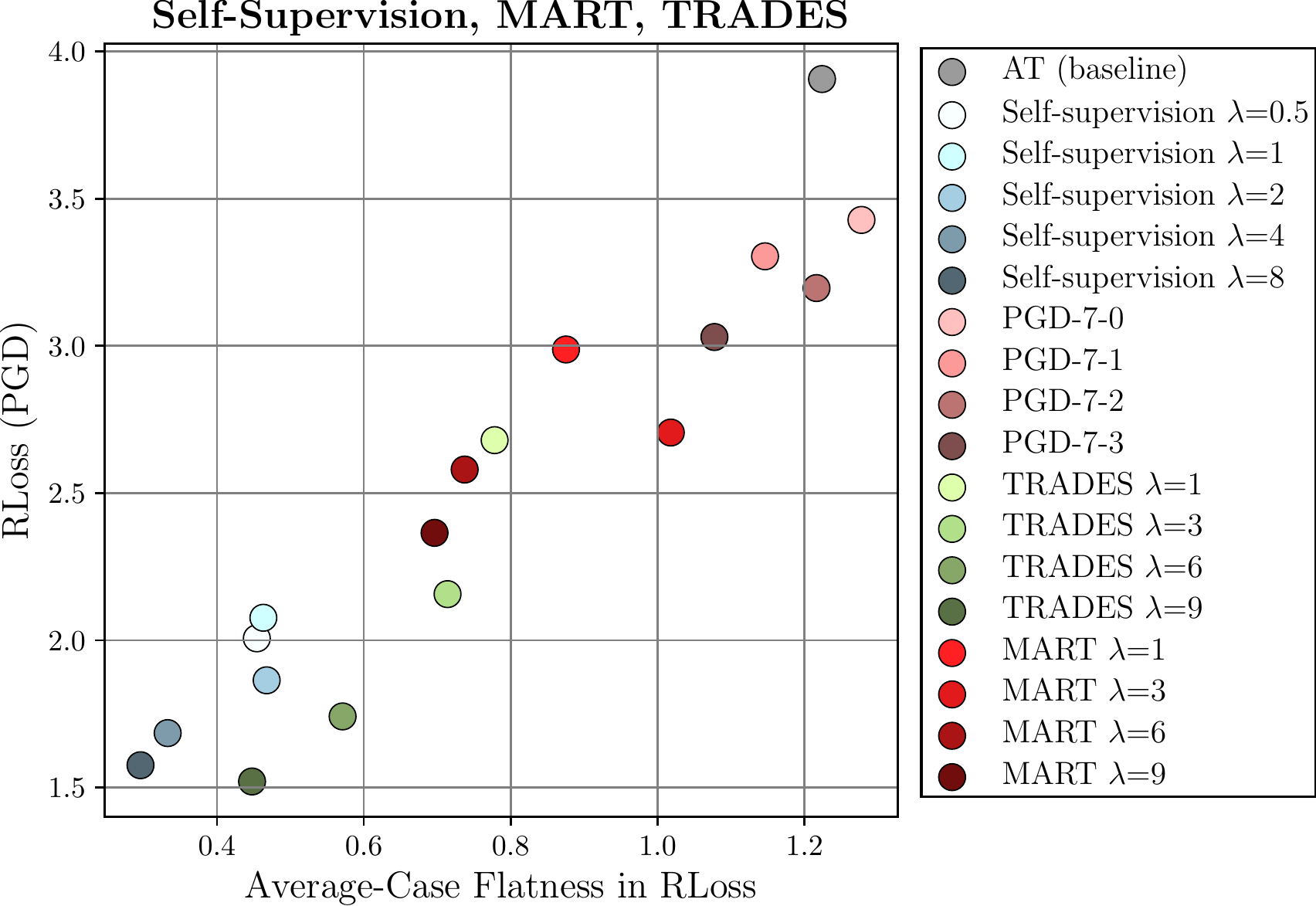}
	\end{minipage}
	\begin{minipage}[t]{0.01\textwidth}
		\vspace*{0px}
		
		\hspace*{4px}{\color{black!75}\rule{0.65px}{3.15cm}}
	\end{minipage}
	\begin{minipage}[t]{0.18\textwidth}
		\vspace*{0px}
				 
		\includegraphics[width=1\textwidth]{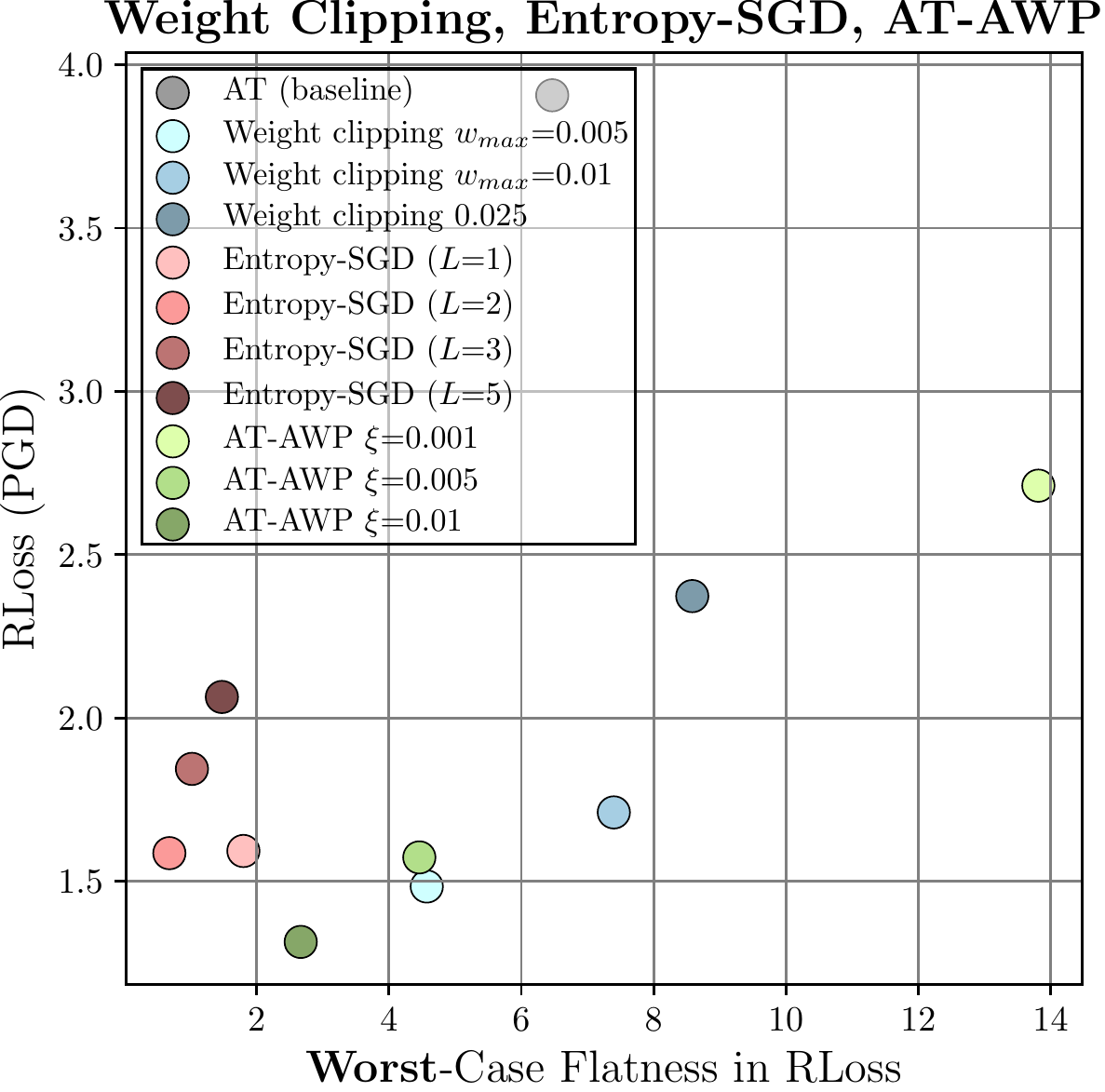}
	\end{minipage}

	\vspace*{-6px}
	\captionof{figure}{\textbf{Robustness and Flatness for Varying Hyper-Parameters:} \textbf{Left:} \RCE (y-axis) plotted against average-case flatness of \RCE (x-axis) for various groups of methods: learning rate schedules (left), label smoothing/noise and weight decay (middle left), weight clipping, Entropy-SGD and AT-AWP (middle right) as well as AT with self-supervision, MART and TRADES (right). As outlined in \secref{sec:supp-methods}, we considered multiple hyper-parameter settings per method and show that favorable hyper-parameters in terms of adversarial robustness also result in improved flatness. That is, in most cases, varying hyper-parameters creates (roughly) a diagonal line in these plots. Interestingly, weight clipping can be considered an outlier: adversarial robustness improves while \emph{average-case} flatness \emph{reduces}. \textbf{Right:} \RCE (y-axis) plotted against \emph{worst-case} flatness in \RCE (x-axis). Here, flatness for weight clipping aligns well with \RCE.}
	\label{fig:supp-flatness-methods}
	\vspace*{12px}
\end{minipage}
\end{figure*}

\textbf{Flatness Throughout Training:}
\figref{fig:supp-methods-flatness-epochs} shows \RCE (y-axis) plotted against average-case flatness in \RCE (x-axis) throughout training, \ie, over epochs ({\color{blue!50!black}dark blue} to {\color{red!50!black}dark red}), for methods not shown in the main paper. Strikingly, using higher $\epsilon{=}\nicefrac{9}{255}$ or alternative activation functions (SiLU \cite{ElfwingNN2018}, GeLU \cite{HendrycksARXIV2016} or MiSH \cite{MisraBMVC2020}) affect neither robust overfitting nor flatness significantly. Interestingly, as discussed in the main paper, label smoothing avoids sharper minima during overfitting, but does \emph{not} avoid an increased \RCE. Methods that consistently reduce or avoid robust overfitting, \eg, weight clipping, label noise, strong weight decay or AutoAugment, avoid both the increase in \RCE as well as worse flatness. Clearly, the observations from the main paper are confirmed: flatness usually reduces alongside \RCE in robust overfitting.

\textbf{Flatness Across Hyper-Parameters:}
In \figref{fig:supp-flatness-methods}, we consider flatness when changing hyper-parameters of selected methods. As before, we plot \RCE (y-axis) against average-case flatness in \RCE (x-axis) for various groups of methods: learning rate schedules (first column), label smoothing/noise and weight decay (second column), methods explicitly improving flatness, \ie, weight clipping, Entropy-SGD and AT-AWP (third column), as well as self-supervision, MART and TRADES (fourth column). Except for weight clipping, hyper-parameter settings with improved adversarial robustness also favor flatter minima. In most cases, this relationship follows a clear, diagonal line. For weight clipping, in contrast, the relationship is reversed: improved flatness reduces \RCE. Thus, \figref{fig:supp-flatness-methods} (fifth column) considers worst-case flatness in \RCE. Here, ``stronger'' weight clipping improves both robustness \emph{and} flatness. This supports our discussion in the main paper: methods need at least ``some kind'' of flatness, average- or worst-case, in order to improve adversarial robustness.

\section{Results in Tabular Form}
\label{sec:supp-results}

\tabref{tab:supp-table-error} and \ref{tab:supp-table-loss} report the quantitative results from all our experiments. Besides flatness in \RCE, we also report both average- and worst-case flatness in (clean) \CE. As described in the main paper, we use $\xi = 0.5$ for average-case flatness and $\xi = 0.003$ for worst-case flatness. In \tabref{tab:supp-table-error}, methods are sorted (in ascending order) by \RTE against AutoAttack \cite{CroceICML2020}. Additionally, we split all methods into four groups: \colorbox{colorbrewer3!15}{good}, \colorbox{colorbrewer5!15}{average}, \colorbox{colorbrewer1!15}{poor} and \colorbox{colorbrewer0!15}{worse} robustness at $57\%$, $60\%$ and $62.8\%$ \RTE. These thresholds correspond roughly to the $30\%$ and $70\%$ percentile of all methods with $\RTE \leq 62.8\%$. As our AT baseline obtains $62.8\%$ \RTE, we group all methods with higher \RTE than $62.8\%$ in \colorbox{colorbrewer0!15}{worse} robustness. In \tabref{tab:supp-table-loss}, methods are sorted (in ascending order) by \RCE against PGD. Finally, \tabref{tab:supp-table-robustbench-error} and \ref{tab:supp-table-robustbench-loss} report \RTE and \RCE, together with our average- and worst-case flatness (of \RCE) measures for the evaluated, pre-trained models from RobustBench \cite{CroceARXIV2020b}.

\newpage
\clearpage
\begin{figure*}
\begin{minipage}{\textwidth}
	\centering
	\scriptsize
	{
	\begin{tabularx}{\textwidth}{|X|c|c|c||c|c|c||c|c|}
		\hline
		Model & \multicolumn{3}{c||}{\bfseries Test Robustness} & \multicolumn{3}{c||}{\bfseries Train Robustness} & \multicolumn{2}{c|}{\bfseries Flatness}\\
		\hline
		(sorted by \RCE on AA) & \TE & \RTE & \RTE & \TE & \RTE & \RTE & Avg & Worst\\
		(PGD = PGD-$20$, $10$ restarts) & (test) & (test) & (test) & (train) & (train) & (train) & \RCE & \RCE\\
		(AA = AutoAttack \cite{CroceARXIV2020}) && (PGD) & (AA) && (PGD) & (AA) &&\\
		\hline
		\hline
		Carmon et al. [16] & 10.31 & 37.6 & 40.8 & 1.93 & 16.8 & 19.2 & 0.7 & 0.34\\
		Engstrom et al. [36] & 12.97 & 45.3 & 49.2 & 6.71 & 33.1 & 36.3 & 0.23 & 0.51\\
		Pang et al. [93] & 14.87 & 36.6 & 45.8 & 7.79 & 20.5 & 28.6 & 0.08 & 0.07\\
		Wang [238] & 12.5 & 37.1 & 42.8 & 8.07 & 24.8 & 32.1 & 0.61 & 0.34\\
		Wong et al. [131] & 16.66 & 54.4 & 57.6 & 11.86 & 44.9 & 49.2 & 0.3 & 0.16\\
		Wu et al. [133] & 14.64 & 41.5 & 43.9 & 2.2 & 14.5 & 16.5 & 0.49 & 0.09\\
		Zhang et al. [148] & 15.08 & 44.1 & 46.4 & 7.83 & 29.9 & 33.6 & 0.61 & 0.43\\
		Zhang et al. [149] & 15.48 & 43 & 47.2 & 4.85 & 26.3 & 30.1 & 0.51 & 0.13\\
		\hline
	\end{tabularx} 
	}
	\vspace*{-8px}
	\captionof{table}{\textbf{RobustBench \cite{CroceARXIV2020b}: \TE, \RTE and Flatness in \RCE:} \TE and \RTE on train and test examples as well as average- and worst-case flatness in \RCE for pre-trained models from RobustBench. In contrast to \tabref{tab:supp-table-error}, the RobustBench models were obtained using early stopping.}
	\label{tab:supp-table-robustbench-error}
	\vspace*{12px}
\end{minipage}
\begin{minipage}{\textwidth}
	\centering
	\scriptsize
	{
	\begin{tabularx}{\textwidth}{|X|c|c|c||c|c|c||c|c|}
		\hline
		Model & \multicolumn{3}{c||}{\bfseries Test Robustness} & \multicolumn{3}{c||}{\bfseries Train Robustness} & \multicolumn{2}{c|}{\bfseries Flatness}\\
		\hline
		(sorted by \RCE on AA) & \CE & \RCE & \RCE & \CE & \RCE & \RCE & Avg & Worst\\
		(PGD = PGD-$20$, $10$ restarts) & (test) & (test) & (test) & (train) & (train) & (train) & \RCE & \RCE\\
		(AA = AutoAttack \cite{CroceARXIV2020}) && (PGD) & (AA) && (PGD) & (AA) &&\\
		\hline
		\hline
		Carmon et al. [16] & 0.53 & 1.02 & 0.63 & 0.36 & 0.62 & 0.41 & 0.7 & 0.34\\
		Engstrom et al. [36] & 0.44 & 1.25 & 0.59 & 0.29 & 0.82 & 0.41 & 0.23 & 0.51\\
		Pang et al. [93] & 1.84 & 1.98 & 1.86 & 1.8 & 1.91 & 1.8 & 0.08 & 0.07\\
		Wang [128] & 0.64 & 1.11 & 0.73 & 0.54 & 0.9 & 0.6 & 0.61 & 0.34\\
		Wong et al. [131] & 0.57 & 1.37 & 0.73 & 0.46 & 1.11 & 0.61 & 0.3 & 0.16\\
		Wu et al. [133] & 0.63 & 1.13 & 0.72 & 0.37 & 0.61 & 0.41 & 0.49 & 0.09\\
		Zhang et al. [148] & 0.55 & 1.19 & 0.66 & 0.39 & 0.83 & 0.48 & 0.61 & 0.43\\
		Zhang et al. [149] & 0.85 & 1.27 & 0.93 & 0.71 & 1.01 & 0.76 & 0.51 & 0.13\\
		\hline
	\end{tabularx} 
	}
	\vspace*{-8px}
	\captionof{table}{\textbf{RobustBench \cite{CroceARXIV2020b}: \emph{\CE, \RCE} and Flatness in \RCE:} \CE and \RCE on train and test examples as well as average- and worst-case flatness in \RCE for pre-trained models from RobustBench. In contrast to \tabref{tab:supp-table-loss}, the RobustBench models were obtained using early stopping.}
	\label{tab:supp-table-robustbench-loss}
	\vspace*{-6px}
\end{minipage}
\end{figure*}
\FloatBarrier
\begin{table*}[t]
	\centering
	\vspace*{-0.5cm}
	\scriptsize
	{
	\begin{tabularx}{\textwidth}{|X|c|c|c||c|c|c||c|c||c|c|c|c|}
		\hline
		Model & \multicolumn{3}{c||}{\bfseries Test Robustness} & \multicolumn{3}{c||}{\bfseries Train Robustness} & \multicolumn{2}{c||}{\bfseries Early Stopping} & \multicolumn{4}{c|}{\bfseries Flatness}\\
		\hline
		(sorted by \RTE on AA) & \TE & \RTE & \RTE & \TE & \RTE & \RTE & \RTE & \RTE & Avg & Worst & Avg & Worst\\
		(PGD = PGD-$20$, $10$ restarts) & (test) & (test) & (test) & (train) & (train) & (train) & (stop) & (stop) & \CE & \CE & \bfseries\RCE & \bfseries\RCE\\
		(AA = AutoAttack \cite{CroceARXIV2020}) && (PGD) & (AA) && (PGD) & (AA) & (PGD) & (AA) &&&&\\
		\hline
		\hline
		\rowcolor{colorbrewer3!15}\hspace*{2px} +Unlabeled & 16.96 & 45.9 & 48.9 & 12.6 & 38.6 & 43.2 & 45.3 & 48.9 & 0.12 & 4.64 & 0.32 & 1.2\\
		\rowcolor{colorbrewer3!15}\hspace*{2px} Cyclic $\times2$ & 19.66 & 51.2 & 53.6 & 7.64 & 32.3 & 35.4 & 51 & 53.6 & 0.09 & 3.93 & 0.35 & 1.5\\
		\rowcolor{colorbrewer3!15}\hspace*{2px} AutoAugment & 16.89 & 49.5 & 54.0 & 12.25 & 42.8 & 47.9 & 49.5 & 53.5 & 0.13 & 15.01 & 0.49 & 0.69\\
		\rowcolor{colorbrewer3!15}\hspace*{2px} AT-AWP $\xi{=}0.01$ & 21.4 & 50.7 & 54.3 & 13.52 & 37.4 & 43.1 & 48.9 & 53.6 & 0.12 & 6.17 & 0.35 & 2.68\\
		\rowcolor{colorbrewer3!15}\hspace*{2px} AT-AWP $\xi{=}0.005$ & 20.05 & 52.5 & 55 & 7.34 & 28.1 & 31.8 & 50.8 & 53.3 & 0.15 & 6.98 & 0.54 & 4.46\\
		\rowcolor{colorbrewer3!15}\hspace*{2px} Label noise $\tau{=}0.4$ & 20.56 & 52.4 & 55 & 9.66 & 32.8 & 36.8 & 51.2 & 54.8 & 0.11 & 3.95 & 0.21 & 0.96\\
		\rowcolor{colorbrewer3!15}\hspace*{2px} TRADES $\lambda{=}9$ & 23.03 & 52.4 & 55 & 2.92 & 16.4 & 18.8 & 49.7 & 53 & 0.19 & 5.04 & 0.45 & 3.08\\
		\rowcolor{colorbrewer3!15}\hspace*{2px} Cyclic $\times3$ & 20.04 & 53.1 & 55.2 & 5.62 & 26.9 & 30.6 & 53.1 & 55.2 & 0.1 & 4.1 & 0.53 & 0.93\\
		\rowcolor{colorbrewer3!15}\hspace*{2px} Cyclic & 22.42 & 53.2 & 55.4 & 13.09 & 39.5 & 43.5 & 53.2 & 55.4 & 0.07 & 2.6 & 0.22 & 0.41\\
		\rowcolor{colorbrewer3!15}\hspace*{2px} Label noise $\tau{=}0.5$ & 22.71 & 51.3 & 55.4 & 15.04 & 40.4 & 45.5 & 51.3 & 55.4 & 0.09 & 0.43 & 0.16 & 0.13\\
		\rowcolor{colorbrewer3!15}\hspace*{2px} Label noise $\tau{=}0.3$ & 19.9 & 54.2 & 56.2 & 5.47 & 26.9 & 30 & 51.8 & 55.5 & 0.15 & 3.37 & 0.33 & 0.93\\
		\rowcolor{colorbrewer3!15}\hspace*{2px} Weight clipping $w_{max}{=}0.005$ & 21.39 & 54.1 & 56.5 & 10.19 & 35.6 & 39 & 54.1 & 56.5 & 0.74 & 10.49 & 0.41 & 4.58\\
		\rowcolor{colorbrewer3!15}\hspace*{2px} TRADES $\lambda{=}6$ & 21.68 & 55.3 & 56.7 & 1.74 & 13.5 & 15.8 & 50.1 & 53.4 & 0.21 & 5.12 & 0.57 & 2.26\\
		\rowcolor{colorbrewer3!15}\hspace*{2px} Cyclic $\times4$ & 19.85 & 55.2 & 56.9 & 4.01 & 23.1 & 26 & 55.1 & 56.9 & 0.16 & 6.65 & 0.62 & 0.8\\
		\hline
		\rowcolor{colorbrewer5!15}\hspace*{2px} Self-supervision $\lambda{=}4$ & 17.1 & 55.3 & 57.1 & 5.76 & 41.9 & 45 & 55.3 & 56.8 & 0.12 & 5.59 & 0.34 & 2.64\\
		\rowcolor{colorbrewer5!15}\hspace*{2px} Adam & 25.84 & 53.9 & 57.5 & 18.87 & 47.9 & 52.3 & 53.9 & 57.5 & 0.22 & 2.65 & 0.56 & 0.9\\
		\rowcolor{colorbrewer5!15}\hspace*{2px} Entropy-SGD ($L{=}2$) & 24.53 & 54.4 & 57.6 & 9.03 & 35.4 & 38.8 & 52.6 & 55.2 & 0.08 & 1.76 & 0.27 & 0.7\\
		\rowcolor{colorbrewer5!15}\hspace*{2px} Self-supervision $\lambda{=}1$ & 15.9 & 56.9 & 58.1 & 1.48 & 28.3 & 31.6 & 55.9 & 57.5 & 0.12 & 6.98 & 0.46 & 3.87\\
		\rowcolor{colorbrewer5!15}\hspace*{2px} Weight decay $0.05$ & 19.32 & 56.2 & 58.1 & 5.03 & 29 & 32.8 & 52 & 54.8 & 0.12 & 5.77 & 0.51 & 3.94\\
		\rowcolor{colorbrewer5!15}\hspace*{2px} Batch size $8$ & 17.73 & 57.1 & 58.2 & 3.46 & 26.8 & 31.4 & 55.6 & 58.2 & 0.32 & 24.01 & 1.55 & 12.27\\
		\rowcolor{colorbrewer5!15}\hspace*{2px} Entropy-SGD ($L{=}1$) & 25.42 & 56 & 58.6 & 12.79 & 42.8 & 46.1 & 53.2 & 56.9 & 0.09 & 3.24 & 0.28 & 1.8\\
		\rowcolor{colorbrewer5!15}\hspace*{2px} Self-supervision $\lambda{=}0.5$ & 16.16 & 58 & 58.6 & 1.26 & 28 & 30.7 & 56.7 & 58.3 & 0.1 & 6.48 & 0.45 & 3.29\\
		\rowcolor{colorbrewer5!15}\hspace*{2px} AT-AWP $\xi{=}0.001$ & 18.75 & 57.3 & 58.7 & 1.34 & 15.1 & 18.3 & 52.1 & 54.6 & 0.34 & 20.42 & 1.44 & 13.82\\
		\rowcolor{colorbrewer5!15}\hspace*{2px} Self-supervision $\lambda{=}2$ & 15.72 & 57.4 & 58.7 & 2.47 & 33.4 & 36.6 & 55.8 & 57.7 & 0.1 & 21.79 & 0.47 & 3.47\\
		\rowcolor{colorbrewer5!15}\hspace*{2px} MART $\lambda{=}9$ & 22.06 & 57 & 58.8 & 3.86 & 16 & 22 & 50 & 55 & 0.18 & 8.08 & 0.7 & 3.42\\
		\rowcolor{colorbrewer5!15}\hspace*{2px} Weight decay $0.01$ & 18.52 & 57.2 & 58.9 & 2.06 & 20.1 & 23.2 & 51.7 & 55.3 & 0.25 & 16.46 & 0.9 & 7.19\\
		\rowcolor{colorbrewer5!15}\hspace*{2px} Batch size $16$ & 18.12 & 58.3 & 59 & 1.82 & 20.4 & 24.5 & 52.5 & 55.6 & 0.33 & 22.11 & 1.41 & 11.39\\
		\rowcolor{colorbrewer5!15}\hspace*{2px} Self-supervision $\lambda{=}8$ & 19.6 & 56.6 & 59 & 12.08 & 50 & 53.3 & 56.6 & 58.6 & 0.11 & 3.59 & 0.29 & 1.76\\
		\rowcolor{colorbrewer5!15}\hspace*{2px} TRADES $\lambda{=}3$ & 20.51 & 57.7 & 59.1 & 0.94 & 13.4 & 15.5 & 52.3 & 54.9 & 0.2 & 19.08 & 0.71 & 3.48\\
		\rowcolor{colorbrewer5!15}\hspace*{2px} Weight decay $0.005$ & 18.79 & 58.2 & 59.4 & 2.03 & 20.2 & 23.9 & 51.8 & 54.3 & 0.26 & 19.67 & 1.2 & 8.35\\
		\rowcolor{colorbrewer5!15}\hspace*{2px} Label noise $\tau{=}0.2$ & 19.45 & 57.5 & 59.5 & 2.34 & 18.8 & 22.2 & 50.2 & 53 & 0.18 & 9.79 & 0.39 & 1.4\\
		\rowcolor{colorbrewer5!15}\hspace*{2px} MART $\lambda{=}3$ & 20.89 & 58.9 & 59.6 & 1.94 & 14.4 & 19.2 & 53.3 & 57.4 & 0.17 & 10.53 & 1.01 & 3.99\\
		\rowcolor{colorbrewer5!15}\hspace*{2px} Weight clipping $w_{max}{=}0.01$ & 19.15 & 58 & 59.6 & 3.28 & 21.5 & 24.8 & 56.7 & 58.5 & 0.66 & 15.1 & 0.26 & 7.41\\
		\rowcolor{colorbrewer5!15}\hspace*{2px} Learning rate $0.2$ & 19.17 & 58.3 & 59.7 & 0.46 & 9.4 & 12.4 & 54.3 & 56.6 & 0.2 & 24.41 & 1.44 & 5.75\\
		\rowcolor{colorbrewer5!15}\hspace*{2px} MiSH & 19.29 & 58.9 & 59.8 & 0.06 & 4.5 & 5.3 & 51.8 & 53.7 & 0.25 & 10.04 & 1.58 & 3.55\\
		\rowcolor{colorbrewer5!15}\hspace*{2px} ``Late'' multi-step & 20.63 & 58.5 & 59.8 & 1.6 & 16.4 & 18.4 & 54.2 & 57.8 & 0.17 & 5.24 & 0.81 & 2.96\\
		\hline
		\rowcolor{colorbrewer1!15}\hspace*{2px} SiLU & 19.45 & 59.7 & 60 & 0.07 & 4.8 & 5.6 & 51.3 & 53.7 & 0.3 & 9.97 & 1.68 & 4.2\\
		\rowcolor{colorbrewer1!15}\hspace*{2px} Weight averaging ($\tau{=}0.9975$) & 19.63 & 59.7 & 60 & 0.19 & 7.9 & 10 & 50.5 & 53 & 0.23 & 15.66 & 1.29 & 6\\
		\rowcolor{colorbrewer1!15}\hspace*{2px} Weight clipping $0.025$ & 18.91 & 59.2 & 60.4 & 0.73 & 12.5 & 15.6 & 52.1 & 54.9 & 0.32 & 17.4 & 0 & 8.61\\
		\rowcolor{colorbrewer1!15}\hspace*{2px} Batch size $32$ & 18.72 & 59.6 & 60.5 & 0.56 & 12 & 14.6 & 53.7 & 55.6 & 0.18 & 19.34 & 1.22 & 7.88\\
		\rowcolor{colorbrewer1!15}\hspace*{2px} Entropy-SGD ($L{=}3$) & 24 & 58.5 & 60.5 & 5.25 & 29.9 & 33.9 & 56.7 & 59.3 & 0.09 & 2.91 & 0.33 & 1.03\\
		\rowcolor{colorbrewer1!15}\hspace*{2px} Label noise $\tau{=}0.1$ & 19.39 & 59 & 60.8 & 1.12 & 14.1 & 17.5 & 51.9 & 55 & 0.2 & 16.75 & 0.69 & 3.55\\
		\rowcolor{colorbrewer1!15}\hspace*{2px} Larger $\epsilon{=}9/255$ & 21.3 & 60.4 & 60.9 & 0.47 & 8.9 & 11.1 & 51.3 & 53.8 & 0.21 & 10.26 & 1.34 & 5.85\\
		\rowcolor{colorbrewer1!15}\hspace*{2px} Label smoothing $\tau{=}0.1$ & 19.55 & 59.6 & 61 & 0.2 & 6.4 & 8.5 & 52.5 & 55 & 0.26 & 8.87 & 0.85 & 2.66\\
		\rowcolor{colorbrewer1!15}\hspace*{2px} MART $\lambda{=}6$ & 21.51 & 58.7 & 61 & 3.21 & 16.1 & 20.8 & 49.2 & 54.7 & 0.18 & 13.52 & 0.74 & 3.17\\
		\rowcolor{colorbrewer1!15}\hspace*{2px} Weight averaging ($\tau{=}0.98$) & 20.01 & 60.6 & 61 & 0.2 & 7.6 & 9.9 & 54.3 & 56.3 & 0.23 & 12.8 & 1.37 & 5.6\\
		\rowcolor{colorbrewer1!15}\hspace*{2px} Weight decay $0.001$ & 19.47 & 59.9 & 61 & 0.36 & 10.4 & 13.3 & 52 & 54.8 & 0.24 & 8.36 & 1.3 & 6.78\\
		\rowcolor{colorbrewer1!15}\hspace*{2px} Batch size $64$ & 19.06 & 60.5 & 61.1 & 0.3 & 9.2 & 11.1 & 51.2 & 54.4 & 0.18 & 23.13 & 1.14 & 5.96\\
		\rowcolor{colorbrewer1!15}\hspace*{2px} GeLU & 20.64 & 60.8 & 61.1 & 0.01 & 2.7 & 3.2 & 54.9 & 56.7 & 0.23 & 14.31 & 1.56 & 4.13\\
		\rowcolor{colorbrewer1!15}\hspace*{2px} Label smoothing $\tau{=}0.3$ & 19.41 & 59.4 & 61.2 & 0.27 & 5.7 & 8 & 51.1 & 54 & 0.29 & 18.42 & 0.65 & 2.72\\
		\rowcolor{colorbrewer1!15}\hspace*{2px} MART $\lambda{=}1$ & 20.51 & 59.4 & 61.2 & 1.04 & 11.4 & 14.7 & 50.3 & 55.4 & 0.17 & 7.97 & 0.87 & 3.1\\
		\rowcolor{colorbrewer1!15}\hspace*{2px} Weight averaging ($\tau{=}0.99$) & 20.41 & 60.3 & 61.4 & 0.19 & 7.8 & 9.6 & 51.7 & 54.2 & 0.22 & 6.12 & 1.44 & 4.98\\
		\rowcolor{colorbrewer1!15}\hspace*{2px} Dropout & 18.91 & 60.5 & 61.6 & 0.58 & 13 & 16.7 & 51.2 & 54.5 & 0.2 & 13.81 & 1.52 & 7.01\\
		\rowcolor{colorbrewer1!15}\hspace*{2px} PGD-14 & 20.8 & 60.6 & 61.6 & 0.22 & 7.1 & 9.3 & 53.6 & 56.1 & 0.27 & 20.9 & 1.48 & 5.35\\

		\rowcolor{colorbrewer1!15}\hspace*{2px} Entropy-SGD ($L{=}5$) & 23.48 & 59.5 & 61.7 & 3.01 & 22.2 & 25.9 & 53.2 & 56.6 & 0.1 & 3.57 & 0.46 & 1.49\\
		\rowcolor{colorbrewer1!15}\hspace*{2px} Ignore incorrect & 18.4 & 60.5 & 61.8 & 0.06 & 6.3 & 9 & 54.4 & 56.4 & 0.21 & 14.65 & 1.68 & 5.93\\
		\rowcolor{colorbrewer1!15}\hspace*{2px} Learning rate $0.1$ & 19.23 & 61.1 & 61.9 & 0.26 & 8.9 & 11.5 & 51.9 & 54.2 & 0.21 & 17.63 & 1.23 & 5.26\\
		\rowcolor{colorbrewer1!15}\hspace*{2px} TRADES $\lambda{=}1$ & 17.54 & 59.5 & 61.9 & 0.15 & 16.6 & 20.7 & 56.6 & 59.6 & 0.16 & 12.68 & 0.78 & 4.3\\
		
		\rowcolor{colorbrewer1!15}\hspace*{2px} Weight averaging ($\tau{=}0.985$) & 20.27 & 61.7 & 62.3 & 0.18 & 7.4 & 9.4 & 55.9 & 58 & 0.22 & 15.66 & 1.35 & 6.51\\
		\rowcolor{colorbrewer1!15}\hspace*{2px} Label smoothing $\tau{=}0.2$ & 20.07 & 60.2 & 62.4 & 0.26 & 5.1 & 7.8 & 51.9 & 54.6 & 0.28 & 9.94 & 0.69 & 2.61\\
		\rowcolor{colorbrewer1!15}\hspace*{2px} Prevent label leaking & 18.38 & 62.1 & 62.4 & 0.38 & 8.6 & 10.8 & 55.3 & 57.7 & 0.22 & 14.62 & 1.48 & 6\\
		\rowcolor{colorbrewer1!15}\hspace*{2px} AT (baseline) & 20.2 & 61 & 62.8 & 0.33 & 8.5 & 10.7 & 52.3 & 54.6 & 0.21 & 21.05 & 1.22 & 6.49\\
		\hline
		\rowcolor{colorbrewer0!15}\hspace*{2px} Const learning rate $0.05$ & 24.96 & 60.7 & 62.9 & 6.17 & 32.9 & 37.8 & 55.4 & 58.9 & 0.09 & 3.52 & 0.44 & 0.9\\
		\rowcolor{colorbrewer0!15}\hspace*{2px} PGD-5 & 20.22 & 61.8 & 62.9 & 0.11 & 7.3 & 10.4 & 55.1 & 57.4 & 0.17 & 20.4 & 1.24 & 4.19\\
		\rowcolor{colorbrewer0!15}\hspace*{2px} Batch size $256$ & 20.86 & 62.6 & 63.3 & 0.28 & 8.2 & 10.3 & 56.9 & 58.4 & 0.3 & 11.22 & 1.35 & 8.33\\
		\rowcolor{colorbrewer0!15}\hspace*{2px} PGD-$7$-$3$ & 17.17 & 61.7 & 63.3 & 0.08 & 19.5 & 25.2 & 51.3 & 58.8 & 0.17 & 7.4 & 1.08 & 5.29\\
		\rowcolor{colorbrewer0!15}\hspace*{2px} Batch size $512$ & 22.58 & 62.9 & 63.5 & 0.64 & 11 & 14.2 & 58.6 & 60 & 0.48 & 23.97 & 1.92 & 16.22\\
		\rowcolor{colorbrewer0!15}\hspace*{2px} Learning rate $0.01$ & 22.83 & 63 & 63.5 & 1.05 & 15.2 & 18 & 57.8 & 59.7 & 0.56 & 23.42 & 2.25 & 16.02\\
		\rowcolor{colorbrewer0!15}\hspace*{2px} No weight decay & 23.37 & 64.8 & 65.7 & 0.23 & 9.2 & 12.7 & 57.1 & 60.3 & 0.66 & 21.05 & 2.53 & 11.75\\
		\rowcolor{colorbrewer0!15}\hspace*{2px} PGD-$7$-$0$ & 14.67 & 63.8 & 65.7 & 0.09 & 23.7 & 30 & 59.4 & 61.4 & 0.11 & 6.86 & 1.28 & 2.8\\
		\rowcolor{colorbrewer0!15}\hspace*{2px} PGD-$7$-$2$ & 16.19 & 63.6 & 65.9 & 0.1 & 20.9 & 28.1 & 58.3 & 62.3 & 0.14 & 22.81 & 1.21 & 2.55\\
		\rowcolor{colorbrewer0!15}\hspace*{2px} PGD-$7$-$1$ & 15.02 & 64.1 & 67.1 & 0.11 & 25.9 & 34.3 & 58.8 & 63.8 & 0.13 & 11.71 & 1.15 & 2.33\\
		\rowcolor{colorbrewer0!15}\hspace*{2px} Const learning rate $0.01$ & 25.87 & 66.7 & 67.4 & 0.67 & 18.5 & 20.7 & 58.4 & 61 & 0.33 & 15.09 & 1.37 & 8.27\\
		\rowcolor{colorbrewer0!15}\hspace*{2px} Const learning rate $0.005$ & 27.24 & 68.3 & 69.2 & 0.42 & 15.5 & 16.7 & 61.1 & 65.5 & 0.59 & 20.63 & 2.06 & 15.74\\
		\hline
	\end{tabularx}
	}
	\vspace*{-8px}
	\caption{\textbf{Results: \TE, \RTE and Flatness in \CE and \RCE.} \TE and \RTE (PGD-$20$ and AutoAttack \cite{CroceICML2020}) on test and train examples, together with average- and worst-case flatness in (clean) \CE and \RCE. Methods sorted by (test) \RTE against AutoAttack and split into \colorbox{colorbrewer3!15}{good}, \colorbox{colorbrewer5!15}{average}, \colorbox{colorbrewer1!15}{poor} and \colorbox{colorbrewer0!15}{worse} robustness at $57\%$, $60\%$ and $62.8\%$ \RTE, see text.}
	\label{tab:supp-table-error}
	\vspace*{-6px}
\end{table*}
\begin{table*}[t]
	\centering
	\vspace*{-0.5cm}
	\scriptsize
	{
	\begin{tabularx}{\textwidth}{|X|c|c|c||c|c|c||c|c|}
		\hline
		Model & \multicolumn{3}{c||}{\bfseries Test Robustness} & \multicolumn{3}{c||}{\bfseries Train Robustness} & \multicolumn{2}{c|}{\bfseries Early Stopping}\\
		\hline
		(sorted by \RCE on PGD) & \CE & \RCE & \RCE & \CE & \RCE & \RCE & \RCE & \RCE\\
		(PGD = PGD-$20$, $10$ restarts) & (test) & (test) & (test) & (train) & (train) & (train) & (stop) & (stop)\\
		(AA = AutoAttack \cite{CroceARXIV2020}) && (PGD) & (AA) && (PGD) & (AA) & (PGD) & (AA)\\
		\hline
		\hline
		+Unlabeled & 0.57 & 1.18 & 0.67 & 0.47 & 0.94 & 0.56 & 1.18 & 0.67\\
		AutoAugment & 0.58 & 1.3 & 0.71 & 0.48 & 1.08 & 0.61 & 1.3 & 0.71\\
		AT-AWP $\xi{=}0.01$ & 0.7 & 1.31 & 0.81 & 0.55 & 0.99 & 0.62 & 1.3 & 0.81\\
		Cyclic & 0.68 & 1.41 & 0.8 & 0.49 & 0.97 & 0.58 & 1.41 & 0.8\\
		Adam & 0.8 & 1.46 & 0.89 & 0.66 & 1.19 & 0.74 & 1.45 & 0.89\\
		Weight clipping $w_{max}{=}0.005$ & 0.77 & 1.48 & 0.91 & 0.53 & 0.99 & 0.62 & 1.47 & 0.9\\
		TRADES $\lambda{=}9$ & 0.77 & 1.52 & 0.9 & 0.33 & 0.58 & 0.37 & 1.42 & 0.9\\
		Label noise $\tau{=}0.4$ & 0.93 & 1.55 & 1.05 & 0.71 & 1.15 & 0.8 & 1.5 & 1.05\\
		Cyclic $\times2$ & 0.6 & 1.55 & 0.74 & 0.32 & 0.76 & 0.42 & 1.55 & 0.74\\
		AT-AWP $\xi{=}0.005$ & 0.59 & 1.57 & 0.74 & 0.29 & 0.66 & 0.38 & 1.36 & 0.74\\
		Self-supervision $\lambda{=}8$ & 0.59 & 1.58 & 0.76 & 0.43 & 1.24 & 0.62 & 1.57 & 0.76\\
		Entropy-SGD ($L{=}2$) & 0.72 & 1.59 & 0.83 & 0.4 & 0.87 & 0.5 & 1.44 & 0.83\\
		Label noise $\tau{=}0.5$ & 1.12 & 1.59 & 1.22 & 1 & 1.39 & 1.08 & 1.59 & 1.22\\
		Entropy-SGD ($L{=}1$) & 0.77 & 1.59 & 0.87 & 0.5 & 1.06 & 0.6 & 1.44 & 0.87\\
		Label noise $\tau{=}0.3$ & 0.78 & 1.62 & 0.94 & 0.45 & 0.91 & 0.55 & 1.47 & 0.94\\
		Weight decay $0.05$ & 0.61 & 1.65 & 0.78 & 0.28 & 0.73 & 0.39 & 1.33 & 0.78\\
		Self-supervision $\lambda{=}4$ & 0.51 & 1.68 & 0.71 & 0.25 & 1.02 & 0.45 & 1.62 & 0.71\\
		Weight clipping $w_{max}{=}0.01$ & 0.62 & 1.71 & 0.83 & 0.22 & 0.61 & 0.31 & 1.59 & 0.83\\
		TRADES $\lambda{=}6$ & 0.7 & 1.74 & 0.86 & 0.2 & 0.44 & 0.25 & 1.4 & 0.86\\
		Cyclic $\times3$ & 0.6 & 1.75 & 0.74 & 0.24 & 0.65 & 0.34 & 1.59 & 0.74\\
		Entropy-SGD ($L{=}3$) & 0.69 & 1.84 & 0.85 & 0.26 & 0.72 & 0.38 & 1.57 & 0.85\\
		Self-supervision $\lambda{=}2$ & 0.47 & 1.86 & 0.69 & 0.13 & 0.8 & 0.33 & 1.53 & 0.69\\
		
		Label noise $\tau{=}0.2$ & 0.68 & 1.89 & 0.9 & 0.22 & 0.63 & 0.32 & 1.4 & 0.9\\
		Cyclic $\times4$ & 0.6 & 1.9 & 0.78 & 0.2 & 0.57 & 0.28 & 1.44 & 0.78\\
		Const learning rate $0.05$ & 0.75 & 1.92 & 0.89 & 0.31 & 0.81 & 0.43 & 1.54 & 0.88\\
		Self-supervision $\lambda{=}0.5$ & 0.48 & 2.01 & 0.71 & 0.09 & 0.67 & 0.27 & 1.6 & 0.71\\
		Entropy-SGD ($L{=}5$) & 0.71 & 2.06 & 0.89 & 0.18 & 0.57 & 0.28 & 1.5 & 0.86\\
		Self-supervision $\lambda{=}1$ & 0.48 & 2.08 & 0.72 & 0.09 & 0.67 & 0.27 & 1.63 & 0.7\\
		Label smoothing $\tau{=}0.3$ & 0.77 & 2.12 & 1.02 & 0.15 & 0.47 & 0.21 & 1.46 & 0.97\\
		TRADES $\lambda{=}3$ & 0.65 & 2.16 & 0.85 & 0.1 & 0.34 & 0.17 & 1.42 & 0.83\\
		Batch size $8$ & 0.56 & 2.22 & 0.78 & 0.17 & 0.64 & 0.3 & 1.86 & 0.76\\
		Label noise $\tau{=}0.1$ & 0.63 & 2.22 & 0.87 & 0.1 & 0.42 & 0.19 & 1.37 & 0.86\\
		Weight decay $0.01$ & 0.58 & 2.23 & 0.78 & 0.12 & 0.47 & 0.22 & 1.35 & 0.78\\
		Label smoothing $\tau{=}0.2$ & 0.71 & 2.26 & 0.98 & 0.09 & 0.35 & 0.14 & 1.44 & 0.89\\
		
		MART $\lambda{=}9$ & 0.7 & 2.36 & 0.93 & 0.24 & 0.48 & 0.3 & 1.41 & 0.93\\
		Weight clipping $0.025$ & 0.57 & 2.37 & 0.81 & 0.06 & 0.32 & 0.14 & 1.39 & 0.81\\
		Weight decay $0.005$ & 0.58 & 2.44 & 0.8 & 0.11 & 0.46 & 0.23 & 1.43 & 0.76\\
		Label smoothing $\tau{=}0.1$ & 0.64 & 2.48 & 0.88 & 0.04 & 0.24 & 0.09 & 1.43 & 0.8\\
		MART $\lambda{=}6$ & 0.71 & 2.58 & 0.93 & 0.21 & 0.45 & 0.27 & 1.4 & 0.93\\
		``Late'' multi-step & 0.66 & 2.63 & 0.87 & 0.09 & 0.36 & 0.17 & 1.47 & 0.87\\
		TRADES $\lambda{=}1$ & 0.56 & 2.68 & 0.81 & 0.04 & 0.38 & 0.18 & 1.74 & 0.74\\
		MART $\lambda{=}3$ & 0.69 & 2.71 & 0.89 & 0.14 & 0.38 & 0.21 & 1.48 & 0.89\\
		AT-AWP $\xi{=}0.001$ & 0.62 & 2.71 & 0.84 & 0.08 & 0.34 & 0.16 & 1.35 & 0.78\\
		Batch size $16$ & 0.57 & 2.78 & 0.83 & 0.1 & 0.46 & 0.22 & 1.63 & 0.74\\
		Learning rate $0.01$ & 0.72 & 2.94 & 0.96 & 0.07 & 0.34 & 0.16 & 1.74 & 0.85\\
		MART $\lambda{=}1$ & 0.7 & 2.99 & 0.94 & 0.08 & 0.29 & 0.15 & 1.44 & 0.94\\
		PGD-$7$-$3$ & 0.55 & 3.03 & 0.8 & 0.04 & 0.48 & 0.19 & 1.7 & 0.73\\
		PGD-$7$-$2$ & 0.54 & 3.2 & 0.79 & 0.03 & 0.48 & 0.21 & 2.12 & 0.71\\
		PGD-$7$-$1$ & 0.51 & 3.3 & 0.75 & 0.04 & 0.61 & 0.25 & 2.43 & 0.68\\
		Batch size $512$ & 0.77 & 3.41 & 1.04 & 0.05 & 0.27 & 0.12 & 1.86 & 0.85\\
		PGD-$7$-$0$ & 0.49 & 3.43 & 0.74 & 0.03 & 0.58 & 0.21 & 2.48 & 0.65\\
		Dropout & 0.66 & 3.44 & 0.9 & 0.04 & 0.3 & 0.13 & 1.44 & 0.76\\
		
		Const learning rate $0.01$ & 0.89 & 3.46 & 1.07 & 0.04 & 0.43 & 0.17 & 1.67 & 0.97\\
		Weight decay $0.001$ & 0.69 & 3.52 & 0.92 & 0.03 & 0.24 & 0.1 & 1.41 & 0.77\\
		Batch size $32$ & 0.67 & 3.54 & 0.91 & 0.04 & 0.27 & 0.11 & 1.69 & 0.73\\
		Batch size $64$ & 0.69 & 3.62 & 0.93 & 0.03 & 0.22 & 0.09 & 1.44 & 0.76\\
		Const learning rate $0.005$ & 0.97 & 3.65 & 1.24 & 0.04 & 0.35 & 0.14 & 1.62 & 1.12\\
		MiSH & 0.69 & 3.65 & 0.92 & 0.01 & 0.1 & 0.04 & 1.45 & 0.73\\
		Learning rate $0.1$ & 0.7 & 3.65 & 0.94 & 0.02 & 0.19 & 0.09 & 1.39 & 0.79\\
		Learning rate $0.2$ & 0.72 & 3.66 & 0.97 & 0.03 & 0.21 & 0.1 & 1.67 & 0.75\\
		Larger $\epsilon{=}9/255$ & 0.78 & 3.69 & 1.01 & 0.03 & 0.19 & 0.09 & 1.45 & 0.79\\
		Prevent label leaking & 0.67 & 3.76 & 0.91 & 0.02 & 0.21 & 0.08 & 1.74 & 0.7\\
		SiLU & 0.71 & 3.81 & 0.96 & 0.01 & 0.11 & 0.04 & 1.47 & 0.73\\
		PGD-14 & 0.76 & 3.84 & 1.05 & 0.02 & 0.15 & 0.07 & 1.52 & 0.78\\
		Weight averaging ($\tau{=}0.9975$) & 0.73 & 3.88 & 1 & 0.02 & 0.17 & 0.08 & 1.34 & 0.81\\
		AT (baseline) & 0.75 & 3.91 & 0.99 & 0.02 & 0.19 & 0.08 & 1.53 & 0.77\\
		Weight averaging ($\tau{=}0.98$) & 0.75 & 3.94 & 1.05 & 0.02 & 0.16 & 0.08 & 1.96 & 0.8\\
		Weight averaging ($\tau{=}0.985$) & 0.75 & 3.95 & 1 & 0.02 & 0.16 & 0.07 & 1.94 & 0.77\\
		Ignore incorrect & 0.71 & 3.99 & 0.96 & 0.01 & 0.16 & 0.07 & 1.65 & 0.7\\
		Weight averaging ($\tau{=}0.99$) & 0.76 & 3.99 & 1.07 & 0.02 & 0.18 & 0.07 & 1.53 & 0.74\\
		Batch size $256$ & 0.79 & 4.02 & 1.07 & 0.02 & 0.18 & 0.08 & 1.74 & 0.81\\
		No weight decay & 0.9 & 4.17 & 1.15 & 0.02 & 0.22 & 0.1 & 1.55 & 0.91\\
		PGD-5 & 0.77 & 4.22 & 0.99 & 0.01 & 0.17 & 0.08 & 1.62 & 0.77\\
		GeLU & 0.82 & 4.27 & 1.06 & 0 & 0.06 & 0.02 & 1.7 & 0.79\\
		\hline
	\end{tabularx} 
	}
	\vspace*{-8px}
	\caption{\textbf{Results: \emph{\CE} and \emph{\RCE}.} \CE and \RCE (PGD-$20$ and AutoAttack \cite{CroceICML2020}) on test and train examples corresponding to the results in \tabref{tab:supp-table-error}.}
	\label{tab:supp-table-loss}
	\vspace*{-6px}
\end{table*}